\def\year{2021}\relax
\documentclass[letterpaper]{article} 
\usepackage{aaai21}  
\usepackage{times}  
\usepackage{helvet} 
\usepackage{courier}  
\usepackage[hyphens]{url}  
\usepackage{graphicx} 
\usepackage{natbib}  
\usepackage{caption} 
\usepackage{color,soul}

\usepackage{kotex} 
\usepackage{amsmath}
\usepackage{amssymb}
\usepackage[utf8]{inputenc}
\usepackage[ruled,vlined]{algorithm2e}
\usepackage{subcaption}
\usepackage{hhline}
\usepackage{graphics}

\title{Understanding Catastrophic Overfitting in Single-step Adversarial Training}
\author {
    Hoki Kim\thanks{Equal contribution.}, 
    Woojin Lee\footnotemark[1], 
    Jaewook Lee\thanks{Corresponding author.} \\
}
\affiliations {
    Seoul National University, Seoul, Korea \\
    ghrl9613@snu.ac.kr, wj926@snu.ac.kr, jaewook@snu.ac.kr

}

\begin{document}

\maketitle

\begin{abstract}
Although fast adversarial training has demonstrated both robustness and efficiency, the problem of ``catastrophic overfitting'' has been observed. This is a phenomenon in which, during single-step adversarial training, robust accuracy against projected gradient descent (PGD) suddenly decreases to 0\% after a few epochs, whereas robust accuracy against fast gradient sign method (FGSM) increases to 100\%. 
In this paper, we demonstrate that catastrophic overfitting is very closely related to the characteristic of single-step adversarial training which uses only adversarial examples with the maximum perturbation, and not all adversarial examples in the adversarial direction, which leads to decision boundary distortion and a highly curved loss surface. Based on this observation, we propose a simple method that not only prevents catastrophic overfitting, but also overrides the belief that it is difficult to prevent multi-step adversarial attacks with single-step adversarial training. 
\end{abstract}

\section{Introduction}
Adversarial examples are perturbed inputs that are designed to deceive machine learning classifiers by adding adversarial noises to the original data. Although such perturbations are sufficiently subtle and undetectable by humans, they result in an incorrect classification. 
Since deep-learning models were found to be vulnerable to adversarial examples \cite{szegedy2013intriguing}, a line of work was proposed to mitigate the problem and improve robustness of the models. 
Among the numerous defensive methods, projected gradient descent (PGD) adversarial training \cite{madry2017towards} is one of the most successful approaches for achieving robustness against adversarial attacks. Although PGD adversarial training serves as a strong defensive algorithm, because it relies on a multi-step adversarial attack, a high computational cost is required for multiple forward and back propagation during batch training. 

\begin{figure}[t]
\centering
\includegraphics[width=0.9\columnwidth]{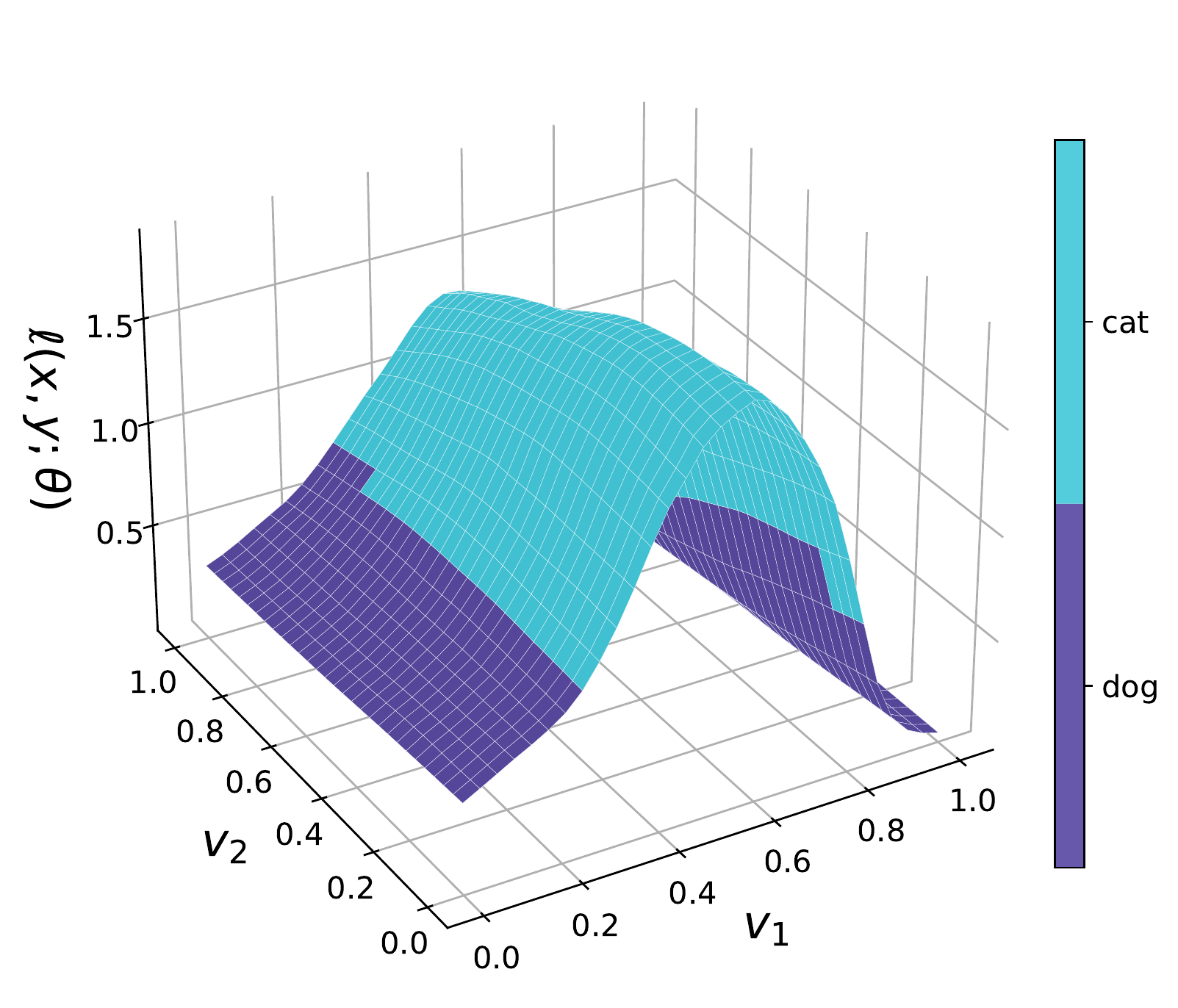} 
\caption{Visualization of distorted decision boundary. The origin indicates the original image $x$, the label of which is ``dog''. In addition, $v_1$ is the direction of a single-step adversarial perturbation and $v_2$ is a random direction. The adversarial image $x+v_1$ is classified as the correct label, although there is distorted interval where $x + k\cdot v_1$ is misclassified even when $k$ is less than 1. Due to this decision boundary distortion, single-step adversarial training becomes vulnerable to multi-step adversarial attacks.}
\label{fig:fast_after_co}
\end{figure}

To overcome this issue, other studies \cite{shafahi2019adversarial,wong2020fast} on reducing the computational burden of adversarial training using single-step adversarial attacks \cite{goodfellow2014explaining} have been proposed. 
Among them, inspired by \citet{shafahi2019adversarial}, \citet{wong2020fast} suggested fast adversarial training, which is a modified version of fast gradient sign method (FGSM) adversarial training designed to be as effective as PGD adversarial training.

Fast adversarial training has demonstrated both robustness and efficiency; however, it suffers from the problem of ``catastrophic overfitting,'' which is a phenomenon that robustness against PGD suddenly decreases to 0\%, whereas robustness against FGSM rapidly increases. \citet{wong2020fast} first discovered this issue and suggested the use of early stopping to prevent it. Later, it was found that catastrophic overfitting also occurs in different single-step adversarial training methods such as free adversarial training \cite{andriushchenko2020understanding}.

In this regard, few attempts have been made to discover the underlying reason for catastrophic overfitting and methods proposed to prevent this failure \cite{andriushchenko2020understanding, vivek2020single, li2020towards}. However, these approaches were computationally inefficient or did not provide a fundamental reason for the problem. 

In this study, we first analyze the differences before and after catastrophic overfitting. We then identify the relationship between distortion of the decision boundary and catastrophic overfitting. Unlike the previous notion in which a larger perturbation implies a stronger attack, we discover that sometimes a smaller perturbation is sufficient to fool the model, whereas the model is robust against larger perturbations during single-step adversarial training. We call this phenomenon ``decision boundary distortion.'' 

Figure \ref{fig:fast_after_co} shows an example of decision boundary distortion by visualizing the loss surface. The model is robust to perturbations when the magnitude of the attack is equal to the maximum perturbation $\epsilon$, but not to other smaller perturbations. When decision boundary distortion occurs, the model becomes more robust against a single-step adversarial attack but reveals fatal weaknesses to multi-step adversarial attacks and leads to catastrophic overfitting.

Through extensive experiments, we empirically discovered the relationship between single-step adversarial training and decision boundary distortion, and found that the problem of single-step adversarial training is a fixed magnitude of the perturbation, not the direction of the attack. Based on this observation, we present a simple algorithm that determines the appropriate magnitude of the perturbation for each image and prevents catastrophic overfitting.

\textbf{}

\textbf{Contributions.}
\begin{itemize} 
    \item We discovered a ``decision boundary distortion'' phenomenon that occurs during single-step adversarial training and the underlying connection between decision boundary distortion and catastrophic overfitting. 
    \item We suggest a simple method that prevents decision boundary distortion by searching the appropriate step size for each image. This method not only prevents catastrophic overfitting, but also achieves near 100\% accuracy for the training examples against PGD. 
    \item We evaluate robustness of the proposed method against various adversarial attacks (FGSM, PGD, and AutoAttack \cite{croce2020reliable}) and demonstrate the proposed method can provide sufficient robustness without catastrophic overfitting. 
    
\end{itemize}

\section{Background and Related Work}
\subsection{Adversarial Robustness}

There are two major movements for building a robust model: provable defenses and adversarial training. 


A considerable number of studies related to provable defenses of deep-learning models have been published. Provable defenses attempt to provide provable guarantees for robust performance, such as linear relaxations \cite{wong2018provable, zhang2019towards}, interval bound propagation \cite{gowal2018effectiveness, lee2020lipschitz}, and randomized smoothing \cite{cohen2019certified, salman2019provably}. However, provable defenses are computationally inefficient and show unsatisfied performance compared to adversarial training.

Adversarial training is an approach that augments adversarial examples generated by adversarial attacks \cite{goodfellow2014explaining,madry2017towards,tramer2017ensemble}. 
Because this approach is simple and achieves high empirical robustness for various attacks, it has been widely used and developed along with other deep learning methods such as mix-up \cite{zhang2017mixup,lamb2019interpolated,pang2019mixup} and unsupervised training \cite{alayrac2019labels, najafi2019robustness, carmon2019unlabeled}. 




In this study, we focus on adversarial training. Given an example $(x,y)\sim \mathcal{D}$, let $\ell(x,y;\theta)=\ell(f_\theta(x), y)$ denote the loss function of a deep learning model $f$ with parameters $\theta$. Then, adversarial training with a maximum perturbation $\epsilon$ can be formalized as follows: 
\begin{equation}
    \min_{\theta}\mathbb{E}_{(x,y)\sim \mathcal{D}} [ \max_{\delta \in \mathcal{B}(x, \epsilon)} \ell(x+\delta, y;\theta) ]
\label{eq:minmax}
\end{equation}
A perturbation $\delta$ is in $\mathcal{B}(x, \epsilon)$ that denotes the $\epsilon$-ball around an example $x$ with a specific distance measure. The most used distance measures are $L_0, L_2$, and $L_\infty$. In this study, we use $L_\infty$ for such a measure.

However, the above optimization is considered as NP-hard because it contains a non-convex min-max problem. Thus, instead of the inner maximization problem, adversarial attacks are used to find the perturbation $\delta$.

\textbf{Fast gradient sign method (FGSM)} \cite{goodfellow2014explaining} is the simplest adversarial attack, which uses a sign of a gradient to find an adversarial image $x'$. Because FGSM requires only one gradient, it is considered the least expensive adversarial attack \cite{goodfellow2014explaining, madry2017towards}.
\begin{equation}
    x' = x+\epsilon\cdot\text{sgn}(\nabla_x \ell(x, y; \theta))
\end{equation}


\textbf{Projected gradient descent (PGD)} \cite{madry2017towards} uses multiple gradients to generate more powerful adversarial examples. With a step size $\alpha$, PGD can be formalized as follows:
\begin{equation}
    x^{t+1} = \Pi_{\mathcal{B}(x, \epsilon)} (x^t + \alpha\cdot\text{sgn} (\nabla_x \ell(x, y; \theta)))
\end{equation}
where $\Pi_{\mathcal{B}(x, \epsilon)}$ refers the projection to the $\epsilon$-ball $\mathcal{B}(x, \epsilon)$. Here, $x^t$ is an adversarial example after $t$-steps. A large number of steps allows us to explore more areas in $\mathcal{B}(x, \epsilon)$. Note that PGD$n$ corresponds to PGD with $n$ steps (or iterations). For instance, PGD7 indicates that the number of PGD steps is 7.

\subsection{Single-step Adversarial Attack versus Multi-step Adversarial Attack}

Single-step adversarial training was previously believed to be a non-robust method because it produces nearly 0\% accuracy against PGD \cite{madry2017towards}. Moreover, the model trained using FGSM has been confirmed to have typical characteristics, such as gradient masking, which indicates that a single-step gradient is insufficient to find a decent adversarial examples \cite{tramer2017ensemble}. 
For the above reasons, a number of studies have been conducted on multi-step adversarial attacks.

Contrary to this perception, however, free adversarial training \cite{shafahi2019adversarial} has achieved a remarkable performance with a single-step gradient using redundant batches and accumulative perturbations. Following \citet{shafahi2019adversarial}, \citet{wong2020fast} proposed fast adversarial training using FGSM with a uniform random initialization. Fast adversarial training shows an almost equivalent performance to those of PGD \cite{madry2017towards} and free adversarial training \cite{shafahi2019adversarial}. 
\begin{equation}
\begin{split}
    & \eta = \text{Uniform}(-\epsilon, \epsilon)
    \\ & \delta = \eta +\alpha\cdot\text{sgn}(\nabla_\eta \ell(x+\eta, y; \theta))
    \\ & x' = x+\delta
\end{split}
\end{equation}

\subsection{Catastrophic Overfitting}

Although fast adversarial training performs well in a short time, a previously undiscovered phenomenon has been identified. That is, after a few epochs with single-step adversarial training, robustness of the model against PGD decreases sharply. This phenomenon is called catastrophic overfitting. Fast adversarial training \cite{wong2020fast} uses early stopping to temporally avoid catastrophic overfitting by tracking robustness accuracy against PGD on the training batches.

To apply early stopping, robustness against PGD must be continuously confirmed. Furthermore, standard accuracy does not yield the maximum potential \cite{andriushchenko2020understanding}. To resolve these shortcomings and gain a deeper understanding of catastrophic overfitting, a line of work has been proposed. \citet{vivek2020single} identified that catastrophic overfitting arises with early overfitting to FGSM. To prevent this type of overfitting, the authors introduced dropout scheduling and demonstrated stable adversarial training for up to 100 epochs. In addition, \citet{li2020towards} trained a model with FGSM at first and then changed it into PGD when there was a large decrease in the PGD accuracy. \citet{andriushchenko2020understanding} found that an abnormal behavior of a single filter leads to a nonlinear model with single-layer convolutional networks. Based on this observation, they proposed a regularization method, GradAlign, which maximizes $\cos(\nabla_x \ell(x,y;\theta), \nabla_{x} \ell(x+\eta,y;\theta))$ and prevents catastrophic overfitting by inducing a gradient alignment.

\begin{figure}[t]
    \centering
    \begin{subfigure}{0.40\textwidth}
    \includegraphics[width=\linewidth]{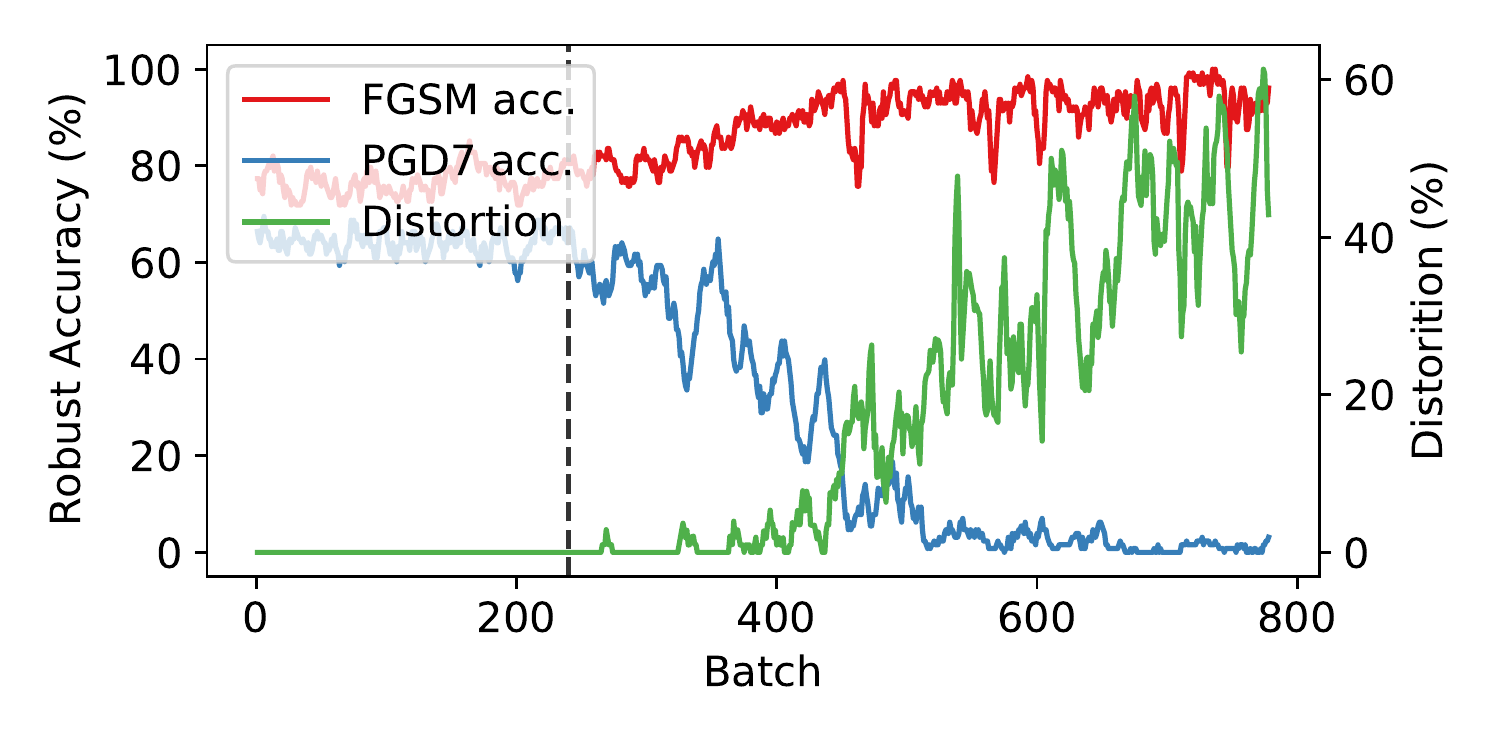}
    \caption{Robust accuracy and distortion}
    \end{subfigure}
    \begin{subfigure}{0.40\textwidth}
    \includegraphics[width=\linewidth]{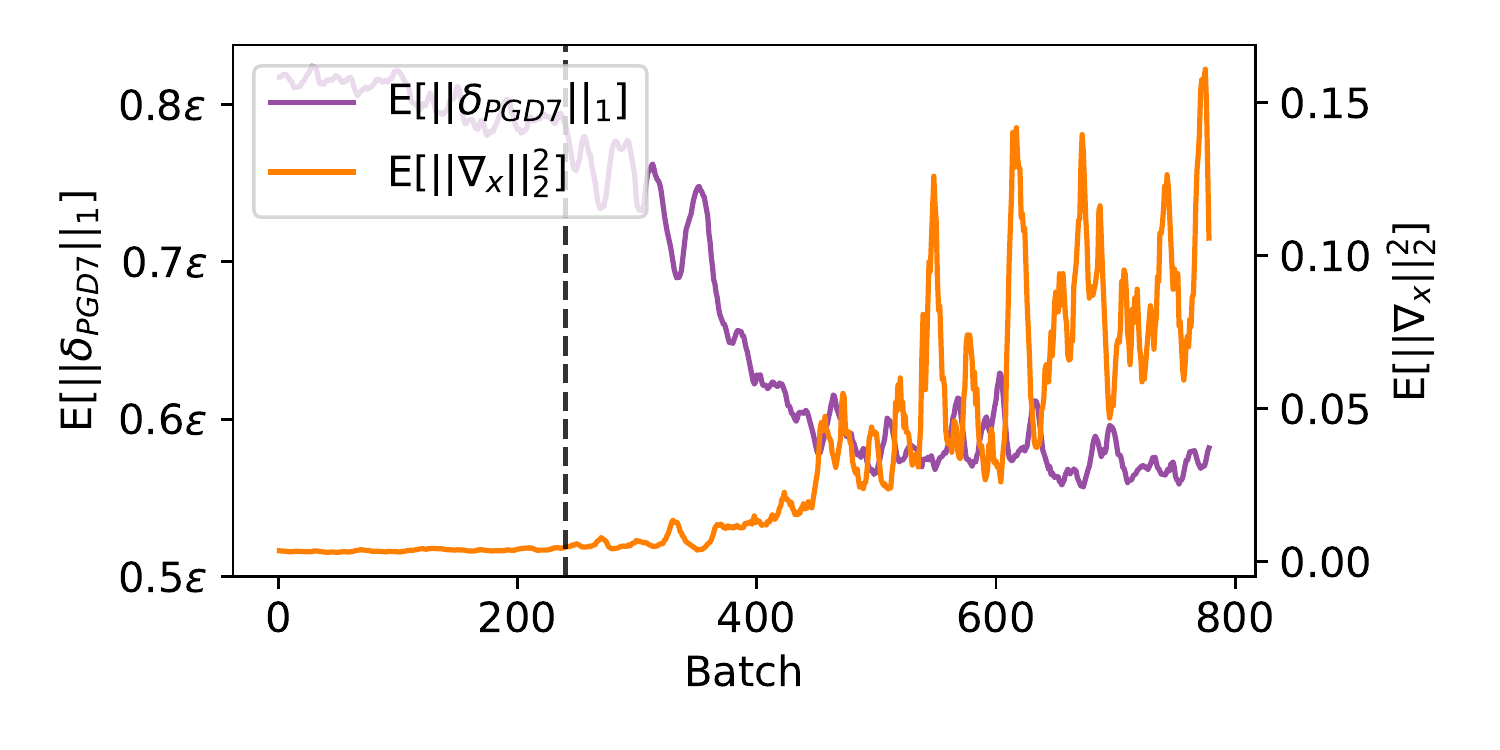}
    \caption{Mean of absolute value of PGD7 perturbations and $L_2$ norm of the gradients of the images}
    \end{subfigure}

    \caption{(CIFAR10) Analysis of catastrophic overfitting. Plot (a) shows robust accuracy of fast adversarial training against FGSM (red) and PGD7 (blue). Distortion (green) denotes the ratio of images in distorted interval in Equation (\ref{eq:distortion}). Plot (b) shows the mean of absolute value of PGD7 perturbation $\mathbb{E}[\vert\vert \delta_{PGD7} \vert\vert_1]$ (purple) and the $L_2$ norm of the gradients of the images $\mathbb{E}[\vert\vert \nabla_x \vert\vert_2]$ (orange). Dashed black lines correspond to the 240th batch, which is the start point of catastrophic overfitting in both plots.}
    \label{fig:batch_trace}
\end{figure}

\begin{figure*}[ht]
\centering
\includegraphics[width=\linewidth]{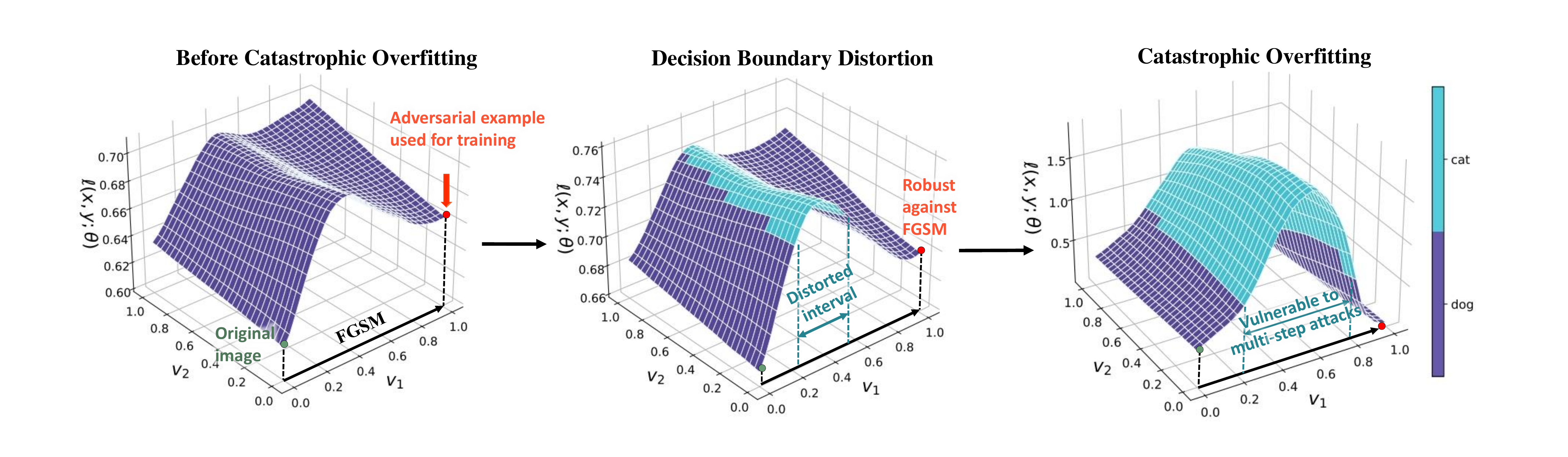} 
\caption{Process of normal decision boundary turns into distorted decision boundary. (Left) The loss surface before catastrophic overfitting with a FGSM adversarial direction $v_1$ and a random direction $v_2$. The red point denotes an adversarial example $x+v_1$ generated from the original image $x$, the label of which is ``dog.'' (Middle) The changed loss surface after learning adversarial example $x+v_1$. Here, $v_1$ is the same vector as that on the left. Distorted interval begins to occur for the first time. (Right) As training continues, distorted decision boundary grows uncontrollably such that robustness against multi-step adversarial attacks decreases.}
\label{fig:fast_before_co}
\end{figure*}

However, even with an increased understanding of catastrophic overfitting and methods for its prevention, a key question remains unanswered:
 
\textit{What characteristic of single-step adversarial attacks is the cause of catastrophic overfitting?}

In this paper, we discuss the cause of catastrophic overfitting in the context of single-step adversarial training. We then propose a new simple method to facilitate stable single-step adversarial training, wherein longer training can produce a higher standard accuracy with sufficient adversarial robustness.

\section{Revisiting Catastrophic Overfitting}
\label{sec:co}
First, to analyze catastrophic overfitting, we start by recording robust accuracy of fast adversarial training on CIFAR-10 \cite{krizhevsky2009learning}. The maximum perturbation $\epsilon$ is fixed to $8/255$. We use FGSM and PGD7 to verify robust accuracy with the same settings $\epsilon=8/255$ and a step size $\alpha=2/255$.

Figure \ref{fig:batch_trace} shows statistics on the training batch when catastrophic overfitting occurs (71st out of 200 epochs). In plot (a), after 240 batches, robustness against PGD7 begins to decrease rapidly; conversely, robustness against FGSM increases. Plot (b) shows the mean of the absolute value of PGD7 perturbation $\mathbb{E}[\vert\vert \delta_{PGD7} \vert\vert_1]$ and squared $L_2$ norm of the gradient of the images $\mathbb{E}[\vert\vert \nabla_x \vert\vert_2]$ of each batch. After catastrophic overfitting, there is a trend of decreasing mean perturbation. This is consistent with the phenomenon in which the perturbations of the catastrophic overfitted model are located away from the maximum perturbation, unlike the model that is stopped early \cite{wong2020fast}. Concurrently, a significant increase in the squared $L_2$ norm of the gradient is also observed. The highest point indicates a large difference, approximately 35 times greater than that before catastrophic overfitting.

These two observations, a low magnitude of perturbations and a high gradient norm, make us wonder what would the loss surface looks like. Figure \ref{fig:fast_before_co} illustrates the progress of adversarial training in which catastrophic overfitting occurs. The loss surface of the perturbed example is shown, where the green spot denotes the original images and the red spot denotes the adversarial example used for adversarial training in the batch. The $v_1$ axis indicates the direction of FGSM, whereas the $v_2$ axis is a random direction. The true label of the original sample is ``dog.'' Hence, the purple area indicates where the perturbed sample is correctly classified, whereas the blue area indicates a misclassified area.

On the left side of Figure \ref{fig:fast_before_co}, we can easily observe that the model is robust against FGSM. However, after training the batch, an interval vulnerable to a smaller perturbation than the maximum perturbation $\epsilon$ appears, whereas the model is still robust against FGSM. This distorted interval implies that the adversarial example with a larger perturbation is weaker than that with a smaller perturbation, which is contrary to the conventional belief that a larger magnitude of perturbation induces a stronger attack. As a result, the model with distorted interval is vulnerable to multi-step adversarial attacks that can search the vulnerable region further inside $\mathcal{B}(x, \epsilon)$. As the training continues, the area of distorted interval increases as shown in the figure on the right. It is now easier to see that the model is now perfectly overfitted for FGSM, yet loses its robustness to the smaller perturbations. We call this phenomenon ``decision boundary distortion.'' 

The evidence of decision boundary distortion is also shown in Figure \ref{fig:batch_trace} (b). When robustness against PGD7 sharply decreases to 0\%, the mean of the absolute value of PGD7 perturbation $\mathbb{E}[\vert\vert \delta_{PGD7} \vert\vert_1]$ decreases. It indicates that, when catastrophic overfitting arises, a smaller perturbation is enough to fool the model than the maximum perturbation $\epsilon$, which implies that distorted interval exists. In addition, during the process of having distorted decision boundary, as shown in the figure on the right, the loss surface inevitably becomes highly curved, which matches the observation of increasing the $L_2$ norm of the gradients of the images $\mathbb{E}[\vert\vert \nabla_x \vert\vert_2]$. This is also consistent with previous research \cite{andriushchenko2020understanding}. \citet{andriushchenko2020understanding} argued that $\nabla_x \ell(x,y;\theta)$ and $\nabla_{x} \ell(x+\eta,y;\theta)$ tend to be perpendicular in catastrophic overfitted models where $\eta$ is drawn from a uniform distribution $U(-\epsilon, \epsilon)$. Considering that a highly curved loss surface implies $(\nabla_x \ell(x,y;\theta))^T(\nabla_{x} \ell(x+\eta,y;\theta))\approx 0$ in high dimensions, the reason why GradAlign \cite{andriushchenko2020understanding} can avoid catastrophic overfitting might be because the gradient alignment leads the model to learn a linear loss surface which reduces the chance of having distorted decision boundary. 

\begin{figure*}[ht]
\centering
\includegraphics[width=0.9\textwidth]{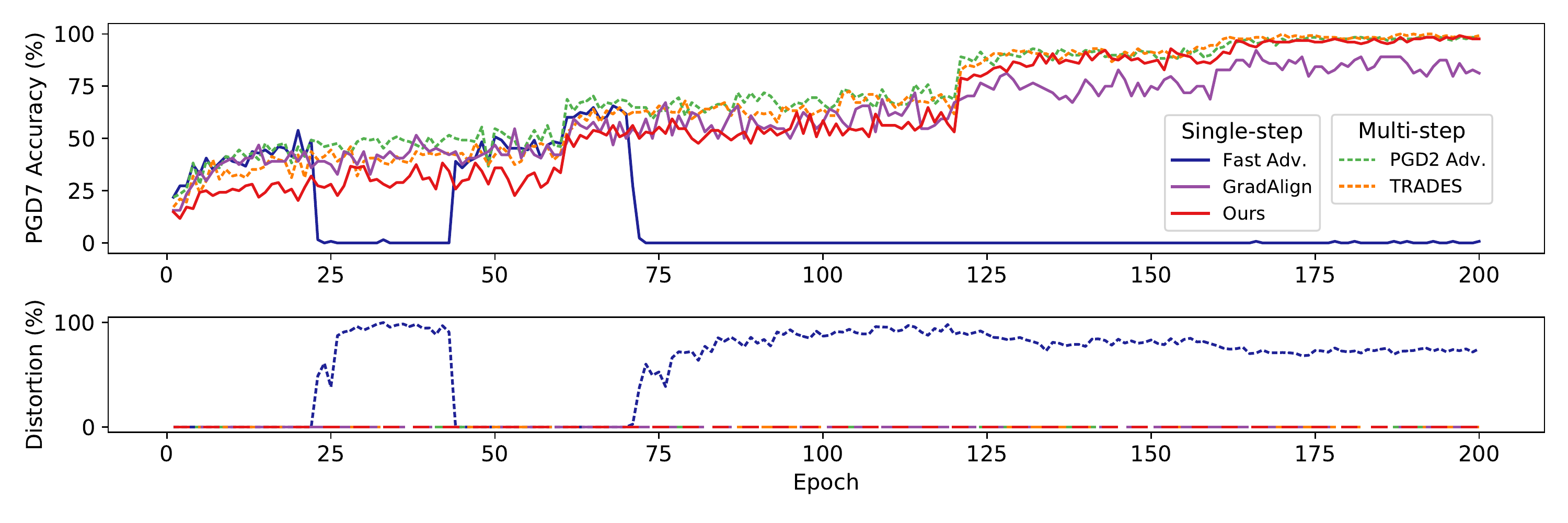} 
\caption{(CIFAR10) Robust accuracy and distortion on the training batch for each epoch. Two multi-step adversarial attacks show zero distortion during the entire training time and reach nearly 100\% PGD7 accuracy. By contrast, fast adversarial training shows high distortion and eventually collapses after the 71st epoch. The proposed method successfully avoids such problems and achieves a high PGD7 accuracy similiar to multi-step adversarial training (Best viewed in color).}
\label{fig:distortion}
\end{figure*}

We next numerically measured the degree of decision boundary distortion. To do so, we first define a new measure distortion $d$. Given a deep learning model $f$ and a loss function $\ell$, distortion $d$ can be formalized as follows:
\begin{equation} \label{eq:distortion}
\begin{split}
    \mathbf{S}_D &= \{x \vert \exists k \in (0, 1) \text{ s.t. } f(x+k\cdot \epsilon \cdot \text{sgn}(\nabla_x \ell))\neq y\} \\
    \mathbf{S}_N &= \{x \vert f(x)=y, f(x+\epsilon\cdot \text{sgn}(\nabla_x \ell)) = y\} \\
     d &= \frac{\vert \mathbf{S}_D \cap \mathbf{S}_N \vert}{\vert \mathbf{S}_N \vert}
\end{split}
\end{equation}
where $(x, y)$ is an example drawn from dataset $\mathcal{D}$. However, because the loss function of the model is not known explicitly, we use a number of samples to estimate distortion $d$. In all experiments, we tested 100 samples in the adversarial direction $\delta=\epsilon \cdot \text{sgn}(\nabla_x \ell)$ for each example. Indeed, we can see that distortion increases in Figure \ref{fig:batch_trace} (a) when catastrophic overfitting arises.

To verify that decision boundary distortion is generally related to catastrophic overfitting, we demonstrate how distortion and robustness against PGD7 change during training. We conducted an experiment on five different models: fast adversarial training (Fast Adv.) \cite{wong2020fast}, PGD2 (PGD2 Adv.)  \cite{madry2017towards}, TRADES \cite{zhang2019theoretically}, GradAlign \cite{andriushchenko2020understanding}, and the proposed method (Ours). All models were tested on $\epsilon = 8/255$. The step size $\alpha$ is set to $\alpha=1.25\epsilon$, $\alpha=1/2\epsilon$, and  $\alpha=1/4\epsilon$ for fast adversarial training, PGD2 adversarial training, and TRADES, respectively. We also conducted same experiment on PGD adversarial training with different number of steps; however, because these show similar results to PGD2 adversarial training, we only included PGD2. TRADES is trained with seven steps.

As the key observation in Figure \ref{fig:distortion}, the point where decision boundary distortion begins in fast adversarial training (22nd epoch) is identical to the point where robustness against PGD7 sharply decreases; that is, catastrophic overfitting occurs. Then, when decision boundary distortion disappears (45th to 72nd epoch), the model immediately recovers robust accuracy. After the 72nd epoch, the model once again suffers a catastrophic overfitting and never regains its robustness with high distortion. Hence, we conclude that there is a close connection between decision boundary distortion and the vulnerability of the model against multi-step adversarial attacks.

\section{Stable Single-Step Adversarial Training} \label{sec:stable}

Based on the results in Section \ref{sec:co}, we assume that distorted decision boundary might be the reason for catastrophic overfitting. Here, we stress that the major cause of distorted decision boundary is that single-step adversarial training uses a point with a fixed distance $\epsilon$ from the original image $x$ as an adversarial image $x'$ instead of an optimal solution of the inner maximum in Equation (\ref{eq:minmax}). Under this linearity assumption, the most powerful adversarial perturbation $\delta$ would be the same as $\epsilon \cdot \text{sgn}(\nabla_x\ell)$ where $\epsilon$ is the maximum perturbation, and the following formula should be satisfied.
\begin{equation}
\begin{split}
    \ell(x+\delta) - \ell(x) & = (\nabla_x \ell)^T \delta
    \\ & = (\nabla_x \ell)^T\epsilon \cdot \text{sgn}(\nabla_x \ell)
    \\ & = \epsilon \vert \vert \nabla_x \ell \vert \vert_1
\end{split}
\end{equation}
However, as confirmed in the previous scetion, decision boundary distortion with a highly curved loss surface has been observed during the training phase, which indicates that $\epsilon$ is no longer the strongest adversarial step size in the direction of $\delta$. Thus, the linear approximation of the inner maximization is not satisfied when distorted decision boundary arises.

To resolve this issue, we suggest a simple fix to prevent catastrophic overfitting by forcing the model to verify the inner interval of the adversarial direction. In this case, the appropriate magnitude of the perturbation should be taken into consideration instead of using $\epsilon$:
\begin{equation}
\begin{split}
& \delta = \epsilon \cdot \text{sgn}(\nabla_x\ell) \\
& \arg\max_{k \in [0,1]}\ell(x+k\cdot  \delta, y;\theta)
\end{split}
\end{equation}
Here, we introduce $k$, which denotes the scaling parameter for the original adversarial direction $\text{sgn}(\nabla_x\ell)$. In contrast to previous single-step adversarial training which uses a fixed size of $k=1$, an appropriate scaling parameter $k^*$ helps the model to train stronger adversarial examples as follows:
\begin{equation}
\begin{split}
    & \delta = \epsilon \cdot  \text{sgn}(\nabla_x\ell)\\
    & k^* = {\min}_{k \in [0, 1]} \{k \vert y \neq f(x+k\cdot\delta;\theta)\}
    \\ &\min_{\theta}\mathbb{E}_{(x,y)\sim \mathcal{D}} [\ell(x+k^* \cdot \delta, y; \theta) ]
\end{split}
\end{equation}
In this way, regardless of the linearity assumption, we can train the model with stronger adversarial examples that induce an incorrect classification in the adversarial direction. Simultaneously, we can also detect distorted decision boundary by inspecting the inside of distorted interval, as shown in Figure \ref{fig:fast_before_co}.

\begin{algorithm}[t]
\SetAlgoLined
\SetKwInOut{Parameter}{Parameter}
\Parameter{$B$ mini-batches, a perturbation size $\epsilon$, a step size $\alpha$, and $c$ check points for a network $f_\theta$}
\For{$i=1, ..., B$}{
    $\eta=\text{Uniform}(-\epsilon, \epsilon)$\\
    $\hat{y}_{i, 0}=f_\theta (x_i+\eta)$\\
    $\delta = \eta + \alpha \cdot \nabla_\eta \ell(\hat{y}_{i, 0}, y_i)$\\
    \For{$j=1, ..., c$}{
       $\hat{y}_{i, j}=f_\theta (x_i+j\cdot\delta/c)$)\\
    }
    $x'_i=x_i + \min(\{k|\hat{y}_{i, k} \neq y_i\} \cup \{1\})\cdot\delta/c$\\
    $\theta=\theta-\nabla_\theta \ell(f_\theta(x'_i), y_i)$
}
 \caption{Stable single-step adversarial training}
 \label{alg:1}
\end{algorithm}

However, because we do not know the explicit loss function of the model, forward propagation is the only approach for checking the adversarial images in the single-step adversarial attack direction. Hence, we propose the following simple method. First, we calculate the single-step adversarial direction $\delta$. Next, we choose multiple checkpoints ($x+\frac{1}{c}\delta , ..., x+\frac{c-1}{c}\delta , x+\delta$). Here, $c$ denotes the number of checkpoints except for the clean image $x$, which is tested in advance during the single-step adversarial attack process. We then feed all checkpoints to the model and verify that the predicted label $\hat{y}_{j}$ matches the correct label $y$ for all checkpoints $x+\frac{j}{c}\delta$ where $j \in \{1, ..., c\}$. Among the incorrect images and the clean image $x$, the smallest $j$ is selected; if all checkpoint are correctly classified, the adversarial image $x'=x+\delta$ is used. Algorithm \ref{alg:1} shows a summary of the proposed method.


Suppose the model has $L$ layers with $n$ neurons. Then, the time complexity of forward propagation is $O(Ln^2)$. Considering that backward propagation has the same time complexity, the generation of one adversarial example requires $O(2Ln^2)$ in total. Thus, with $c$ checkpoints, the proposed method consumes $O((c+4)Ln^2)$ because it requires one adversarial direction $O(2Ln^2)$, forward propagation for $c$ checkpoints $O(cLn^2)$, and one optimization step $O(2Ln^2)$. Compared to PGD2 adversarial training, which demands $O(6Ln^2)$, the proposed method requires more time when $c> 2$. However, the proposed method does not require additional memory for computing the gradients of the checkpoints because we do not need to track a history of variables for backward propagation; hence, larger validation batch sizes can be considered. Indeed, the empirical results describe in Section \ref{section:adv} indicate that the proposed method consumes less time than PGD2 adversarial training under $c \leq 4$.

Figure \ref{fig:distortion} shows that the proposed method successfully avoids catastrophic overfitting despite using a single-step adversarial attack. Furthermore, the proposed model not only achieves nearly 100\% robustness against PGD7, which fast adversarial training cannot accomplish, but also possesses zero distortion until the end of the training. This is the opposite of the common understanding that single-step adversarial training methods cannot perfectly defend the model against multi-step adversarial attacks.

\begin{figure}[t]
\centering
\includegraphics[width=0.9\columnwidth]{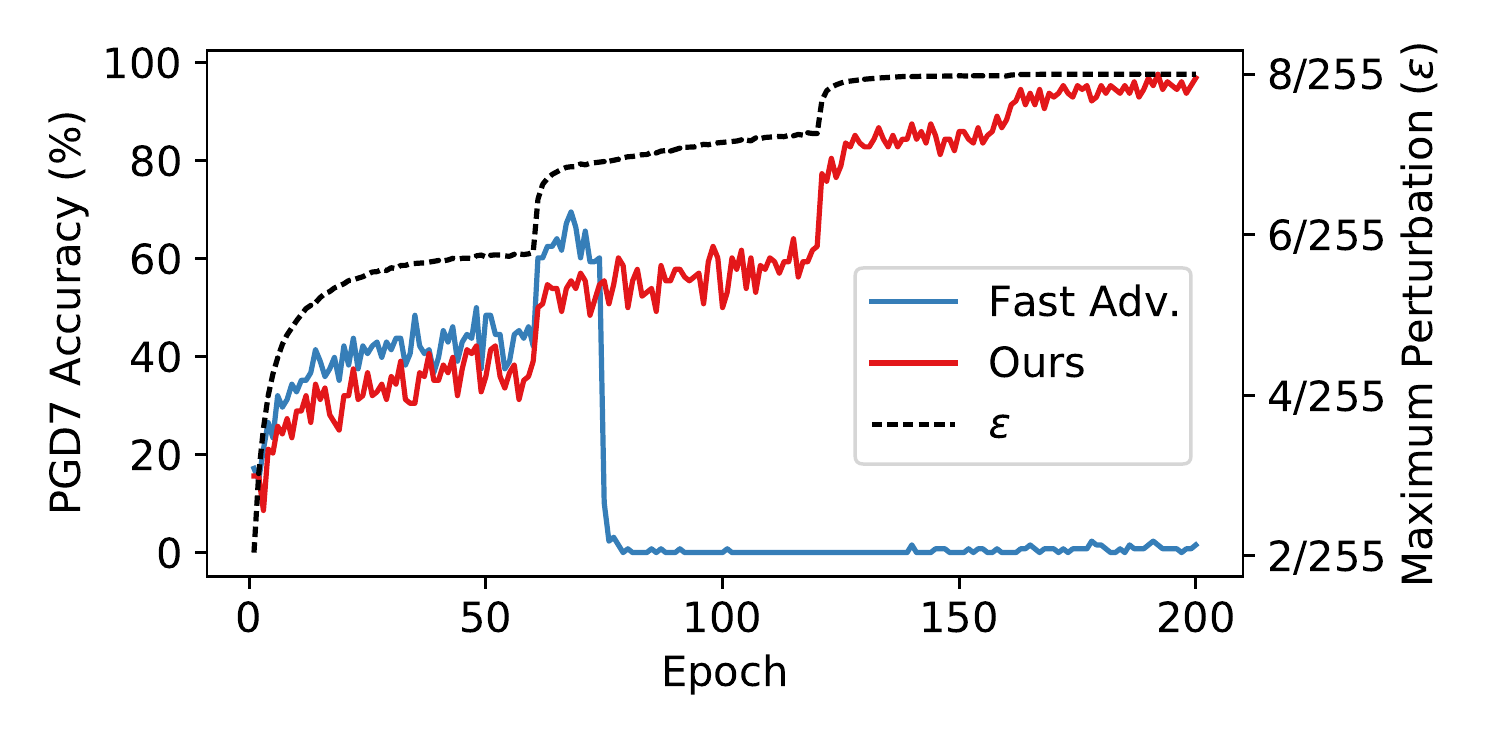}
\caption{(CIFAR10) Comparison of PGD7 accuracy on the training batch between fast adversarial training with $\epsilon$-scheduling and the proposed method. The dashed line indicates the average maximum perturbation $\mathbb{E}\left[\vert\vert\delta\vert\vert_{\infty}\right]$ calculated from the proposed method for each epoch and is used as the maximum perturbation of fast adversarial training.}
\label{fig:eps_schedule}
\end{figure}

The proposed model learns the image with the smallest perturbation among the incorrect adversarial images. In other words, during the initial states, the model outputs incorrect predictions for almost every image such that $\min(\{k|\hat{y}_{i,k} \neq y_i\}\cup\{1\})=0$ in Algorithm \ref{alg:1}. As additional batches are trained, the average maximum perturbation $\mathbb{E}[\vert\vert\delta\vert\vert_{\infty}]$ increases, as in Figure \ref{fig:eps_schedule}, where $\delta=x'-x$ and $x'$ is selected by the proposed method. Thus, the proposed method may appear to simply be a variation of $\epsilon$-scheduling. In order to point out the difference, fast adversarial training with $\epsilon$-scheduling is also considered. For each epoch, we use the average maximum perturbation $\mathbb{E}[\vert\vert\delta\vert\vert_{\infty}]$ calculated from the proposed method as the maximum perturbation $\epsilon$. The result is summarized in Figure \ref{fig:eps_schedule}. 

Notably, $\epsilon$-scheduling cannot help fast adversarial training avoid catastrophic overfitting. The main difference between $\epsilon$-scheduling and the proposed method is that, whereas $\epsilon$-scheduling uniformly applies the same magnitude of the perturbation for every image, the proposed method gradually increases the magnitude of the perturbation appropriately by considering the loss surface of each image. Therefore, in contrast to $\epsilon$-scheduling, the proposed method successfully prevents catastrophic overfitting, despite the same size of the average perturbation used during the training process.




\begin{table*}[ht]
\centering
\caption{Standard and robust accuracy (\%) and training time (hour) on CIFAR10.}
\resizebox{0.80\textwidth}{!}{%
\begin{tabular}{lccccccr}
\hhline{========}
&\textbf{Method} & \textbf{Standard}     & \textbf{FGSM}         & \textbf{PGD50}        & \textbf{Black-box}            & \textbf{AA}           & \textbf{Time (h)} \\ \hhline{========}
\textbf{Multi-step} & PGD2 Adv.            & \textbf{86.6$\pm$0.8} & 49.7$\pm$2.6          & 36.0$\pm$2.3          & \textbf{85.6$\pm$0.8}          & 34.8$\pm$2.1          & 4.5     \\
&PGD4 Adv.            & 86.0$\pm$0.8          & 49.6$\pm$3.0          & 36.7$\pm$2.9          & 85.3$\pm$0.8          & 35.4$\pm$2.6          & 6.7      \\
&PGD7 Adv.            & 84.4$\pm$0.2          & \textbf{51.5$\pm$0.1} & \textbf{40.5$\pm$0.1} & 83.8$\pm$0.2          & \textbf{39.4$\pm$0.2} & 11.1     \\
&TRADES          & 85.3$\pm$0.4          & 50.7$\pm$1.6          & 39.3$\pm$1.9          & 84.4$\pm$0.4 & 38.6$\pm$2.0          & 15.1      \\ \hline
\textbf{Single-step}&Fast Adv.           & 84.5$\pm$4.3          & \textbf{95.1$\pm$6.8} & 0.1$\pm$0.1           & 80.8$\pm$8.7          & 0.0$\pm$0.0           & 3.2      \\
&GradAlign       & 83.9$\pm$0.2          & 44.3$\pm$0.0          & 31.7$\pm$0.2          & 83.3$\pm$0.3          & 30.9$\pm$0.2          & 13.6     \\
&Ours ($c=2$)       & 86.8$\pm$0.3          & 48.3$\pm$0.5          & 32.5$\pm$0.2          & 85.9$\pm$0.1          & 30.9$\pm$0.2
              & 3.5      \\
&Ours ($c=3$)      & 87.7$\pm$0.8          & 50.5$\pm$2.4          & \textbf{33.9$\pm$2.3} & 86.7$\pm$0.9          & \textbf{32.3$\pm$2.2} & 3.9      \\
&Ours ($c=4$)       & \textbf{87.8$\pm$0.9}          & 50.5$\pm$2.3          & 33.7$\pm$2.4          & \textbf{87.0$\pm$0.8}          & 32.2$\pm$2.4          & 4.4      
     \\ \hhline{========}
\end{tabular}
}

\label{table:cifar10}
\end{table*}

\section{Adversarial Robustness}\label{section:adv}

In this section, we conduct a set of experiments on CIFAR10 and Tiny ImageNet
, using PreAct ResNet-18 \cite{he2016deep}. Input normalization and data augmentation including 4-pixel padding, random crop and horizontal flip are applied. We use SGD with a learning rate of $0.01$, momentum of $0.9$ and weight decay of 5e-4. To check whether catastrophic overfitting occurs, we set the total epoch to 200. The learning rate decays with a factor of 0.2 at 60, 120, and 160 epochs. All experiments were conducted on a single NVIDIA TITAN V over three different random seeds. Our implementation in PyTorch \cite{paszke2019pytorch} with Torchattacks \cite{kim2020torchattacks} is available at https://github.com/Harry24k/catastrophic-overfitting.


During the training session, the maximum perturbation $\epsilon$ was set to $8/255$. For PGD adversarial training, we use a step size of $\alpha=\max(2/255, \epsilon/n)$, where $n$ is the number of steps. TRADES uses $\alpha=2/255$ and seven steps for generating adversarial images. Following \citet{wong2020fast}, we use $\alpha=1.25\epsilon$ for fast adversarial training and the proposed method. The regularization parameter $\beta$ for the gradient alignment of GradAlign is set to 0.2 as suggested by \citet{andriushchenko2020understanding}.

\begin{figure}[t]
\centering
\includegraphics[width=0.9\columnwidth]{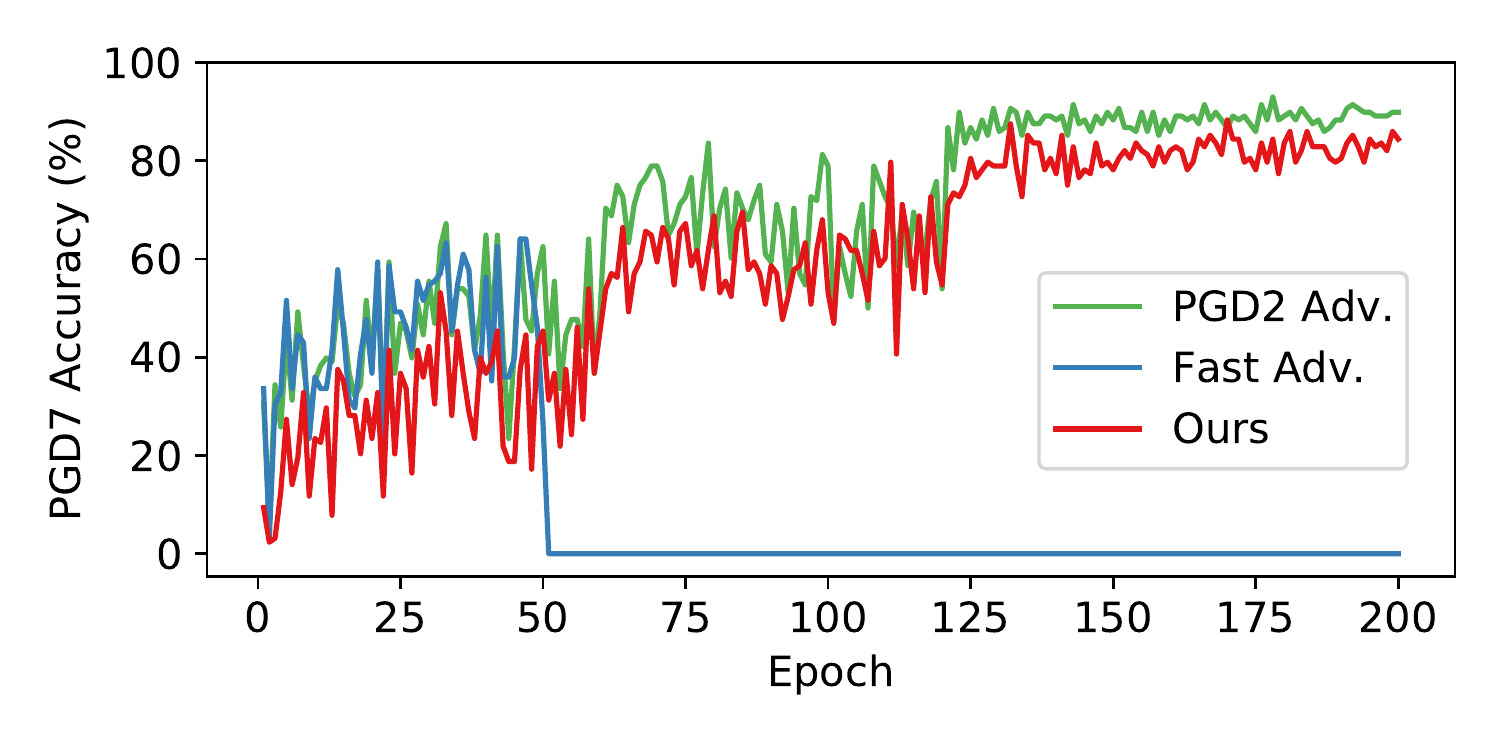}
\caption{(Tiny ImageNet) PGD7 accuracy on the training batch.}
\label{fig:tiny}
\end{figure}

First, we check whether our method shows the same results as those of Tiny ImageNet described in the previous section. Figure \ref{fig:tiny} shows that the proposed method also successfully prevents catastrophic overfitting in a large dataset. PGD7 accuracy decreases rapidly only for fast adversarial training after the 49th epoch, but not for others including the proposed method. The full results with the change in distortion are shown in Appendix \ref{appendix:b}.

\begin{table}[t]
\centering
\caption{Standard accuracy and robustness (\%) and training time (h) on Tiny ImageNet.}
\resizebox{0.45\textwidth}{!}{%
\begin{tabular}{ccccc}
\hline
\textbf{Method} & \textbf{Standard} & \textbf{FGSM} & \textbf{PGD50} & \textbf{Time (h)} \\ \hline
PGD2 Adv.           & 46.3$\pm$1.2              & 14.7$\pm$2.7          & \textbf{10.3$\pm$2.7}   & 27.7               \\
Fast Adv.             & 26.2$\pm$0.7              & \textbf{49.0$\pm$5.7} & 0.0$\pm$0.0            & 19.6               \\
Ours ($c=3$)     & \textbf{49.6$\pm$1.5}     & 12.5$\pm$0.1          & 7.8$\pm$0.1            & 25.7               \\ \hline
\end{tabular}%
}
\label{table:tiny}
\end{table}

We then evaluate robustness on the test set at the end of the training. FGSM and PGD50 with 10 random restarts are used for evaluating robustness of the models. Furthermore, to estimate accurate robustness and detect gradient obfuscation \cite{athalye2018obfuscated}, we also consider PGD50 adversarial images generated from Wide-ResNet 40-10 \cite{zagoruyko2016wide} trained on clean images (Black-box), and AutoAttack (AA) which is one of the latest strong adversarial attacks proposed by \citet{croce2020reliable}.

Tables \ref{table:cifar10} and \ref{table:tiny} summarize the results.
From Table \ref{table:cifar10}, we can see that multi-step adversarial training methods yield more robust models, but generally requires a longer computational time. In particular, TRADES requires over 15 hours, which is 5-times slower than the proposed method.
Among the single-step adversarial training methods, fast adversarial training is computationally efficient, however, because catastrophic overfitting has occurred, it shows 0\% accuracy against PGD50 and AA.

Interestingly, we observe that fast adversarial training achieves a higher accuracy for FGSM adversarial images than clean images in both datasets, which does not appear in other methods. 
The accuracy when applying FGSM on only correctly classified images is 84.4\% on CIFAR10, whereas all other numbers remain almost unchanged when we use attacks on correctly classified clean images. We note that this is another characteristic of the catastrophic overfitted model which we describe in more detail in Appendix \ref{appendix:b}.

The proposed method, by contrast, shows the best standard accuracy and robustness against PGD50, Black-box, and AA with a shorter time. GradAlign also provides sufficient robustness; however, it takes 3-times longer than the proposed method. As shown in Table \ref{table:tiny}, similar results are observed on Tiny ImageNet. We include the results of the main competitors, PGD2 adversarial training, fast adversarial training, and the proposed method with $c=3$ which shows the best performance on CIFAR10. Here again, the proposed method shows high standard accuracy and adversarial robustness close to that of PGD2 adversarial training.
We provide additional experiments with different settings, such as cyclic learning rate schedule in Appendix \ref{appendix:c}. 

\section{Conclusion}
In this study, we empirically showed that catastrophic overfitting is closely related to decision boundary distortion by analyzing their loss surface and robustness during training. Decision boundary distortion provides a reliable understanding of the phenomenon in which a catastrophic overfitted model becomes vulnerable to multi-step adversarial attacks, while achieving a high robustness on the single-step adversarial attacks. Based on these observations, we suggested a new simple method that determines the appropriate magnitude of the perturbation for each image. Further, we evaluated robustness of the proposed method against various adversarial attacks and showed sufficient robustness using single-step adversarial training without the occurrence of any catastrophic overfitting.

\section{Acknowledgments.}
This work was supported by the National Research Foundation of Korea (NRF) grant funded by the Korean government (MSIT) (NRF-2019R1A2C2002358).

\bibliography{ms}

\newpage
\onecolumn
\appendix
\counterwithin{figure}{section}
\counterwithin{table}{section}
\counterwithin{equation}{section}
\section{Distortion and nonlinearity of the loss function}
\label{appendix:a}

In Section \ref{sec:stable}, we mention that single-step adversarial training leads to a highly curved loss surface, and single-step adversarial attacks such as FGSM cannot generate strong adversarial examples because the following linearity assumption is not satisfied.
\begin{equation}
\begin{split}
    \ell(x+\delta) - \ell(x) &  = \epsilon \vert \vert \nabla_x \ell \vert \vert_1
\end{split}
\label{eq:linearity}
\end{equation}

Here, to detect decision boundary distortion from a different perspective of nonlinearity, we propose a new measure $\gamma$ for the nonlinearity of the loss function based on Equation (\ref{eq:linearity}):
\begin{equation}
\begin{split}
    \gamma  = \{\ell(x+\delta) - \ell(x)\} - \epsilon \vert \vert \nabla_x \ell \vert \vert_1
\end{split}
\label{eq:gamma}
\end{equation}

\begin{figure}[ht]
\centering
    \begin{minipage}{0.45\linewidth}
        \begin{subfigure}{\textwidth}
        \includegraphics[width=\linewidth]{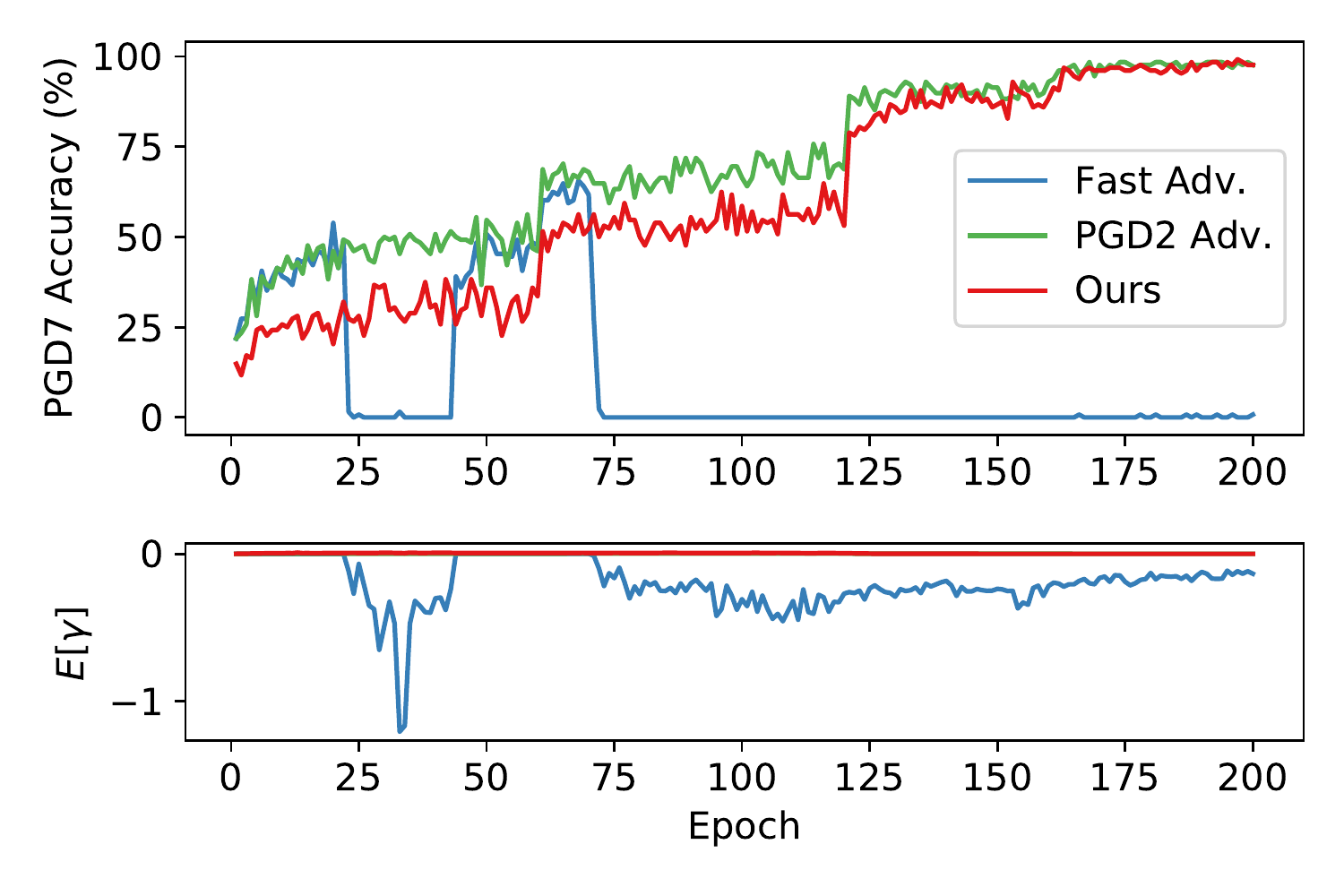}
        \caption{Change in $\mathbb{E}{[\gamma]}$}
        \end{subfigure}
    \end{minipage}
    \begin{minipage}{0.45\linewidth}
            \begin{subfigure}{\textwidth}
            \includegraphics[width=\linewidth]{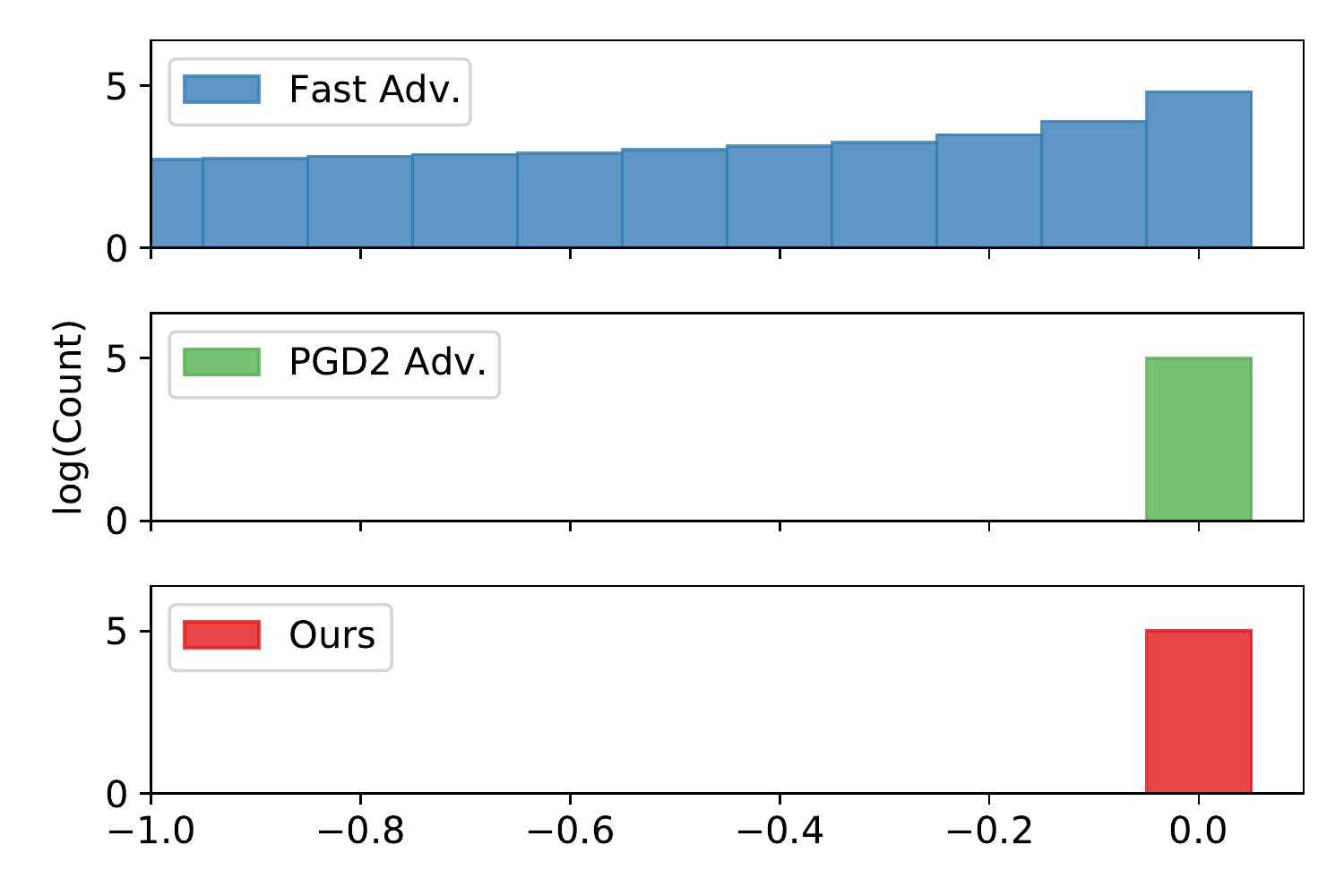}
        \caption{Distribution of $\gamma$}
        \end{subfigure}
    \end{minipage}
\caption{(CIFAR10) Nonlinearity of the loss function. In plot (a), the top figure shows robust accuracy against PGD7 on the entire training images. The bottom figure shows the average of the nonlinearity of the loss function over entire training images. Plot (b) shows the distribution of $\gamma$ for each model at the last epoch.}
\label{fig:gamma}
\end{figure}

In all appendixes, unless specified otherwise, the same settings in Section \ref{section:adv} are used. Figure \ref{fig:gamma} demonstrates the different characteristics of $\gamma$ for each method.

Plot (a) shows the mean value of $\gamma$ for all 50,000 CIFAR10 training images. When catastrophic overfitting occurs, $\mathbb{E}[\gamma]$ becomes negative. The negative $\mathbb{E}[\gamma]$ of fast adversarial training indicates that the loss function becomes nonlinear, which is one of the main characteristics of distorted decision boundary.

Similarly, in plot (b), the distribution of $\gamma$ of fast adversarial training shows a totally different behavior from that of PGD2 adversarial training and the proposed method. Compared to the proposed method and PGD2 adversarial training with few negative values, there are noticeably many negative values in fast adversarial training. These results indicate the loss function $\ell$ of catastrophic overfitted model follows $\ell(x+\delta) - \ell(x) \leq \epsilon \vert \vert \nabla_x \ell \vert \vert_1$, which implies highly nonlinear loss surface.

\section{Visualizing decision boundary distortion} \label{appendix:b}

\begin{figure*}[p]
    \centering
    \begin{minipage}{0.075\linewidth}
        \begin{subfigure}[t]{\textwidth}
        \includegraphics[width=\linewidth]{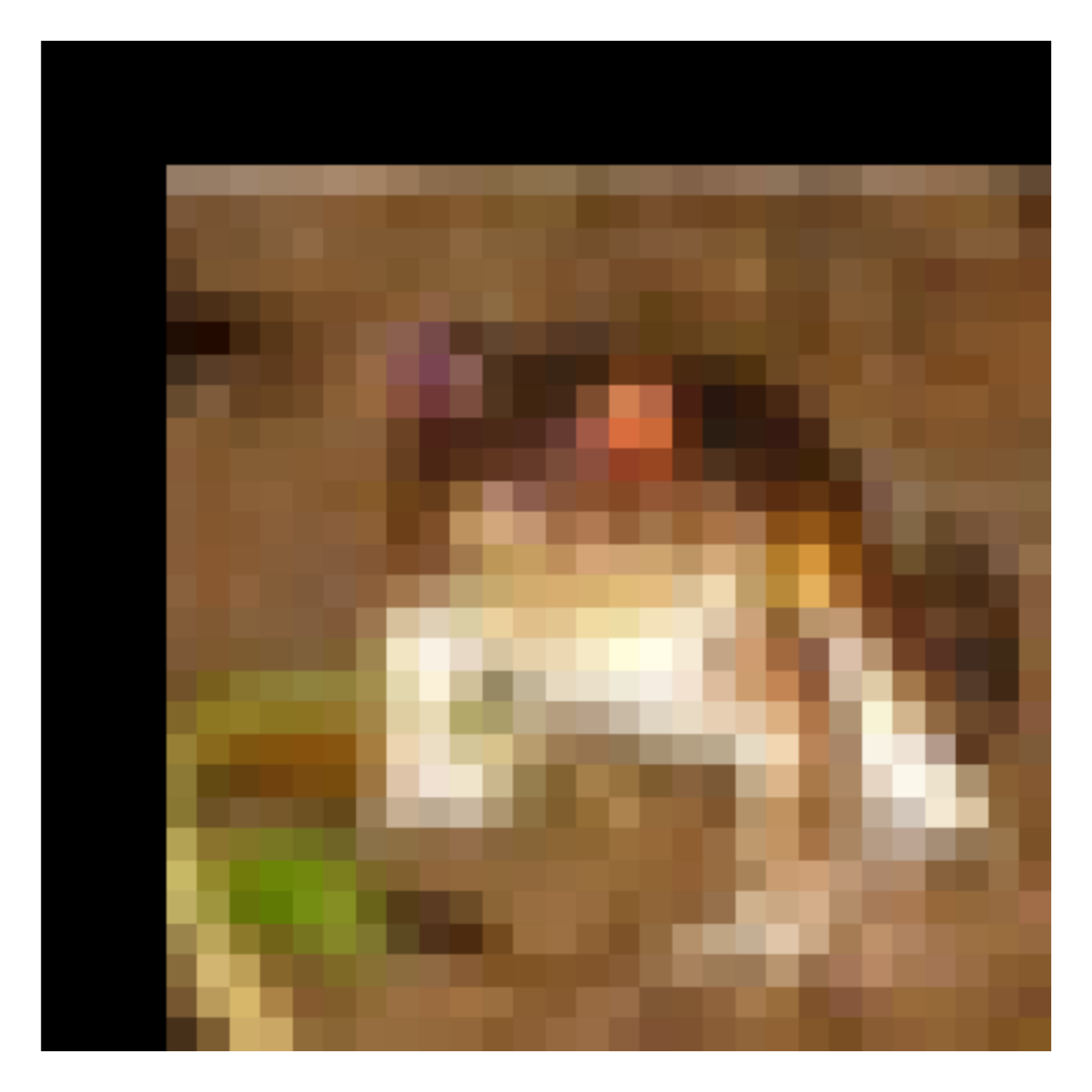}
        \caption{$x$}
        \end{subfigure} \\
        \begin{subfigure}[b]{\textwidth}
        \includegraphics[width=\linewidth]{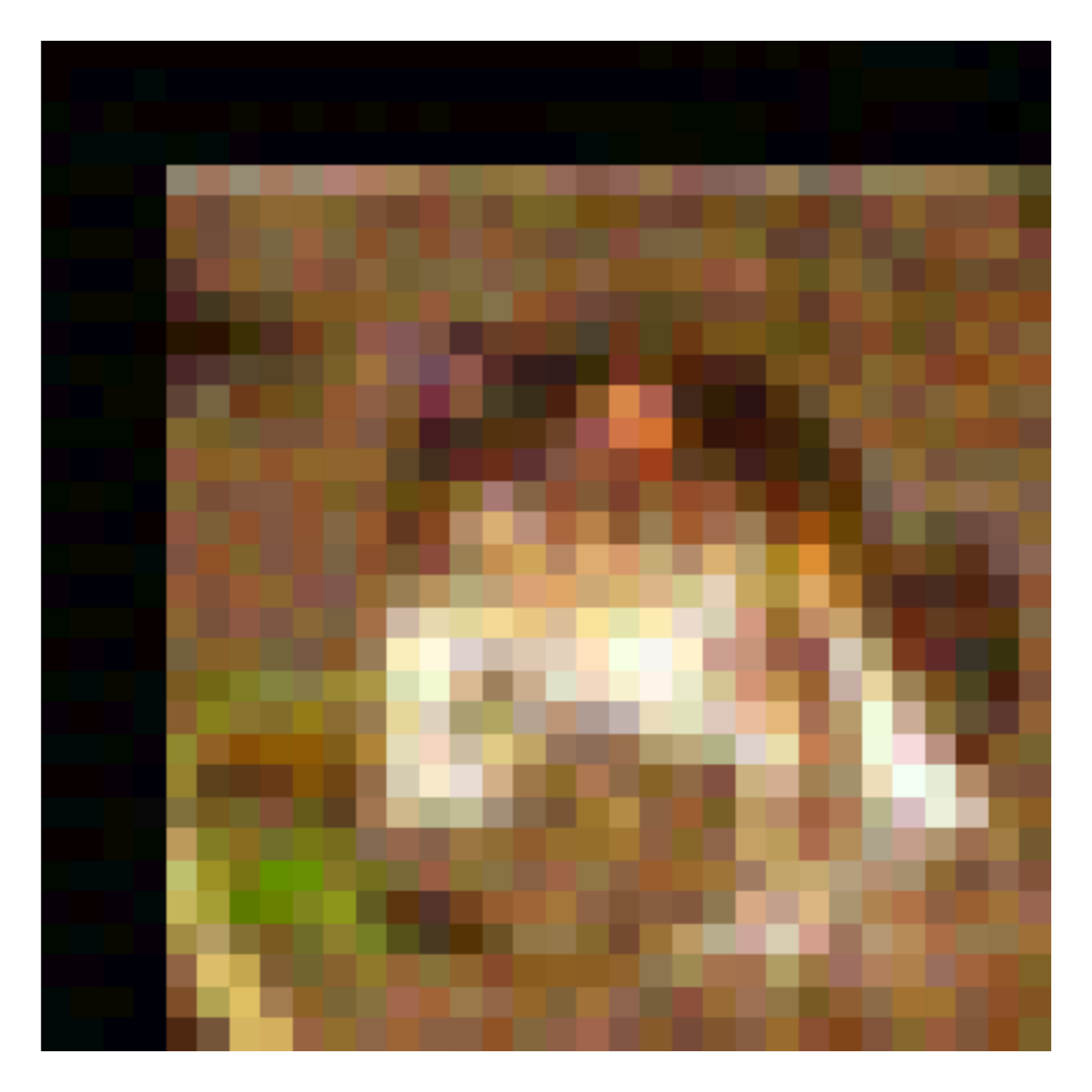}
        \caption{$x'$}
        \end{subfigure}
    \end{minipage}
    \begin{minipage}{0.3009\linewidth}
        \begin{subfigure}{\textwidth}
        \includegraphics[width=\linewidth]{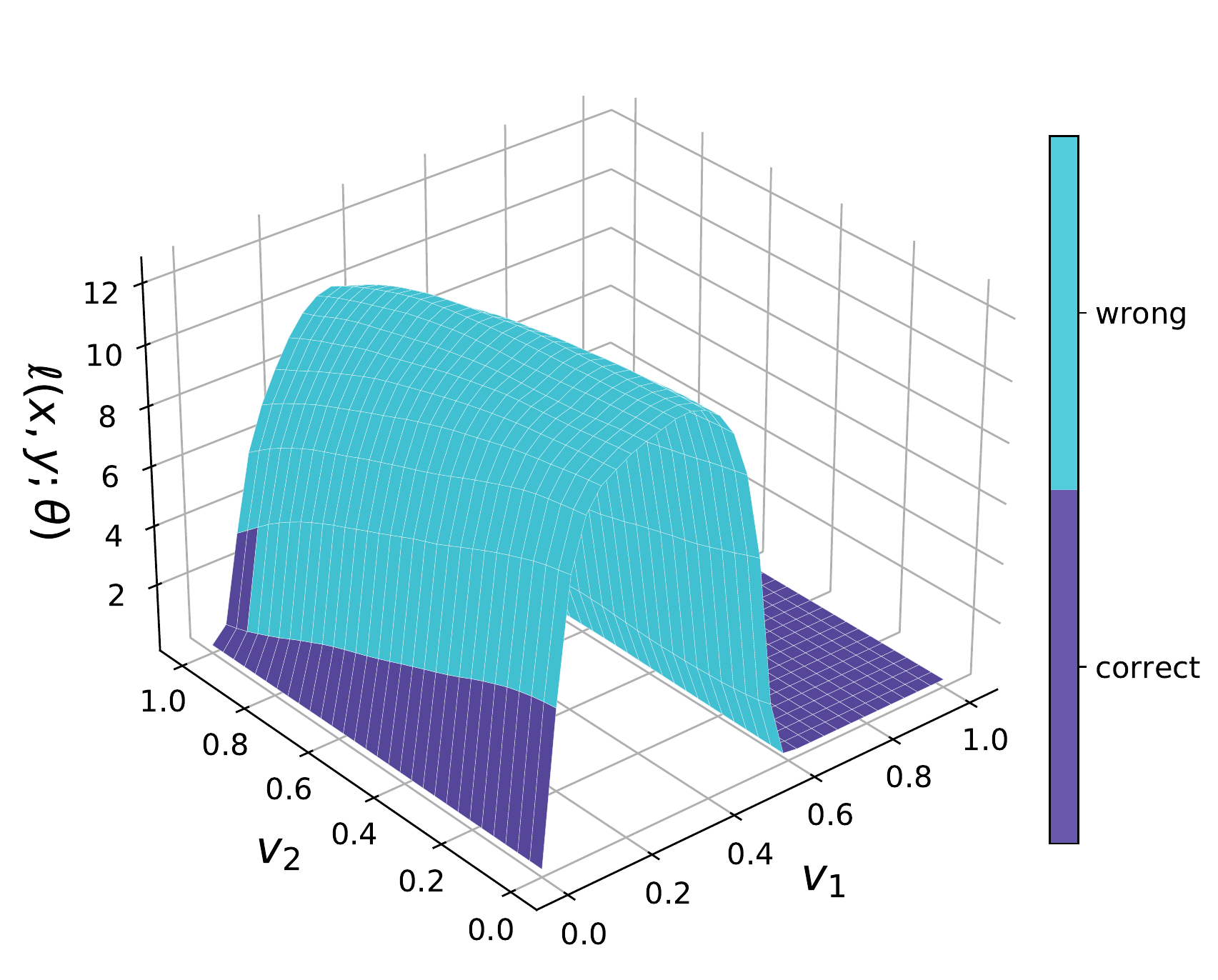}
        \caption{Loss surface}
        \end{subfigure}
    \end{minipage}
    \begin{minipage}{0.075\linewidth}
        \begin{subfigure}[t]{\textwidth}
        \includegraphics[width=\linewidth]{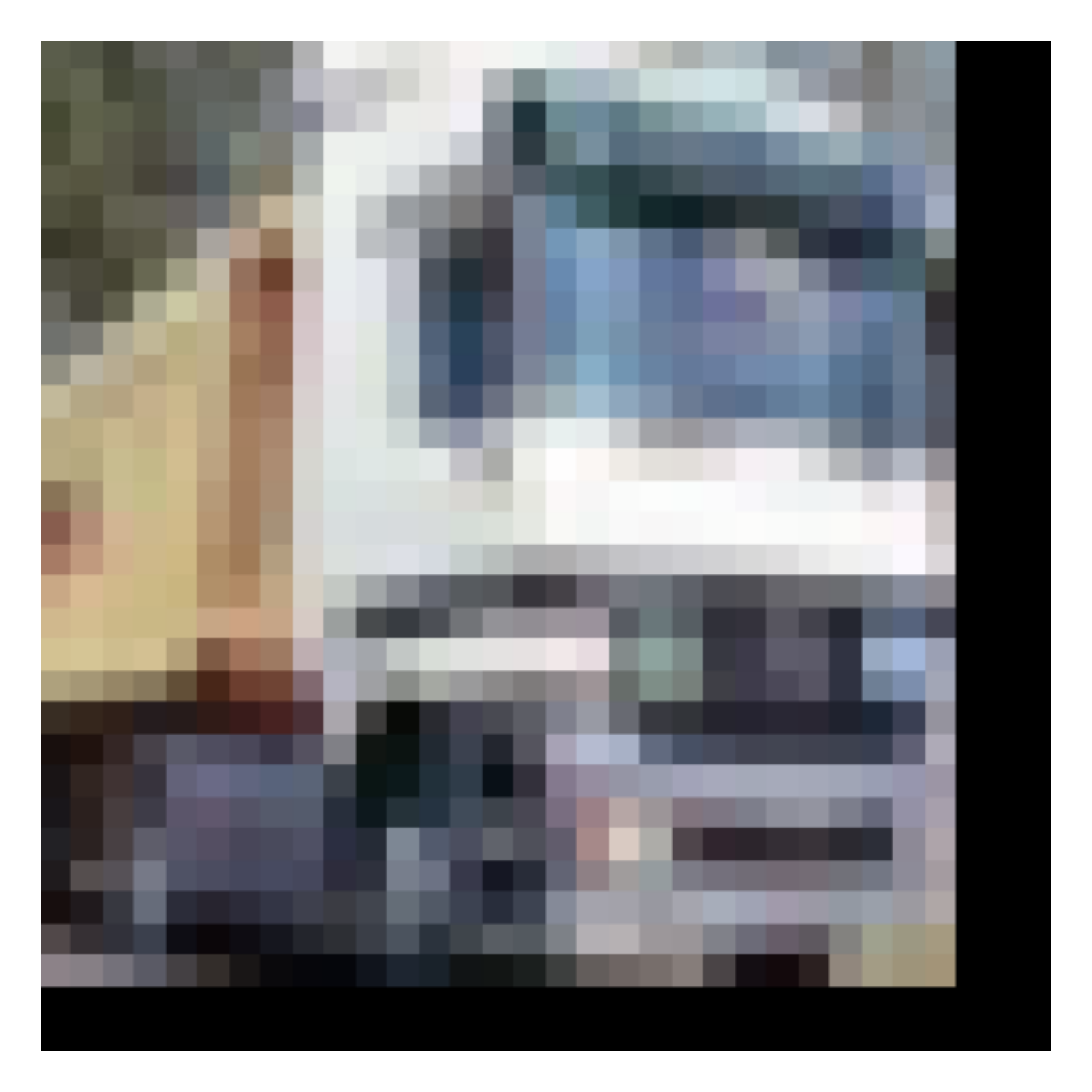}
        \caption{$x$}
        \end{subfigure} \\
        \begin{subfigure}[b]{\textwidth}
        \includegraphics[width=\linewidth]{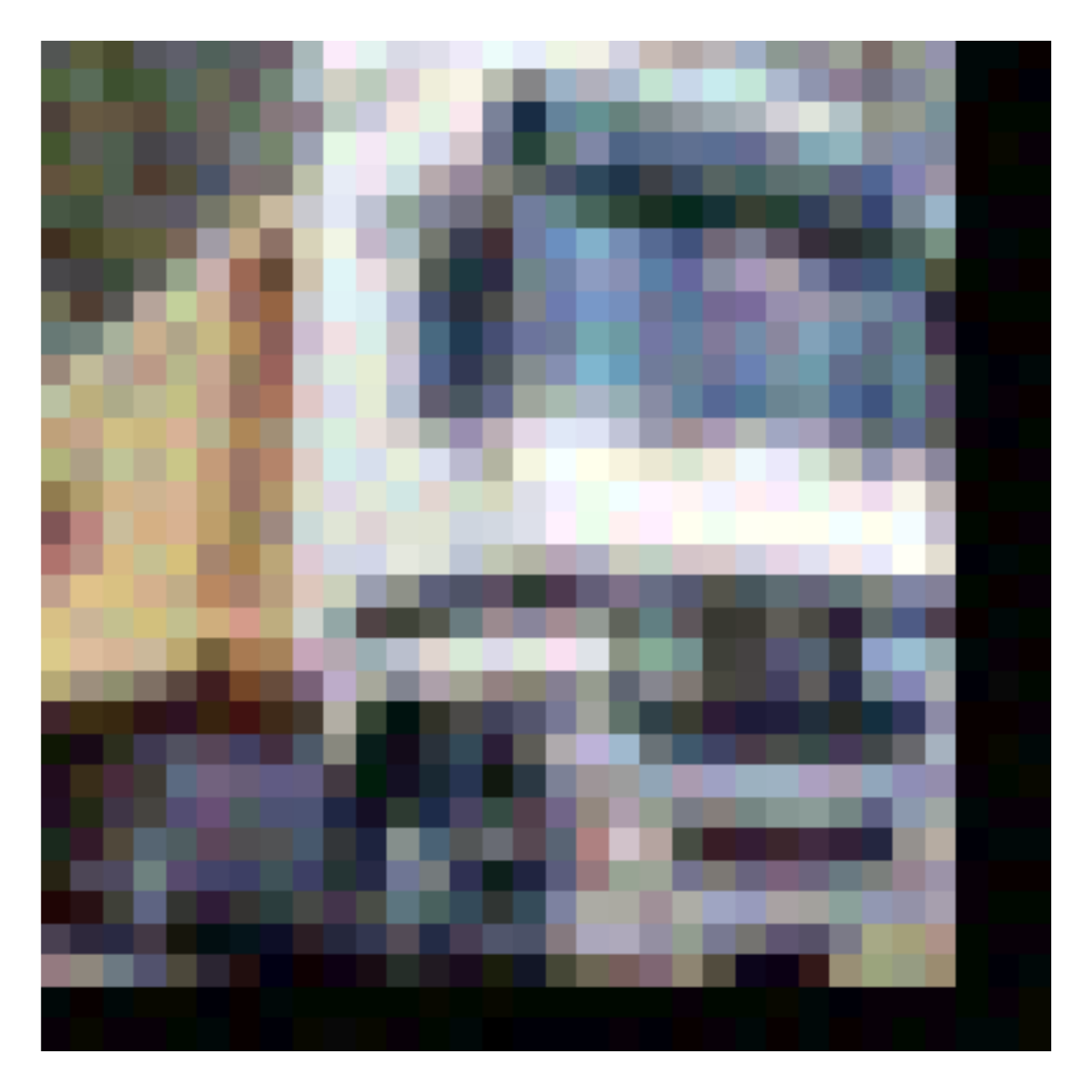}
        \caption{$x'$}
        \end{subfigure}
    \end{minipage}
    \begin{minipage}{0.3009\linewidth}
        \begin{subfigure}{\textwidth}
        \includegraphics[width=\linewidth]{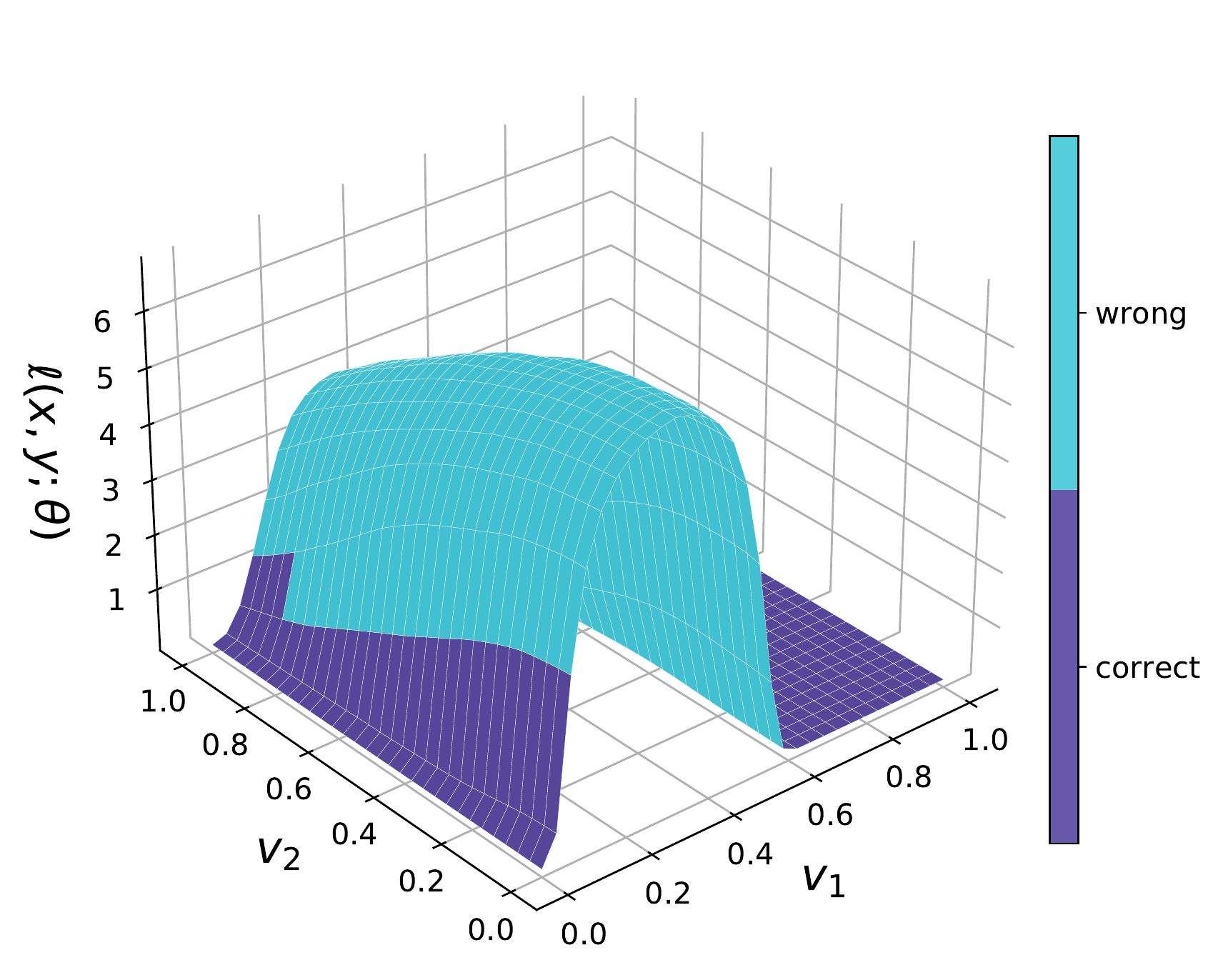}
        \caption{Loss surface}
        \end{subfigure}
    \end{minipage} \\

    \begin{minipage}{0.075\linewidth}
        \begin{subfigure}[t]{\textwidth}
        \includegraphics[width=\linewidth]{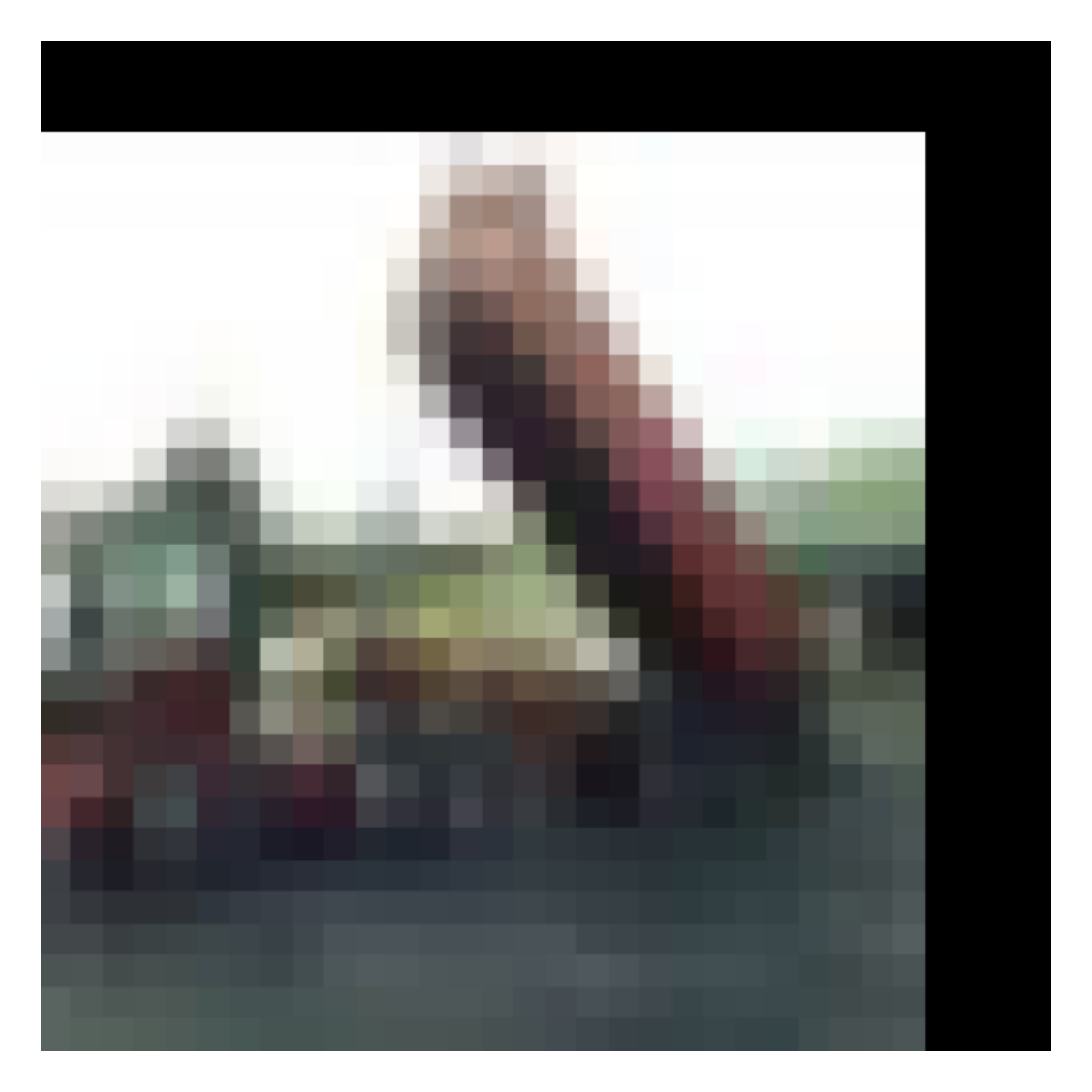}
        \caption{$x$}
        \end{subfigure} \\
        \begin{subfigure}[b]{\textwidth}
        \includegraphics[width=\linewidth]{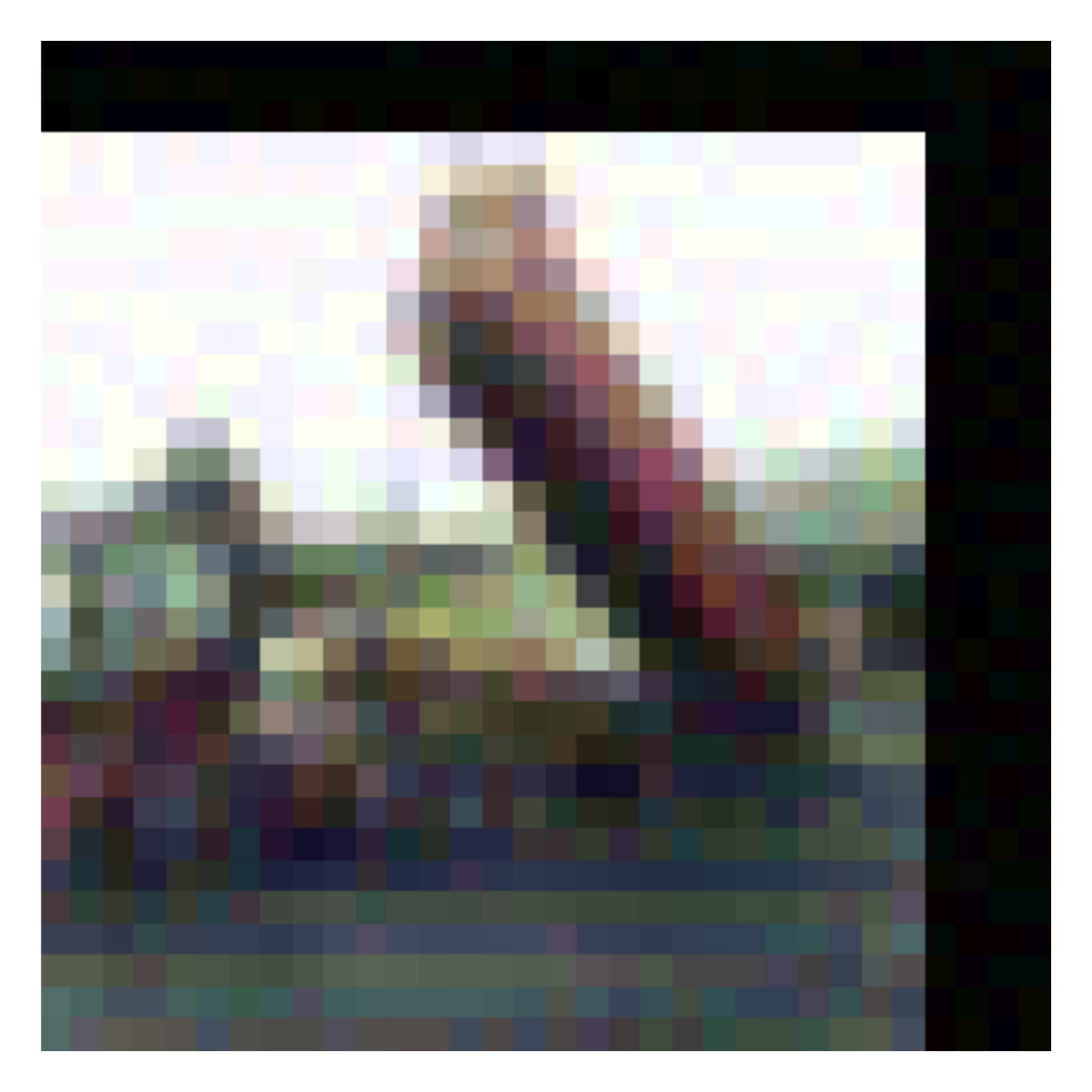}
        \caption{$x'$}
        \end{subfigure}
    \end{minipage}
    \begin{minipage}{0.3009\linewidth}
        \begin{subfigure}{\textwidth}
        \includegraphics[width=\linewidth]{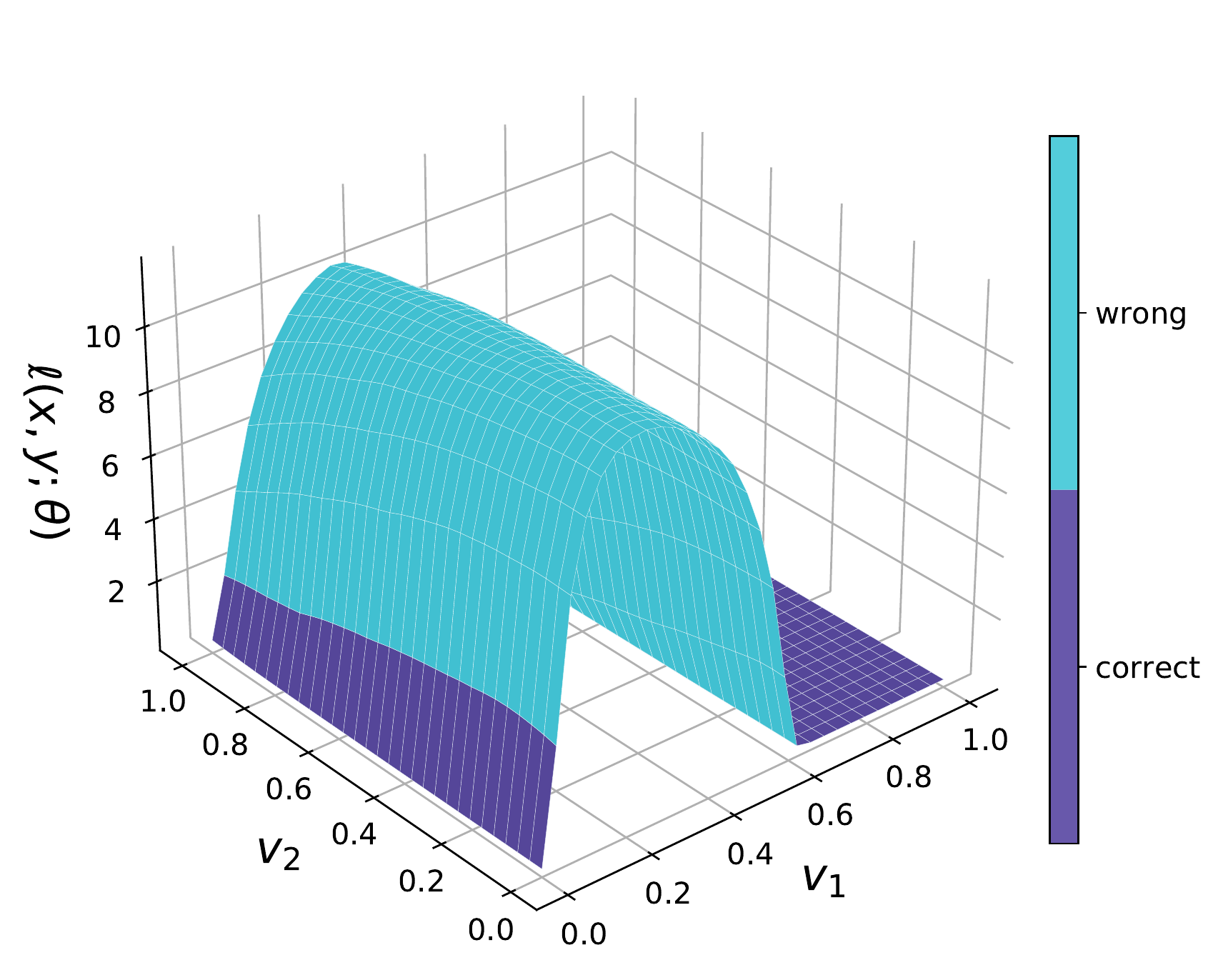}
        \caption{Loss surface}
        \end{subfigure}
    \end{minipage}
    \begin{minipage}{0.075\linewidth}
        \begin{subfigure}[t]{\textwidth}
        \includegraphics[width=\linewidth]{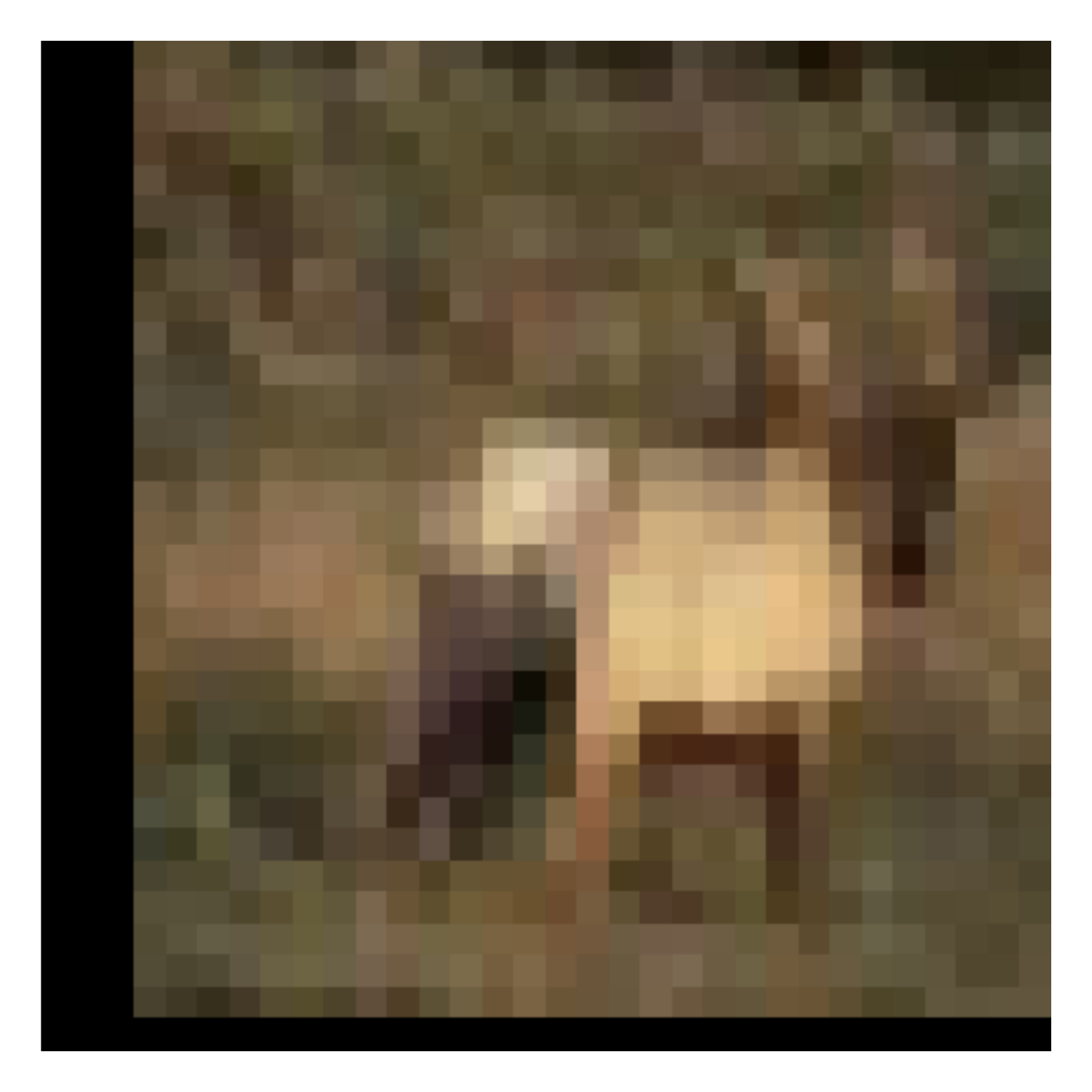}
        \caption{$x$}
        \end{subfigure} \\
        \begin{subfigure}[b]{\textwidth}
        \includegraphics[width=\linewidth]{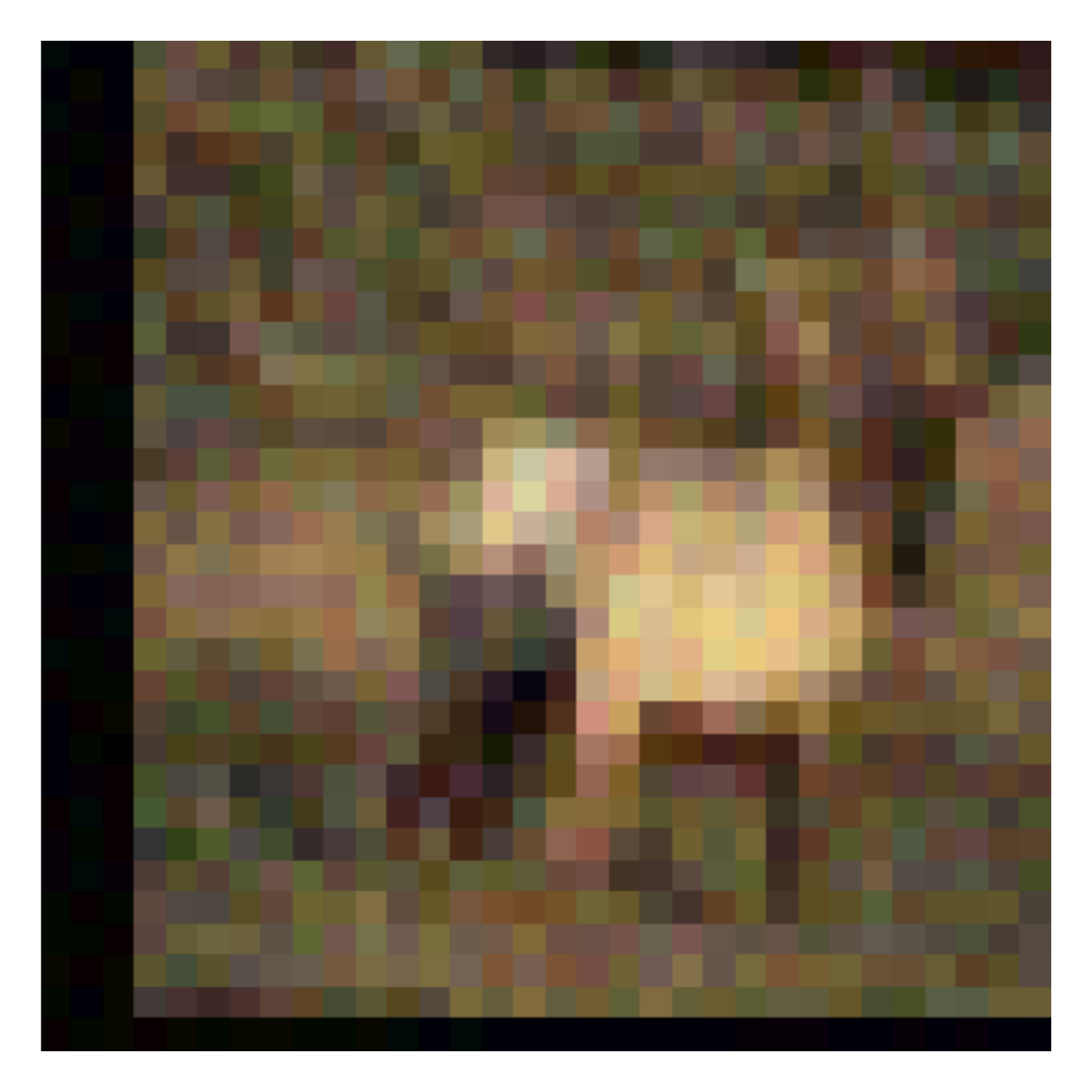}
        \caption{$x'$}
        \end{subfigure}
    \end{minipage}
    \begin{minipage}{0.3009\linewidth}
        \begin{subfigure}{\textwidth}
        \includegraphics[width=\linewidth]{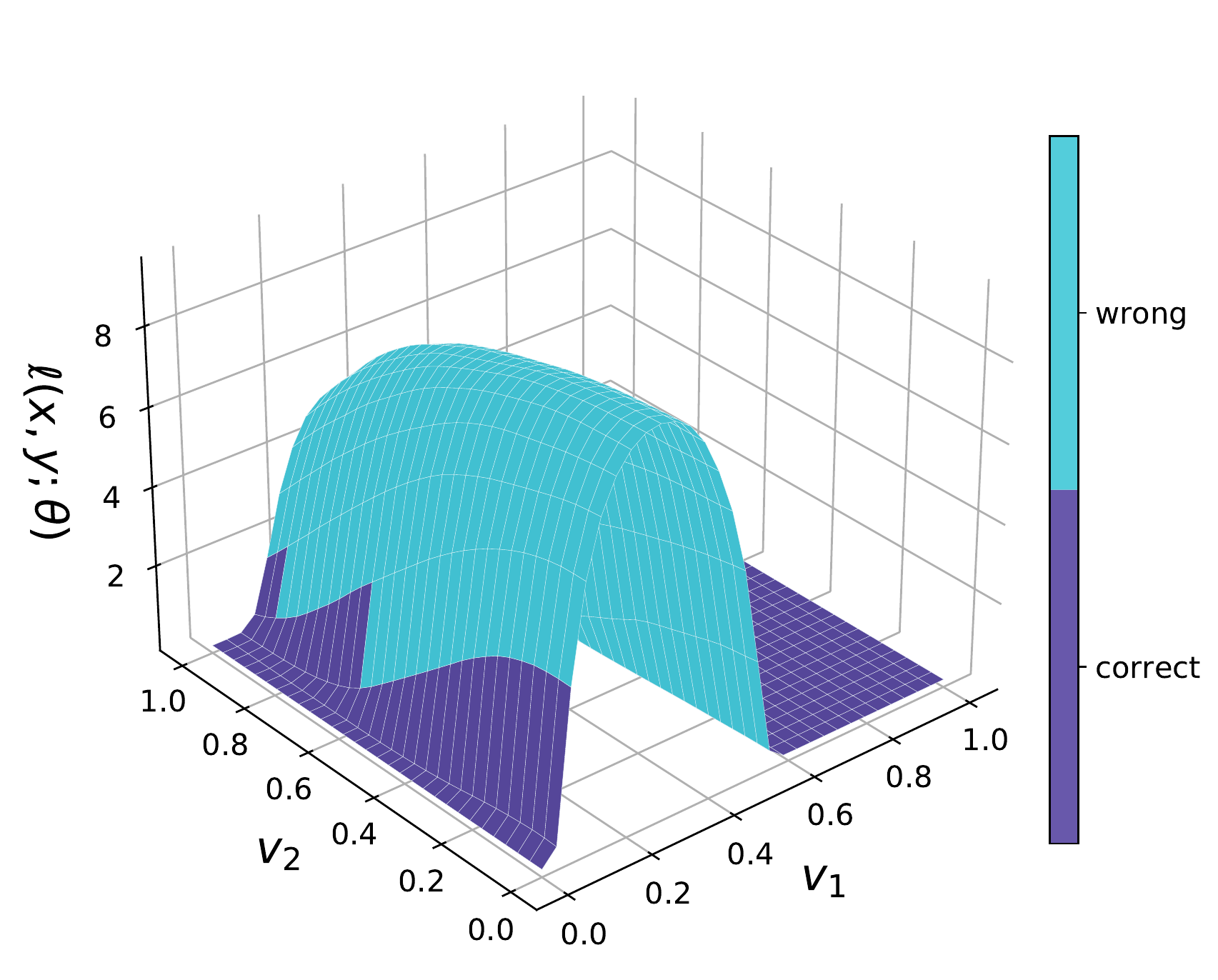}
        \caption{Loss surface}
        \end{subfigure}
    \end{minipage}
    \caption{(CIFAR10) Direction of FGSM adversarial perturbation $v_1$ and random direction $v_2$. Adversarial example $x'=x+v_1$ is generated from original example $x$.}
    \label{fig:fgsm_train}
\end{figure*}

\begin{figure*}[p]
    \centering
    \begin{minipage}{0.075\linewidth}
        \begin{subfigure}[t]{\textwidth}
        \includegraphics[width=\linewidth]{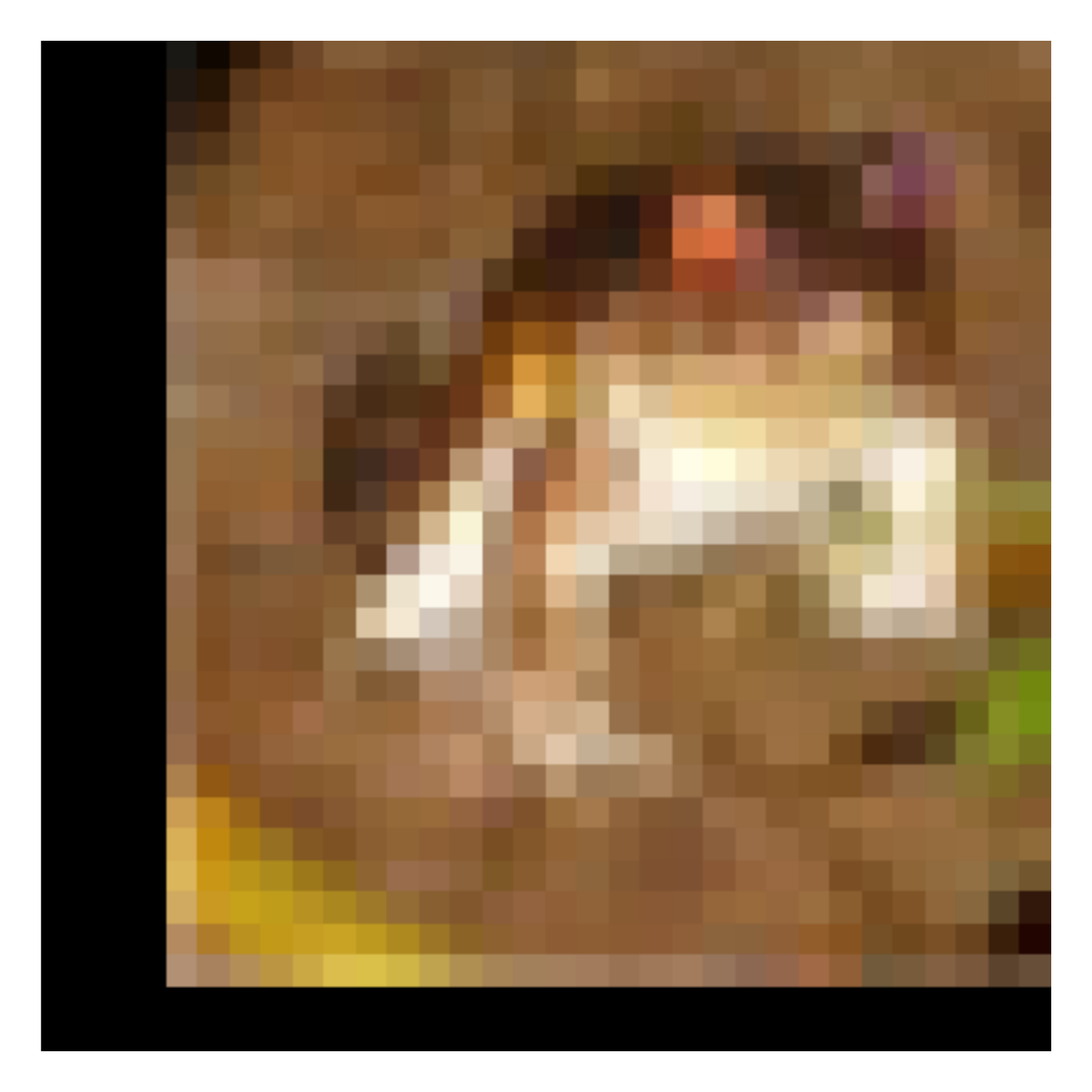}
        \caption{$x$}
        \end{subfigure} \\
        \begin{subfigure}[b]{\textwidth}
        \includegraphics[width=\linewidth]{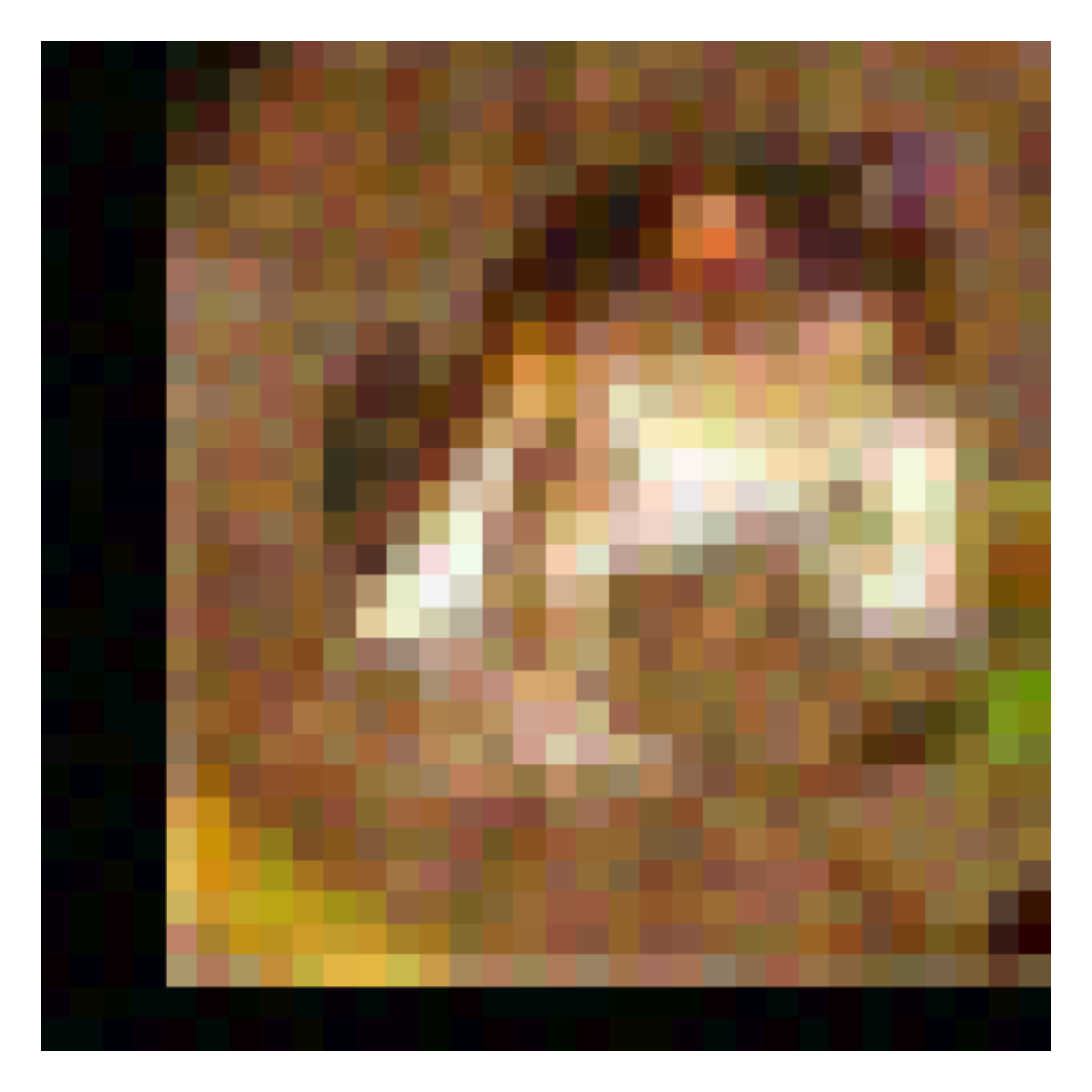}
        \caption{$x'$}
        \end{subfigure}
    \end{minipage}
    \begin{minipage}{0.3009\linewidth}
        \begin{subfigure}{\textwidth}
        \includegraphics[width=\linewidth]{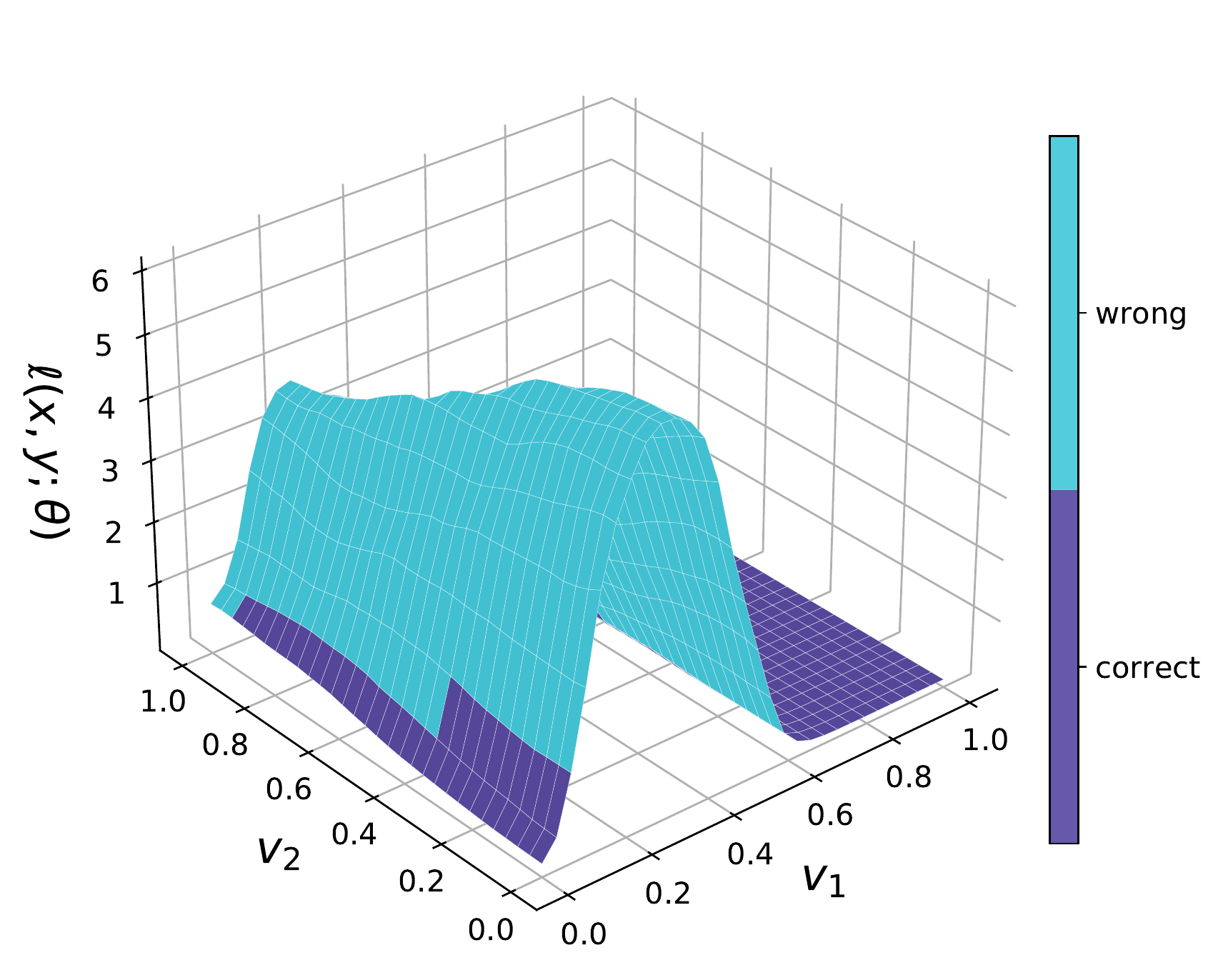}
        \caption{Loss surface}
        \end{subfigure}
    \end{minipage}
    \begin{minipage}{0.075\linewidth}
        \begin{subfigure}[t]{\textwidth}
        \includegraphics[width=\linewidth]{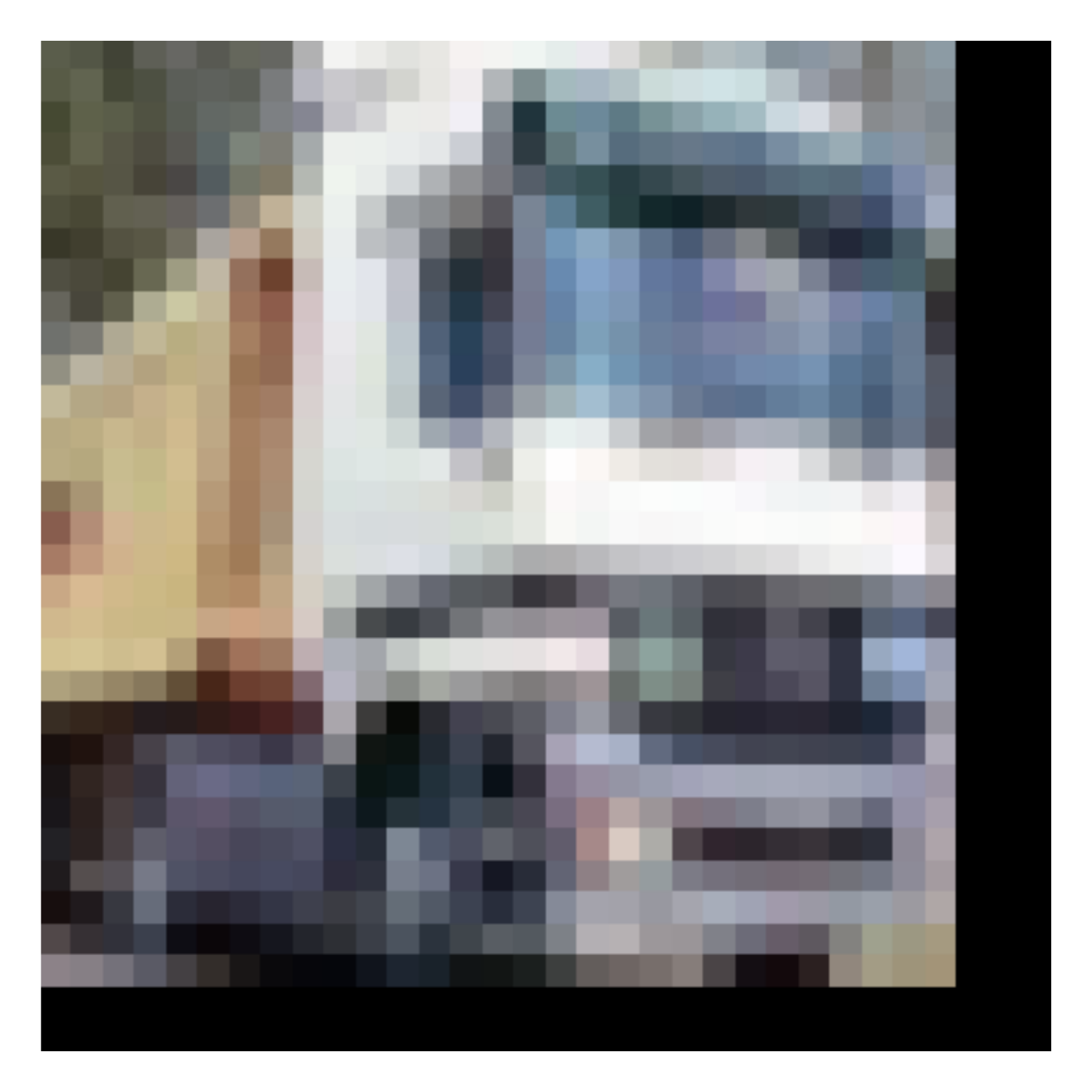}
        \caption{$x$}
        \end{subfigure} \\
        \begin{subfigure}[b]{\textwidth}
        \includegraphics[width=\linewidth]{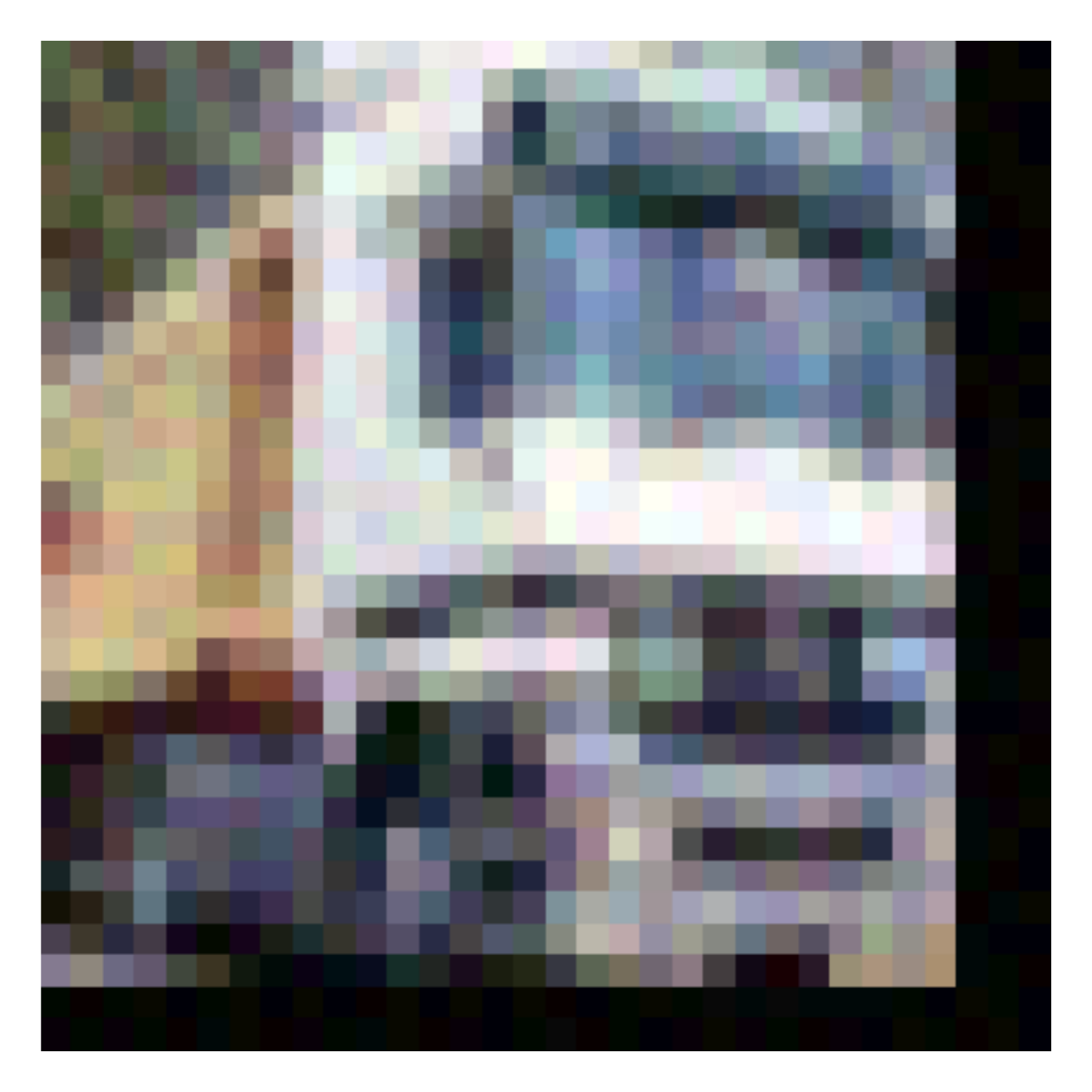}
        \caption{$x'$}
        \end{subfigure}
    \end{minipage}
    \begin{minipage}{0.3009\linewidth}
        \begin{subfigure}{\textwidth}
        \includegraphics[width=\linewidth]{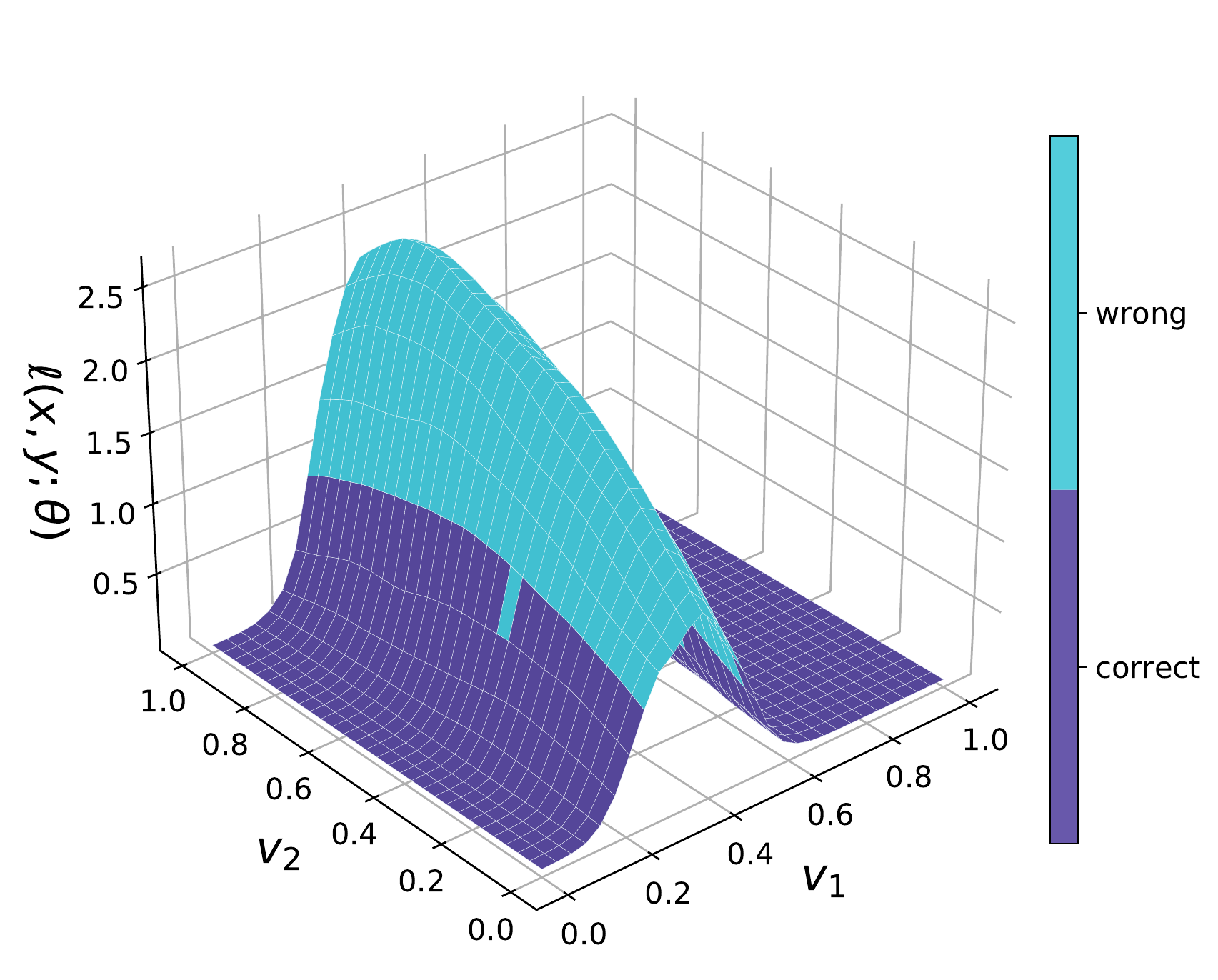}
        \caption{Loss surface}
        \end{subfigure}
    \end{minipage} \\

    \begin{minipage}{0.075\linewidth}
        \begin{subfigure}[t]{\textwidth}
        \includegraphics[width=\linewidth]{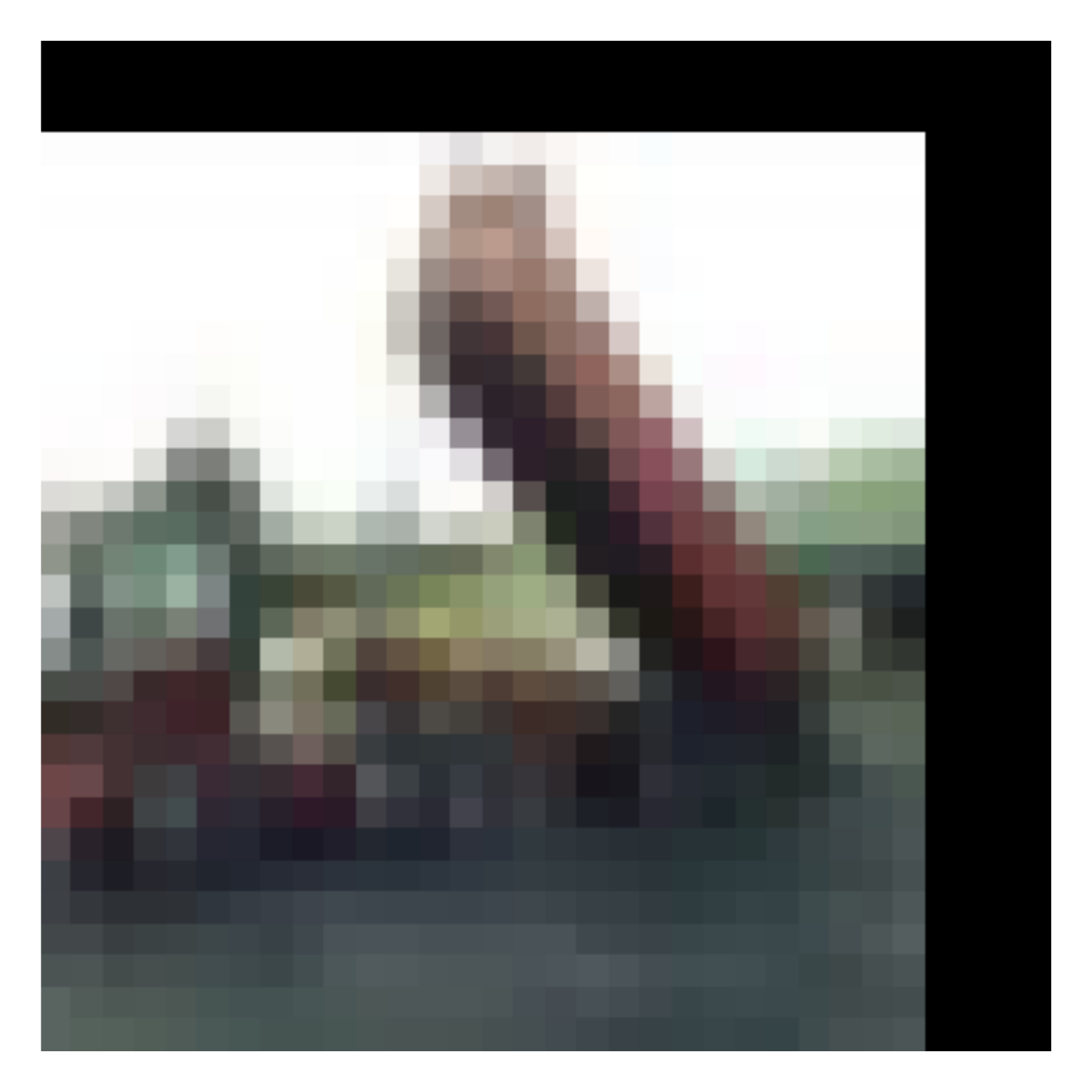}
        \caption{$x$}
        \end{subfigure} \\
        \begin{subfigure}[b]{\textwidth}
        \includegraphics[width=\linewidth]{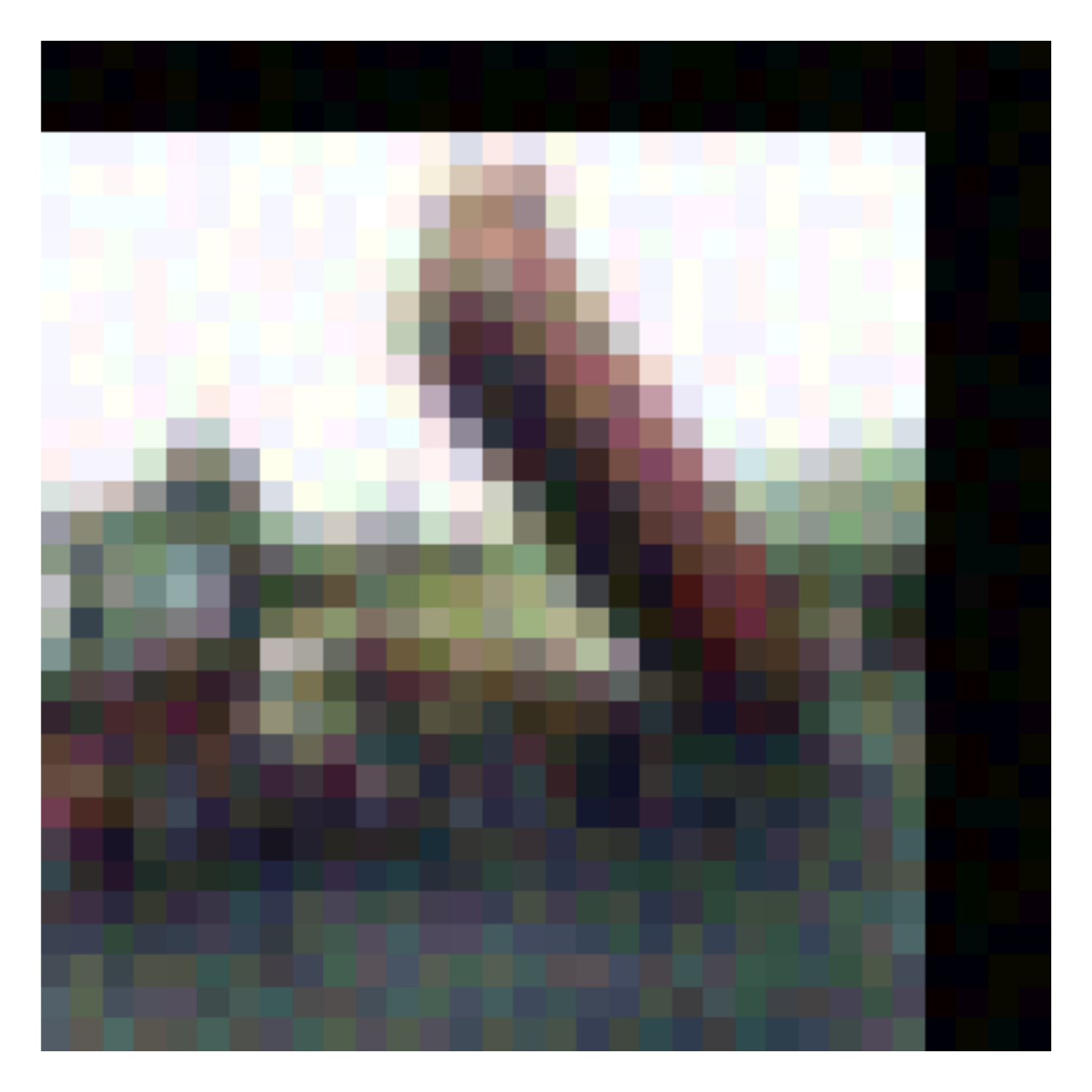}
        \caption{$x'$}
        \end{subfigure}
    \end{minipage}
    \begin{minipage}{0.3009\linewidth}
        \begin{subfigure}{\textwidth}
        \includegraphics[width=\linewidth]{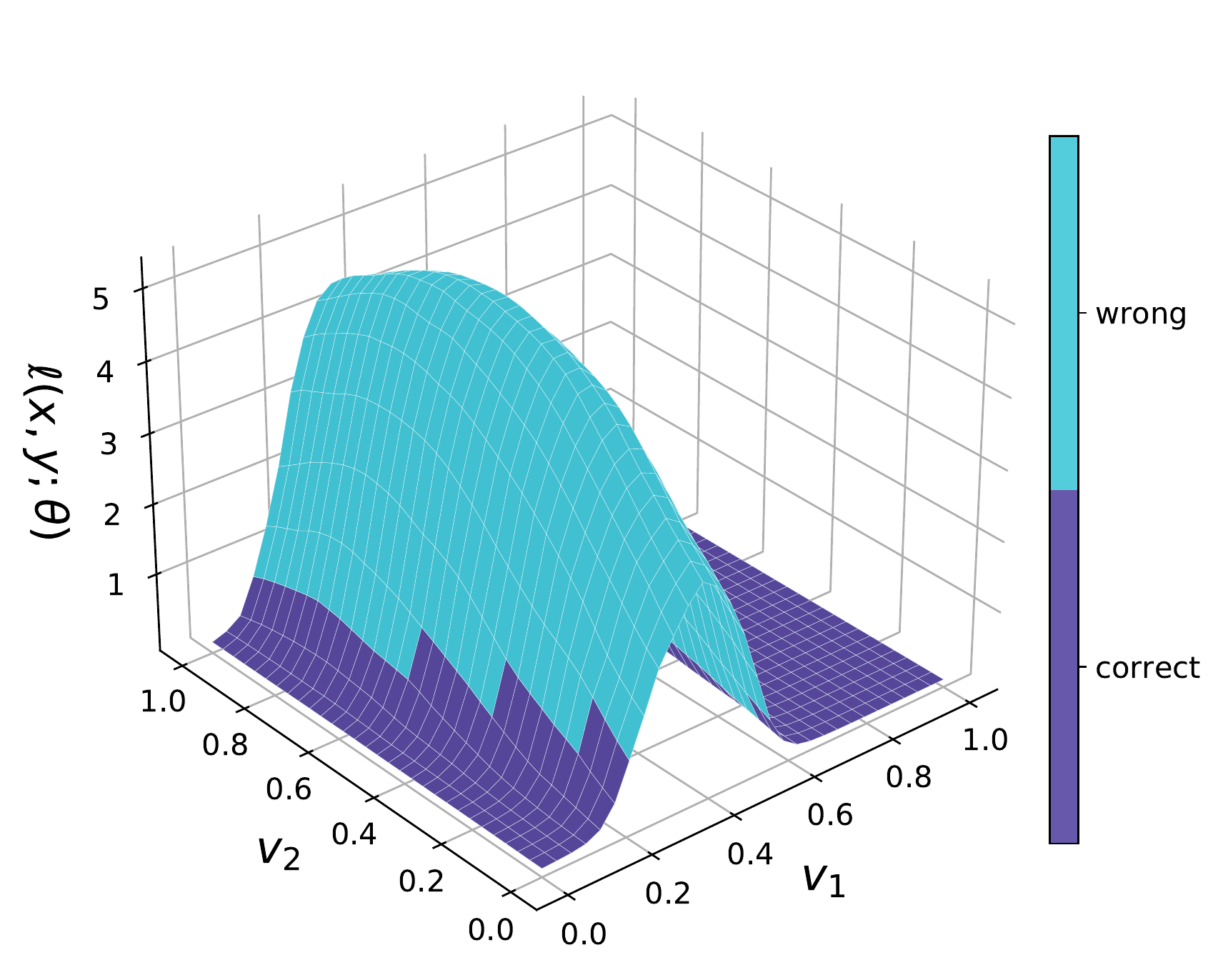}
        \caption{Loss surface}
        \end{subfigure}
    \end{minipage}
    \begin{minipage}{0.075\linewidth}
        \begin{subfigure}[t]{\textwidth}
        \includegraphics[width=\linewidth]{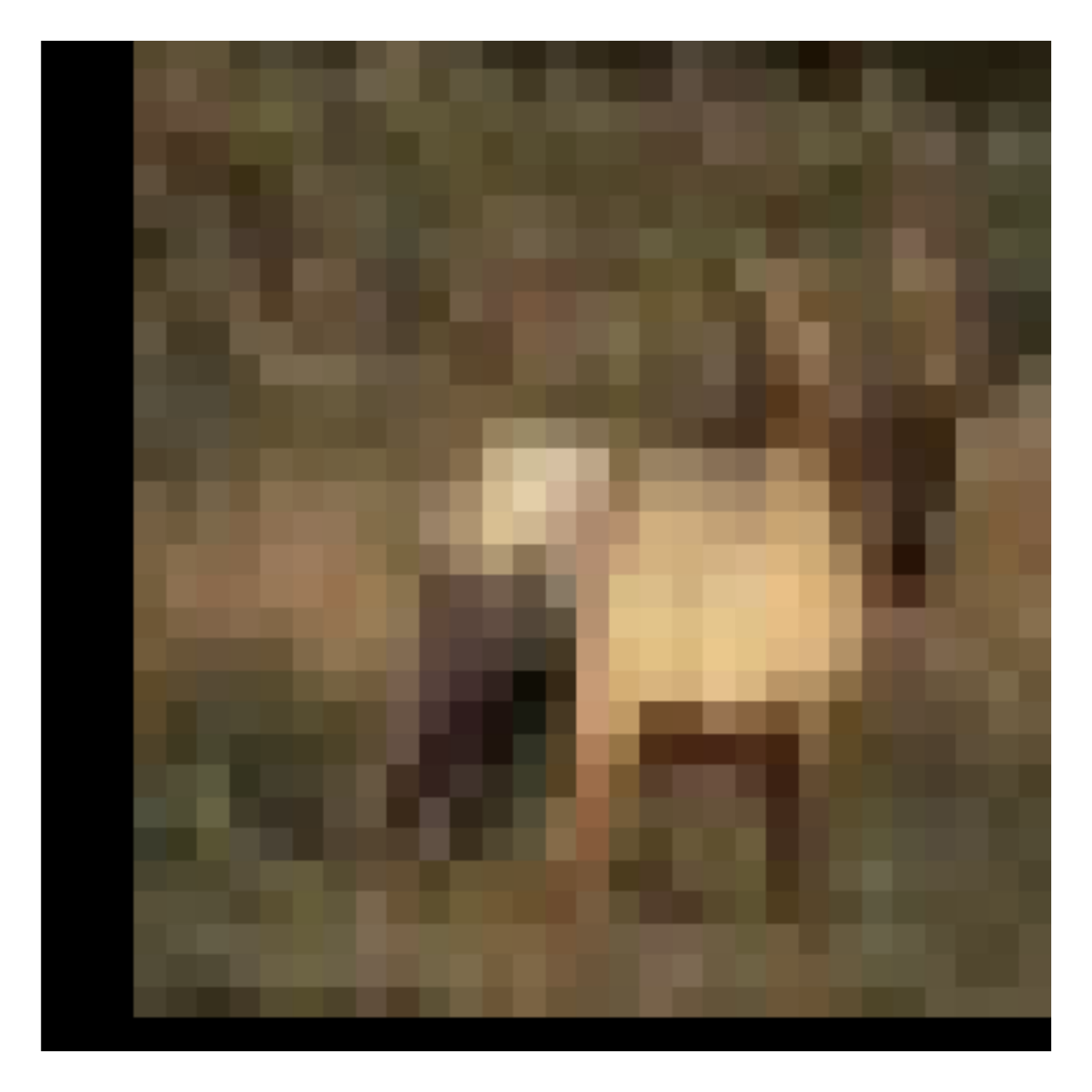}
        \caption{$x$}
        \end{subfigure} \\
        \begin{subfigure}[b]{\textwidth}
        \includegraphics[width=\linewidth]{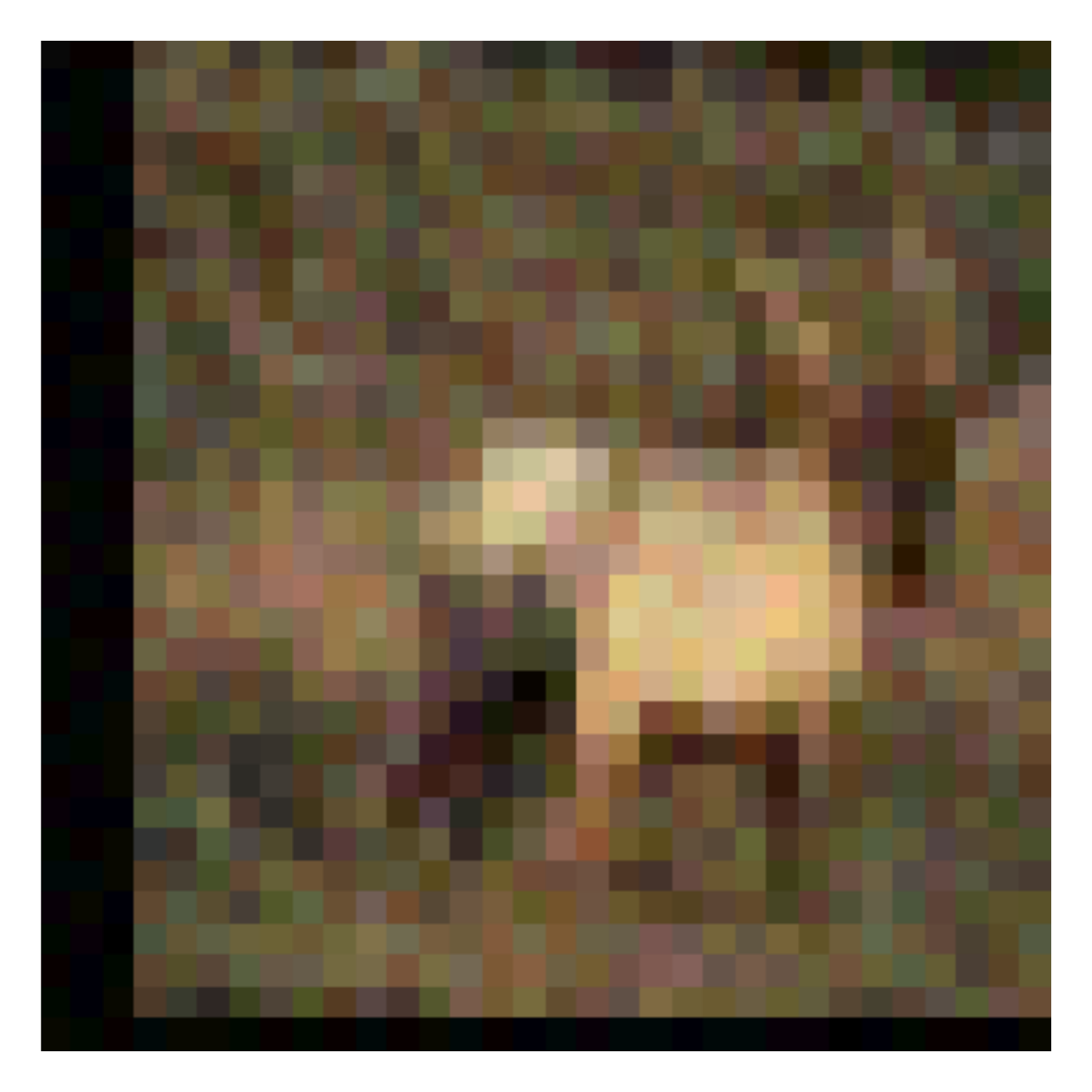}
        \caption{$x'$}
        \end{subfigure}
    \end{minipage}
    \begin{minipage}{0.3009\linewidth}
        \begin{subfigure}{\textwidth}
        \includegraphics[width=\linewidth]{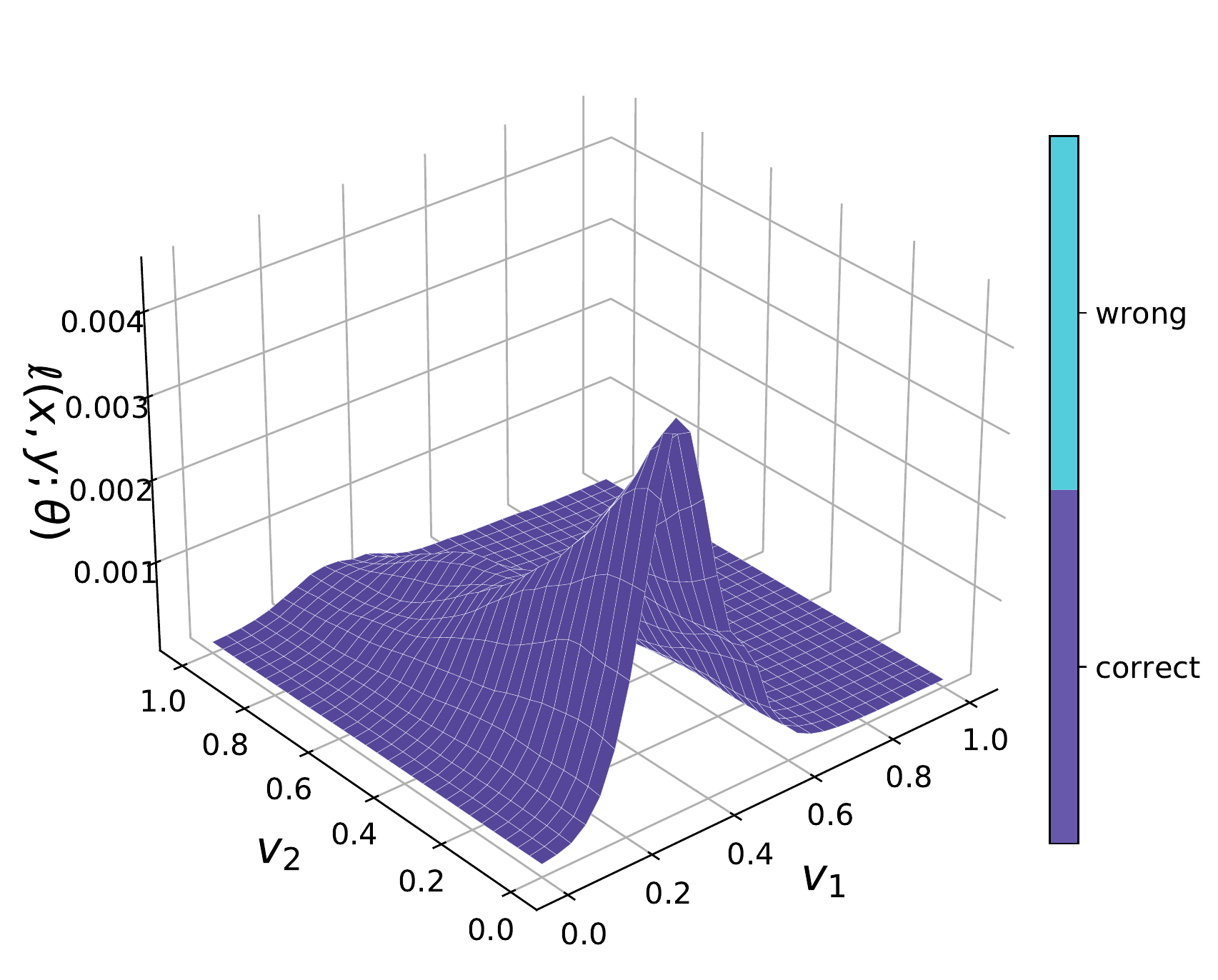}
        \caption{Loss surface}
        \end{subfigure}
    \end{minipage}
    \caption{(CIFAR10) Direction of fast adversarial perturbation $v_1$ and random direction $v_2$. Adversarial example $x'=x+v_1$ is generated from original example $x$.}
    \label{fig:fast_train}
\end{figure*}

\begin{figure*}[p]
    \centering
    \begin{minipage}{0.075\linewidth}
        \begin{subfigure}[t]{\textwidth}
        \includegraphics[width=\linewidth]{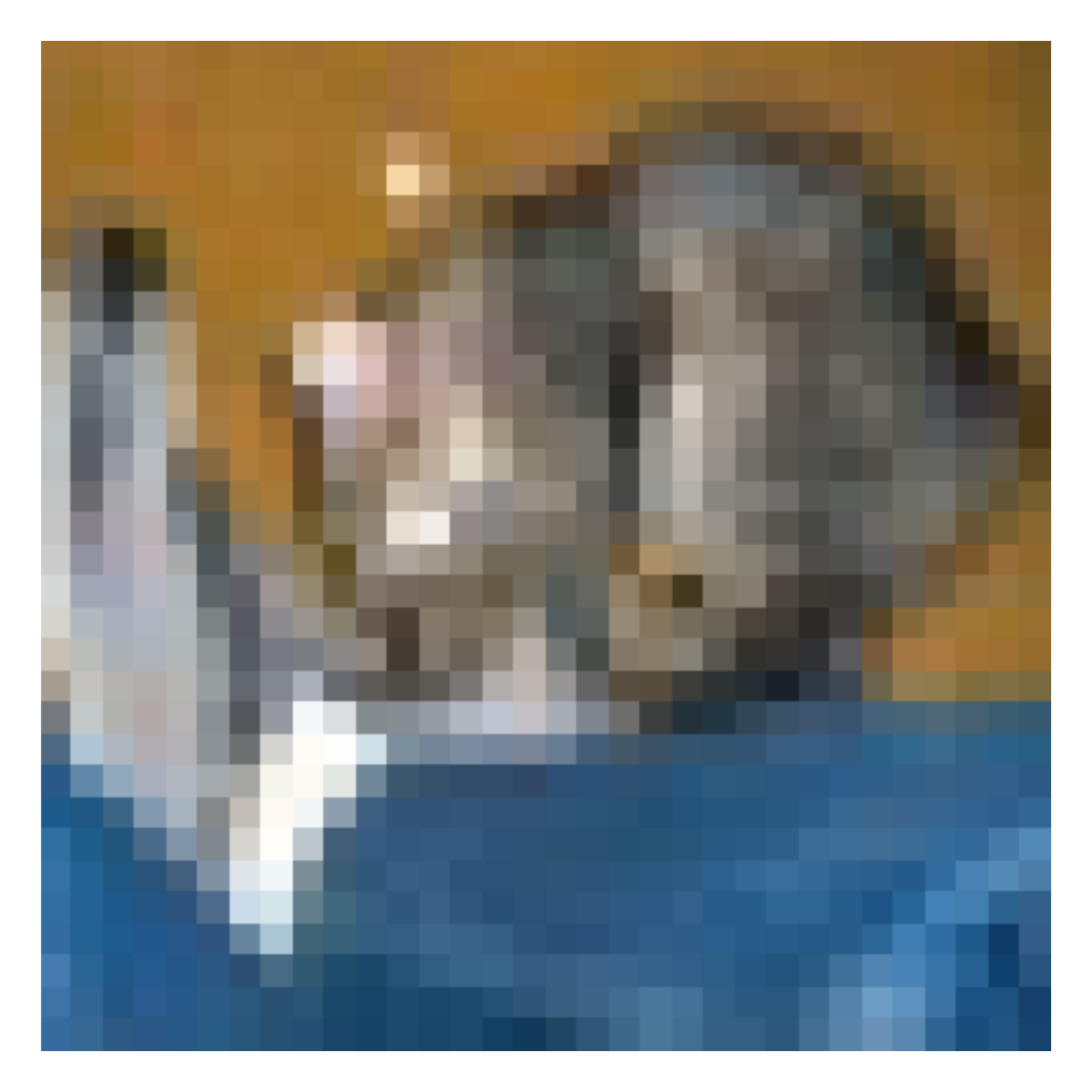}
        \caption{$x$}
        \end{subfigure} \\
        \begin{subfigure}[b]{\textwidth}
        \includegraphics[width=\linewidth]{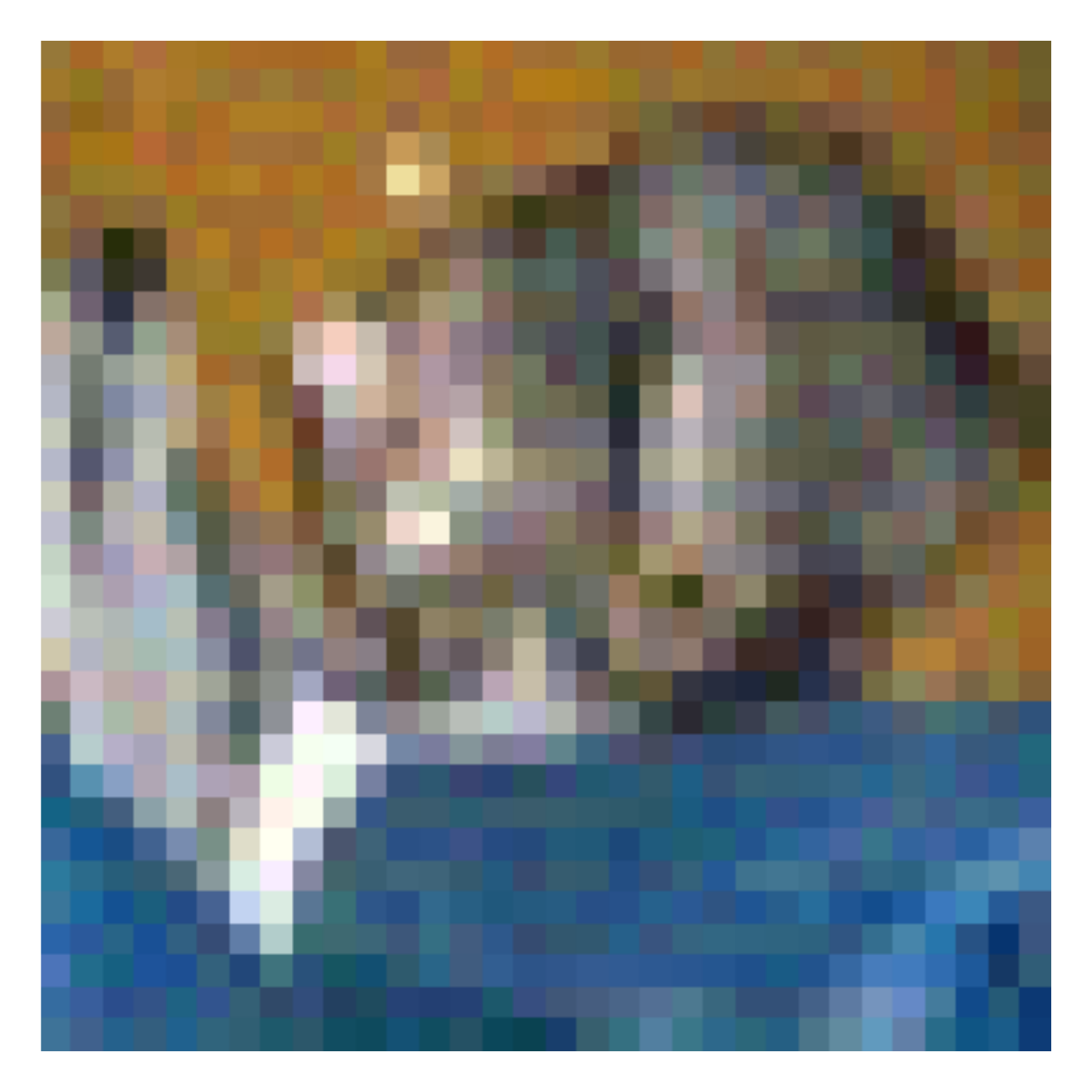}
        \caption{$x'$}
        \end{subfigure}
    \end{minipage}
    \begin{minipage}{0.3009\linewidth}
        \begin{subfigure}{\textwidth}
        \includegraphics[width=\linewidth]{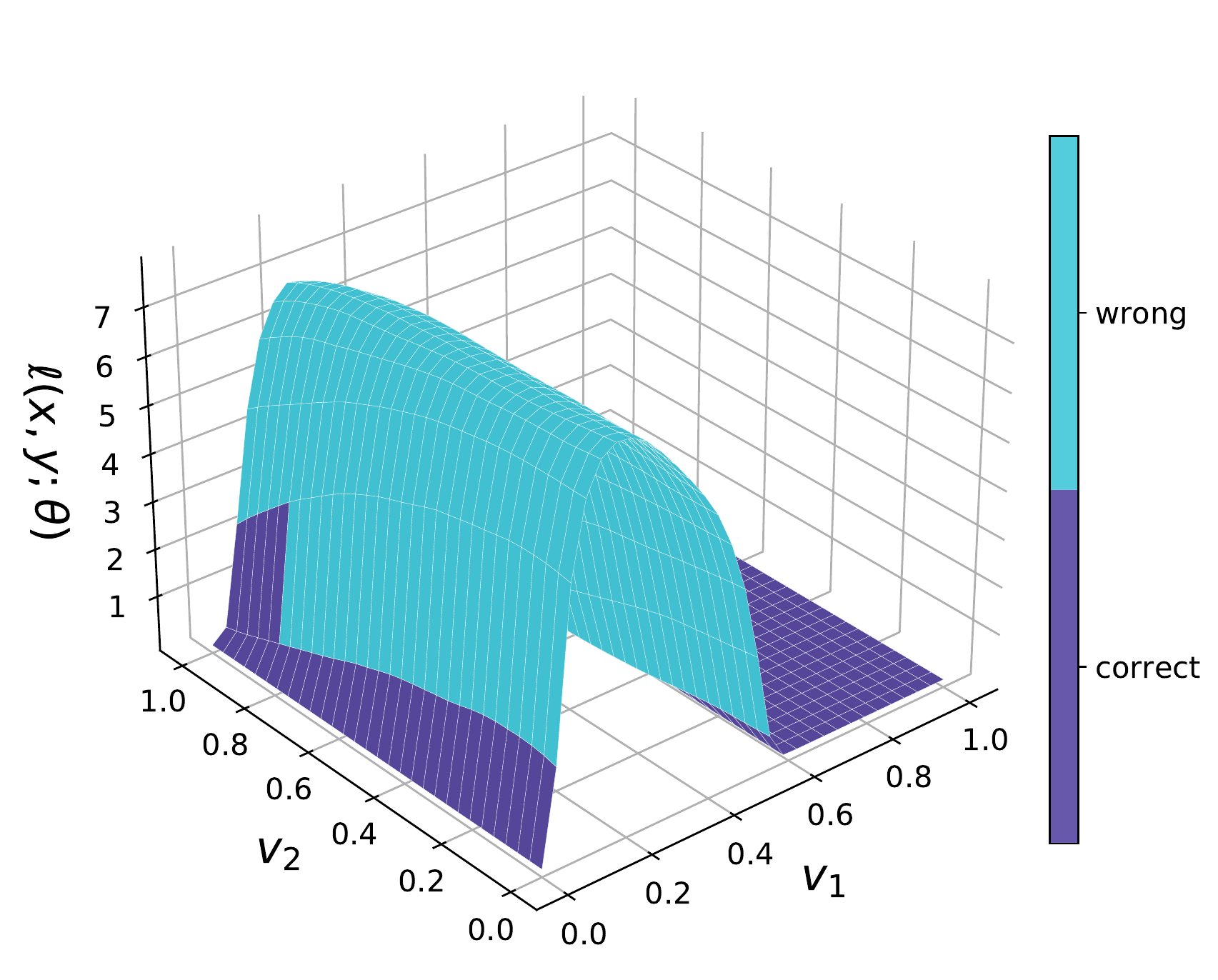}
        \caption{Loss surface}
        \end{subfigure}
    \end{minipage}
    \begin{minipage}{0.075\linewidth}
        \begin{subfigure}[t]{\textwidth}
        \includegraphics[width=\linewidth]{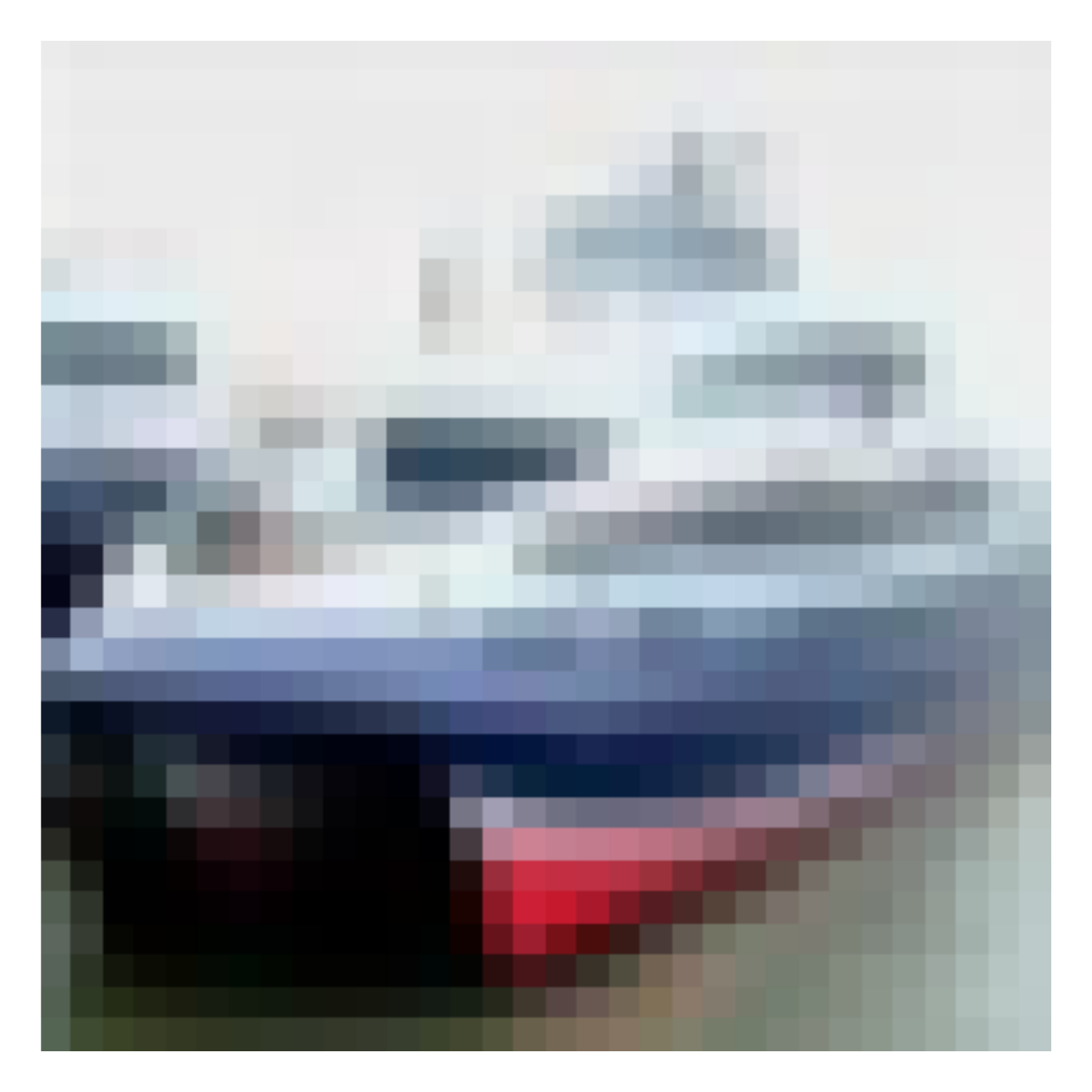}
        \caption{$x$}
        \end{subfigure} \\
        \begin{subfigure}[b]{\textwidth}
        \includegraphics[width=\linewidth]{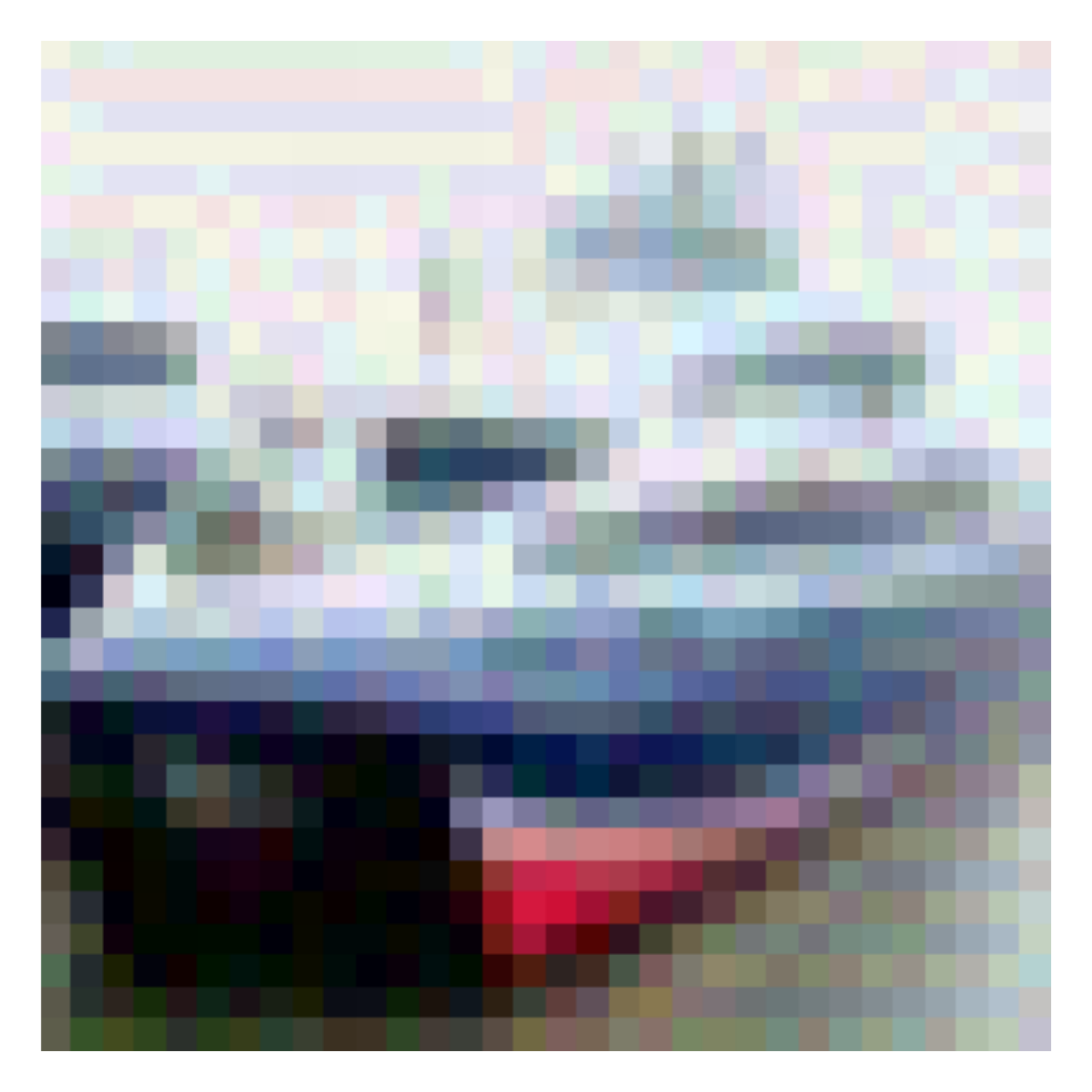}
        \caption{$x'$}
        \end{subfigure}
    \end{minipage}
    \begin{minipage}{0.3009\linewidth}
        \begin{subfigure}{\textwidth}
        \includegraphics[width=\linewidth]{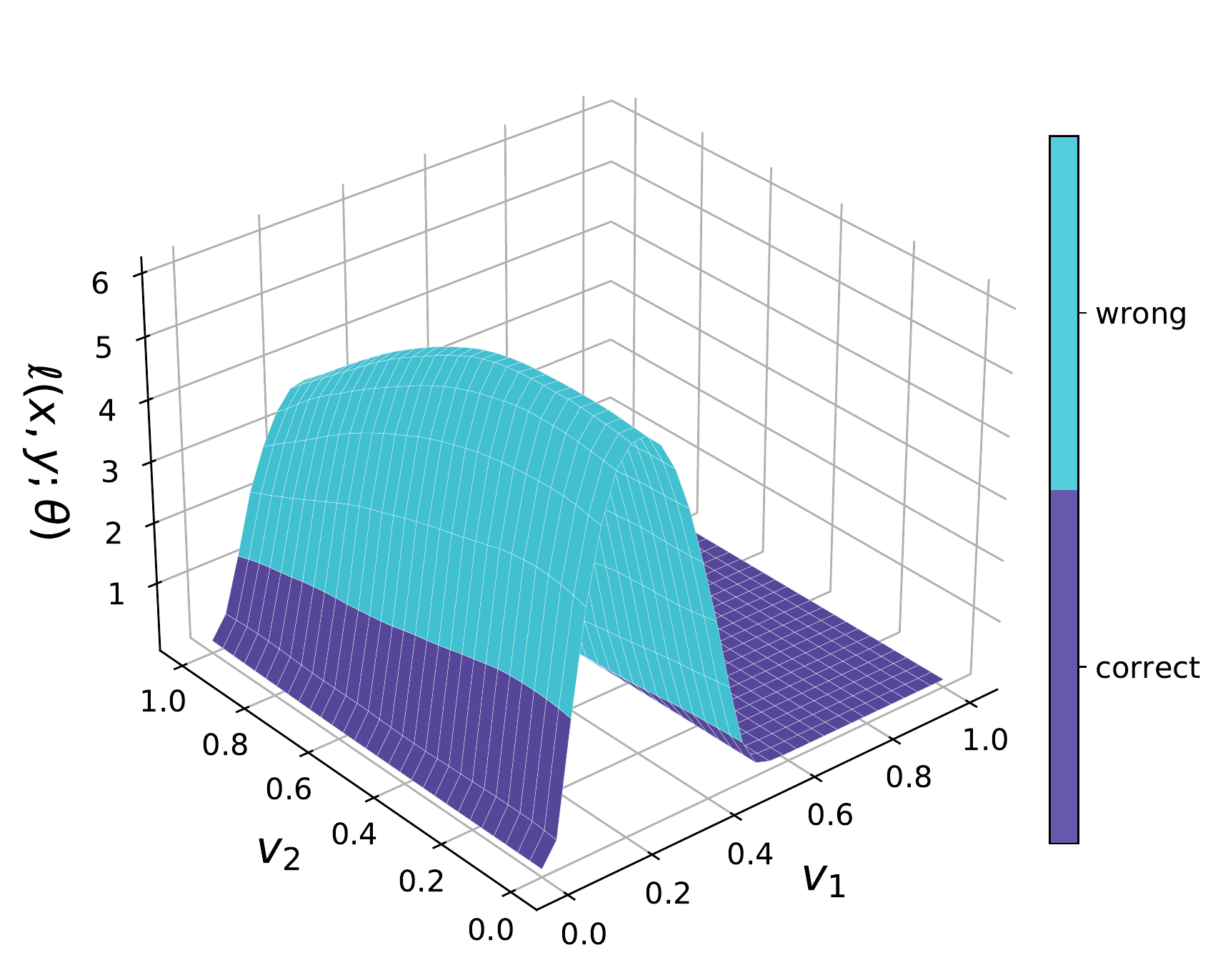}
        \caption{Loss surface}
        \end{subfigure}
    \end{minipage} \\
    
    \begin{minipage}{0.075\linewidth}
        \begin{subfigure}[t]{\textwidth}
        \includegraphics[width=\linewidth]{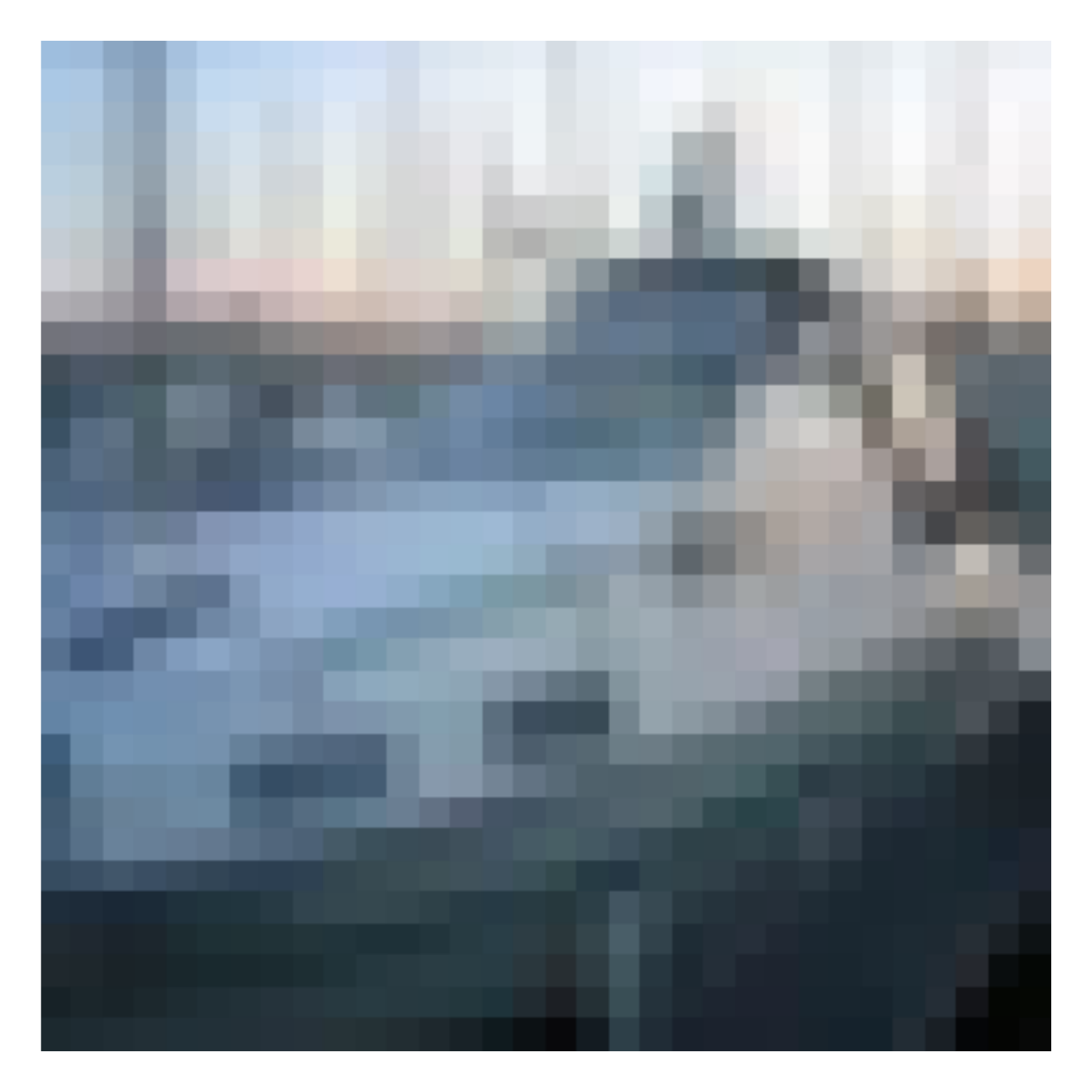}
        \caption{$x$}
        \end{subfigure} \\
        \begin{subfigure}[b]{\textwidth}
        \includegraphics[width=\linewidth]{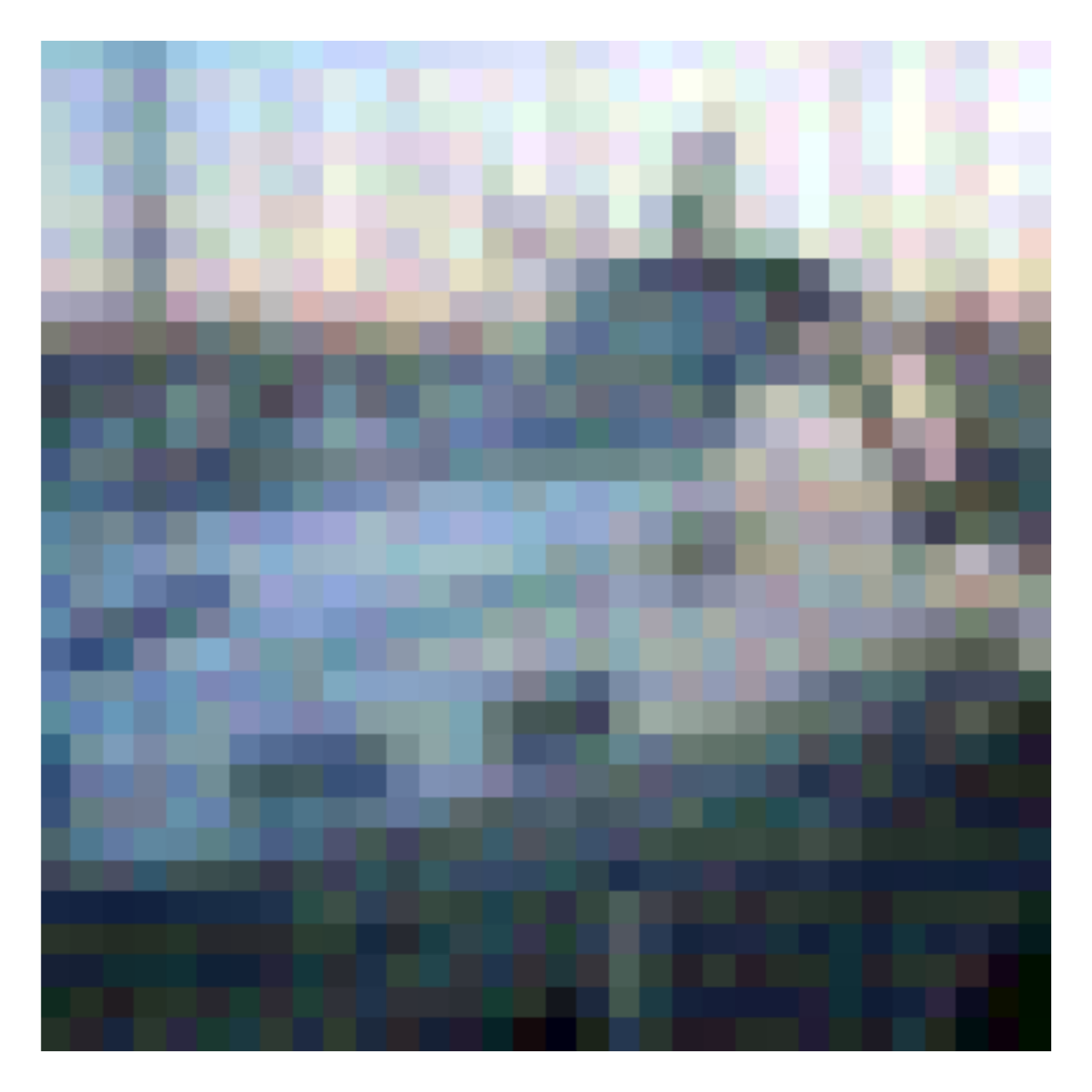}
        \caption{$x'$}
        \end{subfigure}
    \end{minipage}
    \begin{minipage}{0.3009\linewidth}
        \begin{subfigure}{\textwidth}
        \includegraphics[width=\linewidth]{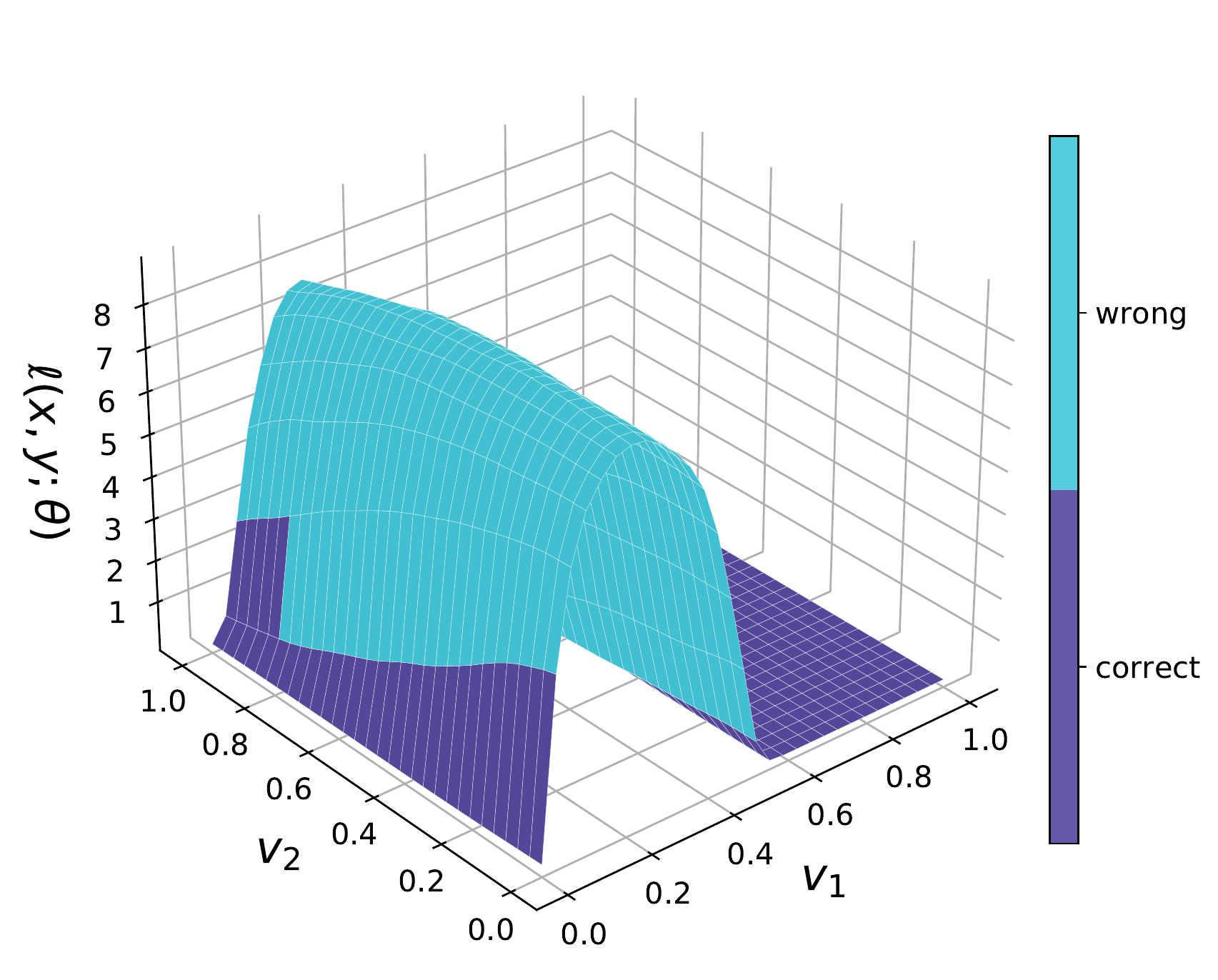}
        \caption{Loss surface}
        \end{subfigure}
    \end{minipage}
    \begin{minipage}{0.075\linewidth}
        \begin{subfigure}[t]{\textwidth}
        \includegraphics[width=\linewidth]{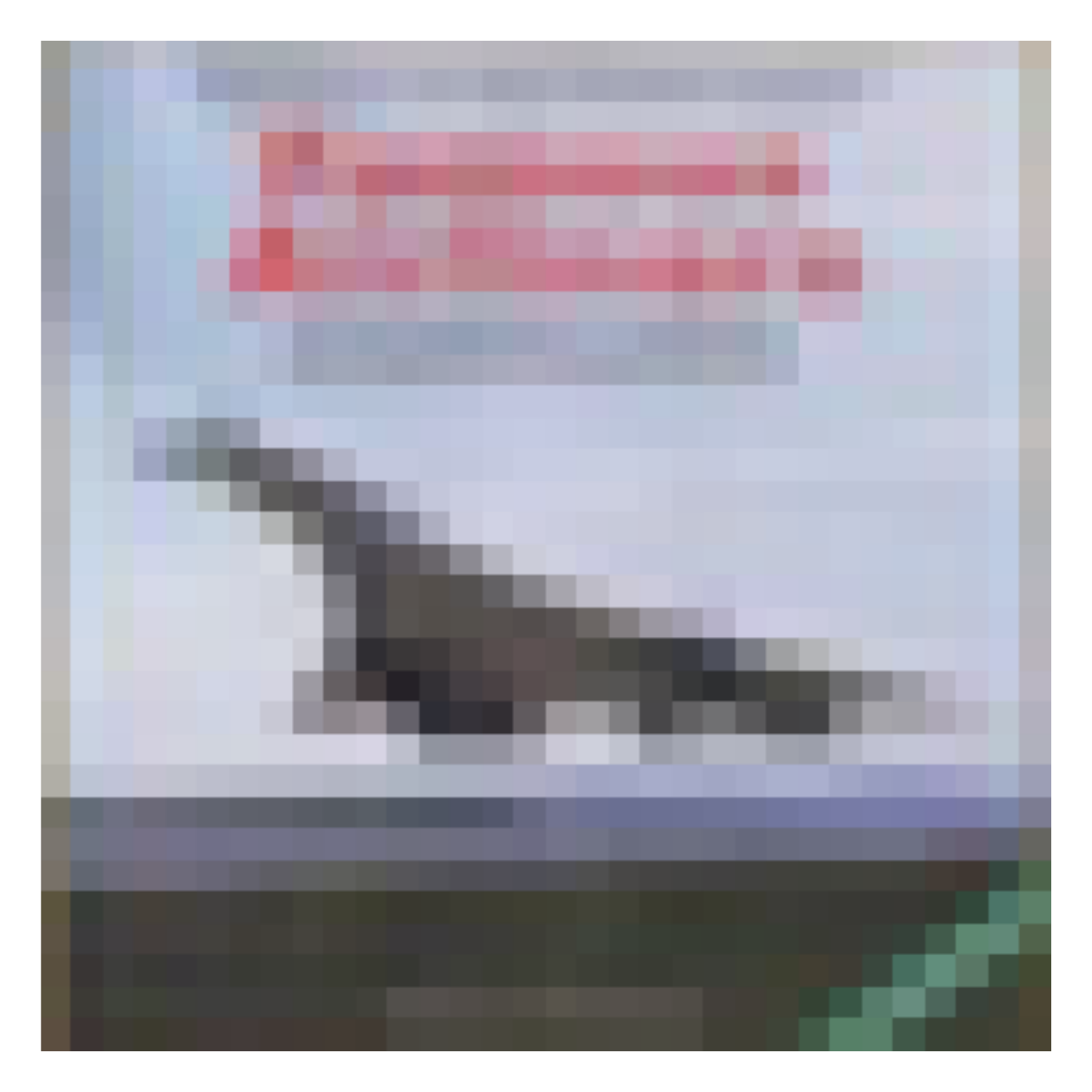}
        \caption{$x$}
        \end{subfigure} \\
        \begin{subfigure}[b]{\textwidth}
        \includegraphics[width=\linewidth]{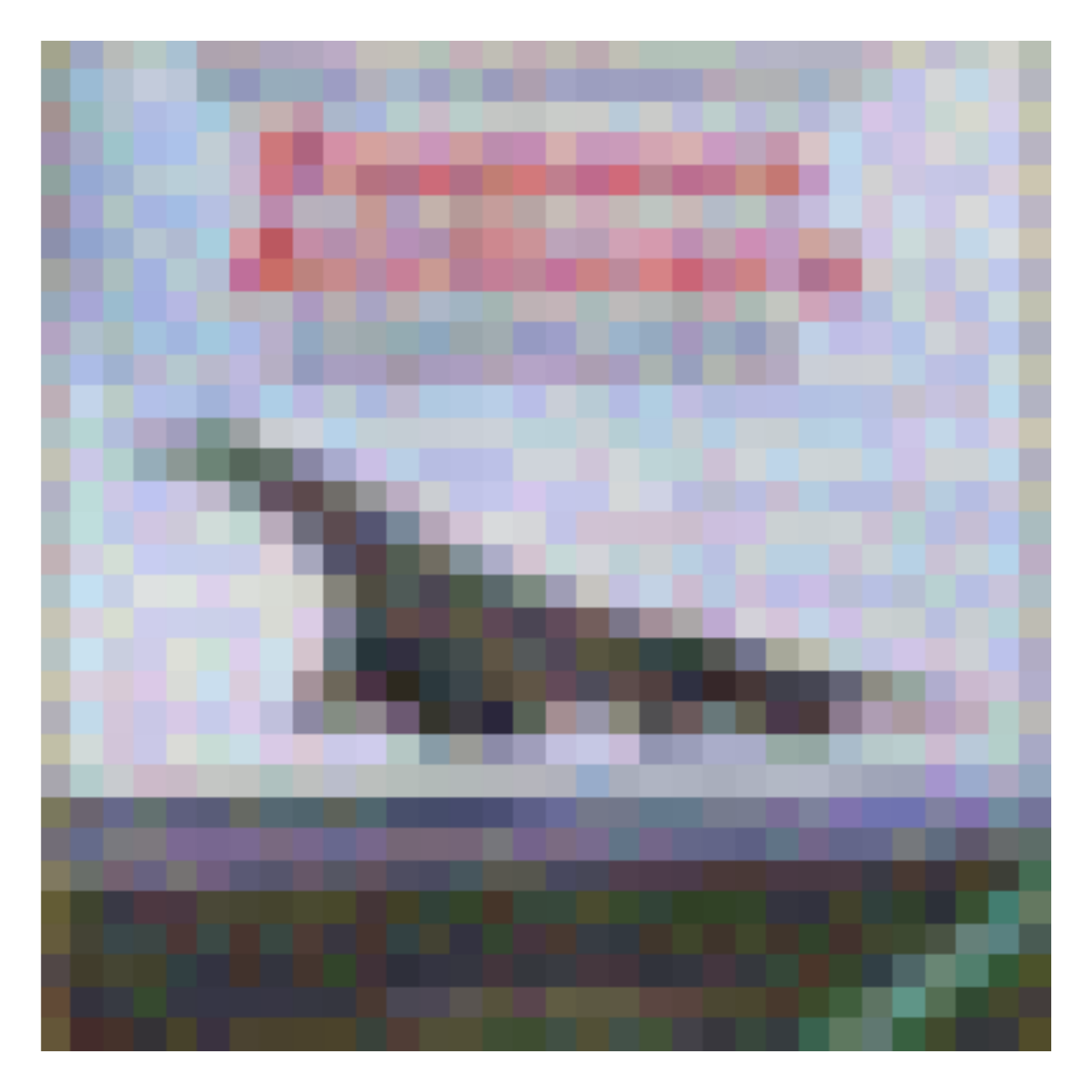}
        \caption{$x'$}
        \end{subfigure}
    \end{minipage}
    \begin{minipage}{0.3009\linewidth}
        \begin{subfigure}{\textwidth}
        \includegraphics[width=\linewidth]{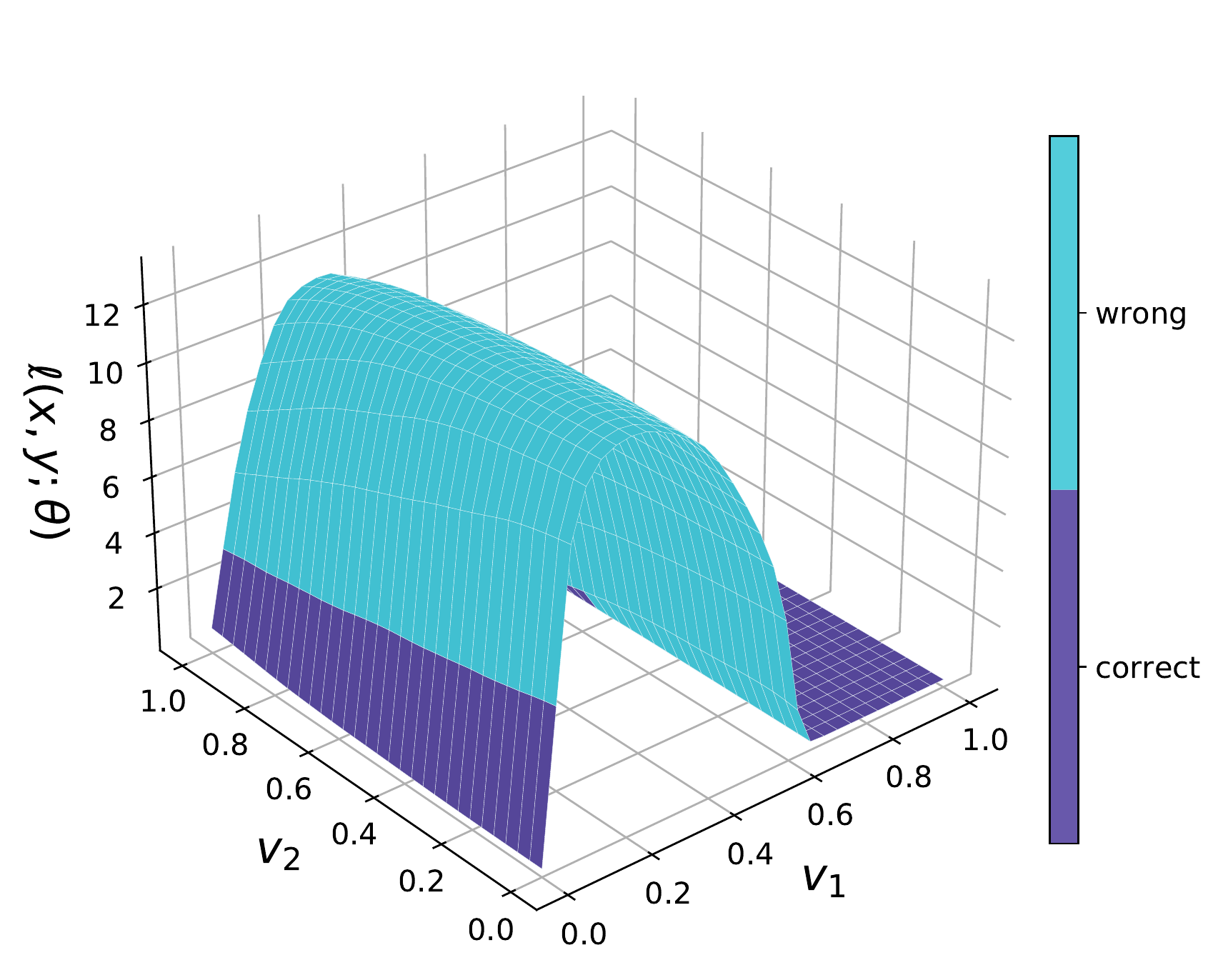}
        \caption{Loss surface}
        \end{subfigure}
    \end{minipage}
    \caption{(CIFAR10) Direction of FGSM adversarial perturbation $v_1$ and random direction $v_2$. Adversarial example $x'=x+v_1$ is generated from original example $x$.}
    \label{fig:fgsm_test}
\end{figure*}

\begin{figure*}[p]
    \centering
    \begin{minipage}{0.075\linewidth}
        \begin{subfigure}[t]{\textwidth}
        \includegraphics[width=\linewidth]{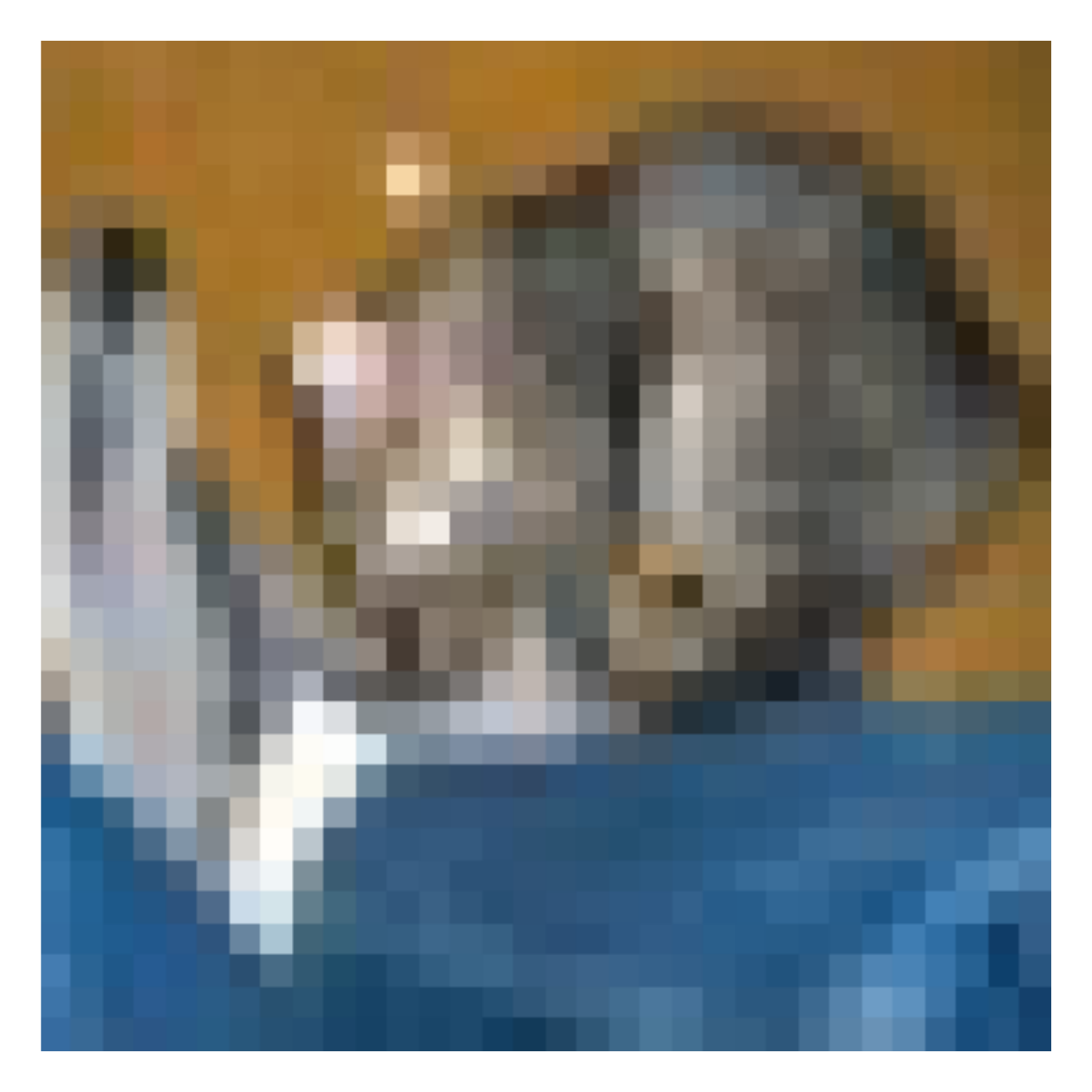}
        \caption{$x$}
        \end{subfigure} \\
        \begin{subfigure}[b]{\textwidth}
        \includegraphics[width=\linewidth]{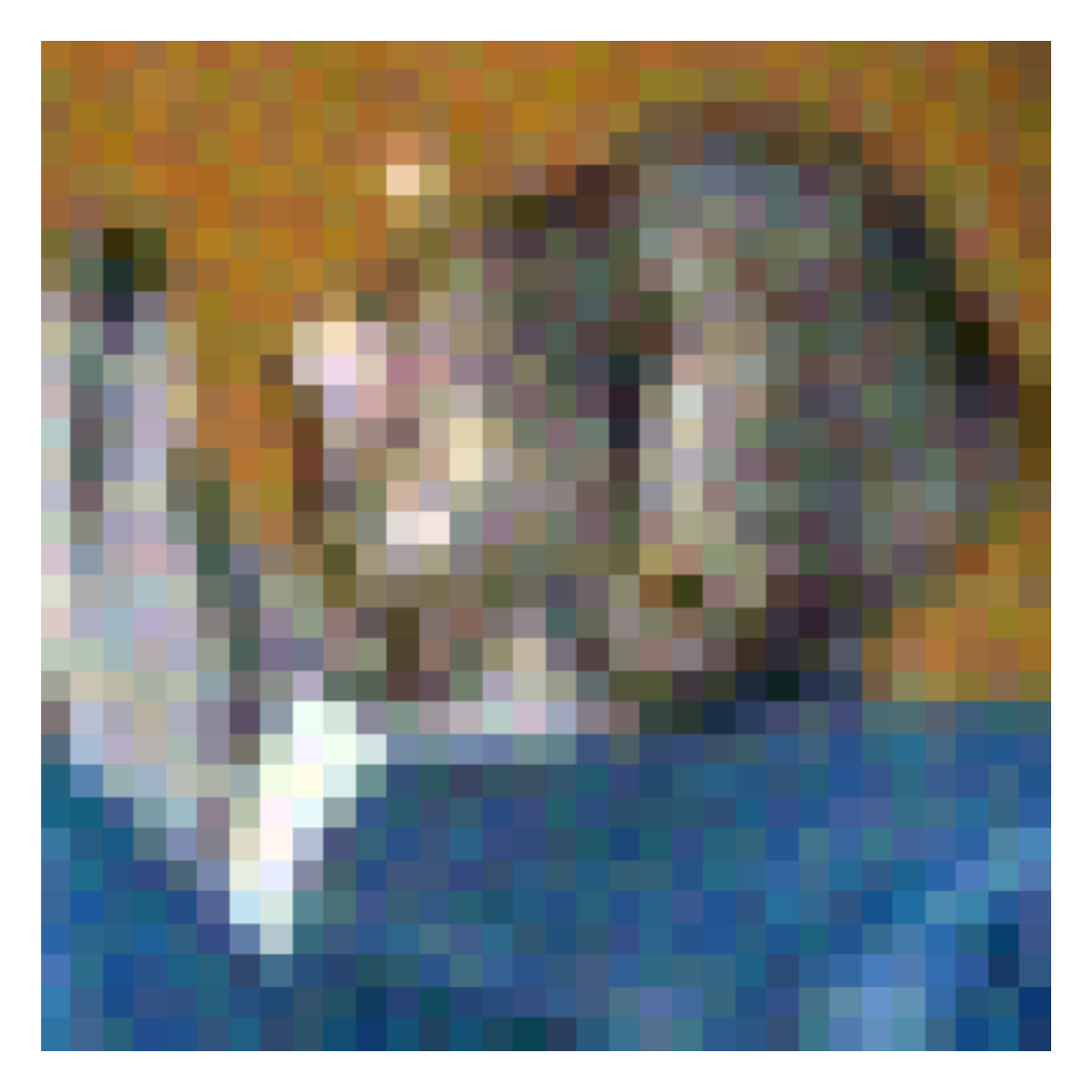}
        \caption{$x'$}
        \end{subfigure}
    \end{minipage}
    \begin{minipage}{0.3009\linewidth}
        \begin{subfigure}{\textwidth}
        \includegraphics[width=\linewidth]{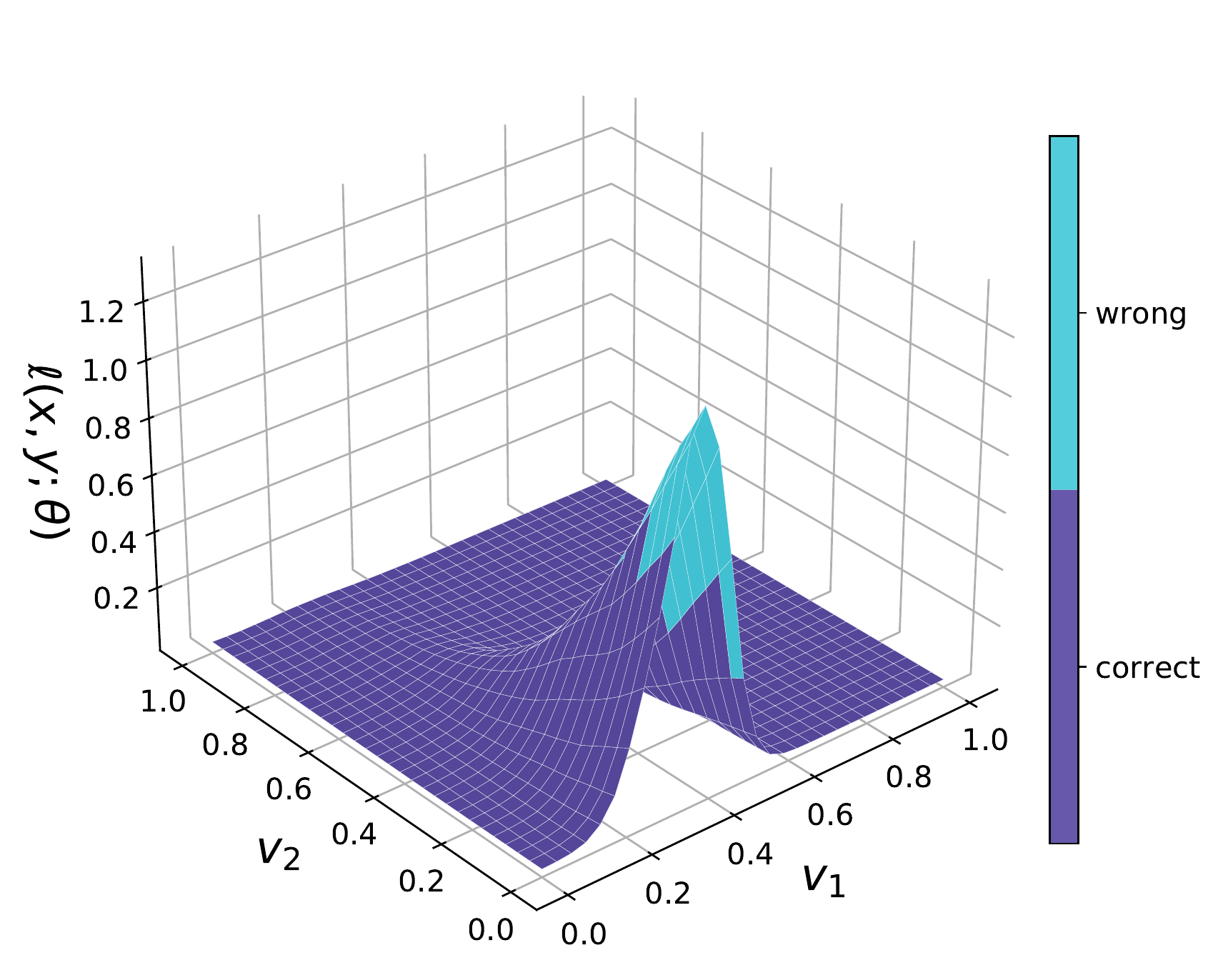}
        \caption{Loss surface}
        \end{subfigure}
    \end{minipage}
    \begin{minipage}{0.075\linewidth}
        \begin{subfigure}[t]{\textwidth}
        \includegraphics[width=\linewidth]{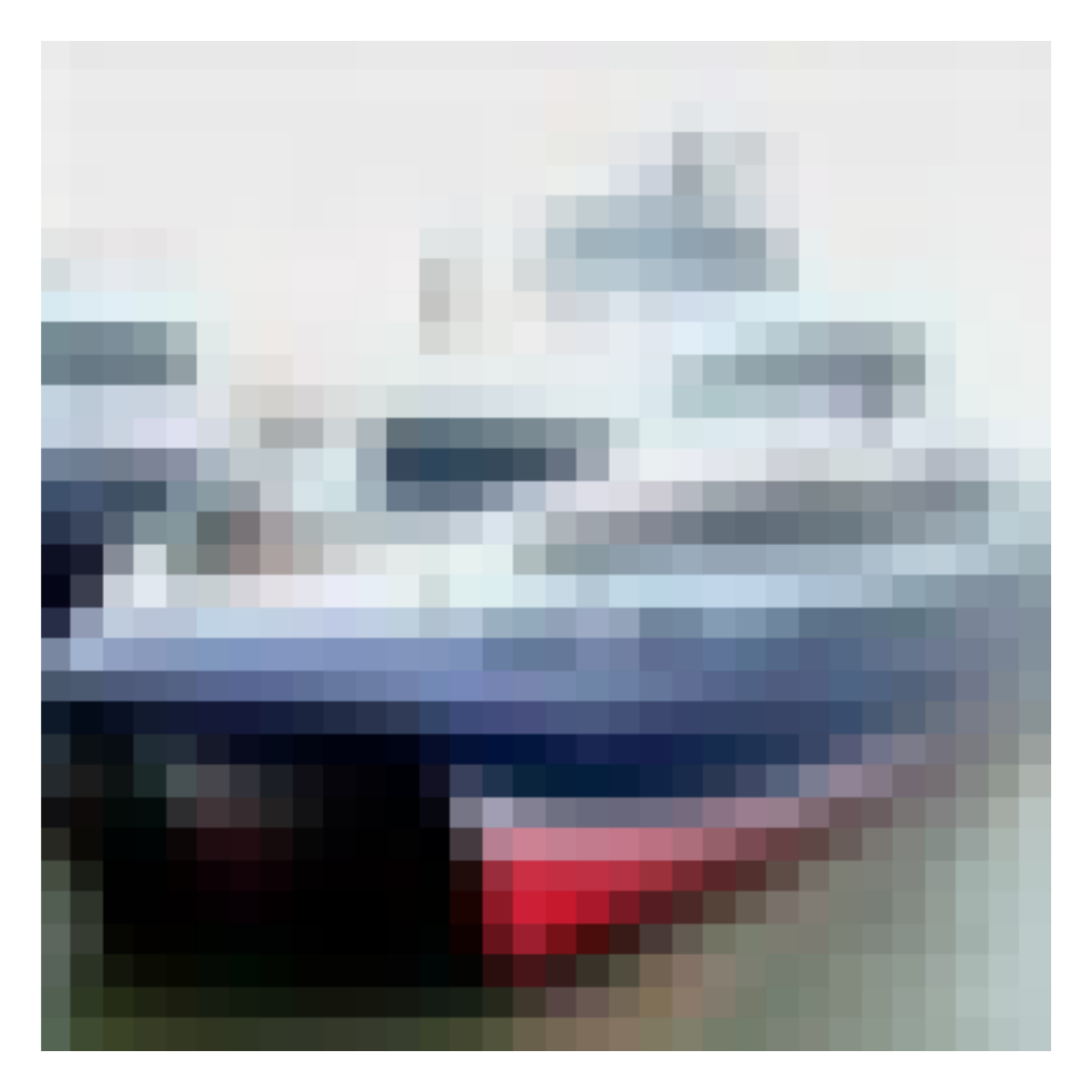}
        \caption{$x$}
        \end{subfigure} \\
        \begin{subfigure}[b]{\textwidth}
        \includegraphics[width=\linewidth]{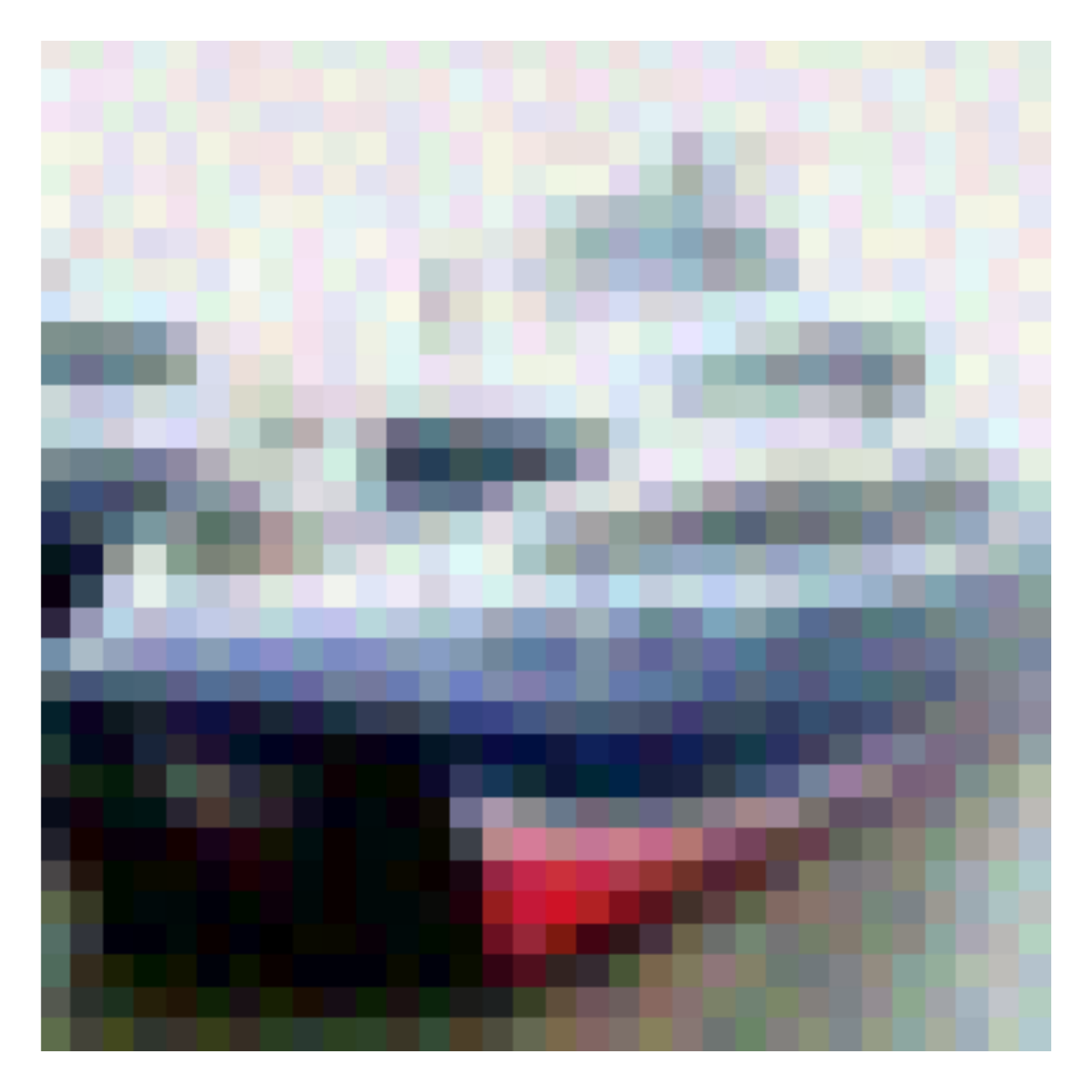}
        \caption{$x'$}
        \end{subfigure}
    \end{minipage}
    \begin{minipage}{0.3009\linewidth}
        \begin{subfigure}{\textwidth}
        \includegraphics[width=\linewidth]{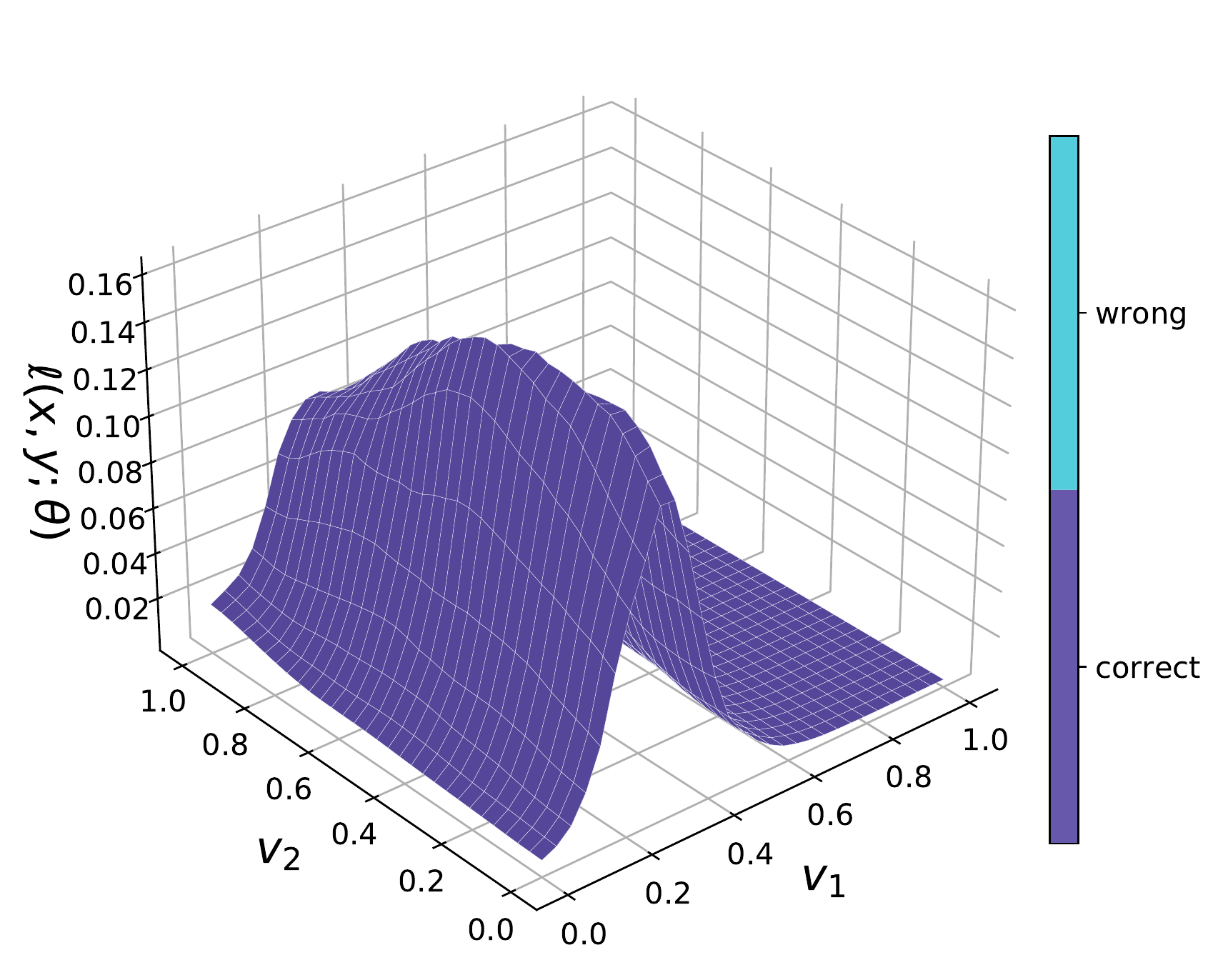}
        \caption{Loss surface}
        \end{subfigure}
    \end{minipage} \\

    \begin{minipage}{0.075\linewidth}
        \begin{subfigure}[t]{\textwidth}
        \includegraphics[width=\linewidth]{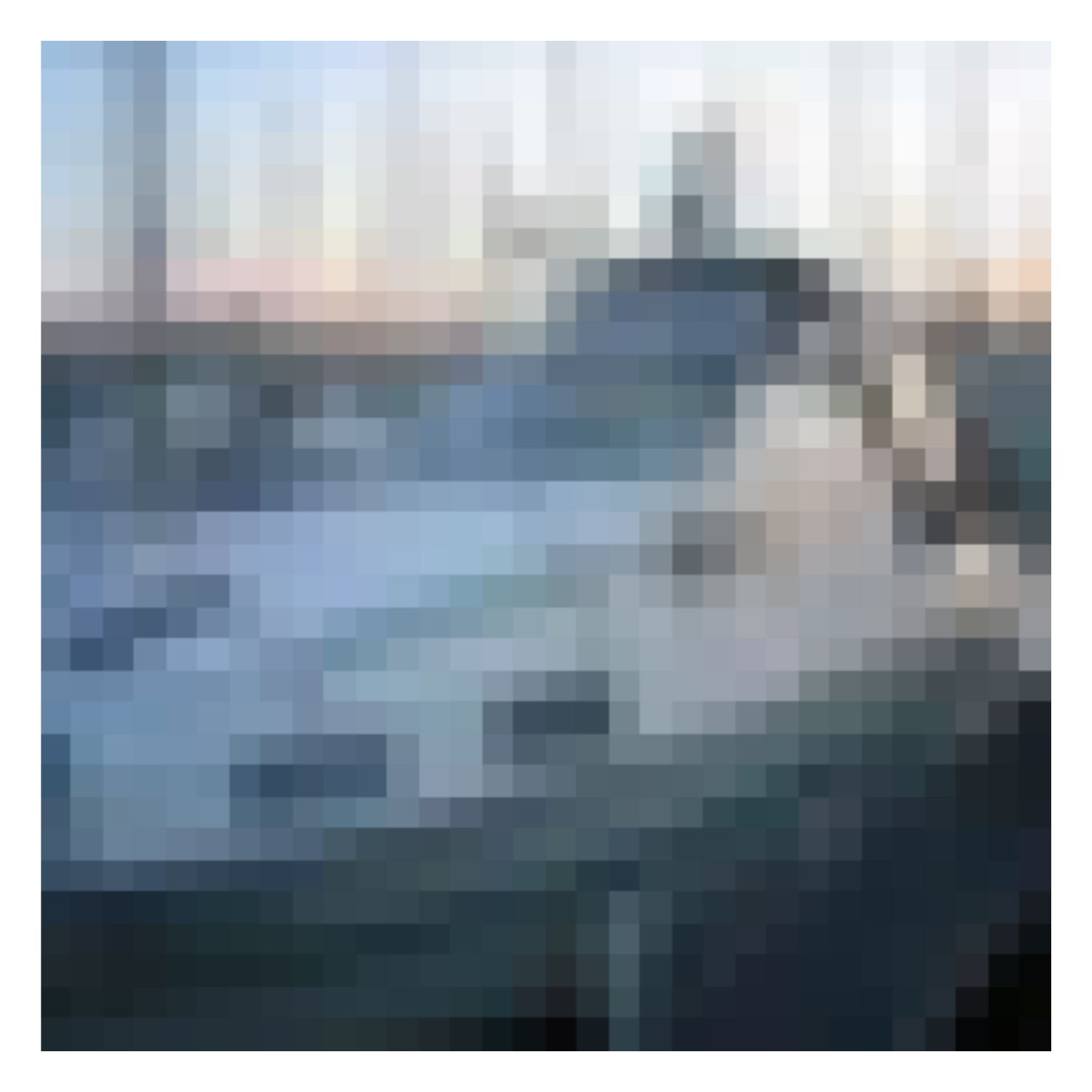}
        \caption{$x$}
        \end{subfigure} \\
        \begin{subfigure}[b]{\textwidth}
        \includegraphics[width=\linewidth]{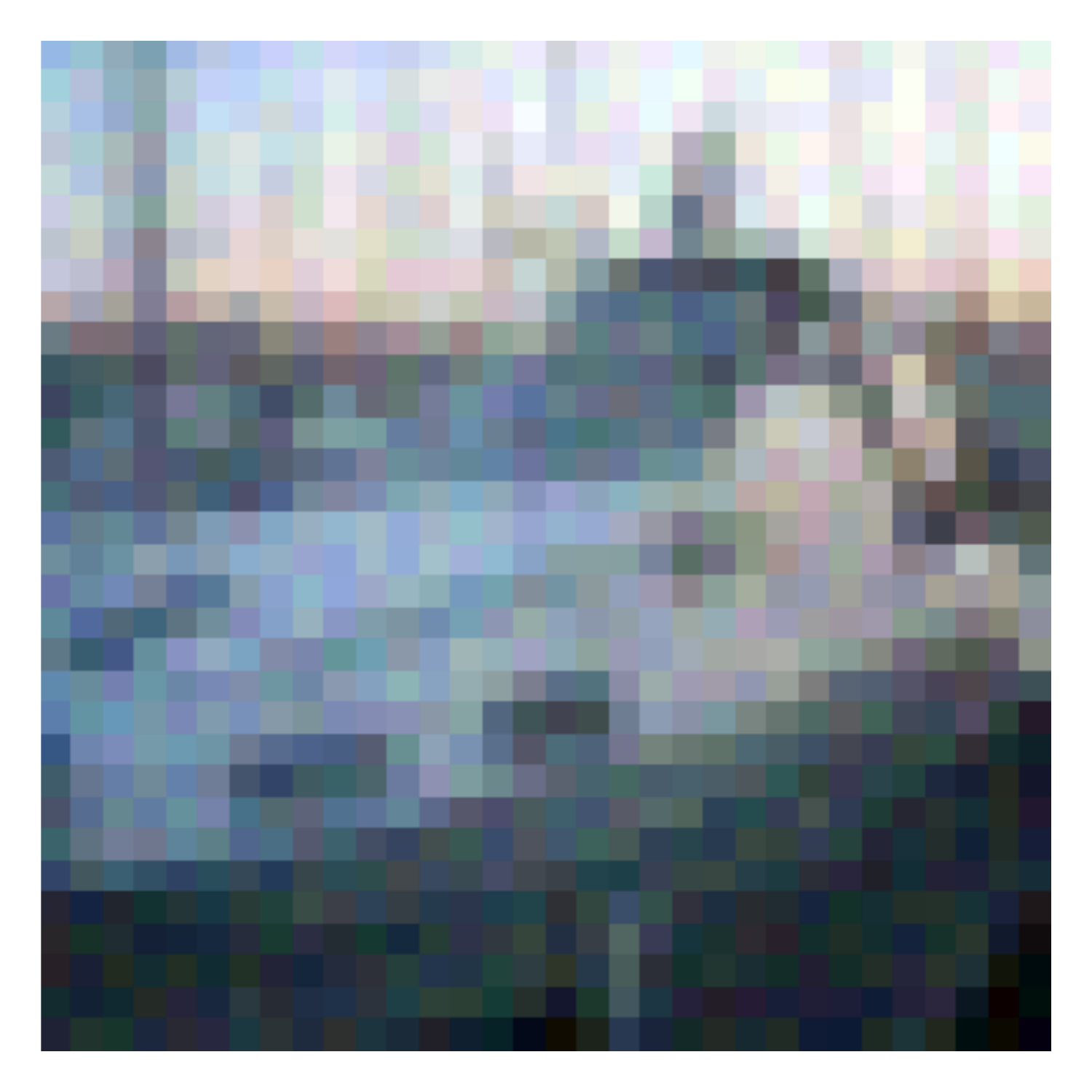}
        \caption{$x'$}
        \end{subfigure}
    \end{minipage}
    \begin{minipage}{0.3009\linewidth}
        \begin{subfigure}{\textwidth}
        \includegraphics[width=\linewidth]{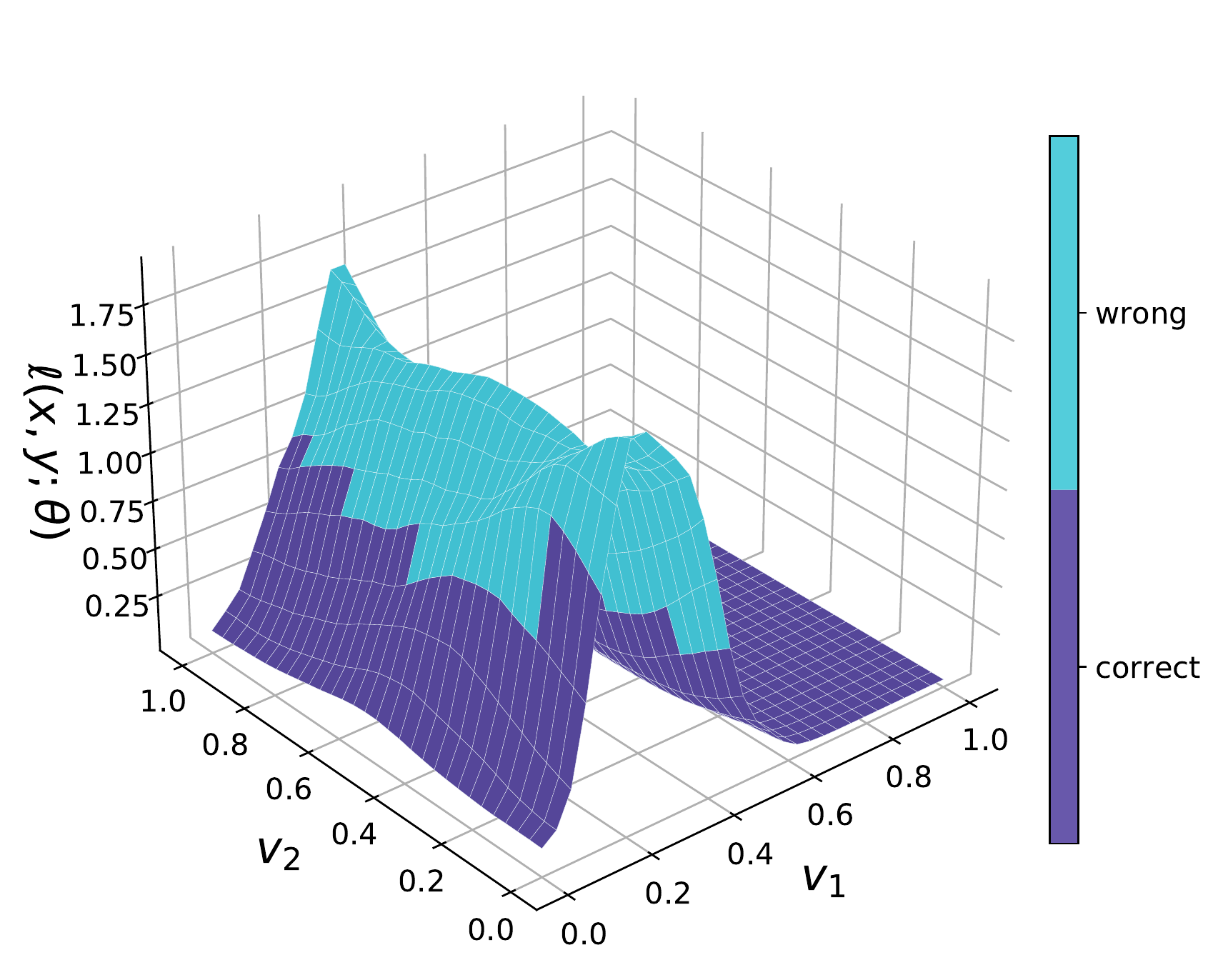}
        \caption{Loss surface}
        \end{subfigure}
    \end{minipage}
    \begin{minipage}{0.075\linewidth}
        \begin{subfigure}[t]{\textwidth}
        \includegraphics[width=\linewidth]{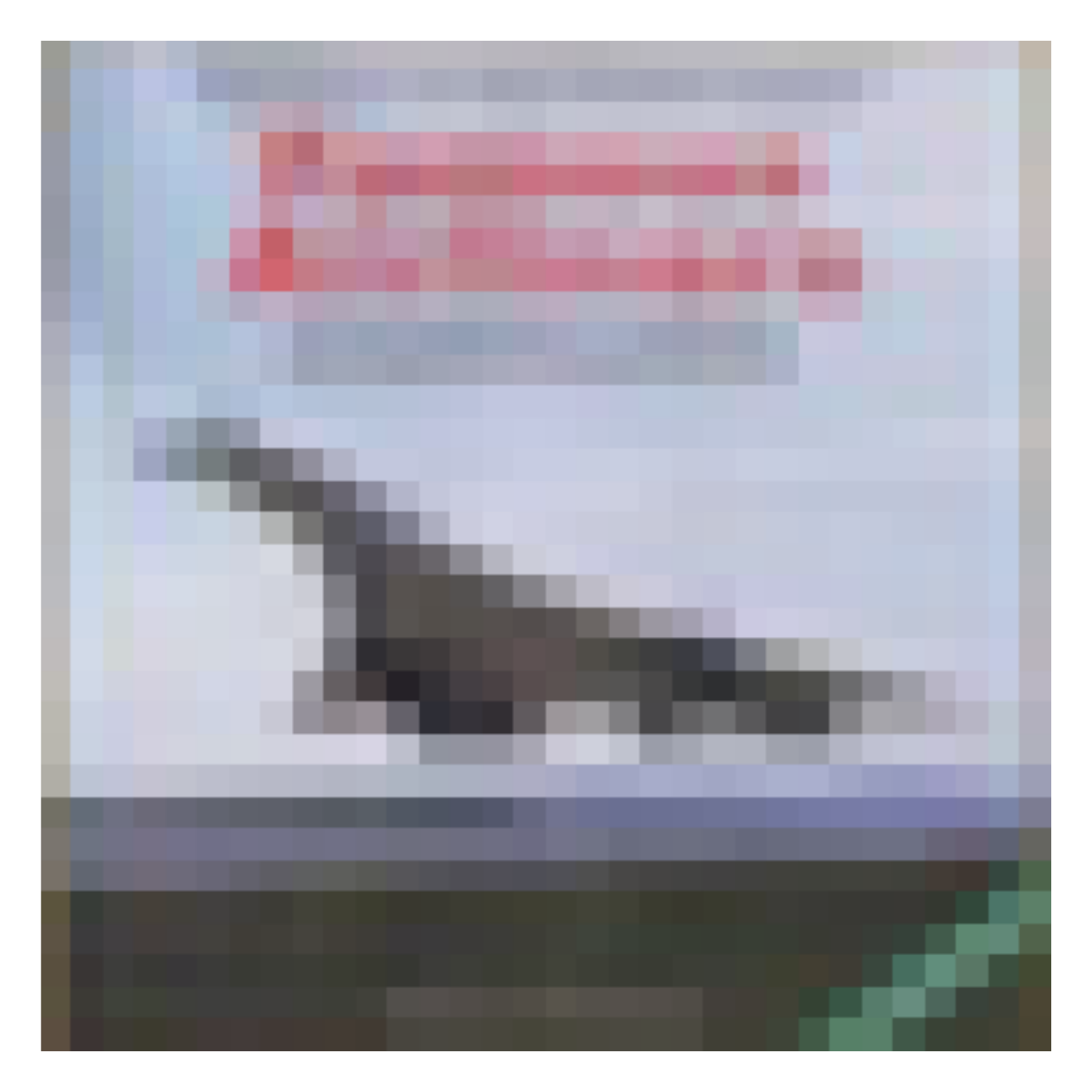}
        \caption{$x$}
        \end{subfigure} \\
        \begin{subfigure}[b]{\textwidth}
        \includegraphics[width=\linewidth]{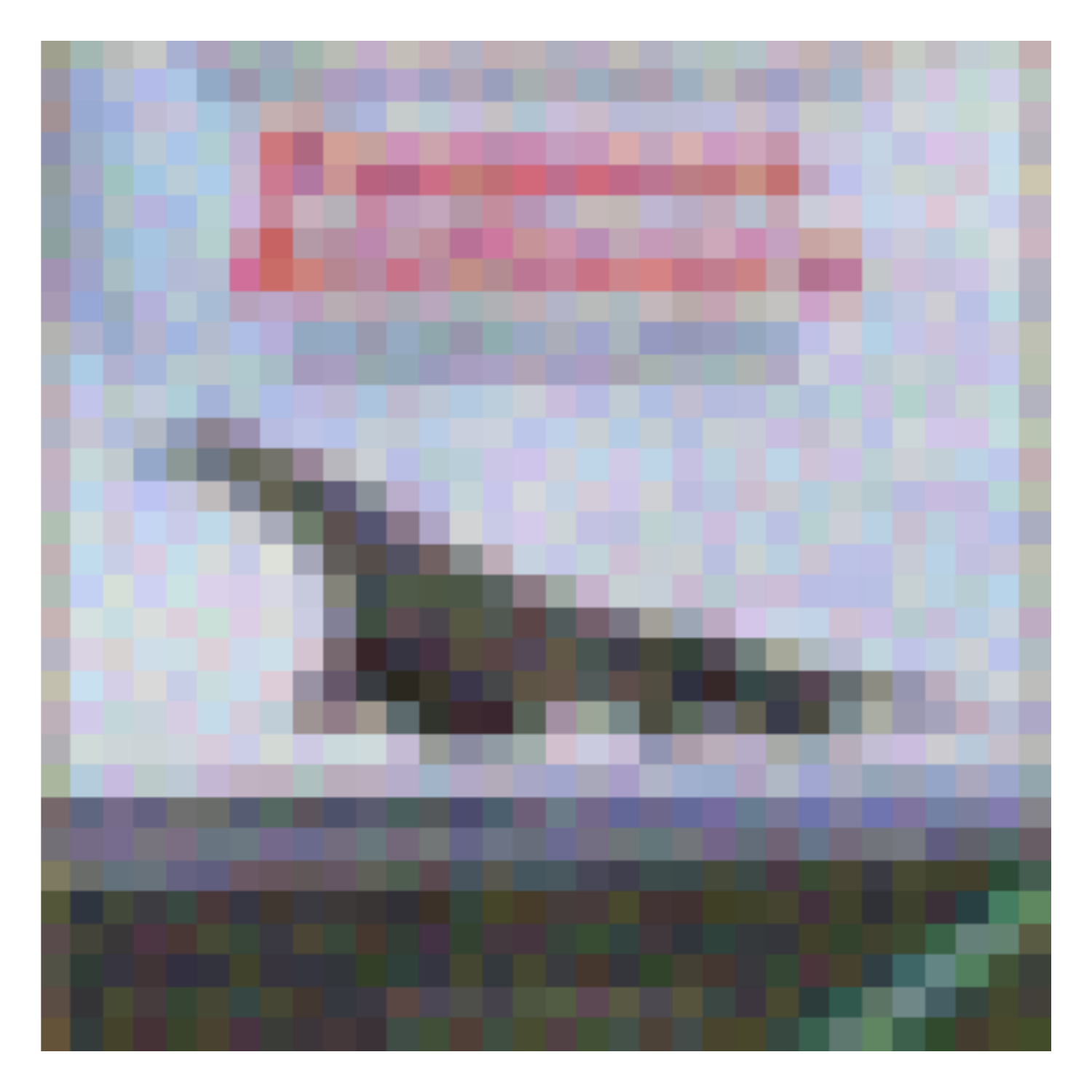}
        \caption{$x'$}
        \end{subfigure}
    \end{minipage}
    \begin{minipage}{0.3009\linewidth}
        \begin{subfigure}{\textwidth}
        \includegraphics[width=\linewidth]{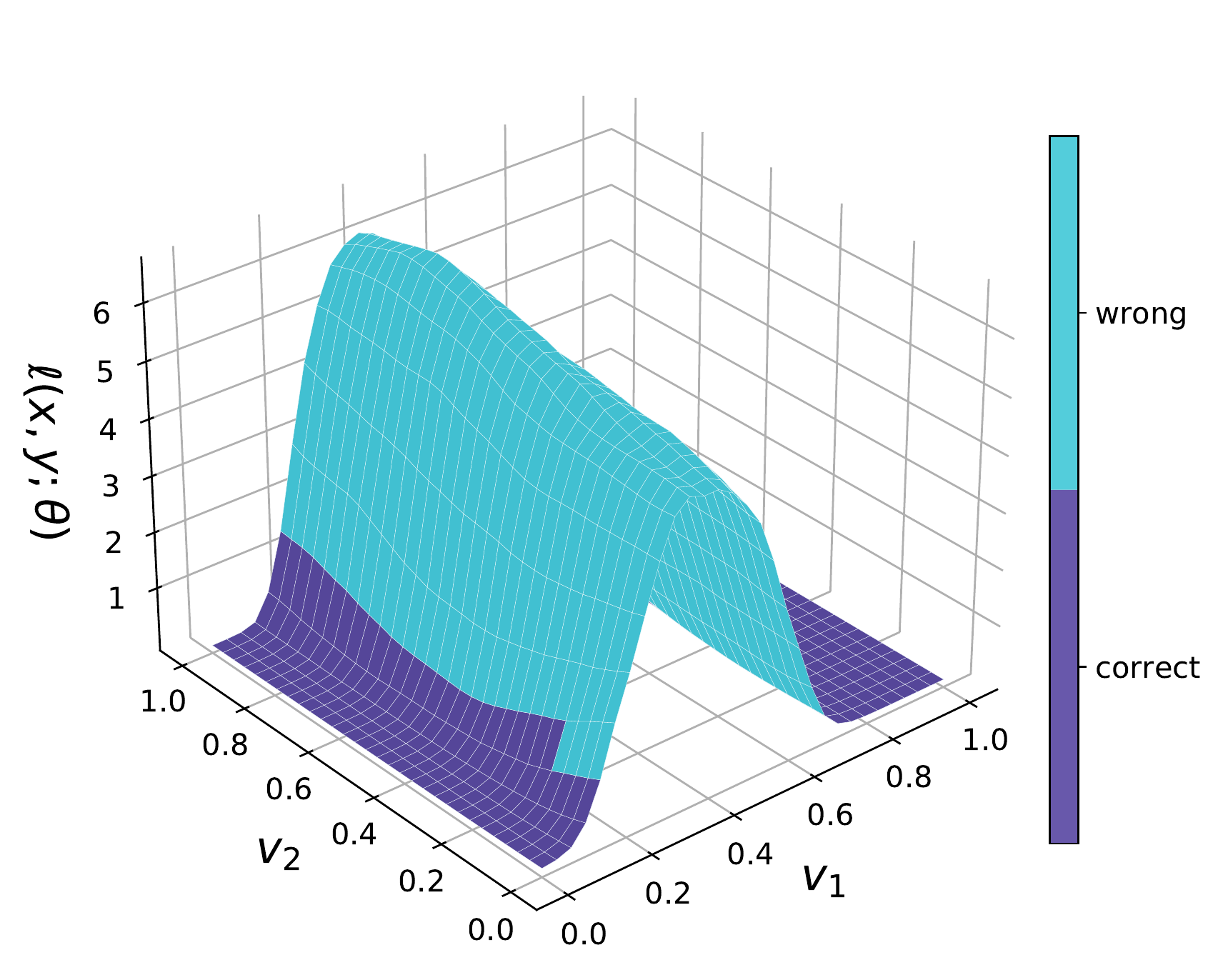}
        \caption{Loss surface}
        \end{subfigure}
    \end{minipage}
    \caption{(CIFAR10) Direction of fast adversarial perturbation $v_1$ and random direction $v_2$. Adversarial example $x'=x+v_1$ is generated from original example $x$.}
    \label{fig:fast_test}
\end{figure*}

\subsection{CIFAR10}
In this section, we aim to provide more figures to describe the relationship between catastrophic overfitting and decision boundary distortion. First, we plot various shapes of distorted decision boundary. We note that it is not difficult to observe distorted decision boundary of the catastrophic overfitted model.

Figure \ref{fig:fgsm_train} shows distorted decision boundary of the fast adversarial trained model on the first four images in the training set of CIFAR10. Although we applied random crop and padding, the loss surface has distorted interval in the direction of their FGSM adversarial perturbation.

In Figure \ref{fig:fast_train}, we plot distorted decision boundary in the direction of perturbation obtained by the attack used in fast adversarial training. Similarly, even though the attack used in fast adversarial training uses a uniform random initialization $U(-\epsilon, \epsilon)$ before adding a sign of gradient, decision boundary distortion is observed for most of the cases. This indicates that single-step adversarial training cannot escape from decision boundary distortion by using a random initialization.

Similar results were found for the test set of CIFAR10 as shown in Figures \ref{fig:fgsm_test} and \ref{fig:fast_test}. Each figure shows the loss surface of the fast adversarial trained model for the first four images in the test set.

\clearpage

\subsection{Tiny ImageNet}

\begin{figure}[t]
\centering
\includegraphics[width=0.9\columnwidth]{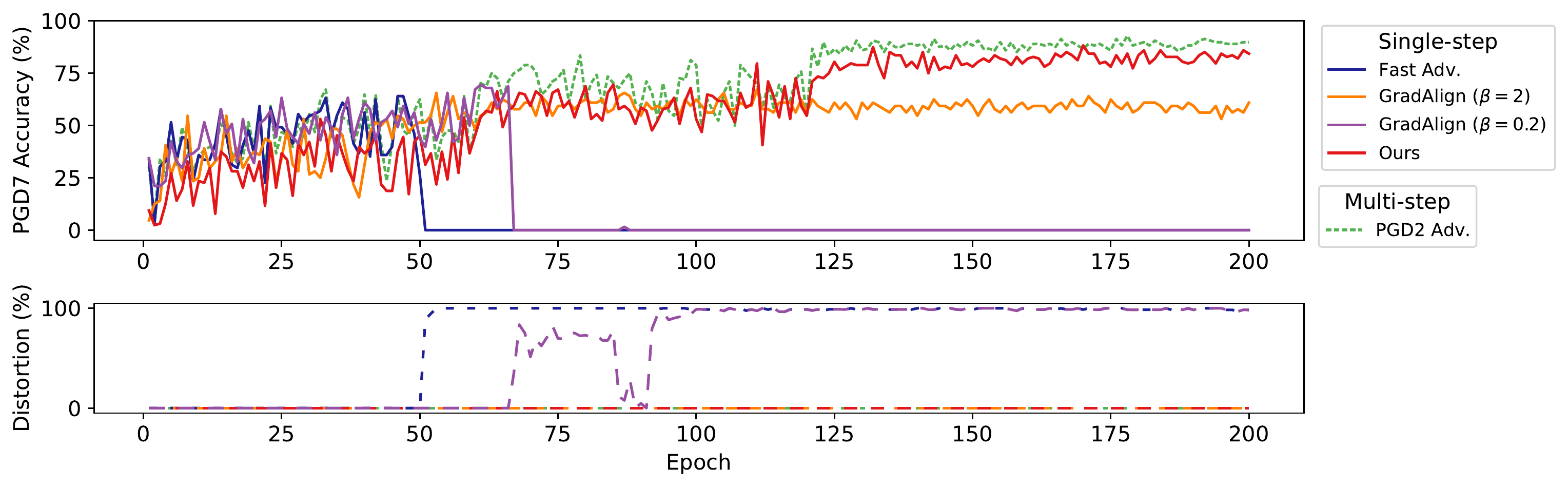}
\caption{(Tiny ImageNet) Robust accuracy and distortion on the training set for each epoch.}
\label{fig:distortion_tiny2}
\end{figure}

To push further, we conducted the same experiment on Tiny ImageNet to observe the relationship between catastrophic overfitting and distorted decision boundary. As shown in Figure \ref{fig:distortion_tiny2}, PGD2 adversarial training shows zero distortion during the training time and achieves nearly 100\% accuracy against PGD7. Among the single-step adversarial training methods, fast adversarial training shows 0\% of robustness against PGD7 with high distortion. The proposed method avoids catastrophic overfitting and achieves a high PGD7 accuracy, similar to multi-step adversarial training. 

\begin{figure*}[t]
    \centering
    \begin{minipage}{0.075\linewidth}
        \begin{subfigure}[t]{\textwidth}
        \includegraphics[width=\linewidth]{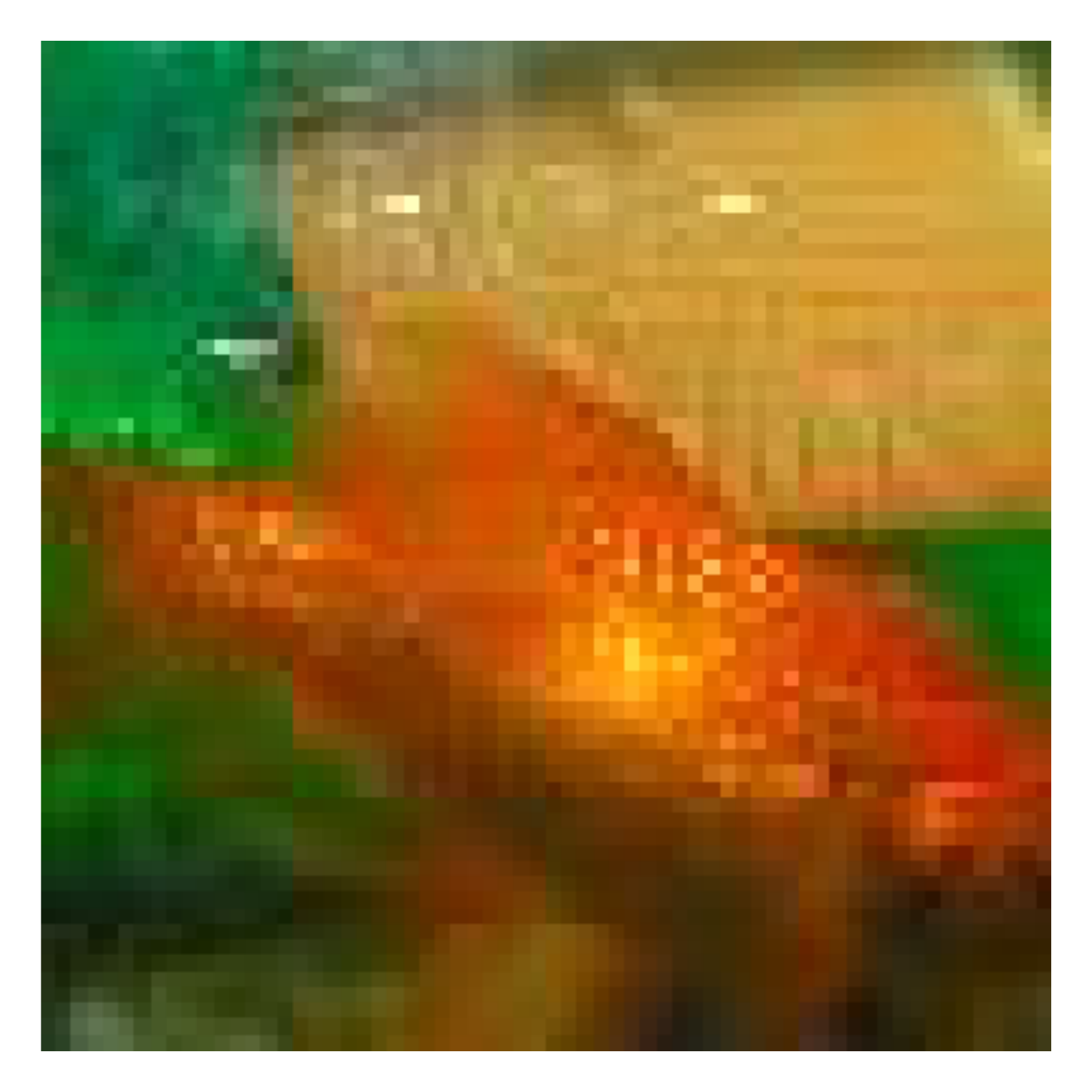}
        \caption{$x$}
        \end{subfigure} \\
        \begin{subfigure}[b]{\textwidth}
        \includegraphics[width=\linewidth]{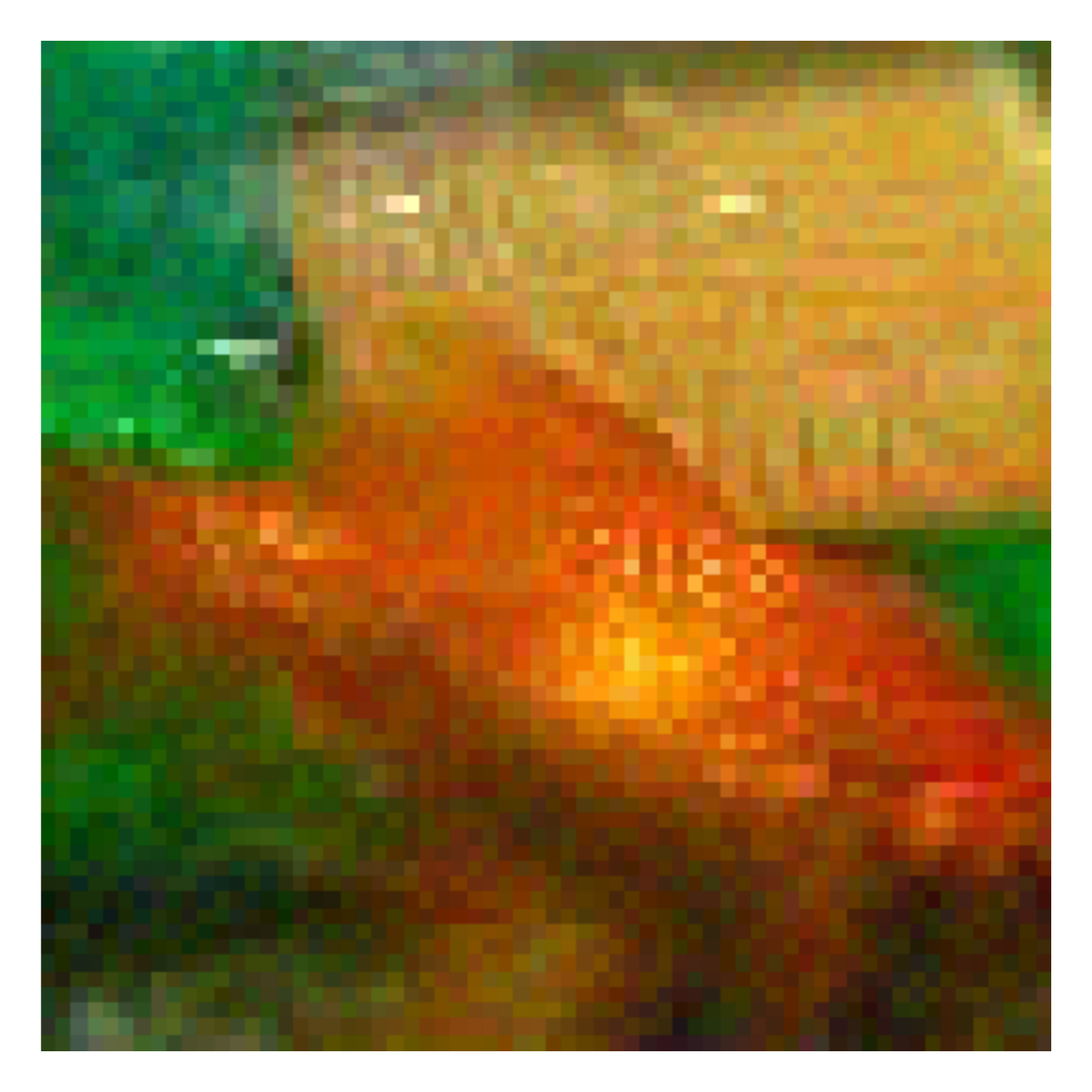}
        \caption{$x'$}
        \end{subfigure}
    \end{minipage}
    \begin{minipage}{0.3009\linewidth}
        \begin{subfigure}{\textwidth}
        \includegraphics[width=\linewidth]{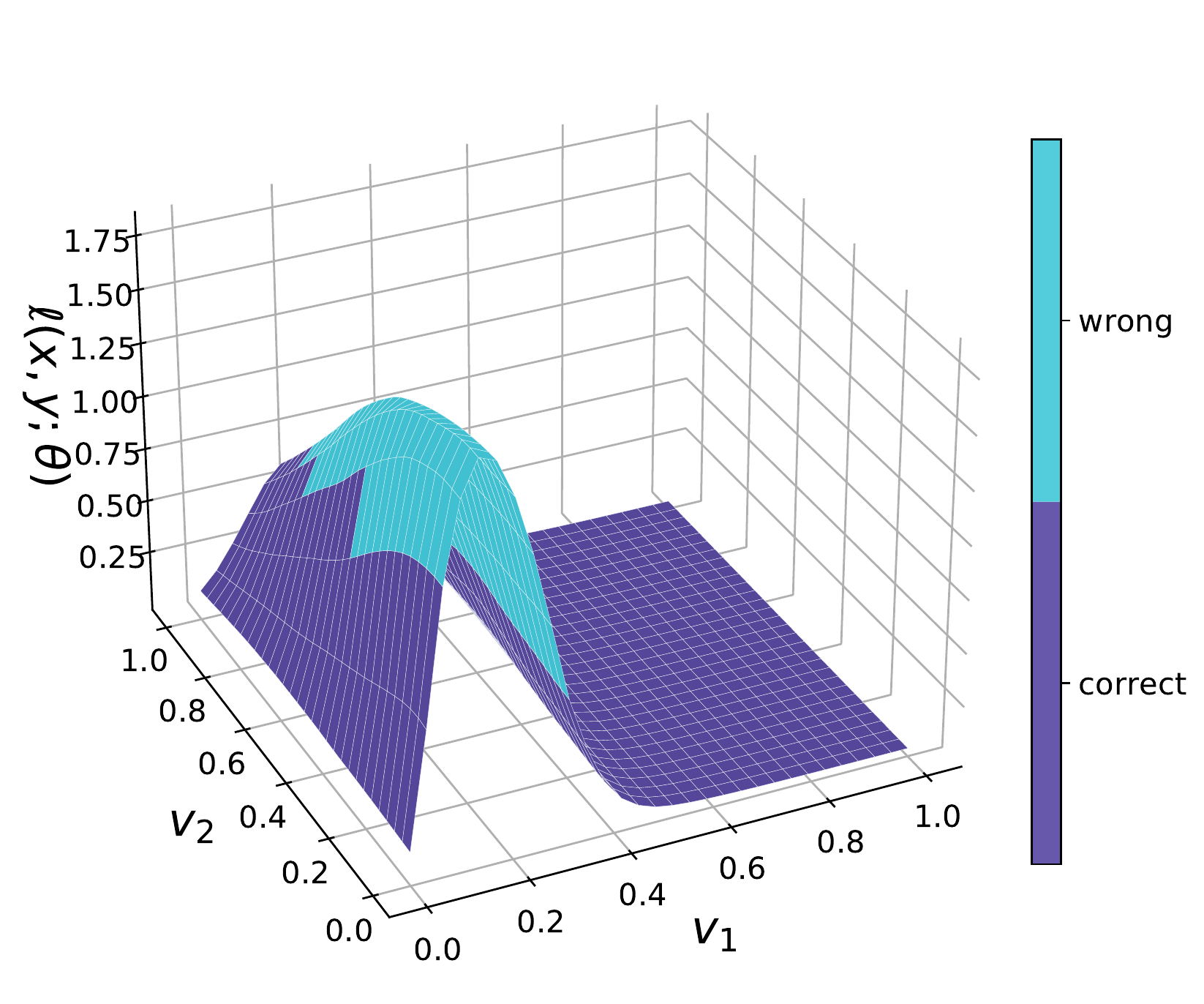}
        \caption{Loss surface (67th epoch)}
        \end{subfigure}
    \end{minipage}
    \begin{minipage}{0.3009\linewidth}
        \begin{subfigure}{\textwidth}
        \includegraphics[width=\linewidth]{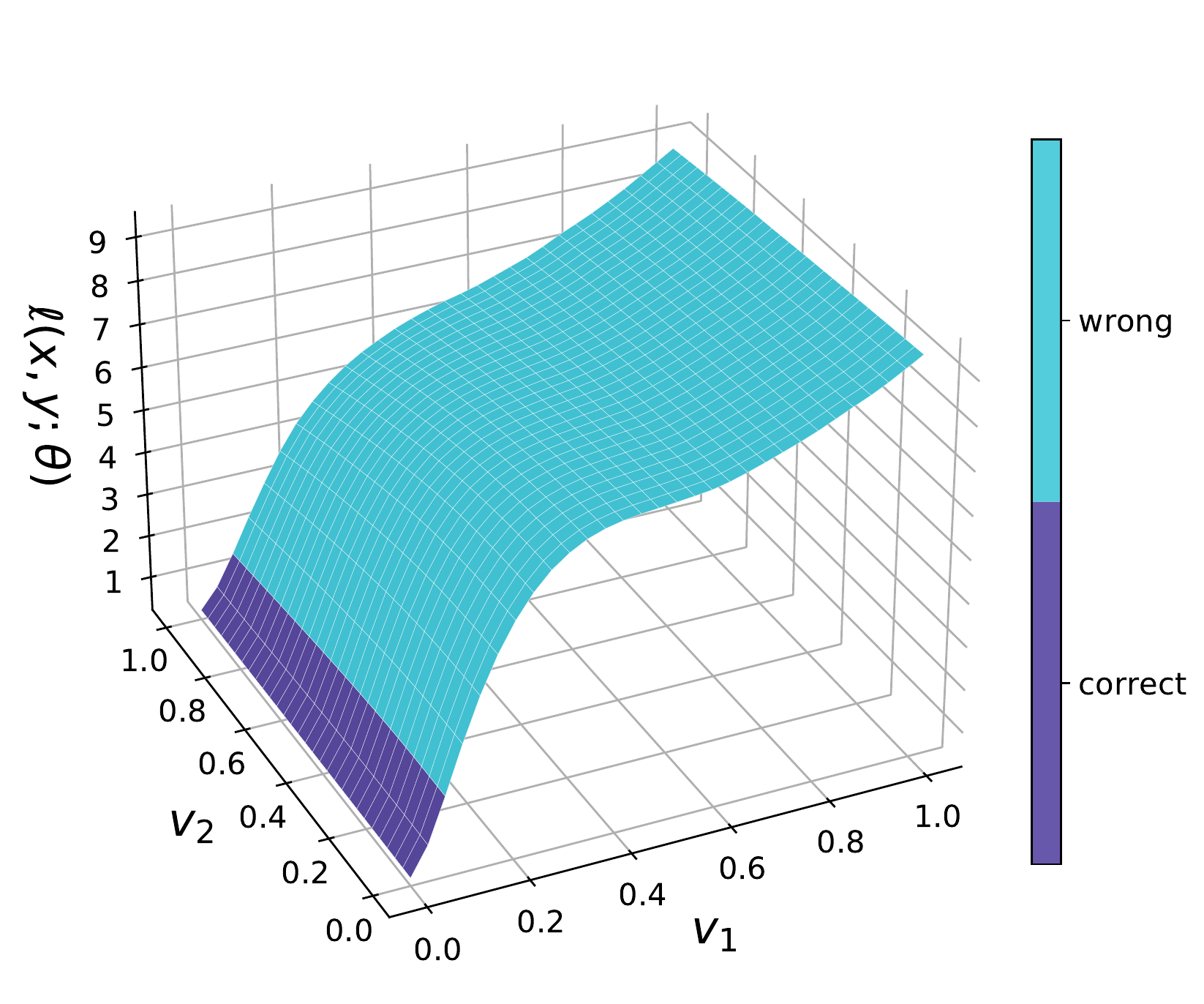}
        \caption{Loss surface (89th epoch)}
        \end{subfigure}
    \end{minipage} \\
    \caption{(Tiny ImageNet) Direction of FGSM adversarial perturbation $v_1$ and random direction $v_2$. Adversarial example $x'=x+v_1$ is generated from original example $x$.}
    \label{fig:grad_tiny}
\end{figure*}

Noteworthy is the fact that GradAlign shows unusual behavior for different regularization parameter $\beta$ for the gradient alignment.
First, contrary to CIFAR10, catastrophic overfitting occurs in GradAlign with $\beta=0.2$ at the 67th epoch on Tiny ImageNet. At 89th epoch, distortion of GradAlign decreases to 0\% even when the accuracy against PGD7 is nearly zero. To investigate this result, we plot the loss surface of GradAlign for each epoch. As in Figure \ref{fig:grad_tiny}, GradAlign with $\beta=0.2$ shows distorted decision boundary at the 67th epoch. However, at the 89th epoch, although gradient alignment works properly after catastrophic overfitting, the model cannot escape from the local minima and thus PGD7 accuracy is still zero.

Considering that the successful value of $\beta$ was different for different maximum perturbation $\epsilon$ as \citet{andriushchenko2020understanding} noted, we also conducted the same experiment with different $\beta$ because Tiny ImageNet is a more complicated dataset than CIFAR10. We additionally use $\beta=2$ and $0.02$. When we use $\beta=2$, GradAlign shows zero-distortion until the end of the training. However, robustness against PGD7 on the training batch cannot achieve near 100\% unlike PGD adversarial training and the proposed method. We think this is an interesting phenomenon that might help the community interested to continue further research in better schemes for single-step adversarial training. Note that the result of GradAlign with $\beta=0.02$ was similar to $\beta=0.2$. 

\begin{figure*}[p]
    \centering
    \begin{minipage}{0.075\linewidth}
        \begin{subfigure}[t]{\textwidth}
        \includegraphics[width=\linewidth]{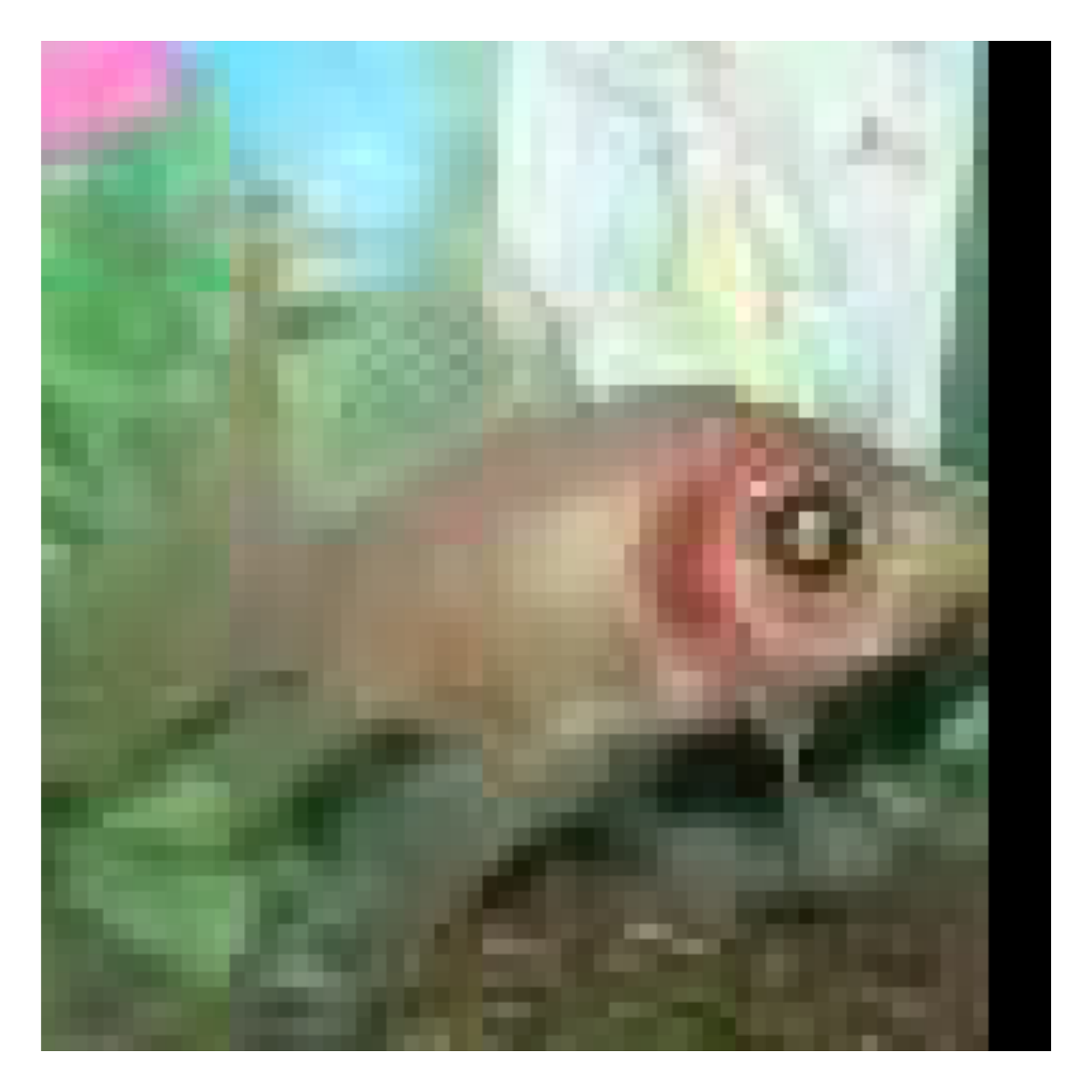}
        \caption{$x$}
        \end{subfigure} \\
        \begin{subfigure}[b]{\textwidth}
        \includegraphics[width=\linewidth]{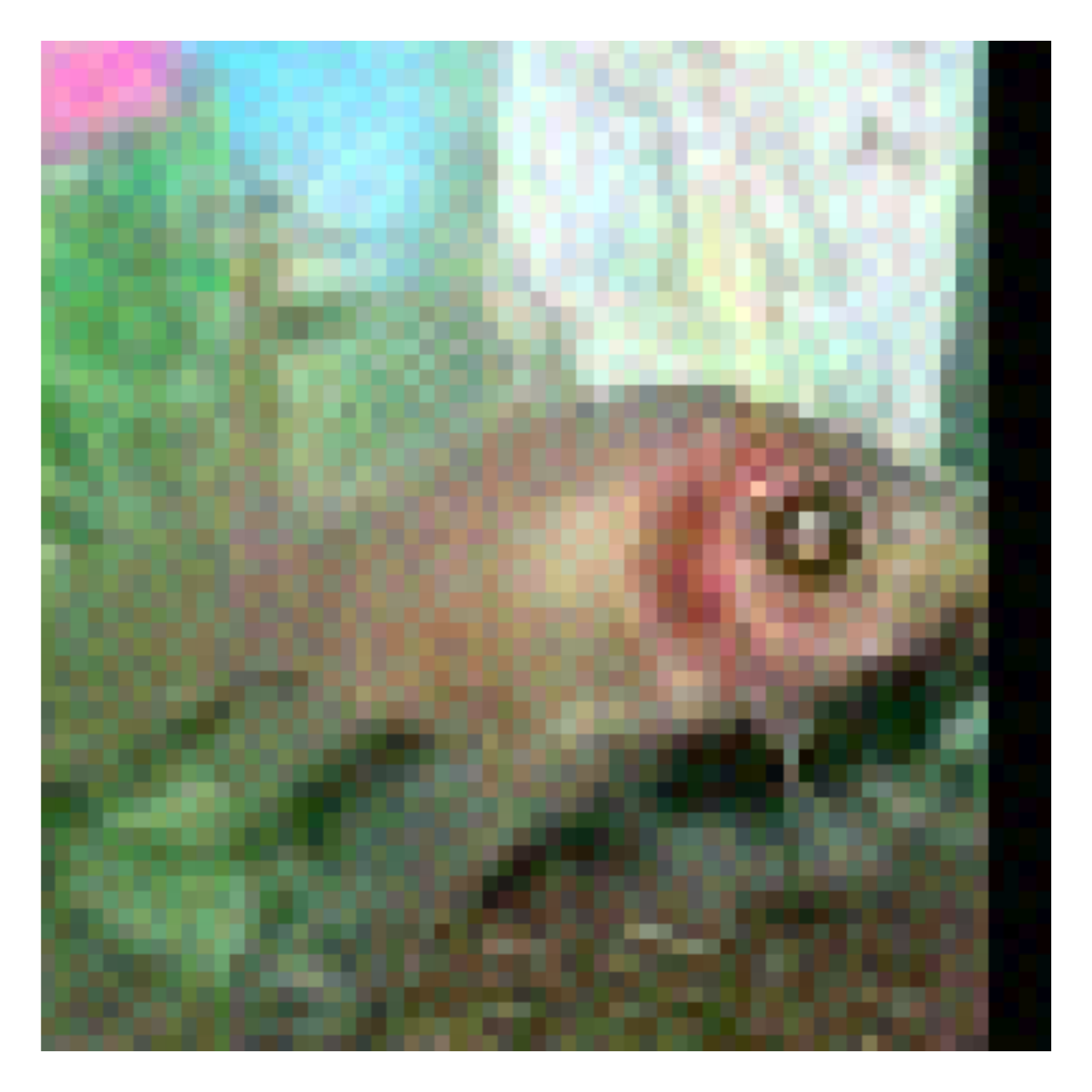}
        \caption{$x'$}
        \end{subfigure}
    \end{minipage}
    \begin{minipage}{0.3009\linewidth}
        \begin{subfigure}{\textwidth}
        \includegraphics[width=\linewidth]{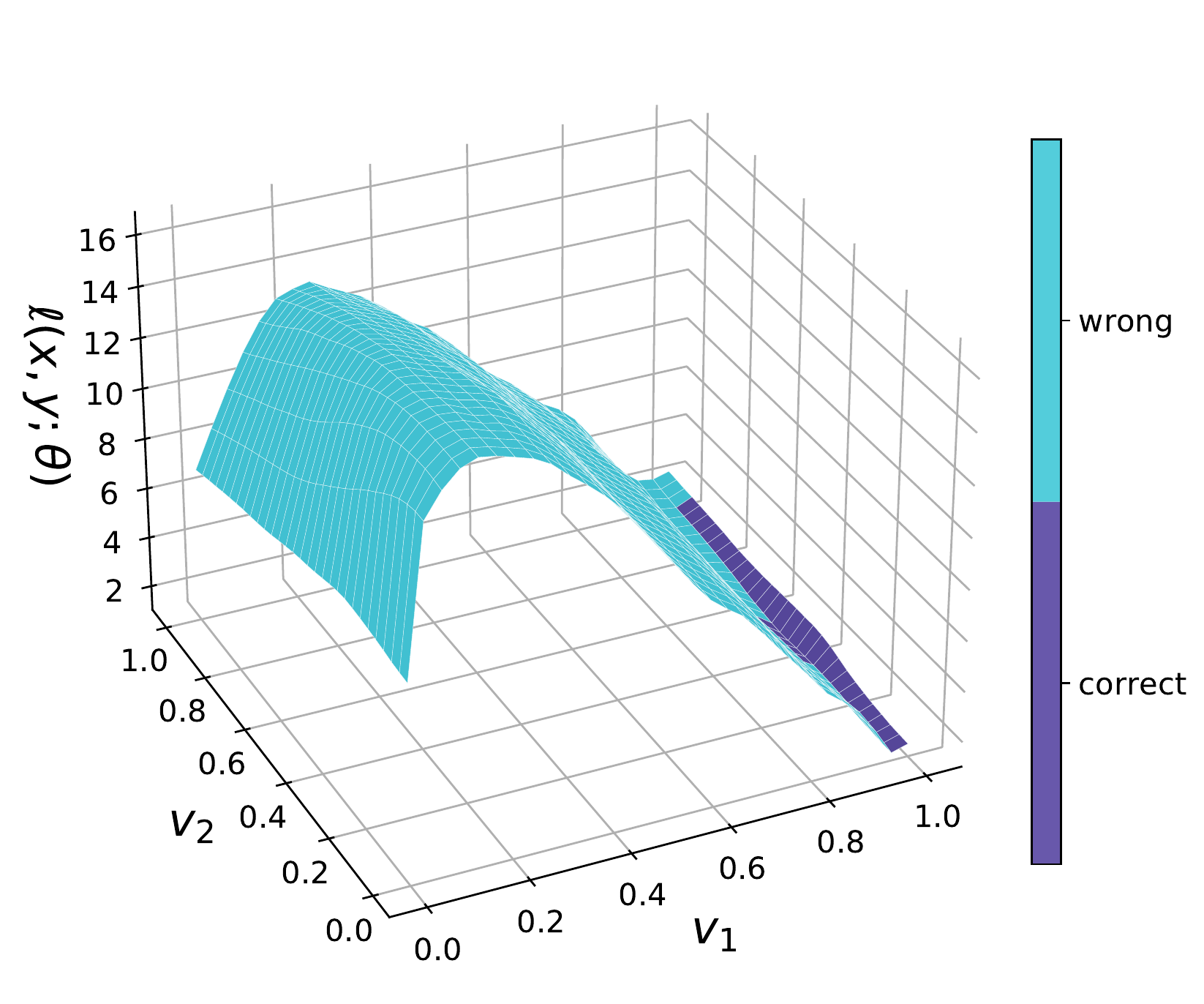}
        \caption{Loss surface}
        \end{subfigure}
    \end{minipage}
    \begin{minipage}{0.075\linewidth}
        \begin{subfigure}[t]{\textwidth}
        \includegraphics[width=\linewidth]{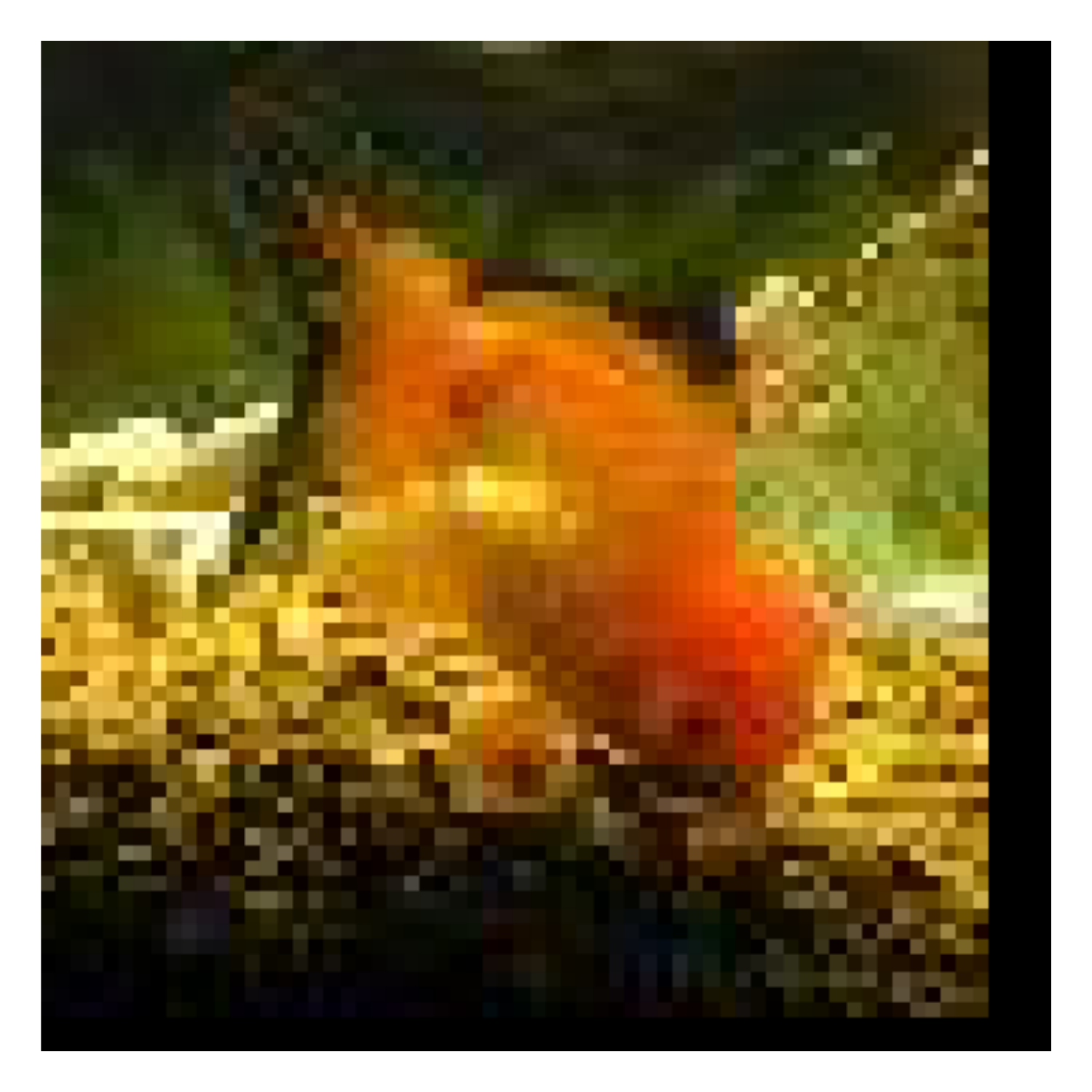}
        \caption{$x$}
        \end{subfigure} \\
        \begin{subfigure}[b]{\textwidth}
        \includegraphics[width=\linewidth]{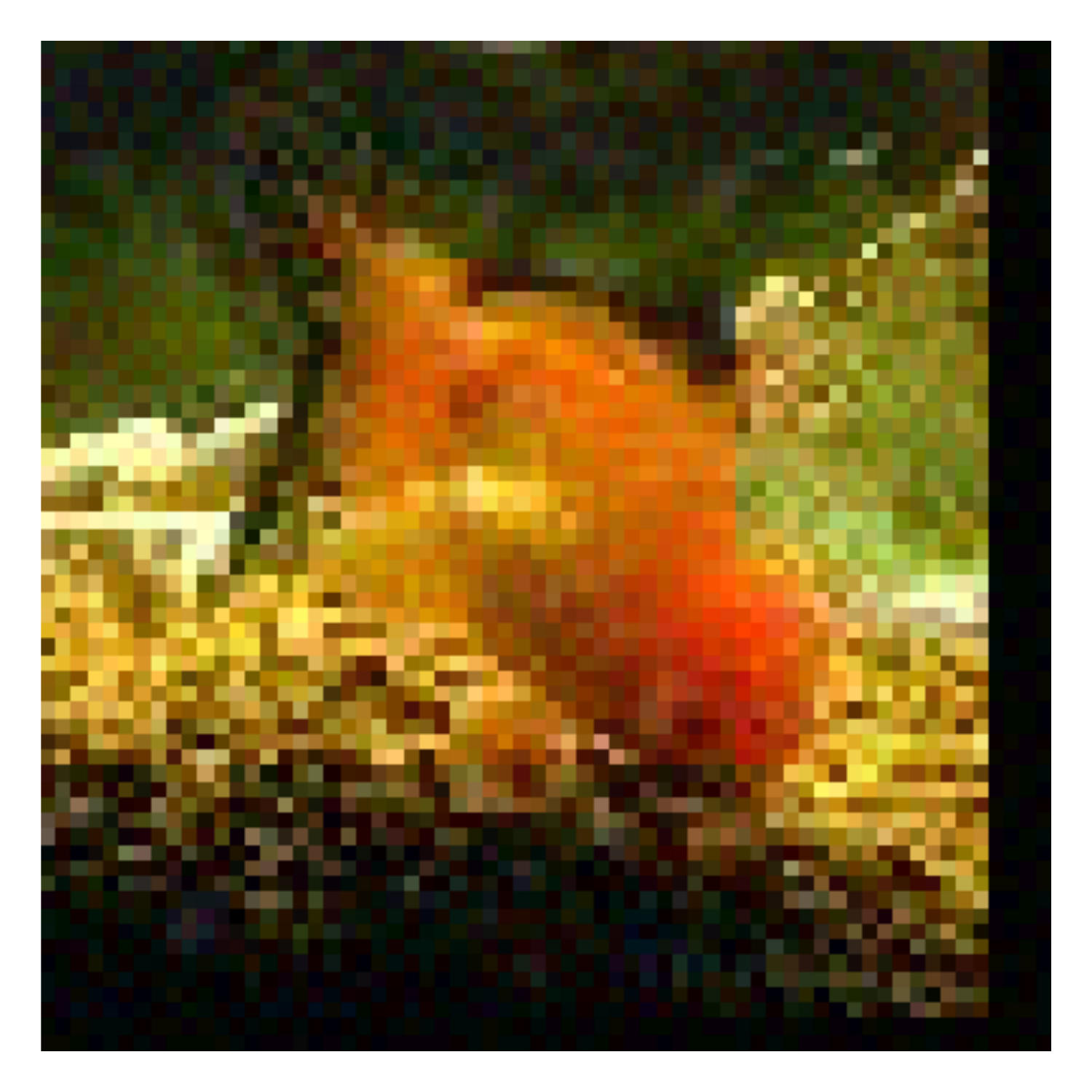}
        \caption{$x'$}
        \end{subfigure}
    \end{minipage}
    \begin{minipage}{0.3009\linewidth}
        \begin{subfigure}{\textwidth}
        \includegraphics[width=\linewidth]{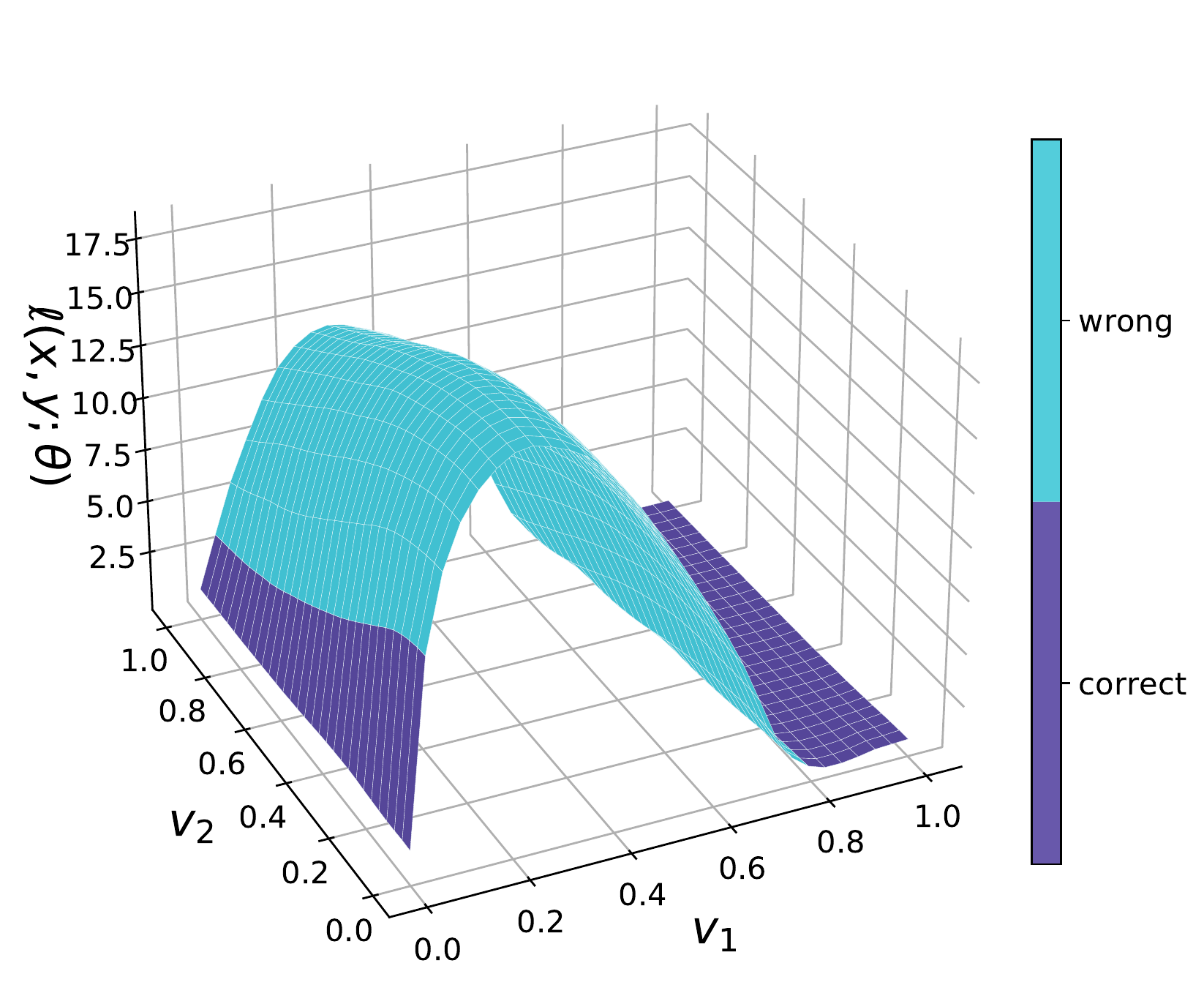}
        \caption{Loss surface}
        \end{subfigure}
    \end{minipage} \\

    \begin{minipage}{0.075\linewidth}
        \begin{subfigure}[t]{\textwidth}
        \includegraphics[width=\linewidth]{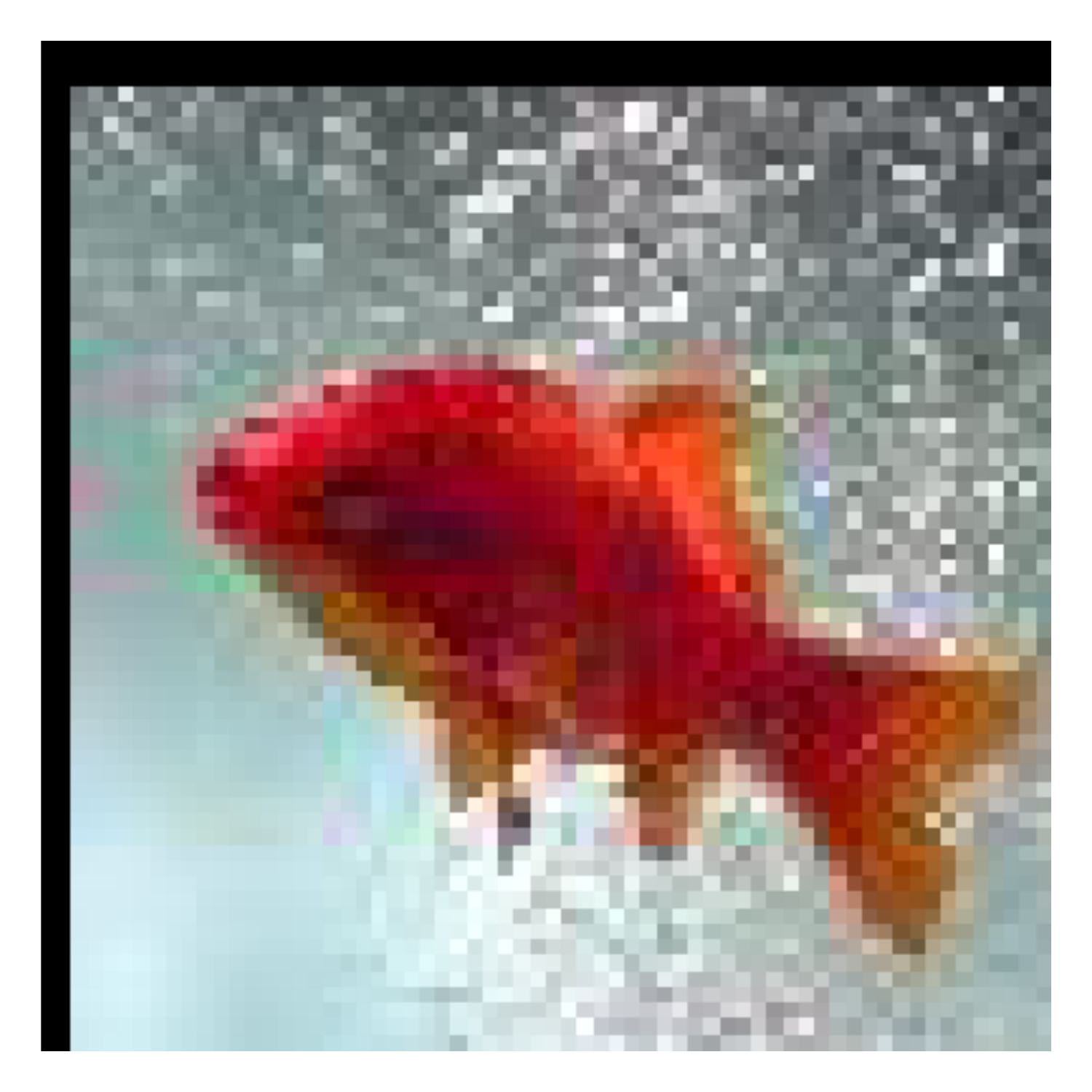}
        \caption{$x$}
        \end{subfigure} \\
        \begin{subfigure}[b]{\textwidth}
        \includegraphics[width=\linewidth]{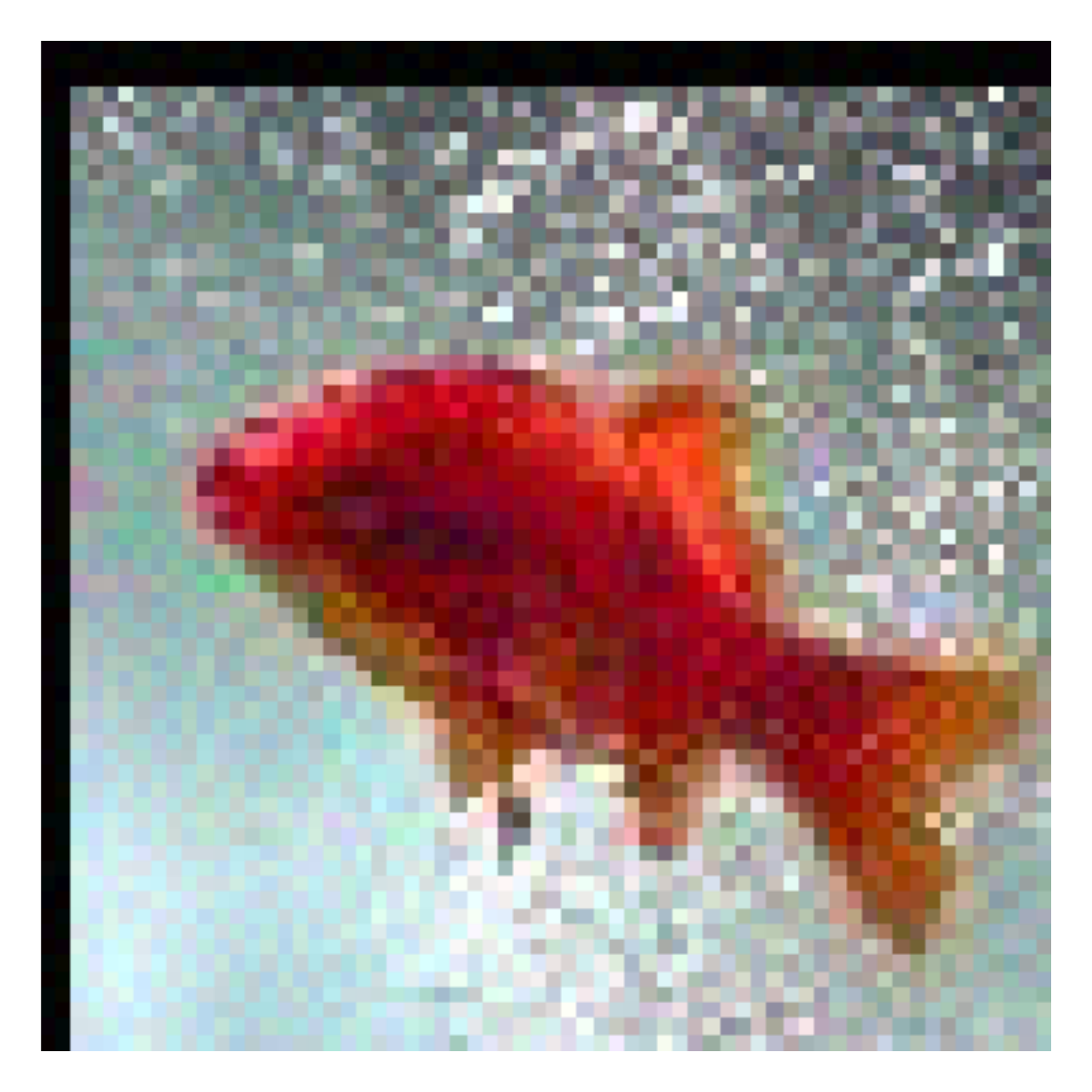}
        \caption{$x'$}
        \end{subfigure}
    \end{minipage}
    \begin{minipage}{0.3009\linewidth}
        \begin{subfigure}{\textwidth}
        \includegraphics[width=\linewidth]{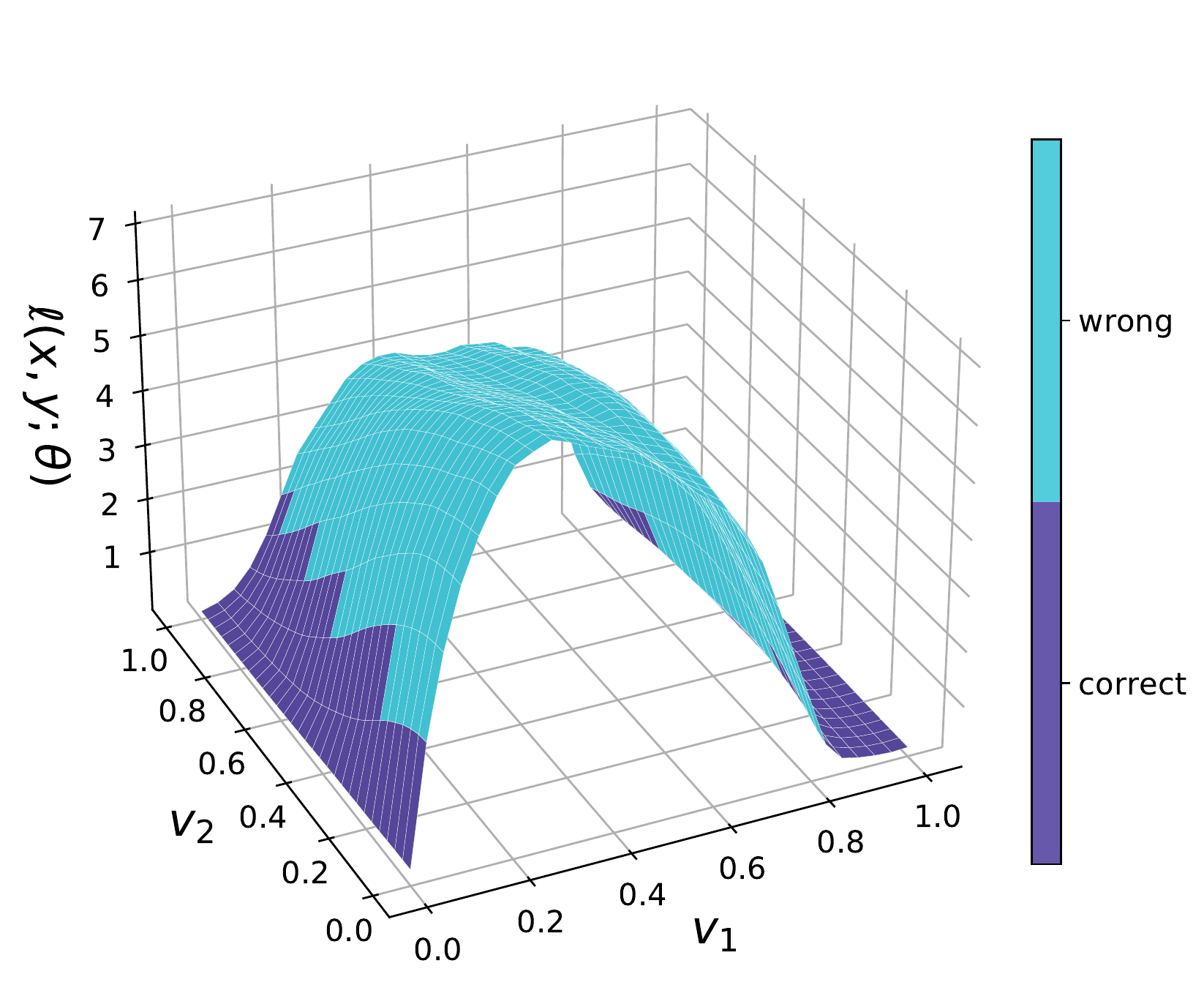}
        \caption{Loss surface}
        \end{subfigure}
    \end{minipage}
    \begin{minipage}{0.075\linewidth}
        \begin{subfigure}[t]{\textwidth}
        \includegraphics[width=\linewidth]{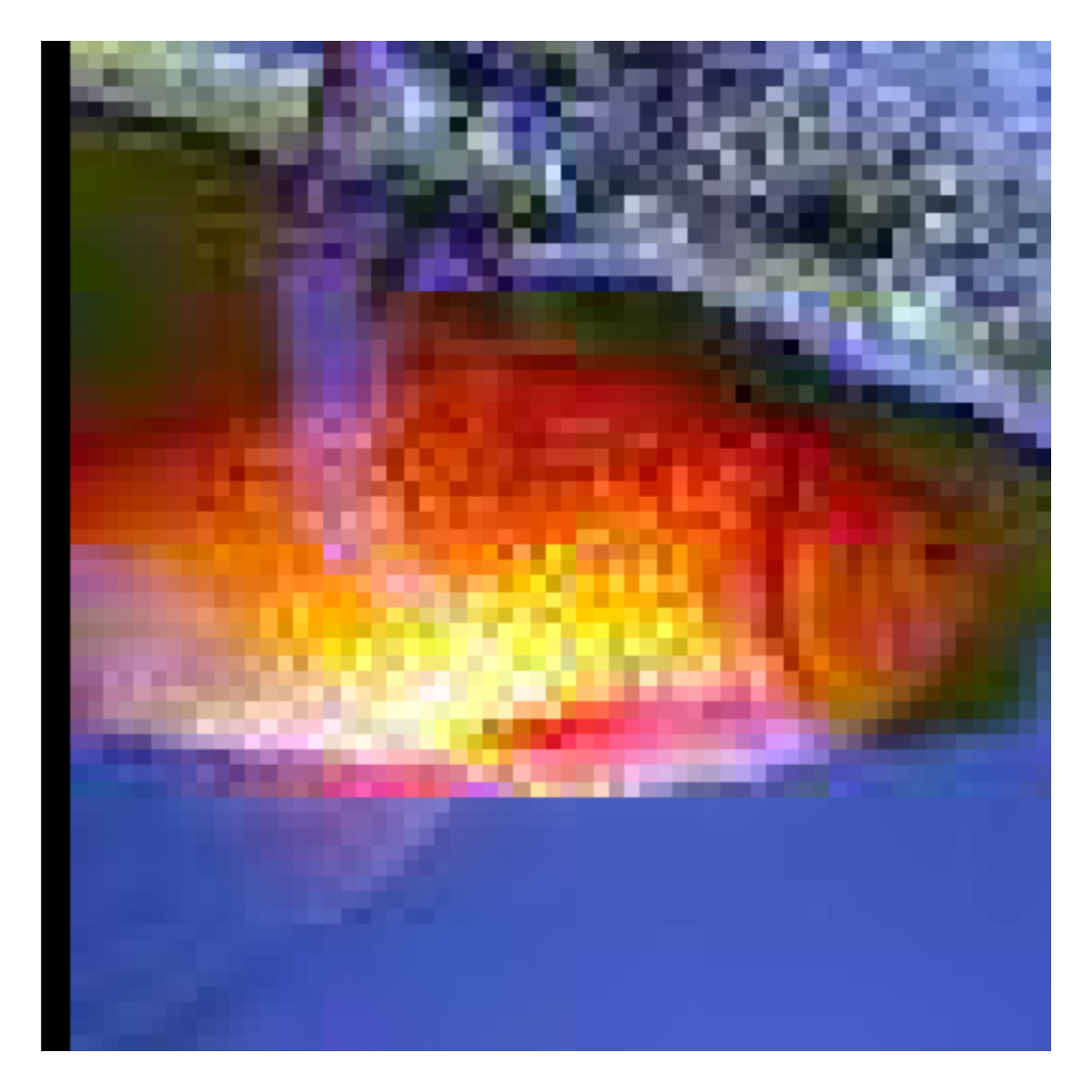}
        \caption{$x$}
        \end{subfigure} \\
        \begin{subfigure}[b]{\textwidth}
        \includegraphics[width=\linewidth]{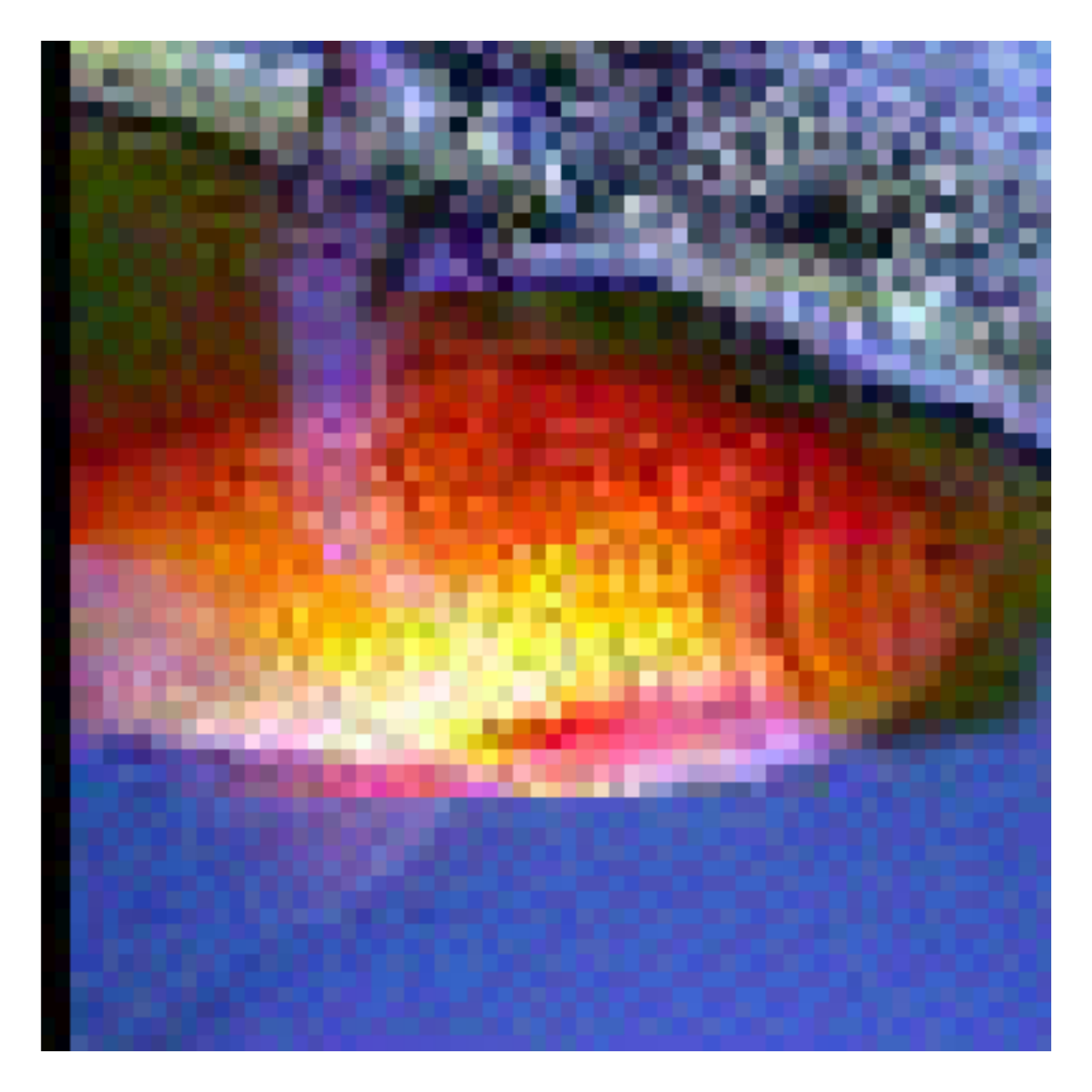}
        \caption{$x'$}
        \end{subfigure}
    \end{minipage}
    \begin{minipage}{0.3009\linewidth}
        \begin{subfigure}{\textwidth}
        \includegraphics[width=\linewidth]{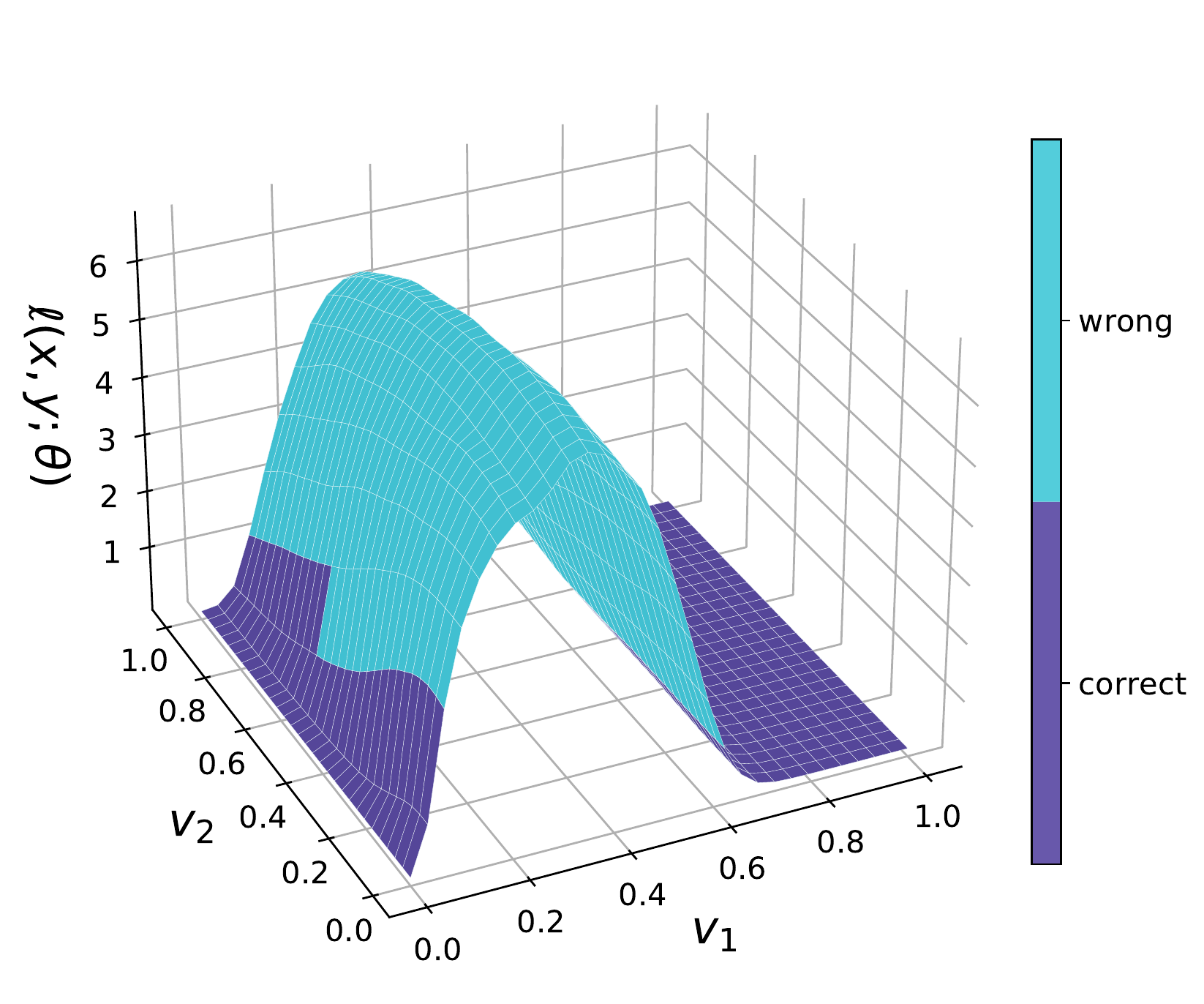}
        \caption{Loss surface}
        \end{subfigure}
    \end{minipage}
    \caption{(Tiny ImageNet) Direction of FGSM adversarial perturbation $v_1$ and random direction $v_2$. Adversarial example $x'=x+v_1$ is generated from original example $x$.}
    \label{fig:fgsm_train_tiny}
\end{figure*}

\begin{figure*}[p]
    \centering
    \begin{minipage}{0.075\linewidth}
        \begin{subfigure}[t]{\textwidth}
        \includegraphics[width=\linewidth]{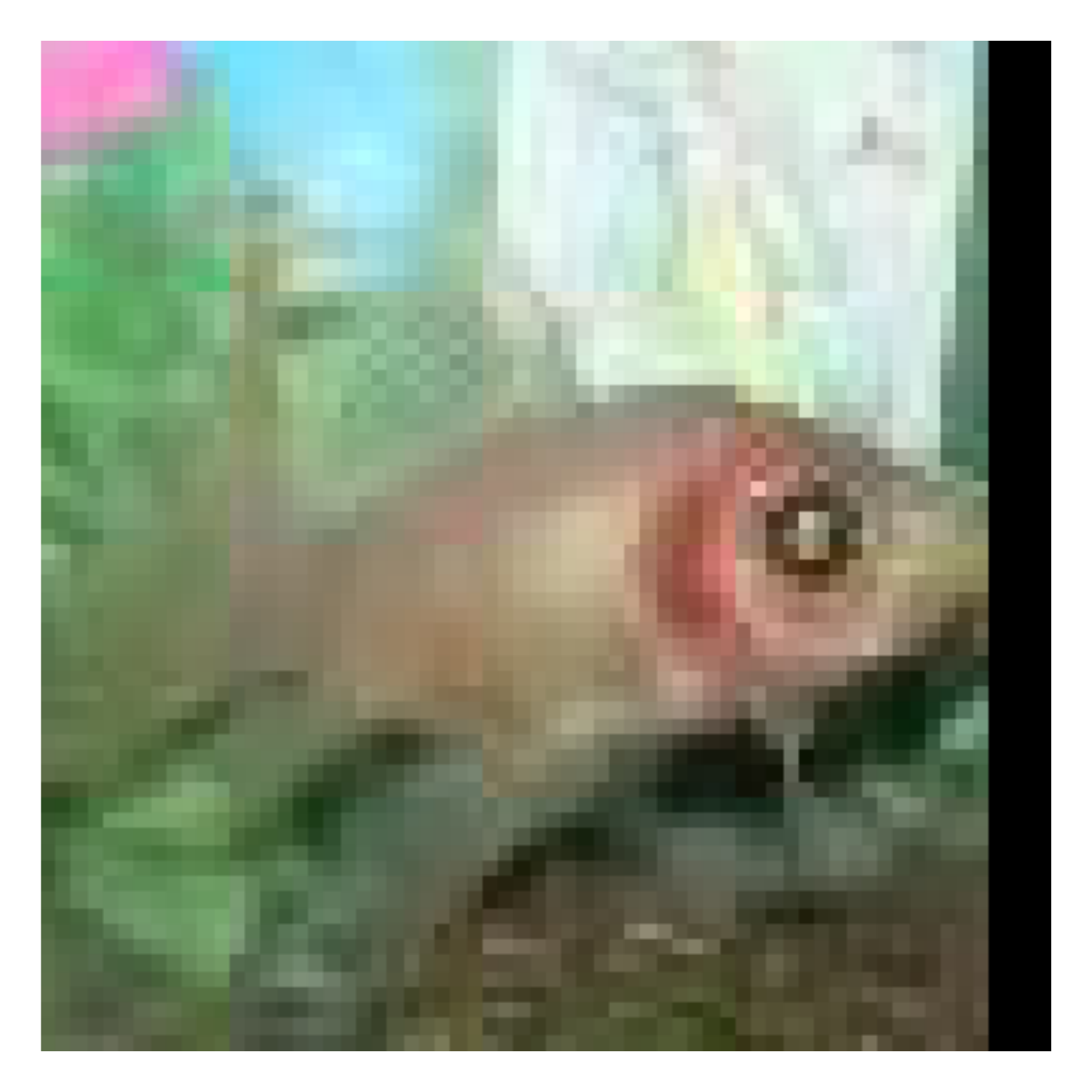}
        \caption{$x$}
        \end{subfigure} \\
        \begin{subfigure}[b]{\textwidth}
        \includegraphics[width=\linewidth]{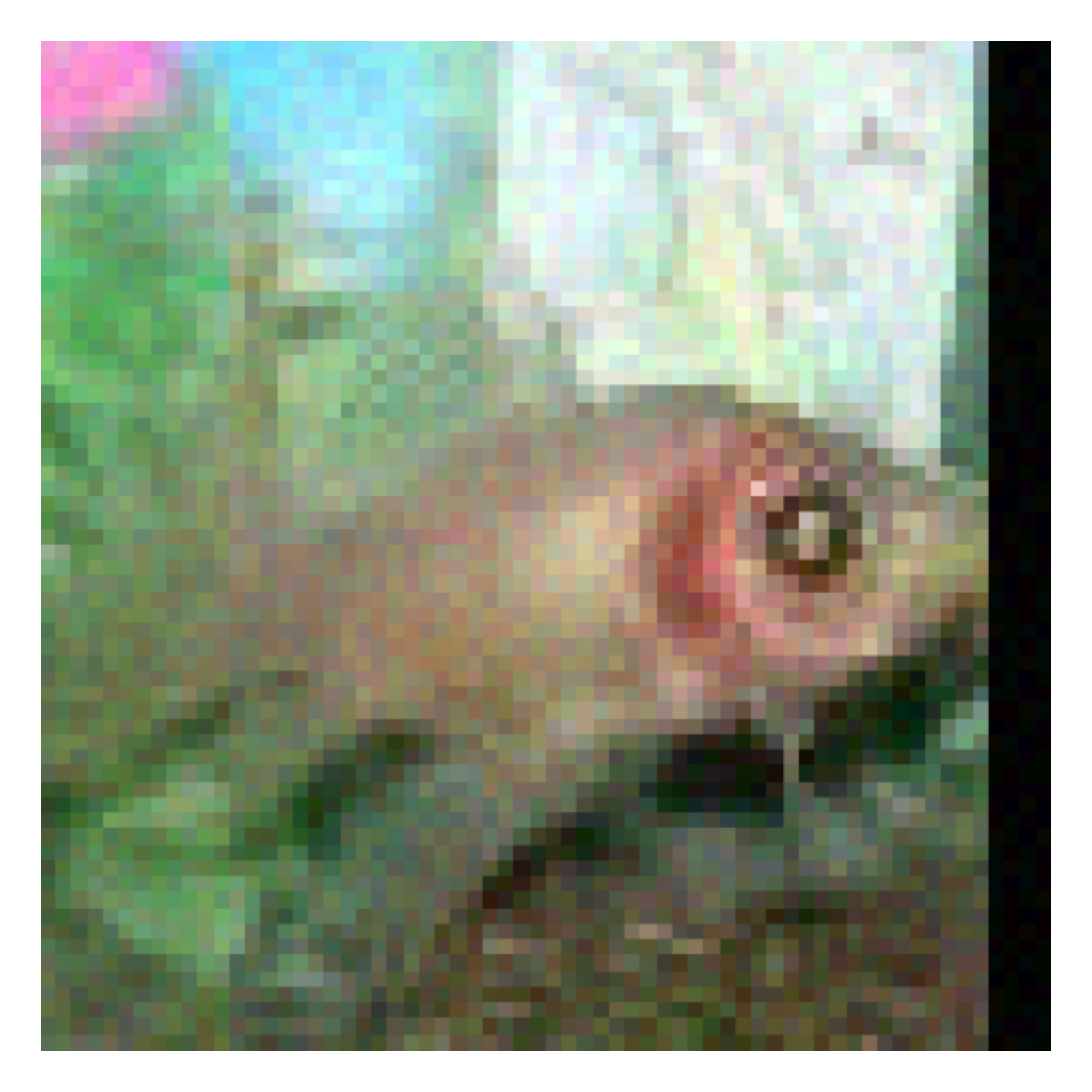}
        \caption{$x'$}
        \end{subfigure}
    \end{minipage}
    \begin{minipage}{0.3009\linewidth}
        \begin{subfigure}{\textwidth}
        \includegraphics[width=\linewidth]{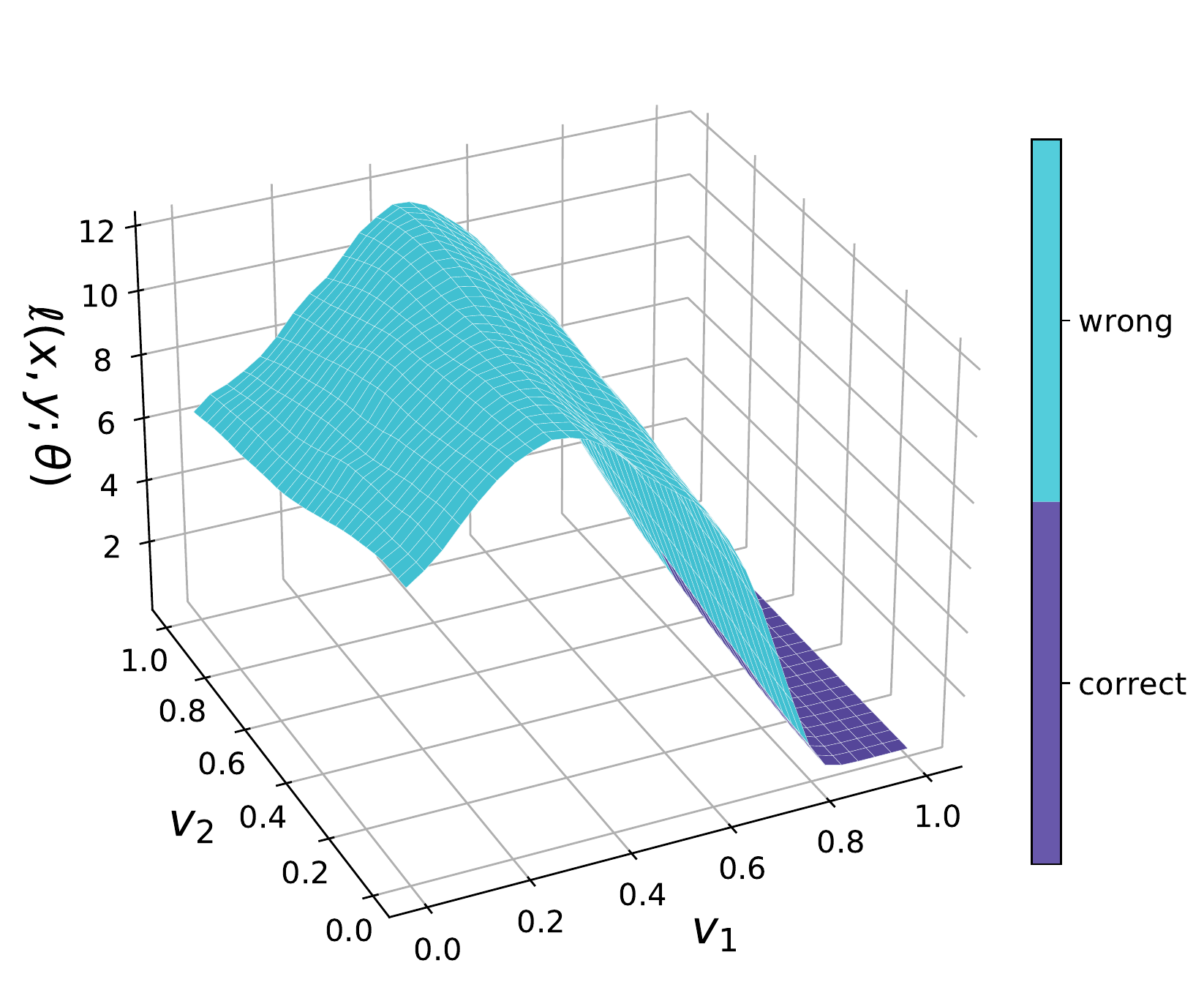}
        \caption{Loss surface}
        \end{subfigure}
    \end{minipage}
    \begin{minipage}{0.075\linewidth}
        \begin{subfigure}[t]{\textwidth}
        \includegraphics[width=\linewidth]{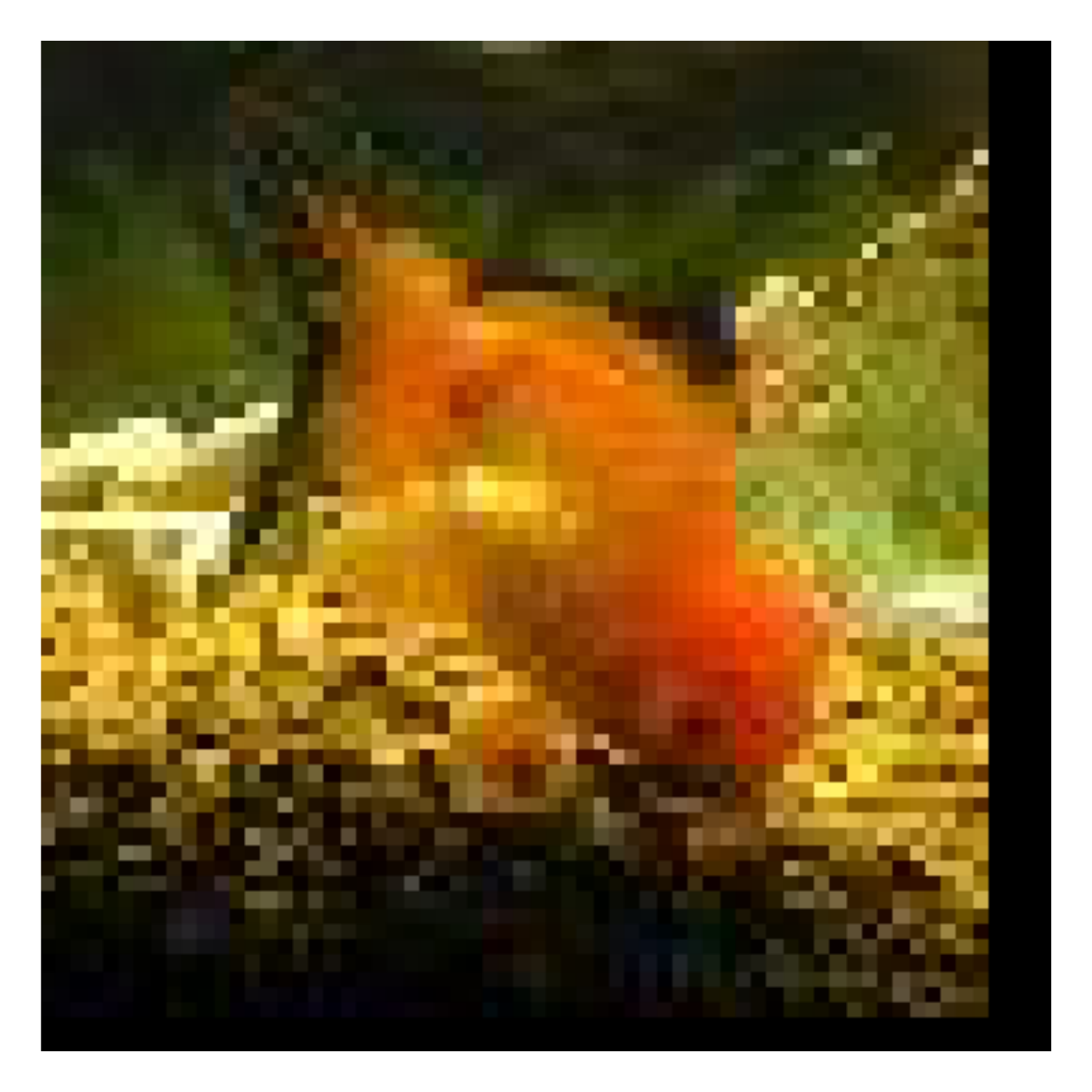}
        \caption{$x$}
        \end{subfigure} \\
        \begin{subfigure}[b]{\textwidth}
        \includegraphics[width=\linewidth]{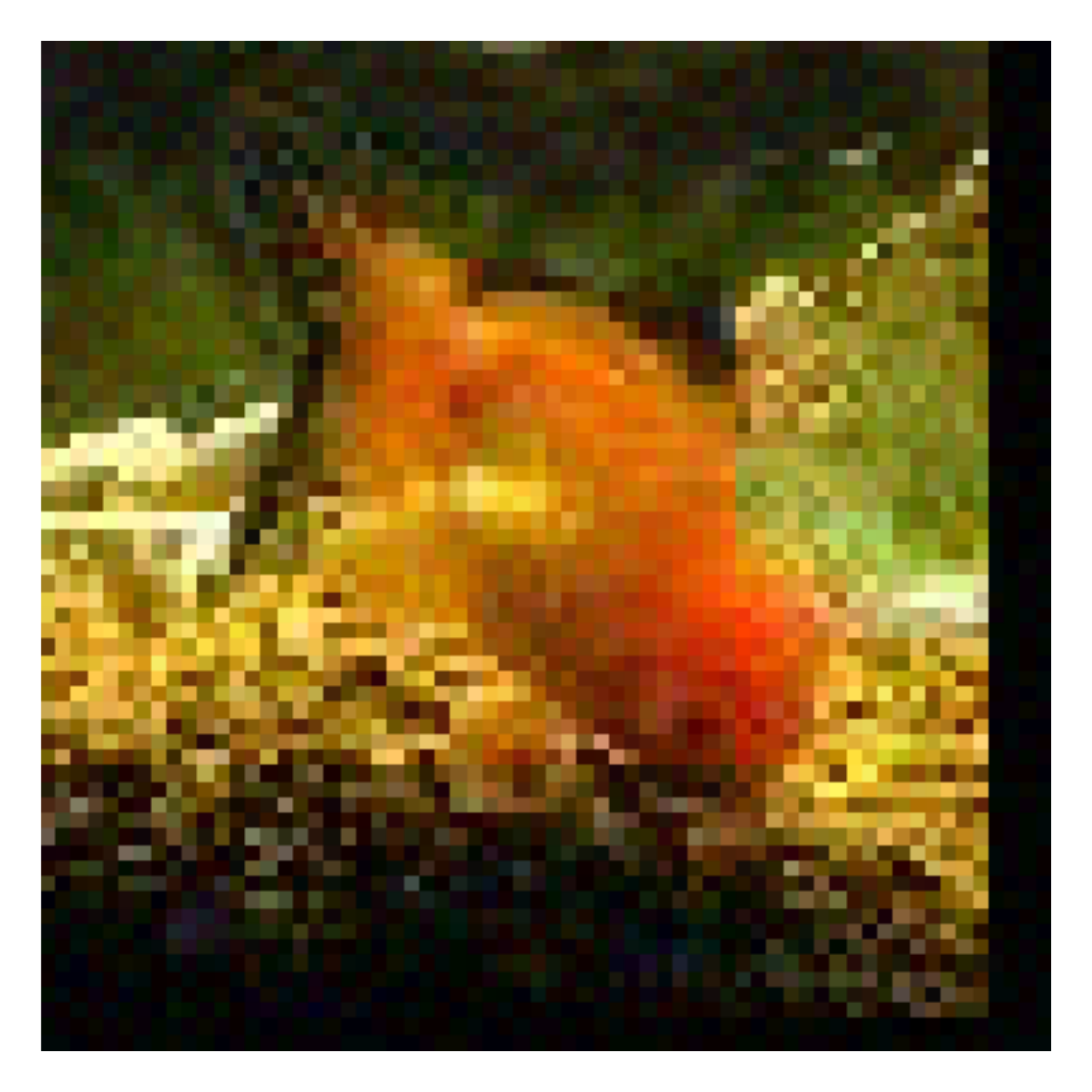}
        \caption{$x'$}
        \end{subfigure}
    \end{minipage}
    \begin{minipage}{0.3009\linewidth}
        \begin{subfigure}{\textwidth}
        \includegraphics[width=\linewidth]{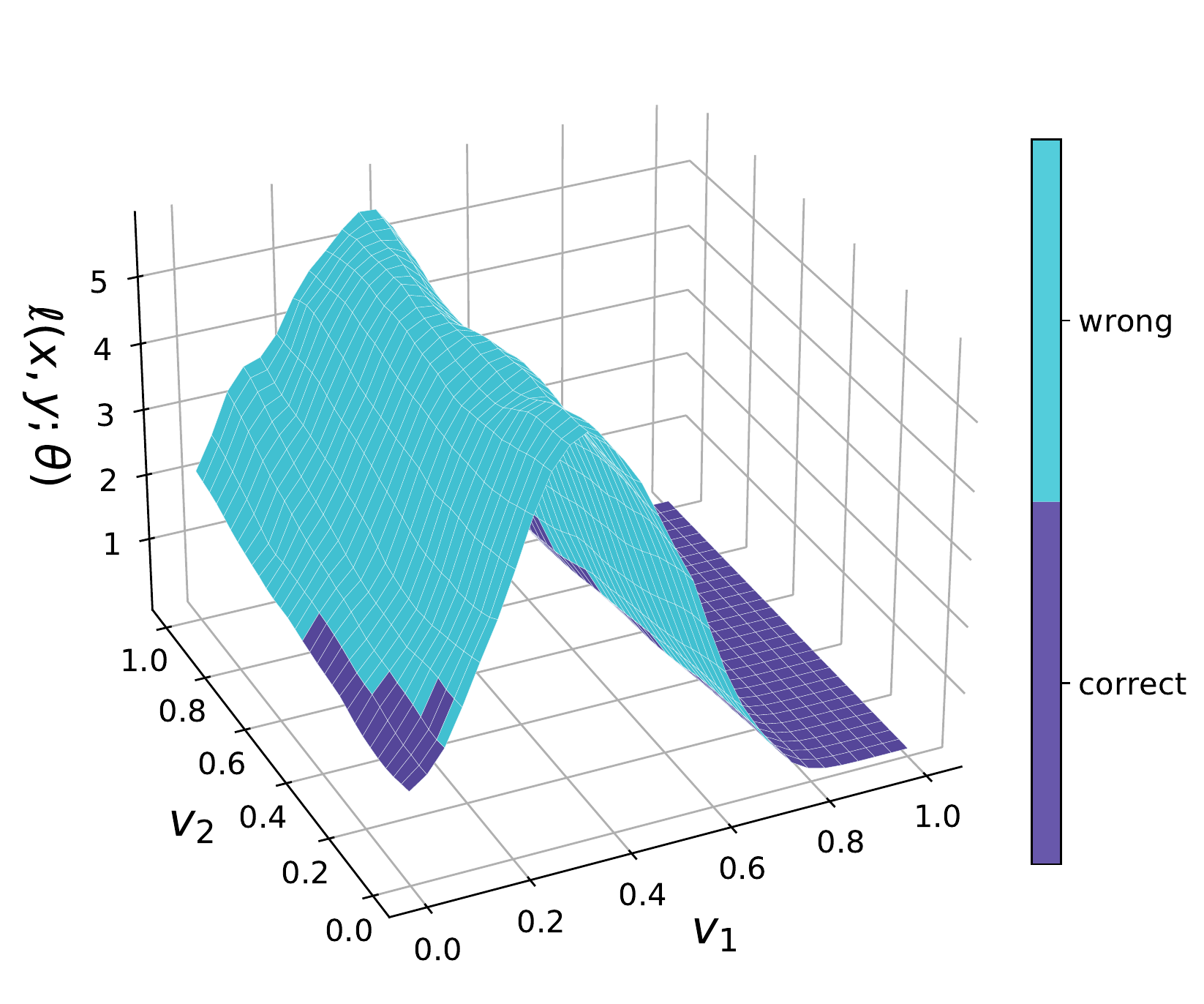}
        \caption{Loss surface}
        \end{subfigure}
    \end{minipage} \\

    \begin{minipage}{0.075\linewidth}
        \begin{subfigure}[t]{\textwidth}
        \includegraphics[width=\linewidth]{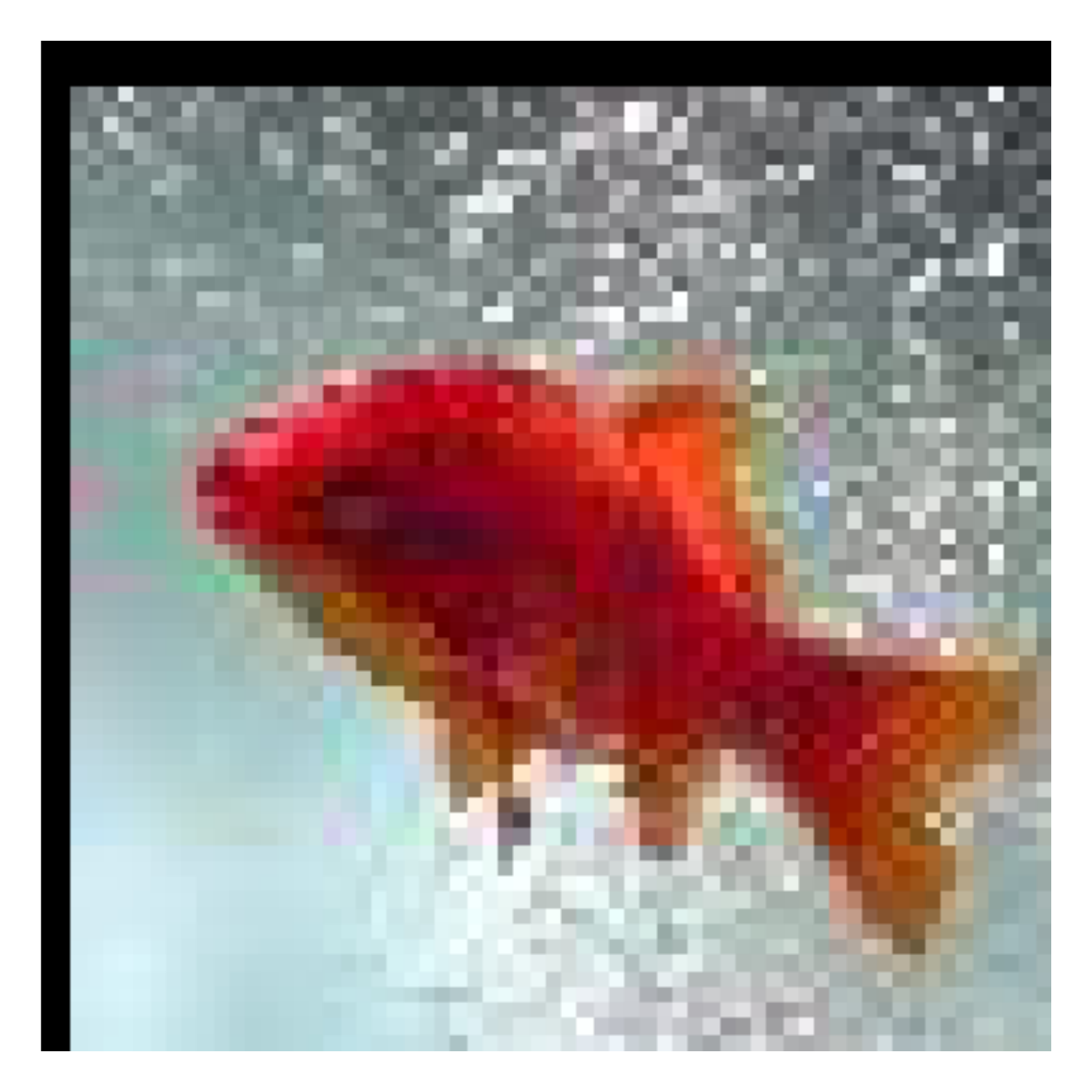}
        \caption{$x$}
        \end{subfigure} \\
        \begin{subfigure}[b]{\textwidth}
        \includegraphics[width=\linewidth]{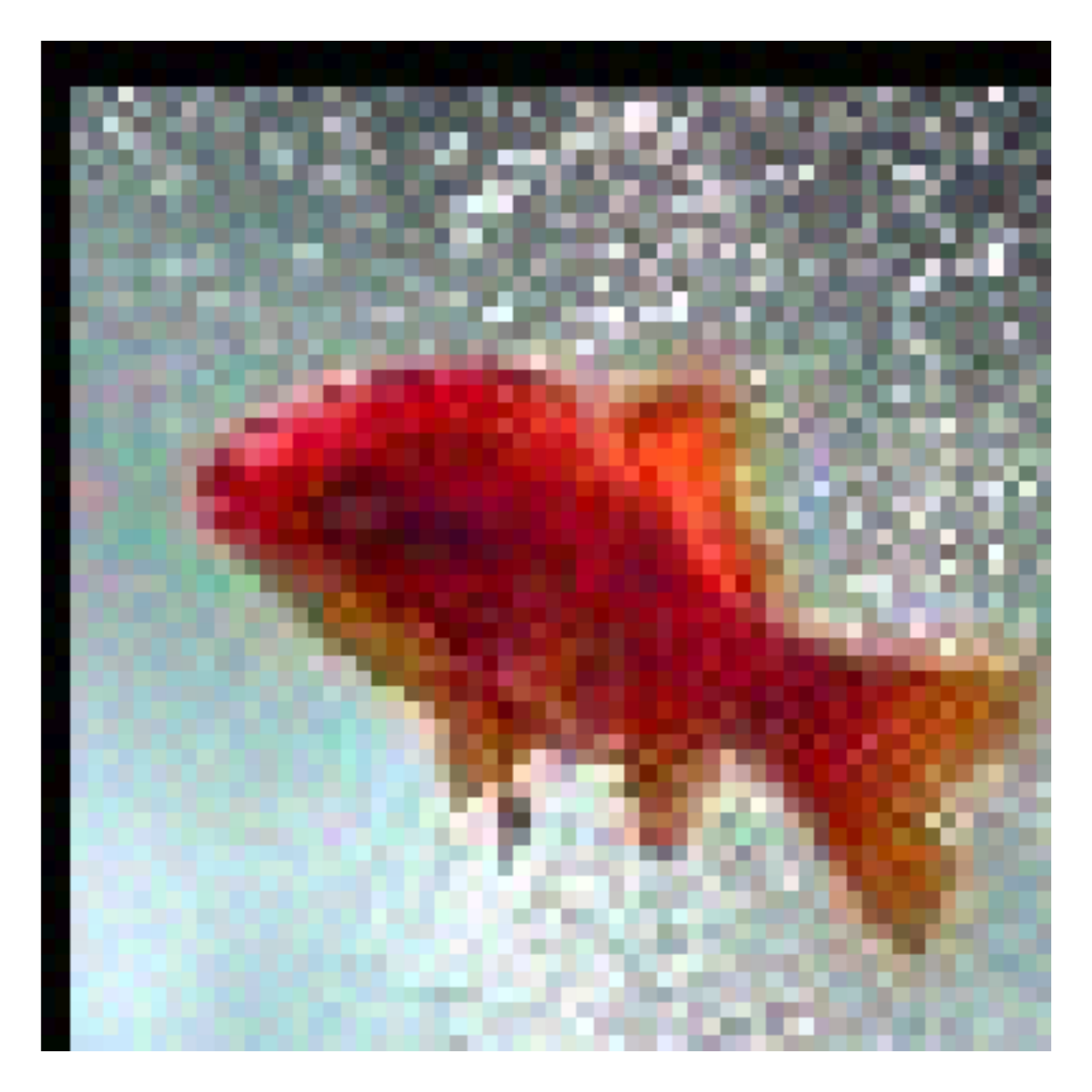}
        \caption{$x'$}
        \end{subfigure}
    \end{minipage}
    \begin{minipage}{0.3009\linewidth}
        \begin{subfigure}{\textwidth}
        \includegraphics[width=\linewidth]{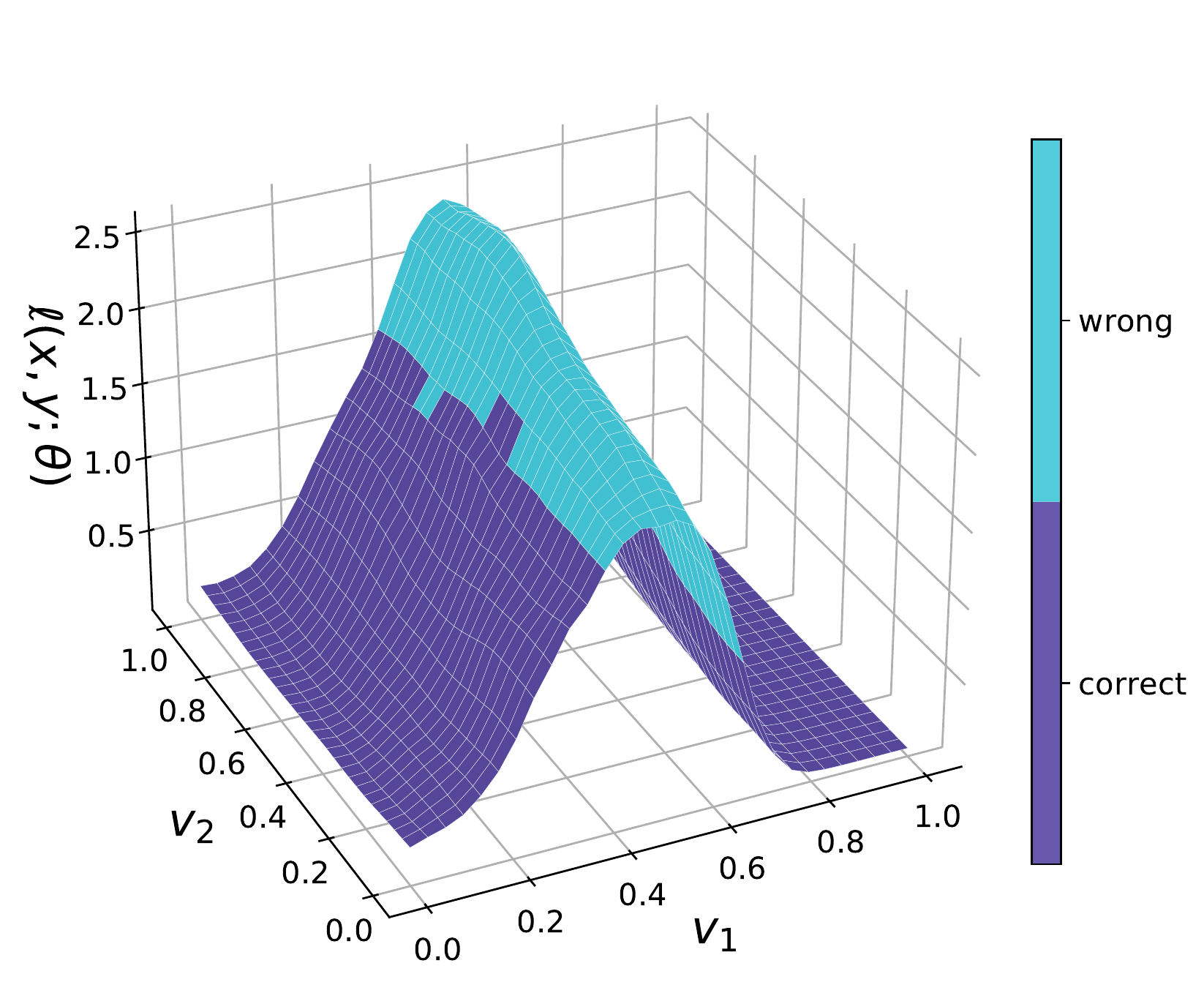}
        \caption{Loss surface}
        \end{subfigure}
    \end{minipage}
    \begin{minipage}{0.075\linewidth}
        \begin{subfigure}[t]{\textwidth}
        \includegraphics[width=\linewidth]{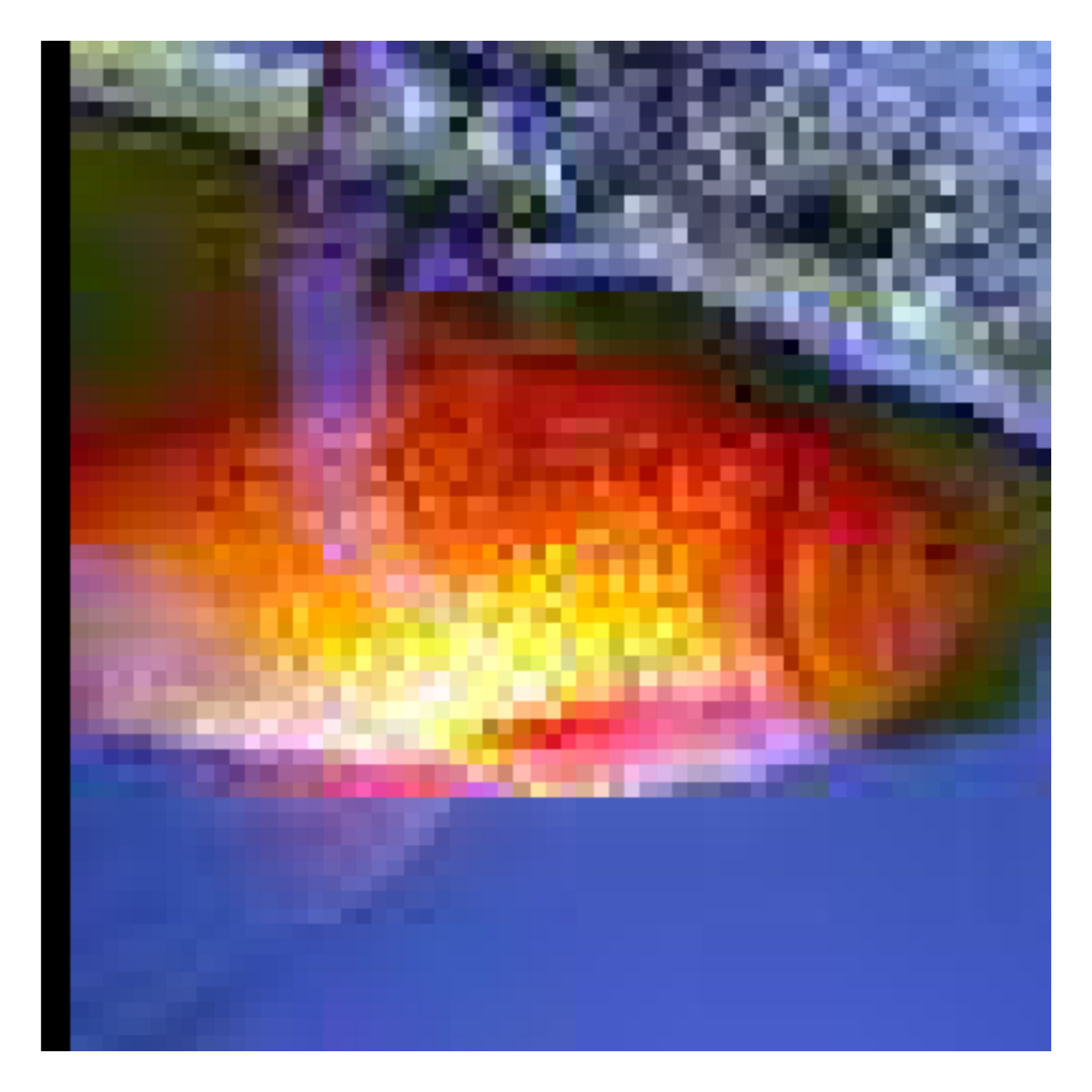}
        \caption{$x$}
        \end{subfigure} \\
        \begin{subfigure}[b]{\textwidth}
        \includegraphics[width=\linewidth]{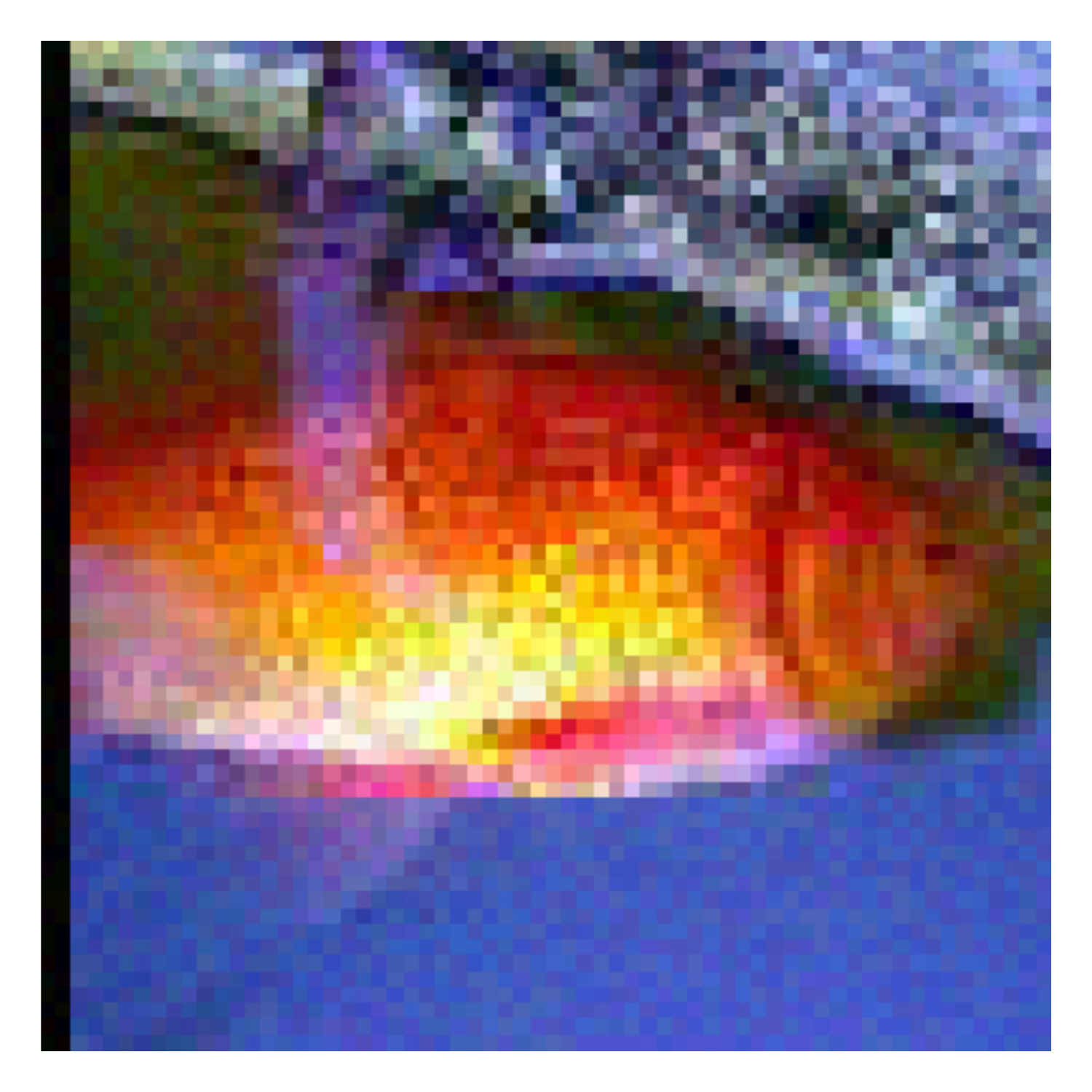}
        \caption{$x'$}
        \end{subfigure}
    \end{minipage}
    \begin{minipage}{0.3009\linewidth}
        \begin{subfigure}{\textwidth}
        \includegraphics[width=\linewidth]{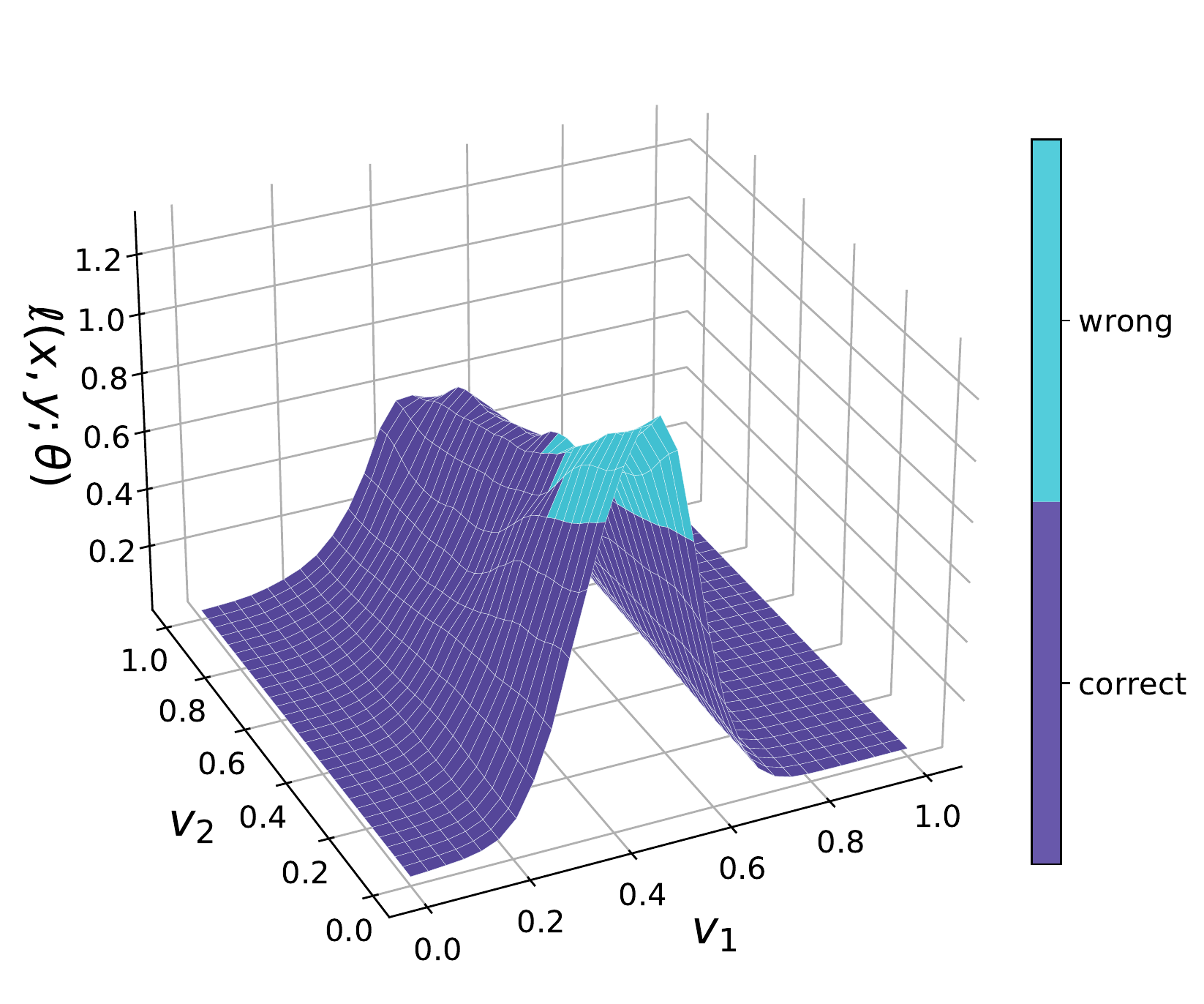}
        \caption{Loss surface}
        \end{subfigure}
    \end{minipage}
    \caption{(Tiny ImageNet) Direction of fast adversarial perturbation $v_1$ and random direction $v_2$. Adversarial example $x'=x+v_1$ is generated from original example $x$.}
    \label{fig:fast_train_tiny}
\end{figure*}

\begin{figure*}[p]
    \centering
    \begin{minipage}{0.075\linewidth}
        \begin{subfigure}[t]{\textwidth}
        \includegraphics[width=\linewidth]{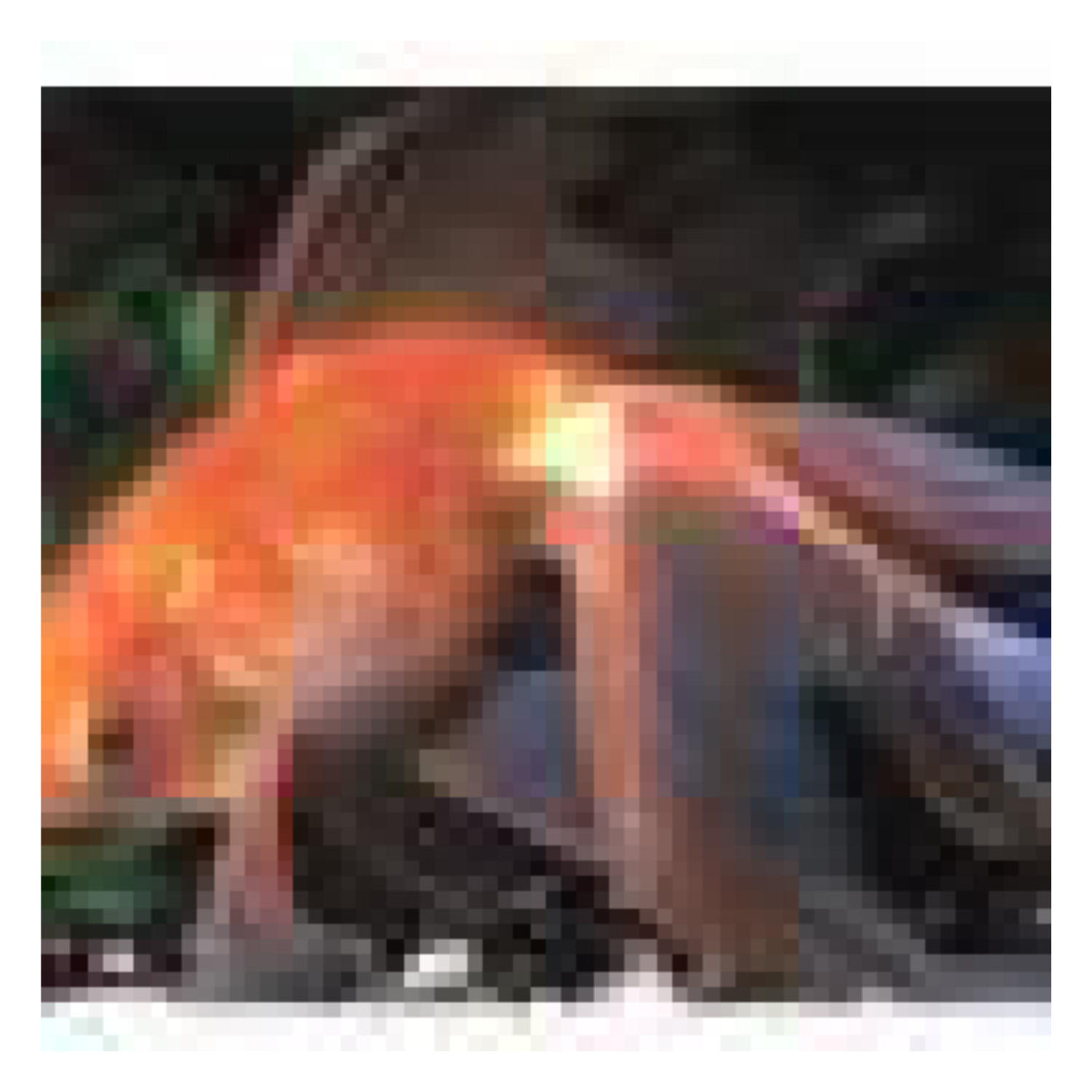}
        \caption{$x$}
        \end{subfigure} \\
        \begin{subfigure}[b]{\textwidth}
        \includegraphics[width=\linewidth]{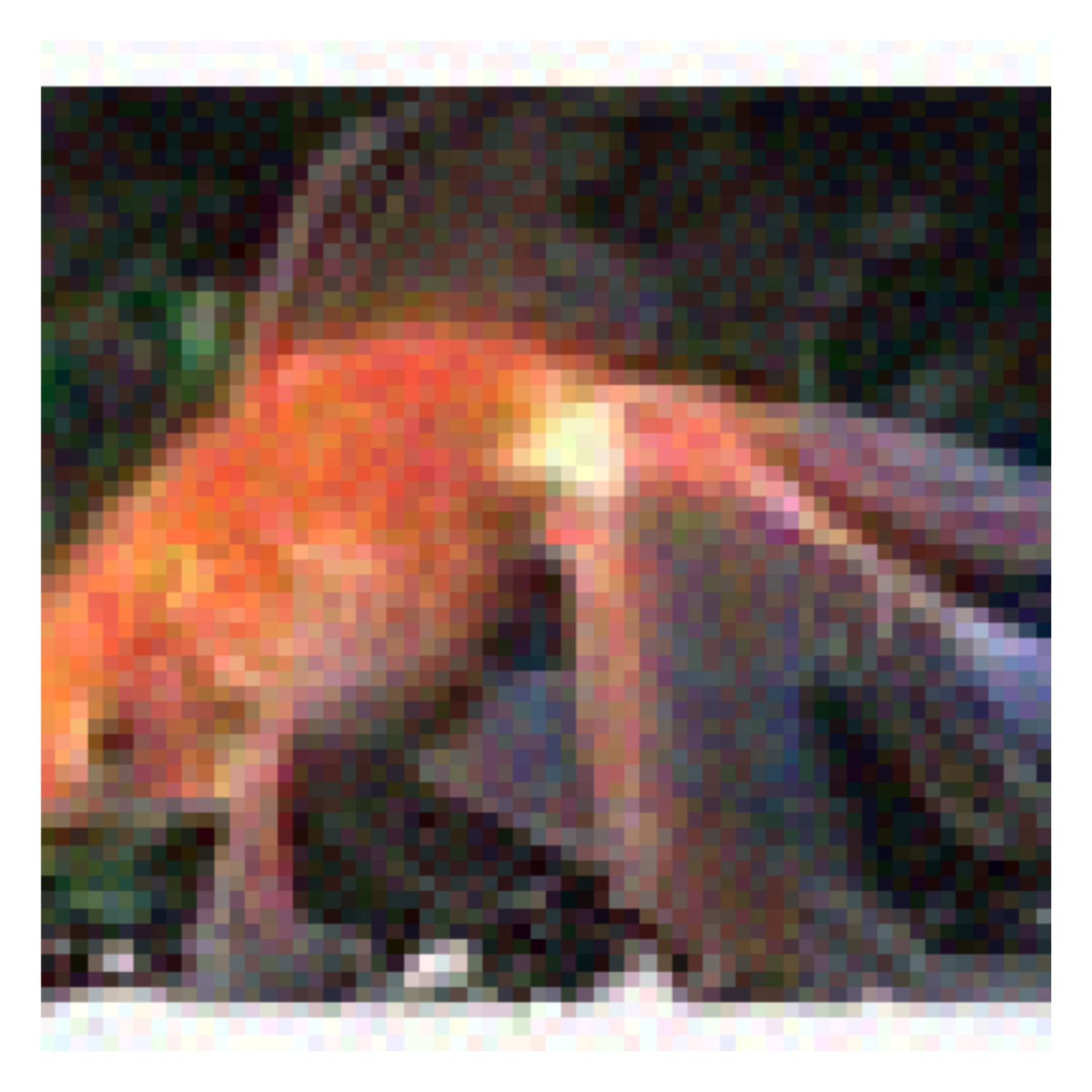}
        \caption{$x'$}
        \end{subfigure}
    \end{minipage}
    \begin{minipage}{0.3009\linewidth}
        \begin{subfigure}{\textwidth}
        \includegraphics[width=\linewidth]{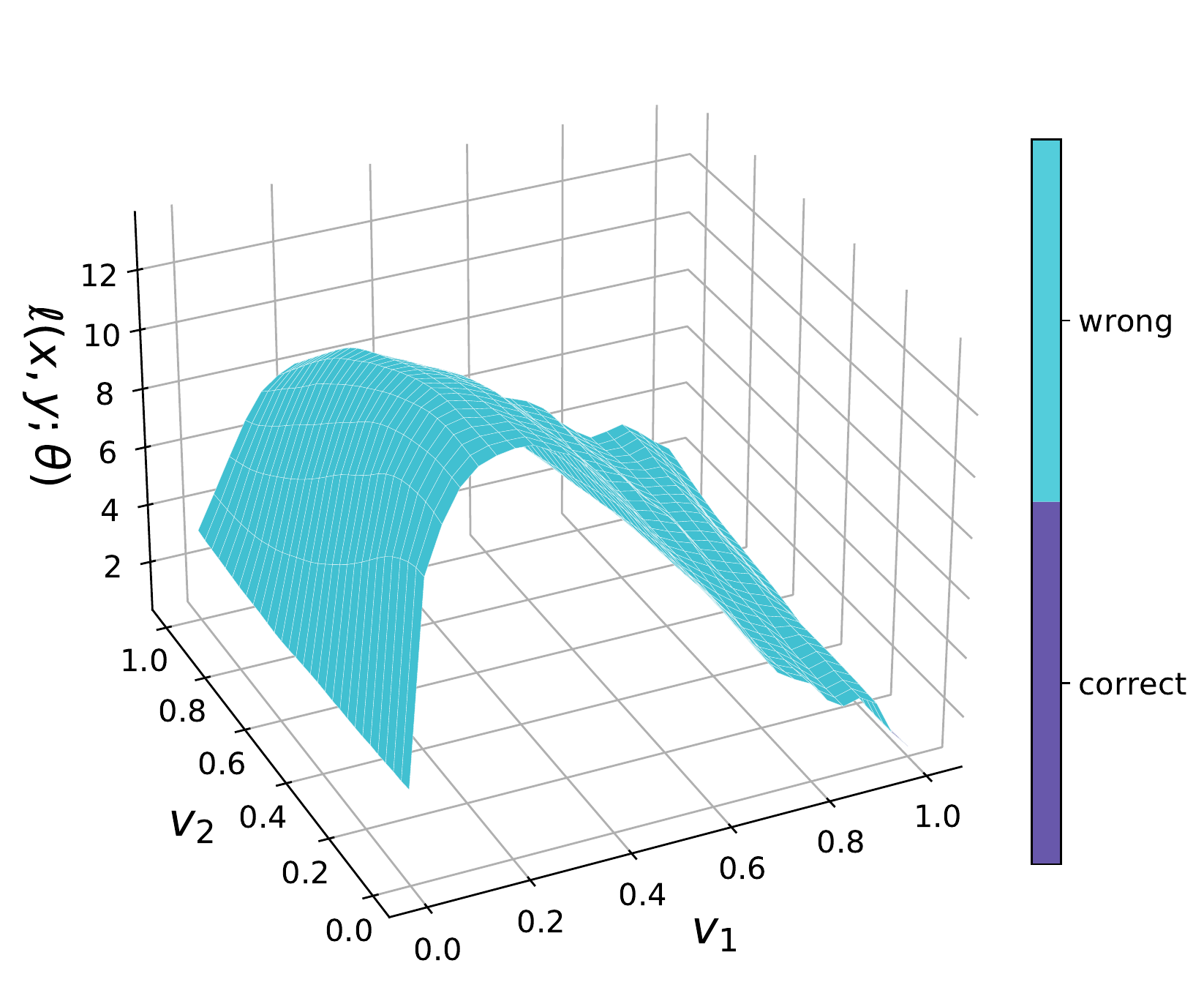}
        \caption{Loss surface}
        \end{subfigure}
    \end{minipage}
    \begin{minipage}{0.075\linewidth}
        \begin{subfigure}[t]{\textwidth}
        \includegraphics[width=\linewidth]{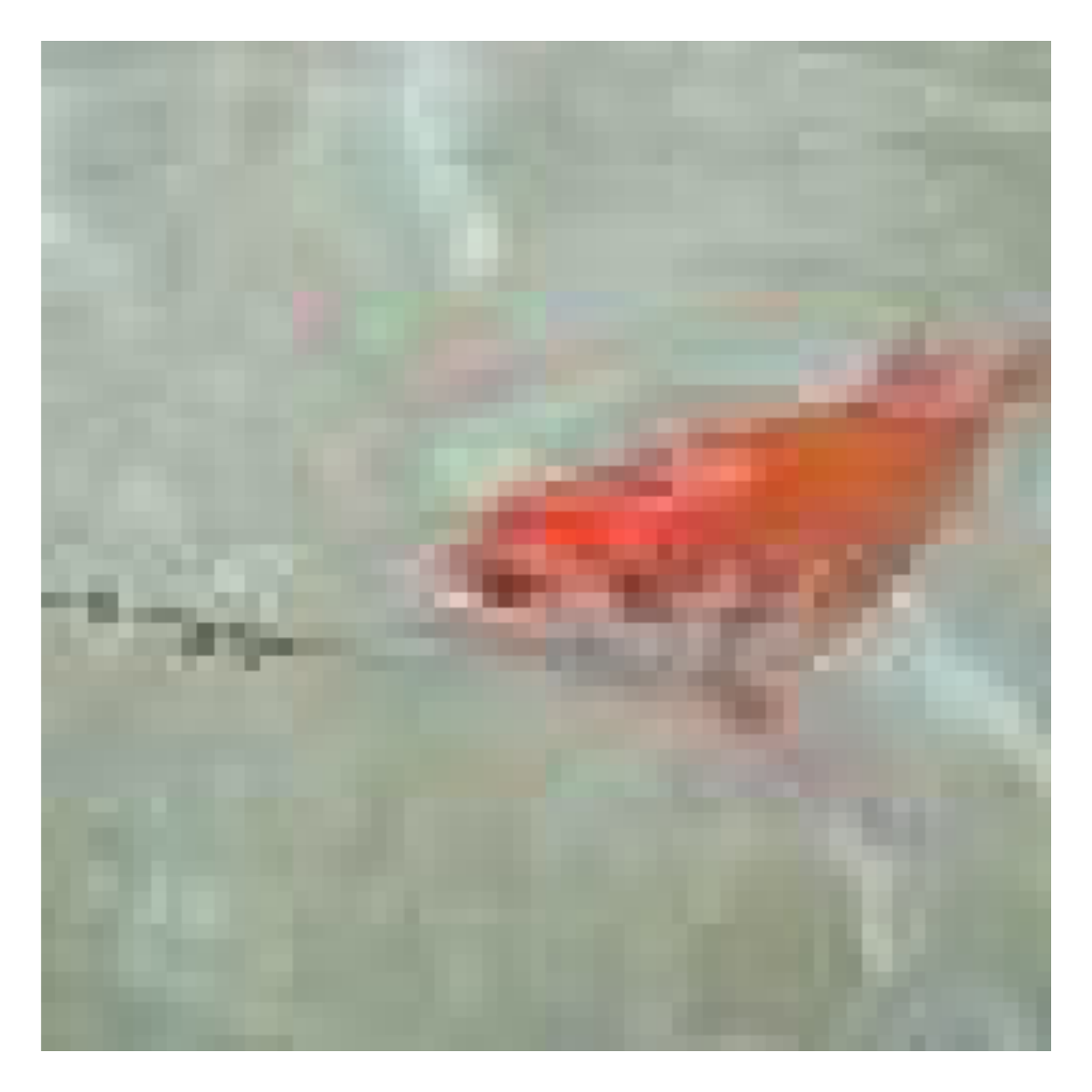}
        \caption{$x$}
        \end{subfigure} \\
        \begin{subfigure}[b]{\textwidth}
        \includegraphics[width=\linewidth]{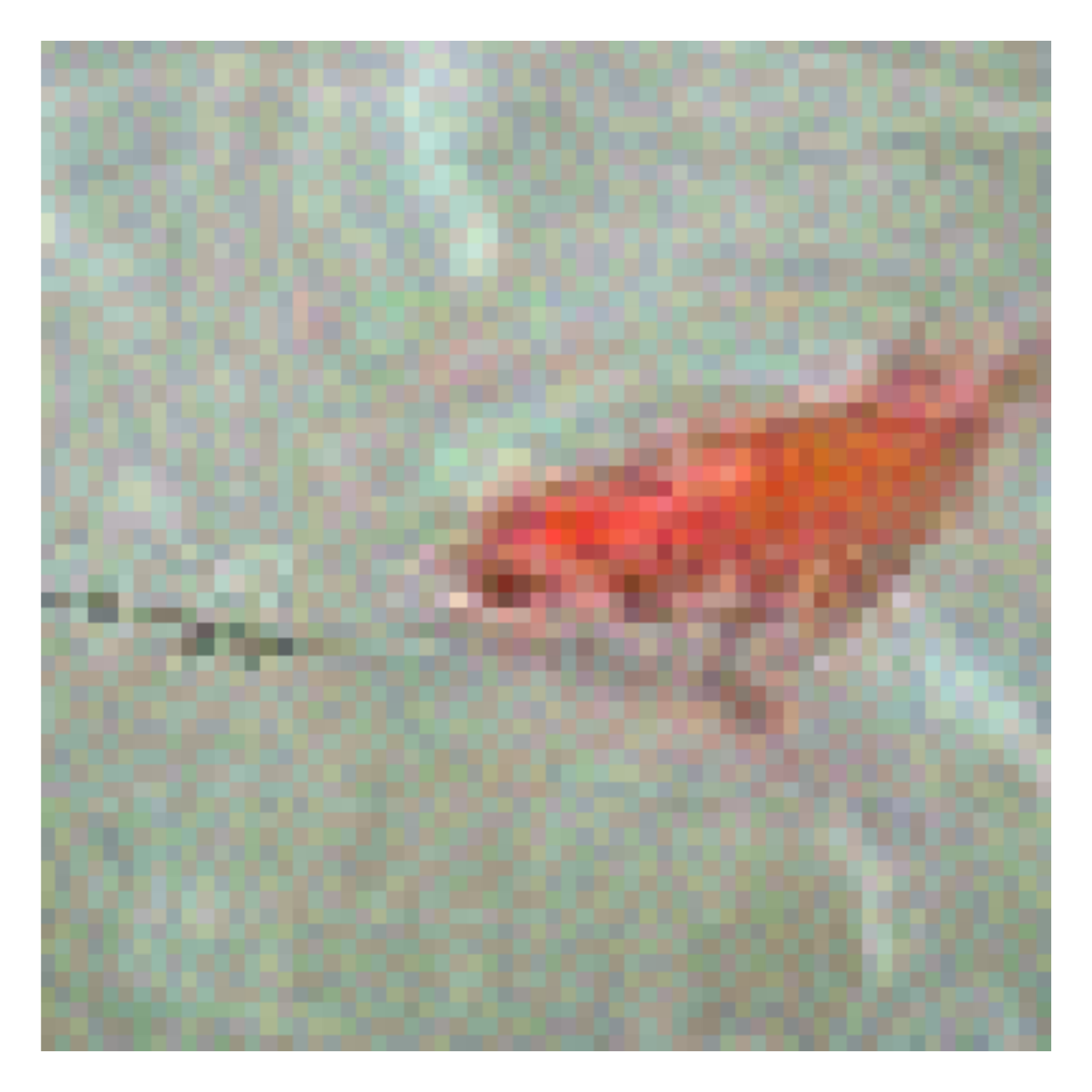}
        \caption{$x'$}
        \end{subfigure}
    \end{minipage}
    \begin{minipage}{0.3009\linewidth}
        \begin{subfigure}{\textwidth}
        \includegraphics[width=\linewidth]{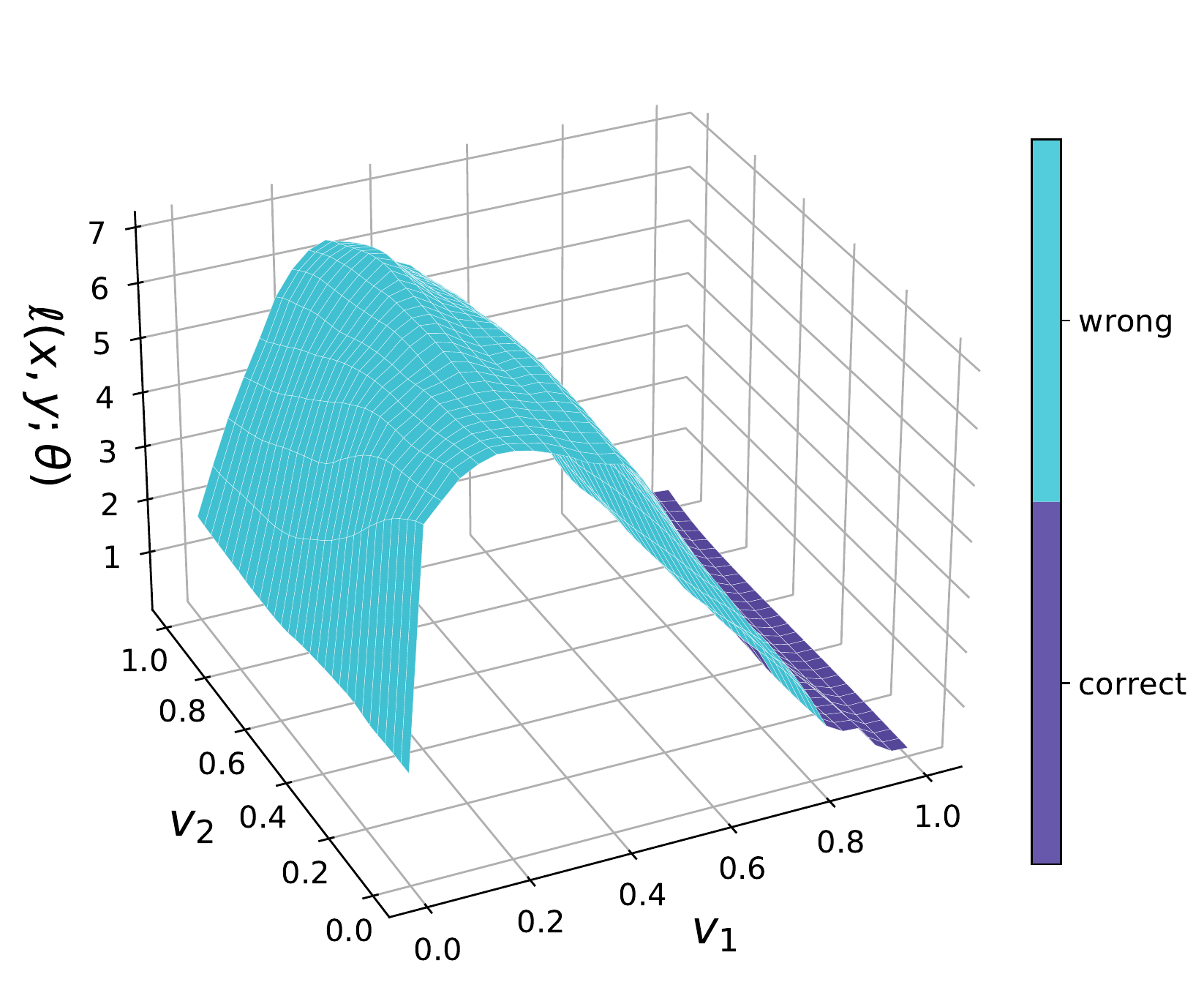}
        \caption{Loss surface}
        \end{subfigure}
    \end{minipage} \\
    
    \begin{minipage}{0.075\linewidth}
        \begin{subfigure}[t]{\textwidth}
        \includegraphics[width=\linewidth]{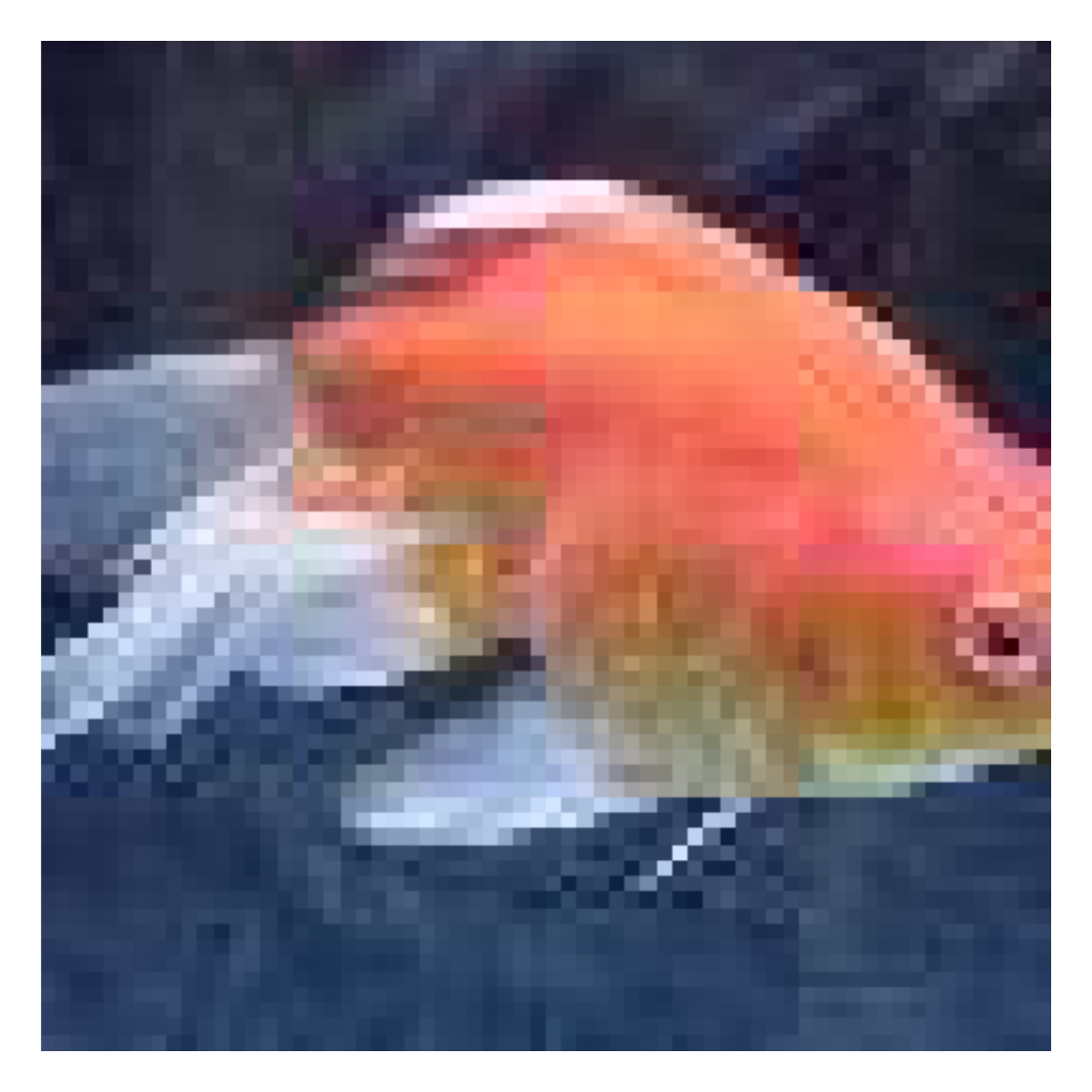}
        \caption{$x$}
        \end{subfigure} \\
        \begin{subfigure}[b]{\textwidth}
        \includegraphics[width=\linewidth]{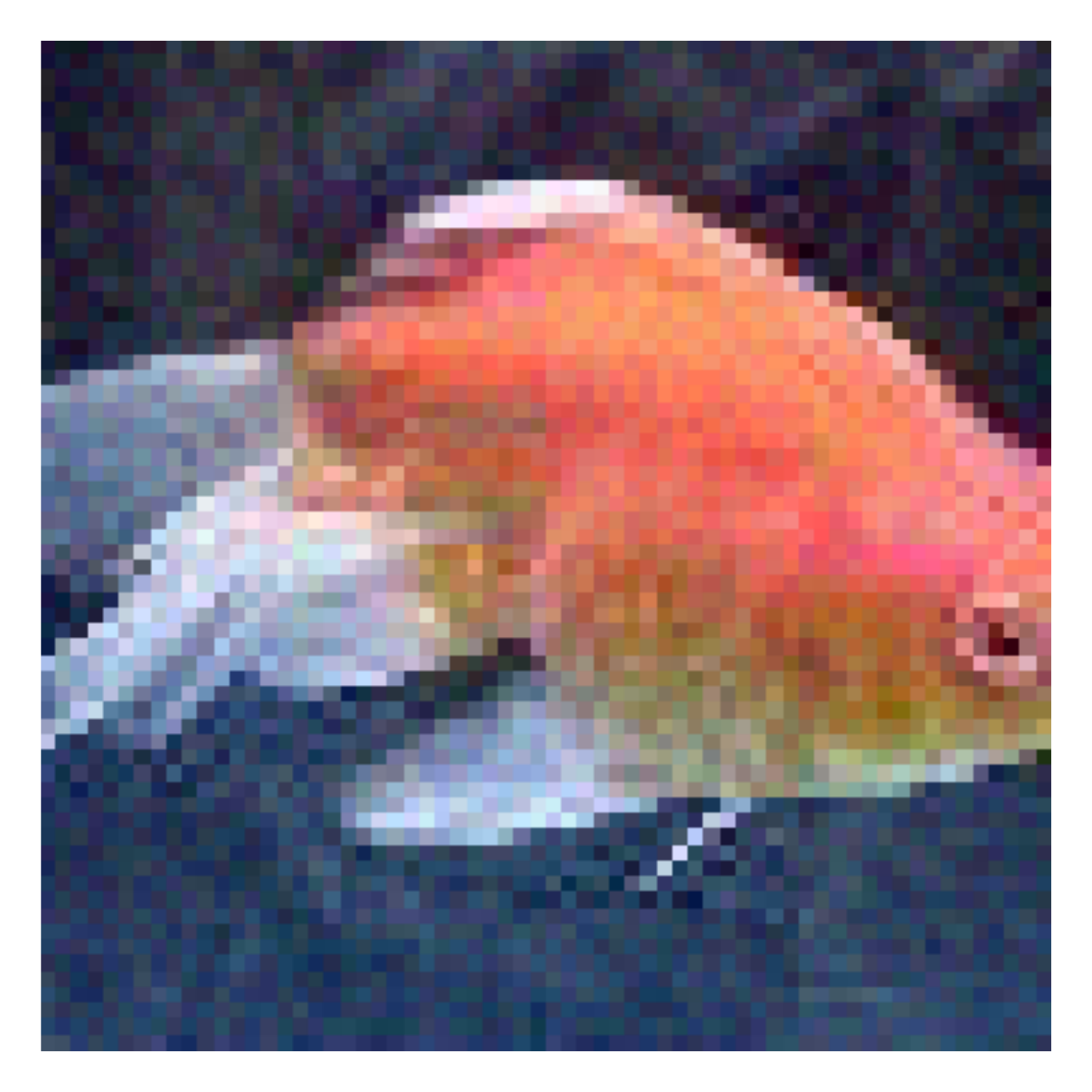}
        \caption{$x'$}
        \end{subfigure}
    \end{minipage}
    \begin{minipage}{0.3009\linewidth}
        \begin{subfigure}{\textwidth}
        \includegraphics[width=\linewidth]{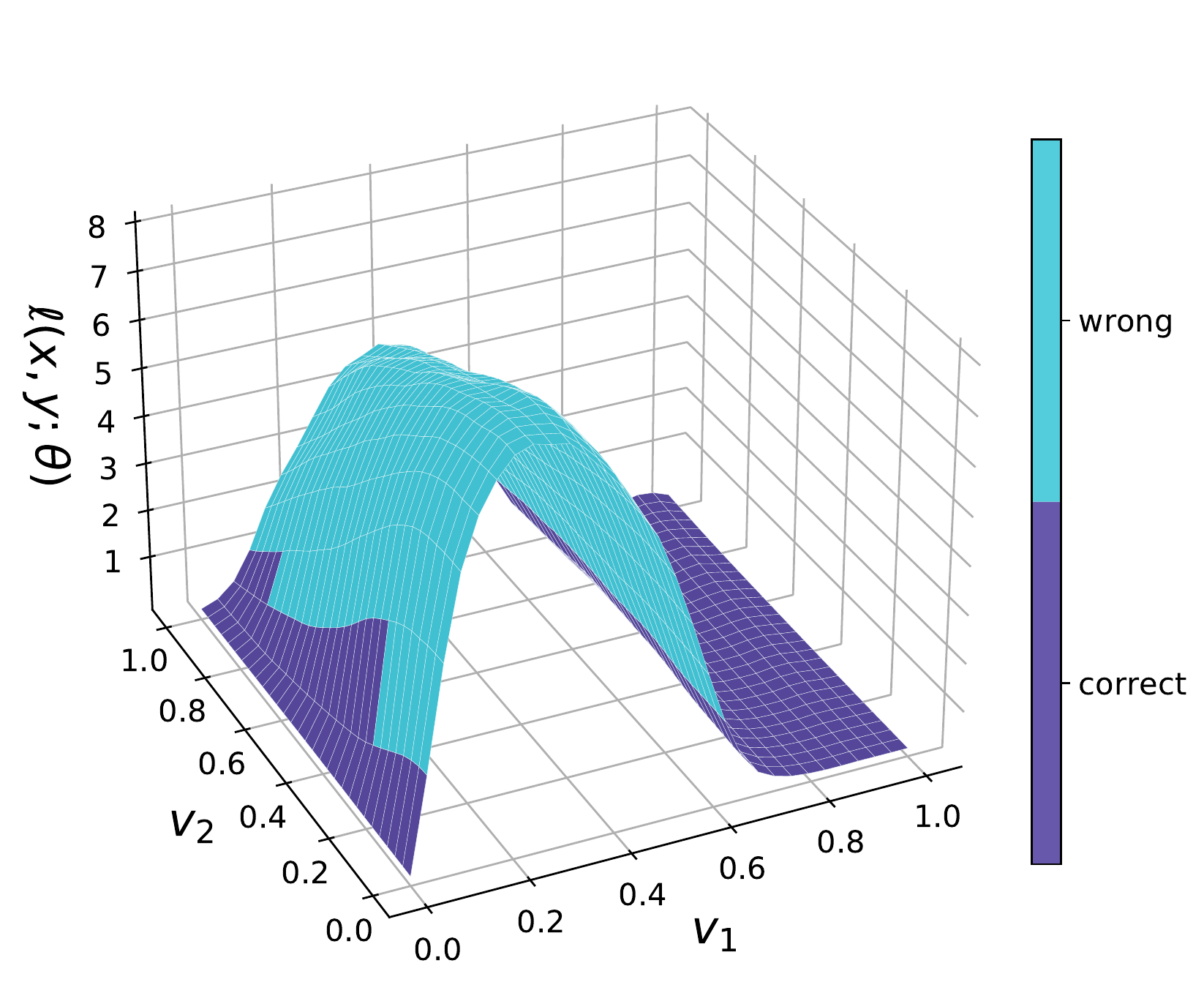}
        \caption{Loss surface}
        \end{subfigure}
    \end{minipage}
    \begin{minipage}{0.075\linewidth}
        \begin{subfigure}[t]{\textwidth}
        \includegraphics[width=\linewidth]{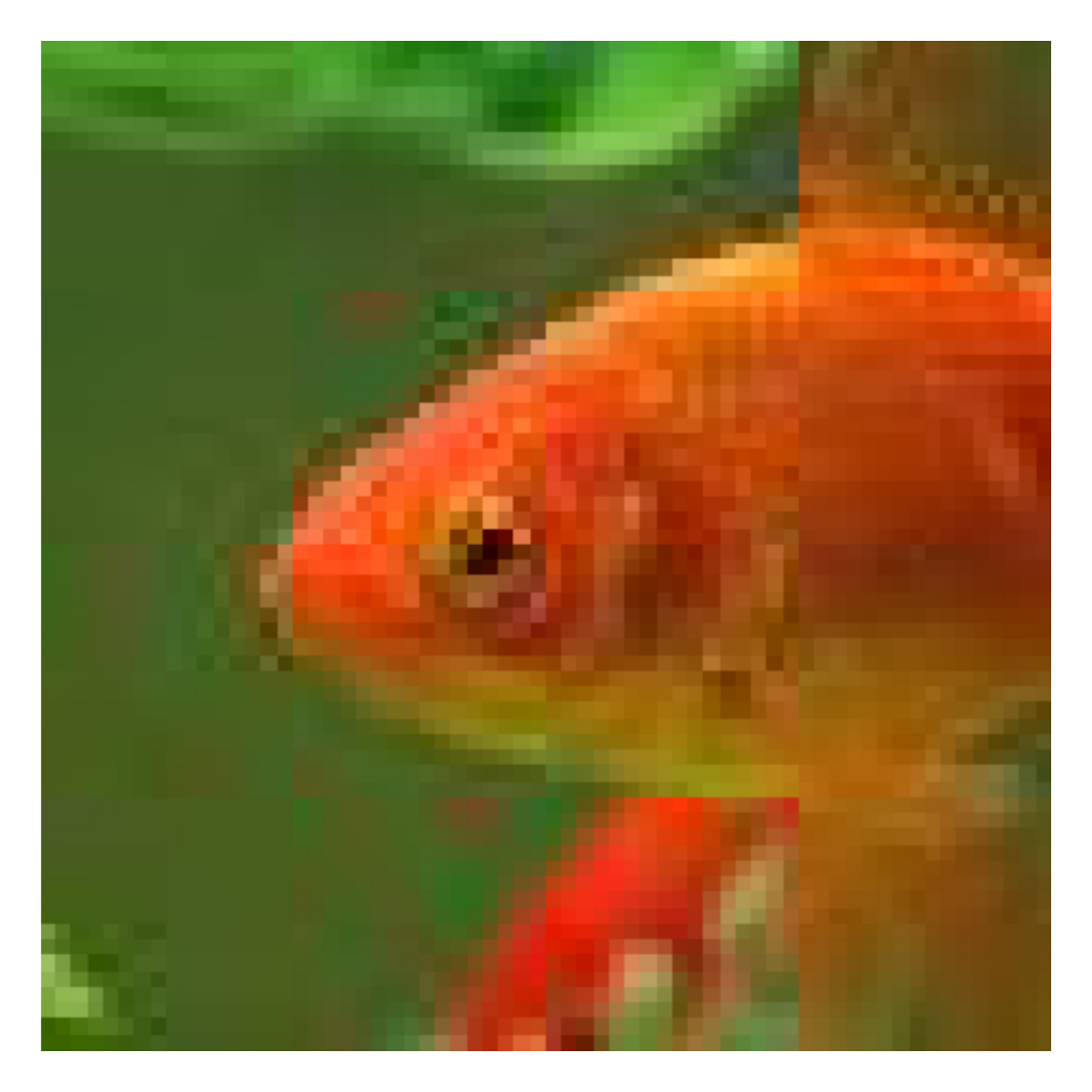}
        \caption{$x$}
        \end{subfigure} \\
        \begin{subfigure}[b]{\textwidth}
        \includegraphics[width=\linewidth]{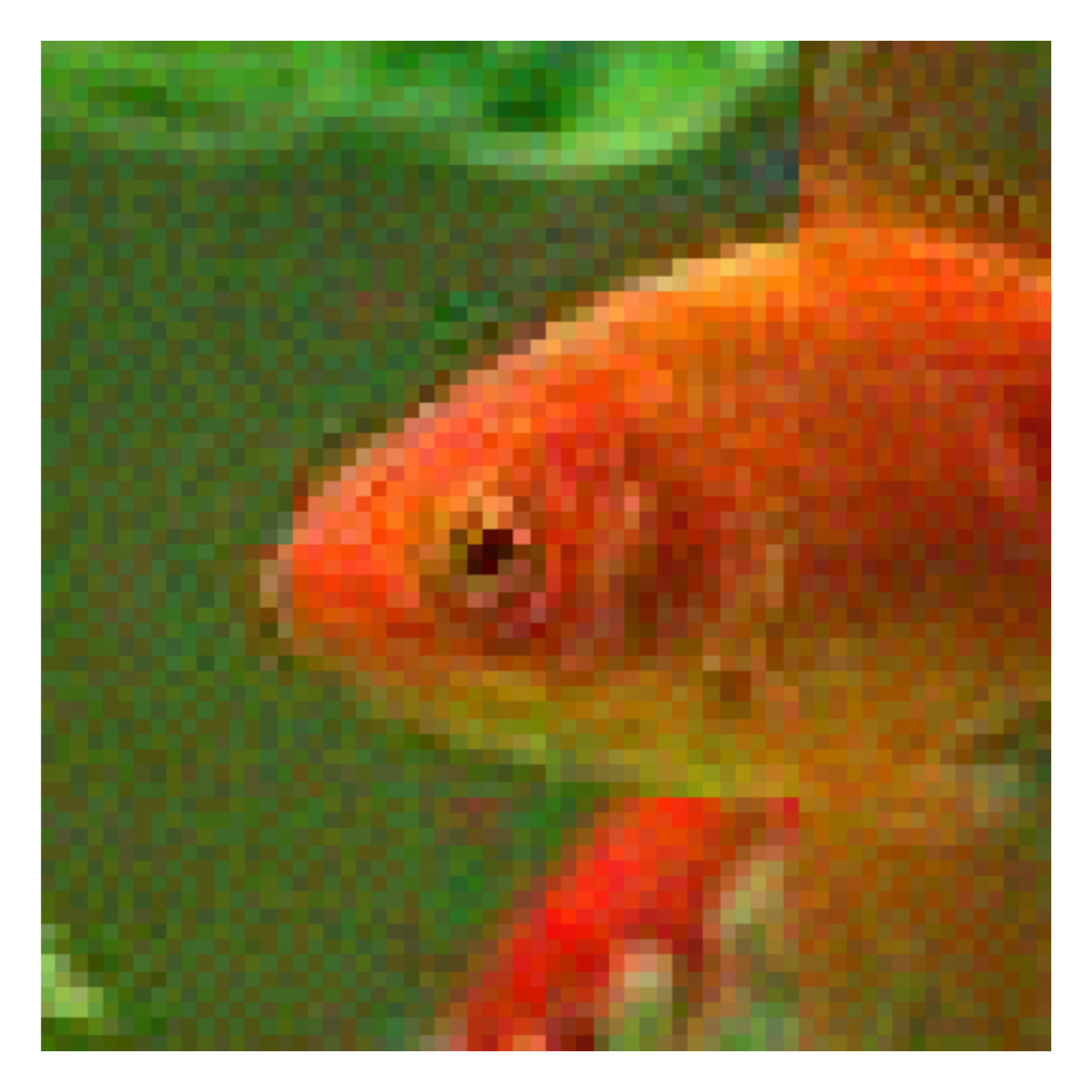}
        \caption{$x'$}
        \end{subfigure}
    \end{minipage}
    \begin{minipage}{0.3009\linewidth}
        \begin{subfigure}{\textwidth}
        \includegraphics[width=\linewidth]{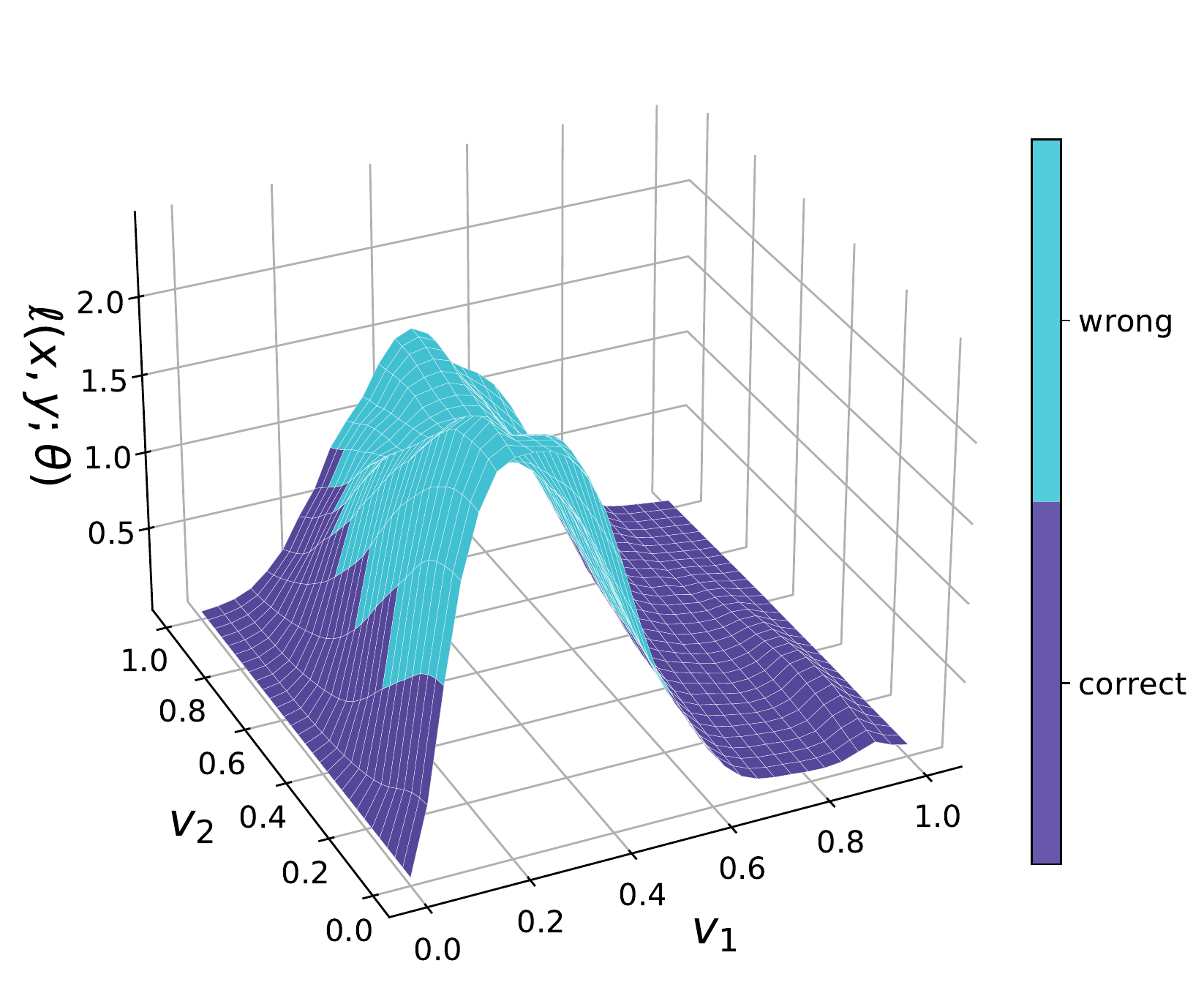}
        \caption{Loss surface}
        \end{subfigure}
    \end{minipage}
    \caption{(Tiny ImageNet) Direction of FGSM adversarial perturbation $v_1$ and random direction $v_2$. Adversarial example $x'=x+v_1$ is generated from original example $x$.}
    \label{fig:fgsm_test_tiny}
\end{figure*}

\begin{figure*}[p]
    \centering
    \begin{minipage}{0.075\linewidth}
        \begin{subfigure}[t]{\textwidth}
        \includegraphics[width=\linewidth]{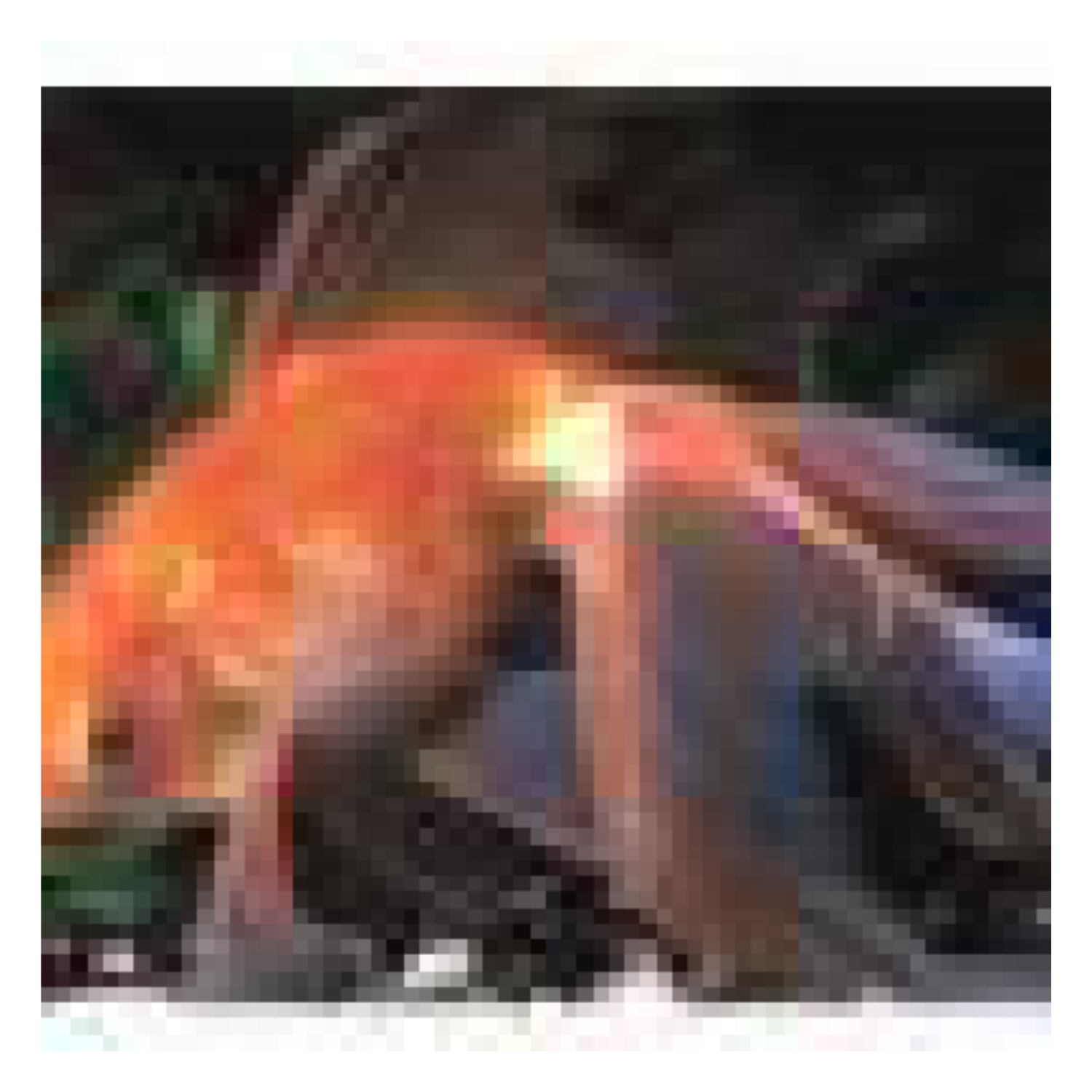}
        \caption{$x$}
        \end{subfigure} \\
        \begin{subfigure}[b]{\textwidth}
        \includegraphics[width=\linewidth]{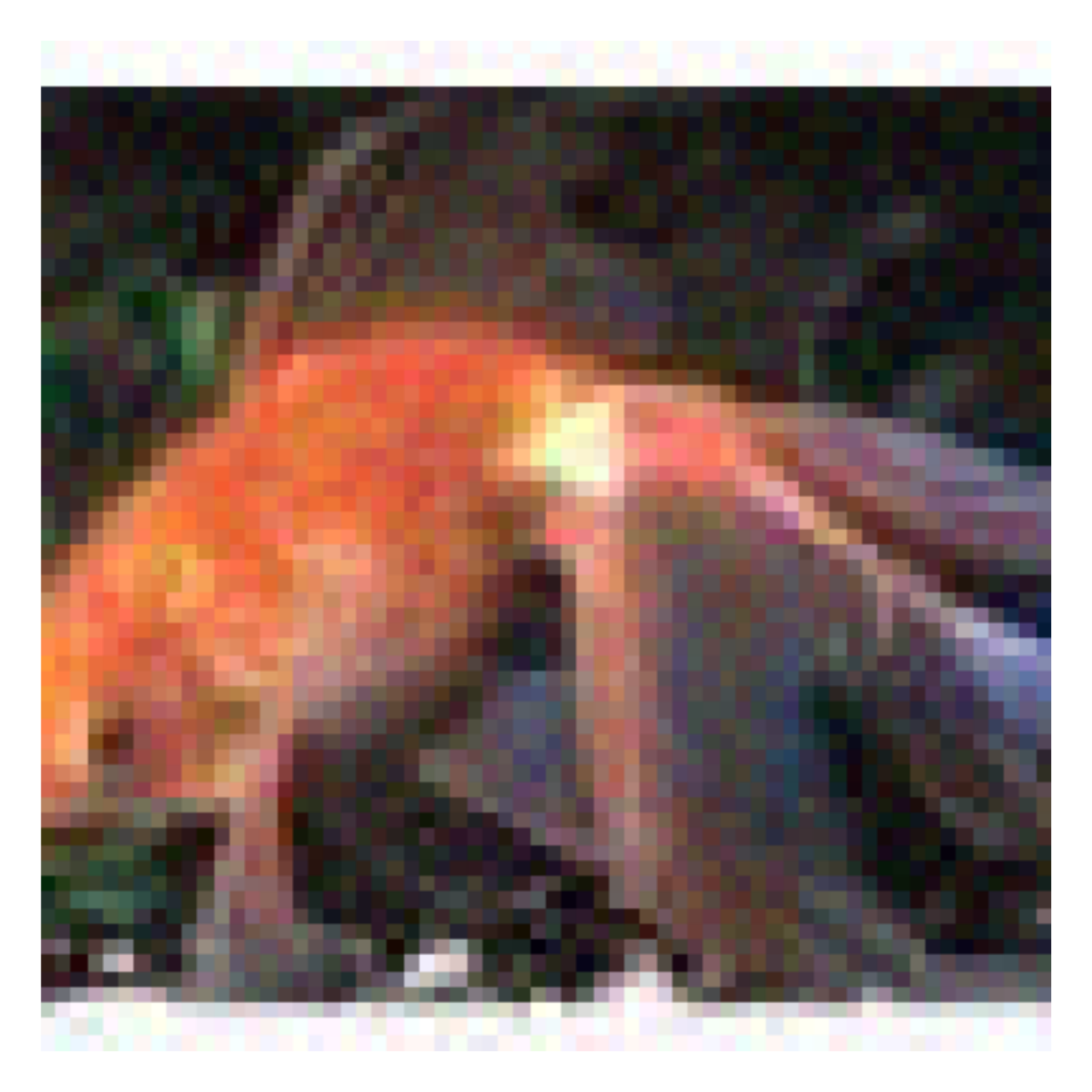}
        \caption{$x'$}
        \end{subfigure}
    \end{minipage}
    \begin{minipage}{0.3009\linewidth}
        \begin{subfigure}{\textwidth}
        \includegraphics[width=\linewidth]{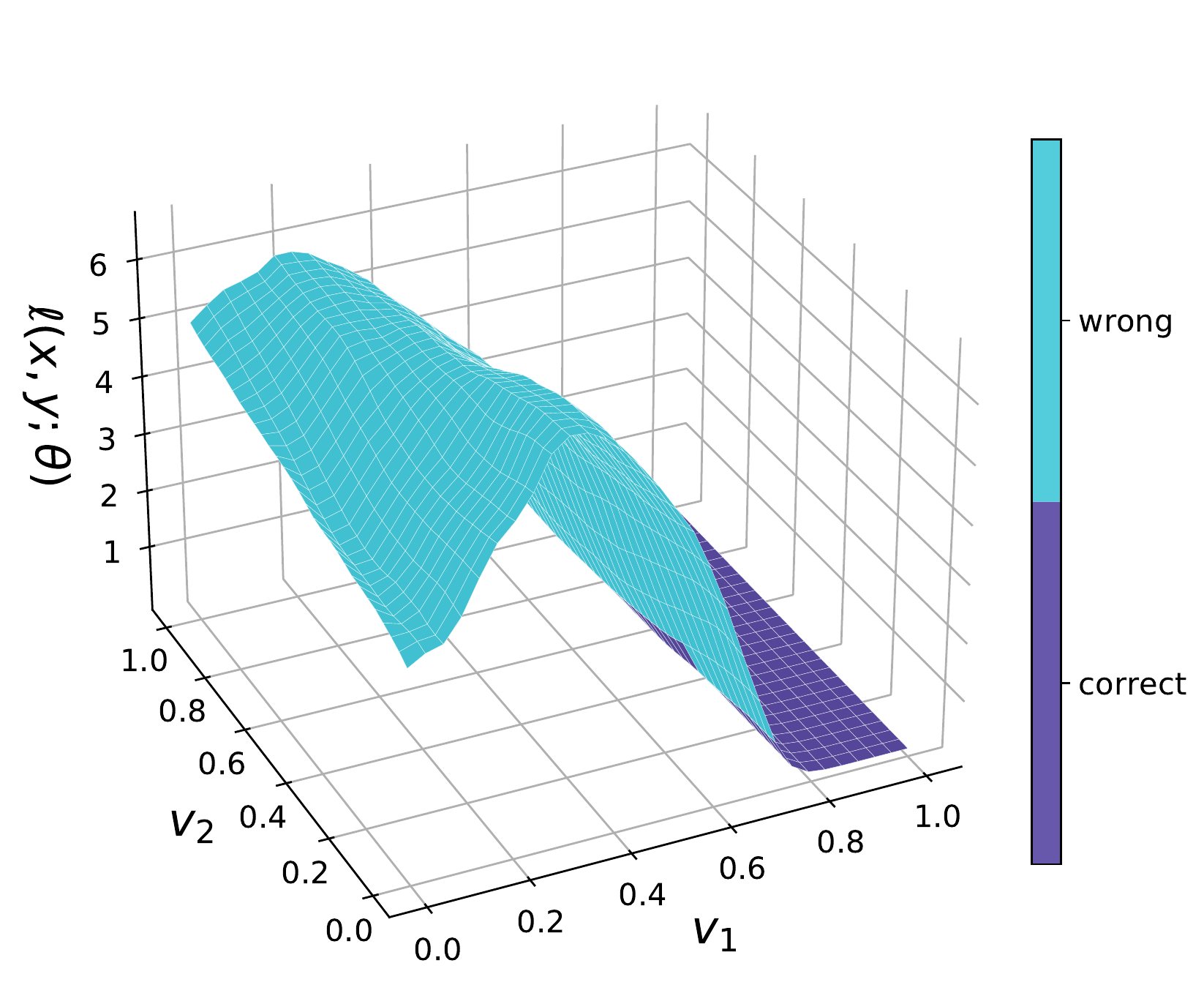}
        \caption{Loss surface}
        \end{subfigure}
    \end{minipage}
    \begin{minipage}{0.075\linewidth}
        \begin{subfigure}[t]{\textwidth}
        \includegraphics[width=\linewidth]{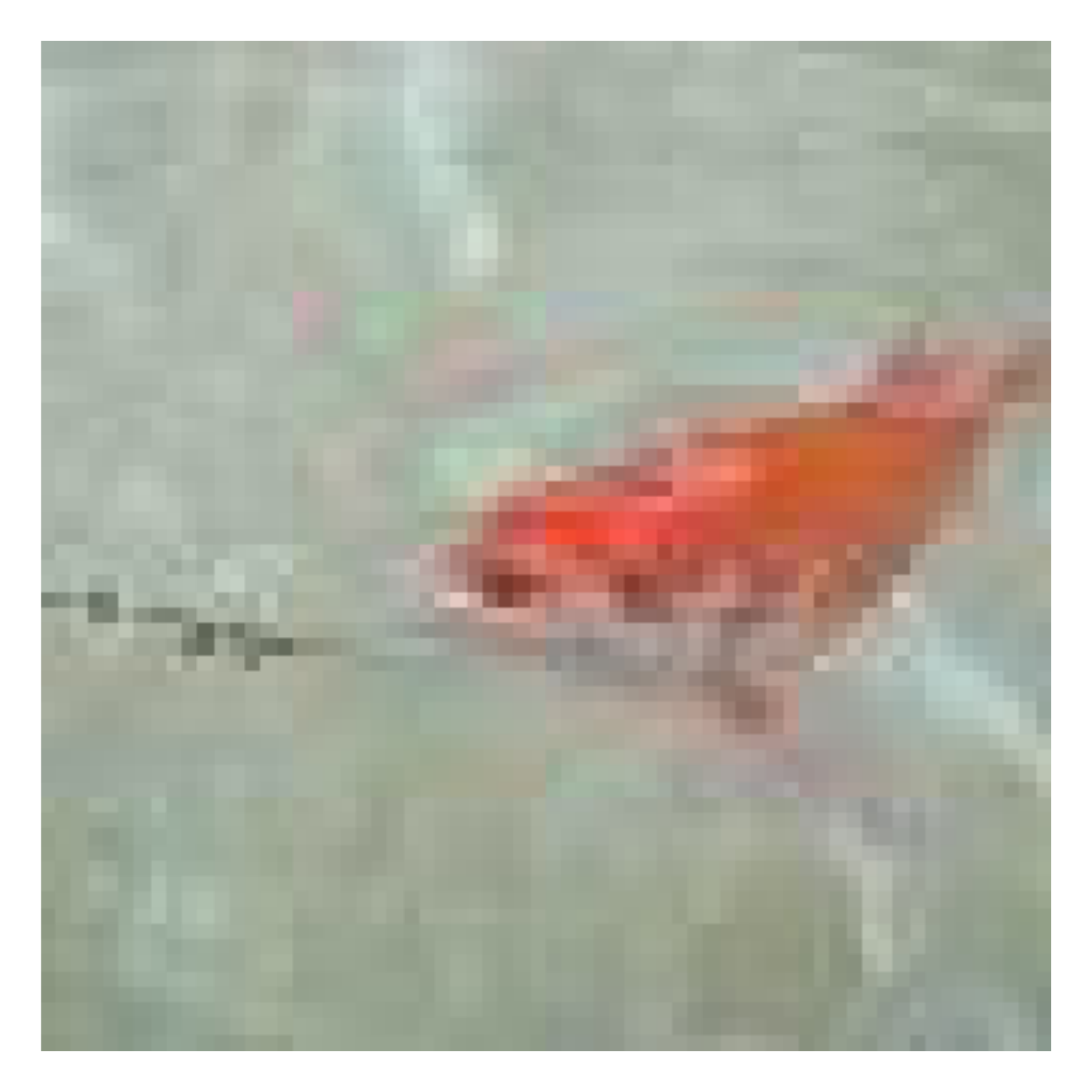}
        \caption{$x$}
        \end{subfigure} \\
        \begin{subfigure}[b]{\textwidth}
        \includegraphics[width=\linewidth]{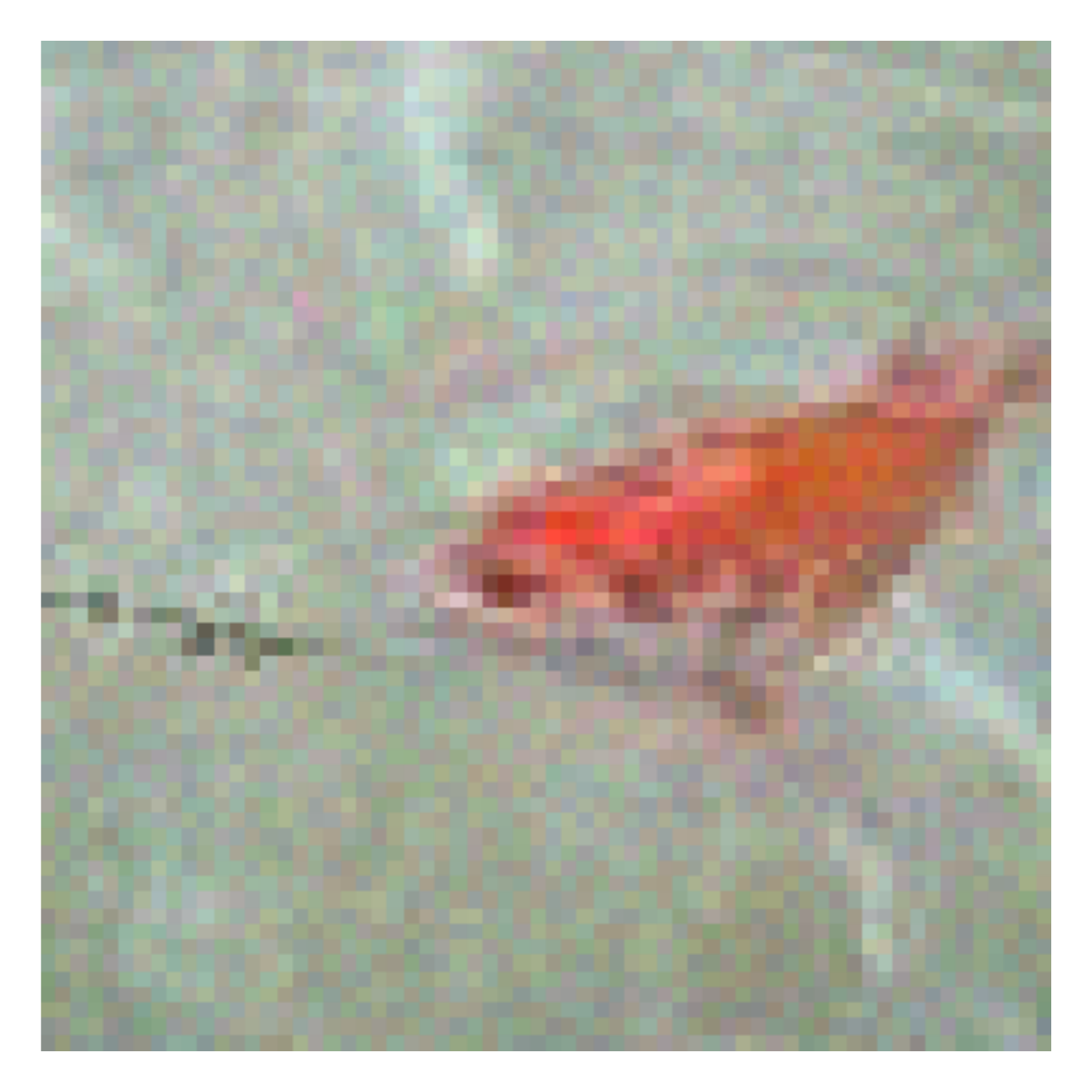}
        \caption{$x'$}
        \end{subfigure}
    \end{minipage}
    \begin{minipage}{0.3009\linewidth}
        \begin{subfigure}{\textwidth}
        \includegraphics[width=\linewidth]{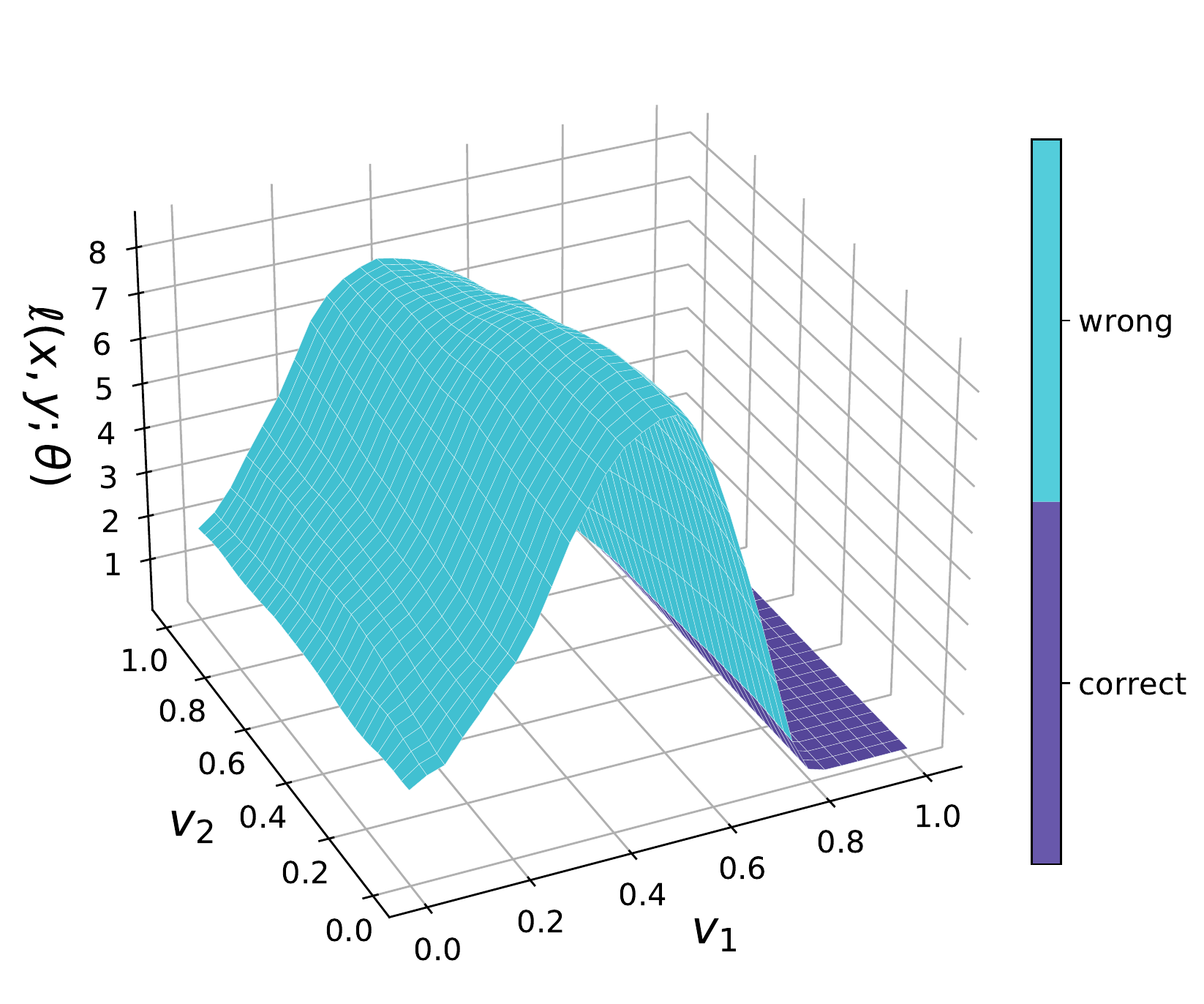}
        \caption{Loss surface}
        \end{subfigure}
    \end{minipage} \\

    \begin{minipage}{0.075\linewidth}
        \begin{subfigure}[t]{\textwidth}
        \includegraphics[width=\linewidth]{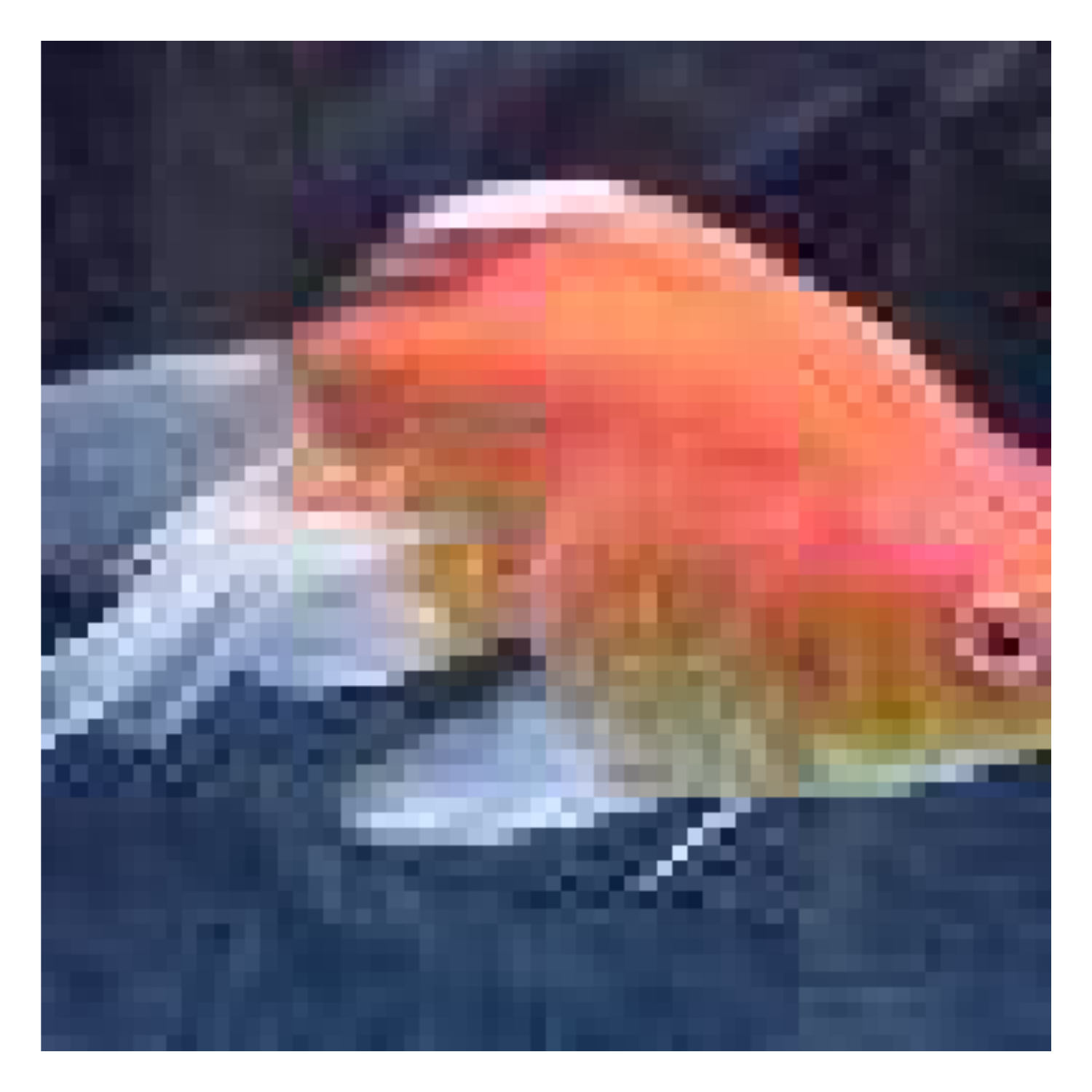}
        \caption{$x$}
        \end{subfigure} \\
        \begin{subfigure}[b]{\textwidth}
        \includegraphics[width=\linewidth]{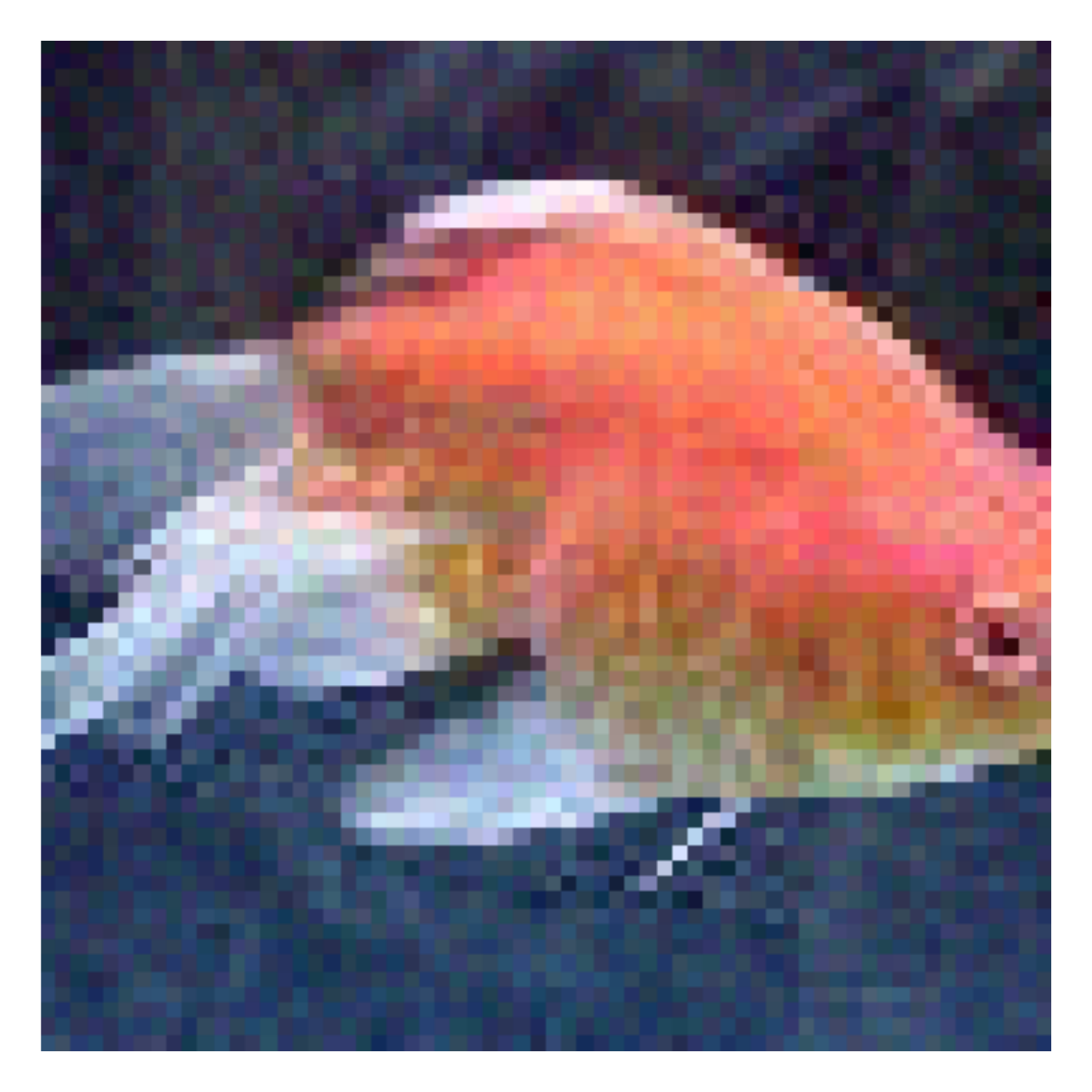}
        \caption{$x'$}
        \end{subfigure}
    \end{minipage}
    \begin{minipage}{0.3009\linewidth}
        \begin{subfigure}{\textwidth}
        \includegraphics[width=\linewidth]{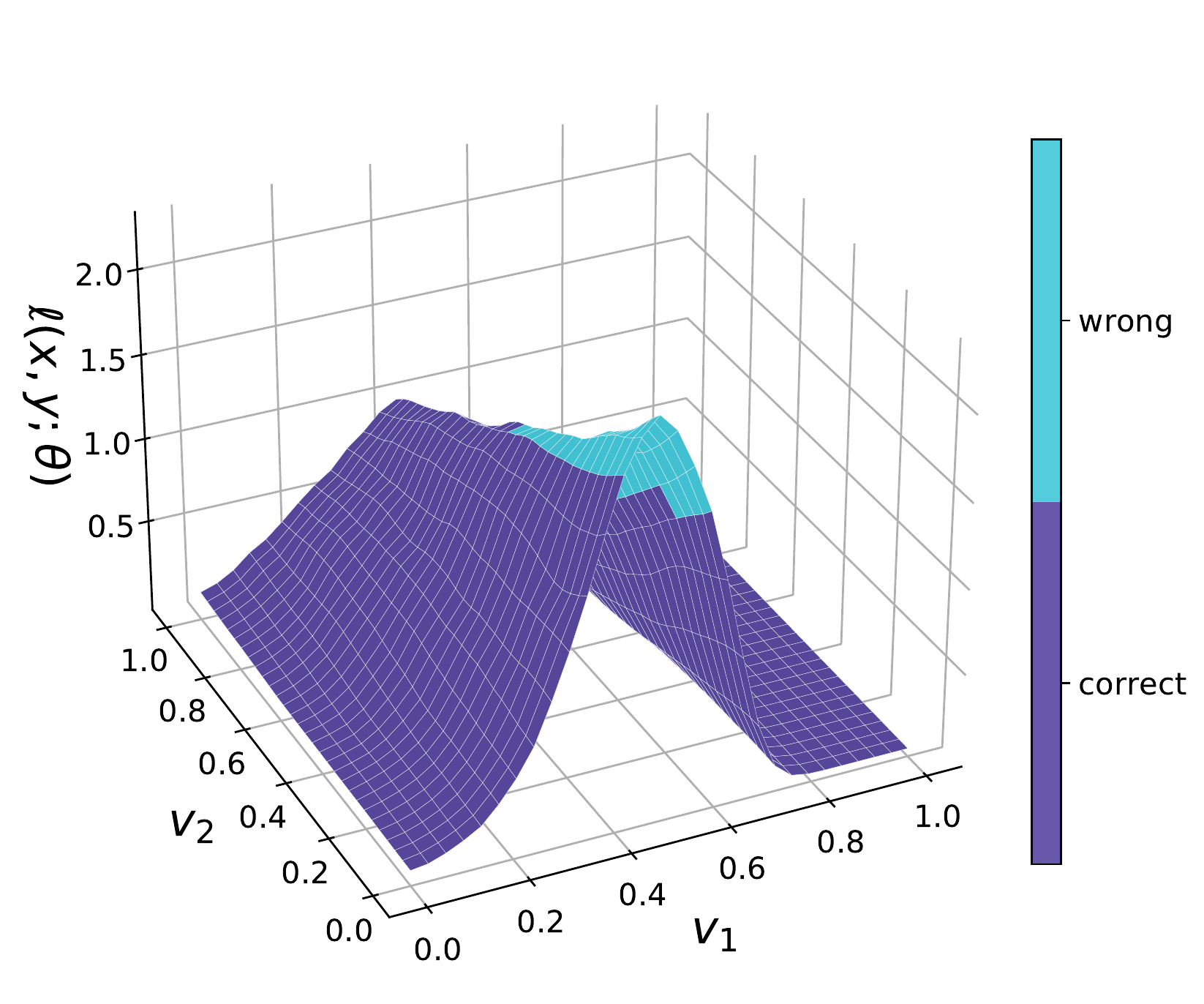}
        \caption{Loss surface}
        \end{subfigure}
    \end{minipage}
    \begin{minipage}{0.075\linewidth}
        \begin{subfigure}[t]{\textwidth}
        \includegraphics[width=\linewidth]{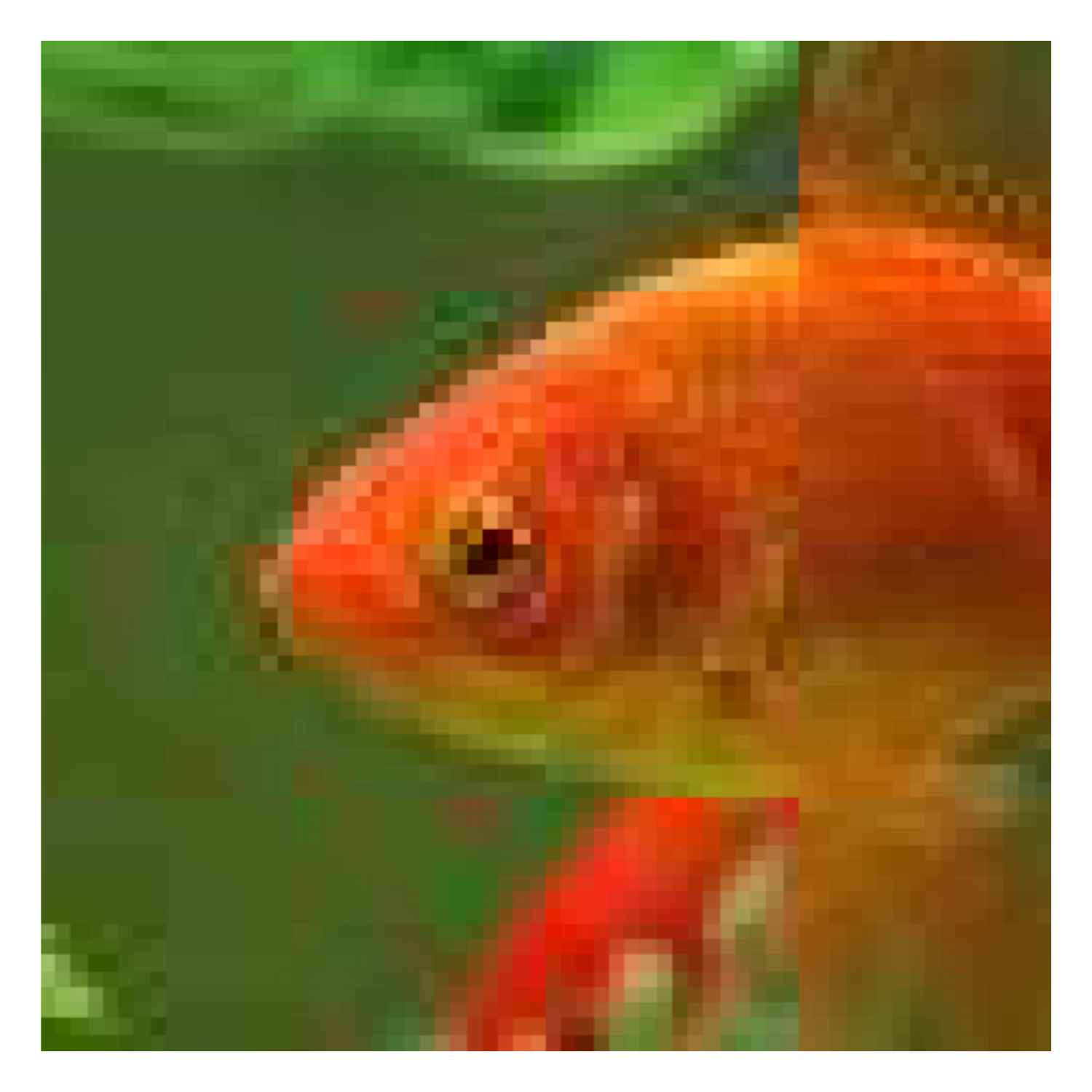}
        \caption{$x$}
        \end{subfigure} \\
        \begin{subfigure}[b]{\textwidth}
        \includegraphics[width=\linewidth]{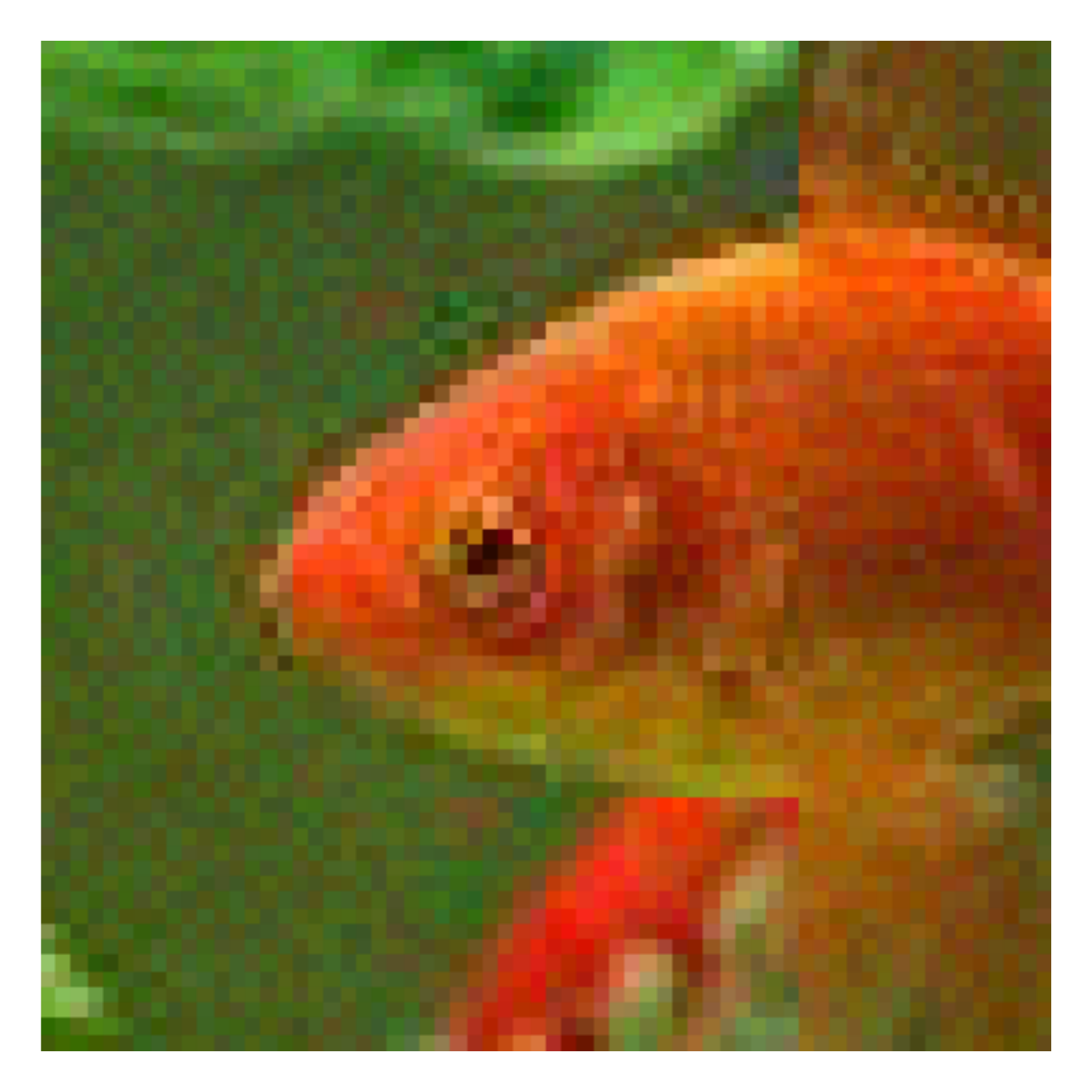}
        \caption{$x'$}
        \end{subfigure}
    \end{minipage}
    \begin{minipage}{0.3009\linewidth}
        \begin{subfigure}{\textwidth}
        \includegraphics[width=\linewidth]{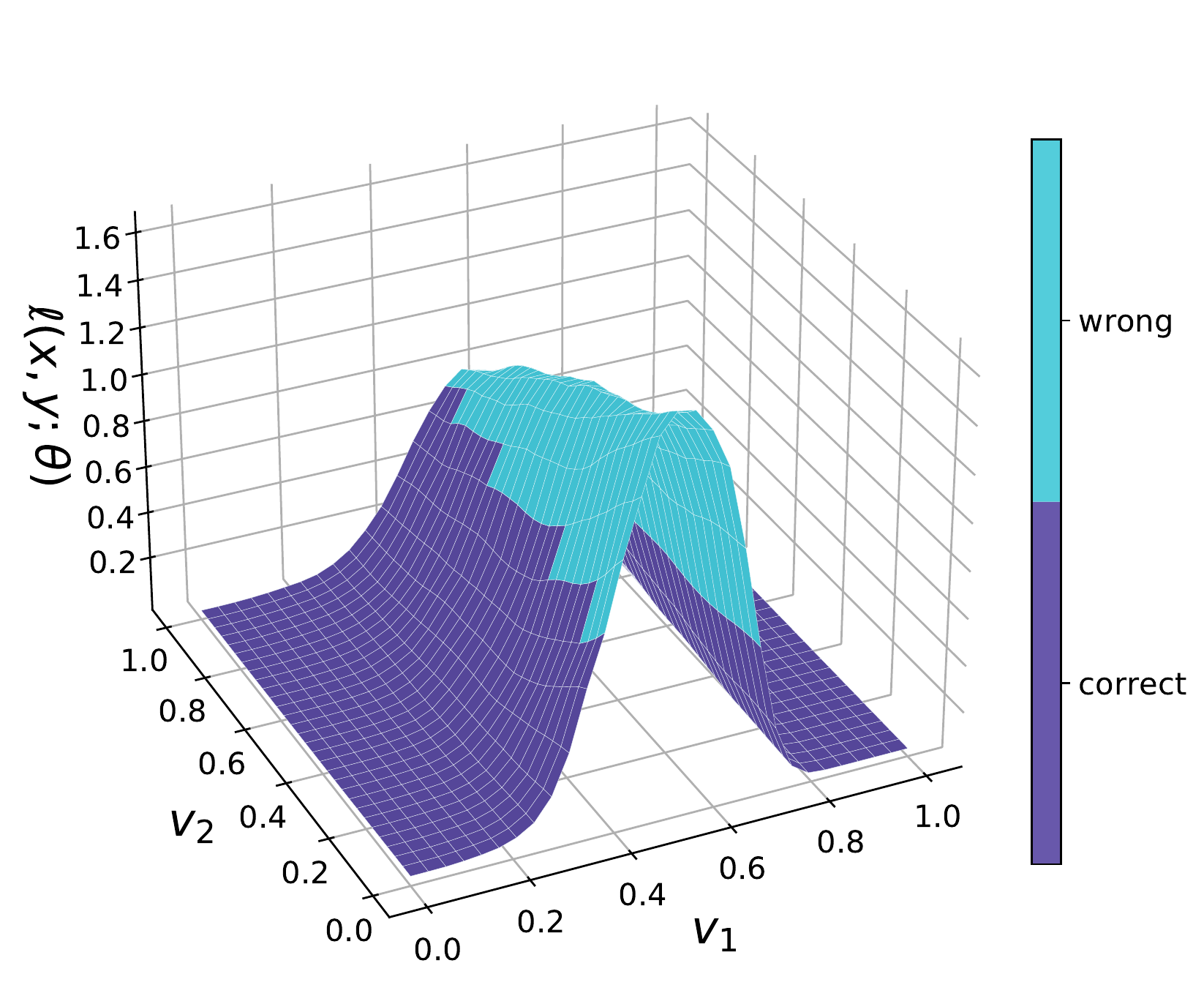}
        \caption{Loss surface}
        \end{subfigure}
    \end{minipage}
    \caption{(Tiny ImageNet) Direction of fast adversarial perturbation $v_1$ and random direction $v_2$. Adversarial example $x'=x+v_1$ is generated from original example $x$.}
    \label{fig:fast_test_tiny}
\end{figure*}

Finally, we plot distorted decision boundary of the fast adversarial trained model on Tiny ImageNet. Figure \ref{fig:fgsm_train_tiny} and Figure \ref{fig:fast_train_tiny} show the results on the training set and Figure \ref{fig:fgsm_test_tiny} and Figure \ref{fig:fast_test_tiny} on the test set. Interestingly, the model occasionally outputs correct answers for the adversarial image but not for the original image, for example, Figure \ref{fig:fast_train_tiny} (c). Due to this unusual behavior, fast adversarial training shows different result when we apply FGSM for only correctly classified images, as shown in Tables \ref{table:cifar10} and \ref{table:tiny}. This can be considered additional evidence that the model is overfitted to only single-step adversarial images. 

\clearpage
\section{Additional experiments} \label{appendix:c}

\subsection{FGSM adversarial training}
\begin{figure}[t]
\centering
    \begin{minipage}{0.39\linewidth}
        \begin{subfigure}{\textwidth}
        \includegraphics[width=\linewidth]{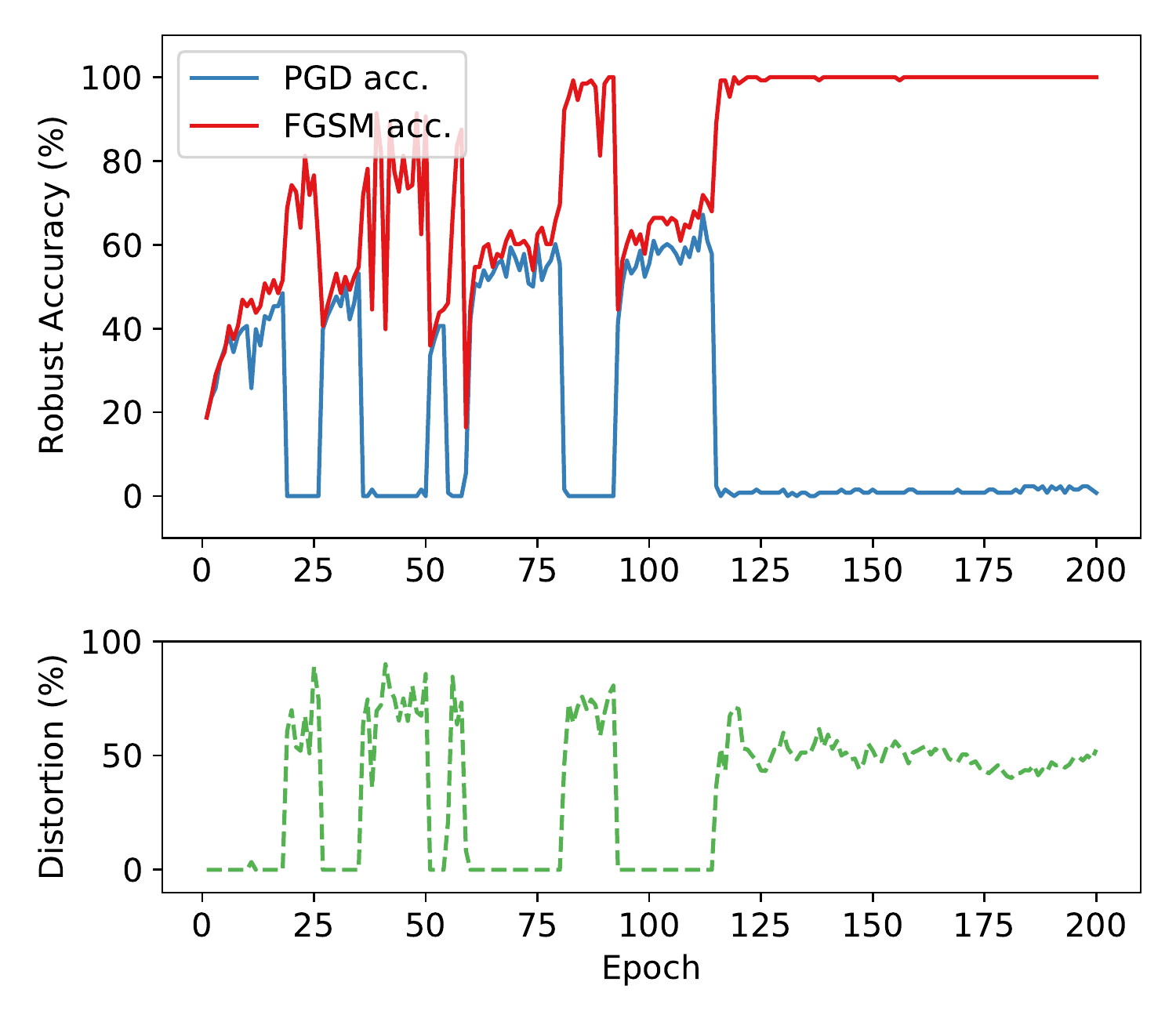} 
        \caption{Robust accuracy and distortion}
        \end{subfigure}
    \end{minipage}
    \begin{minipage}{0.45\linewidth}
            \begin{subfigure}{\textwidth}
            \includegraphics[width=\linewidth]{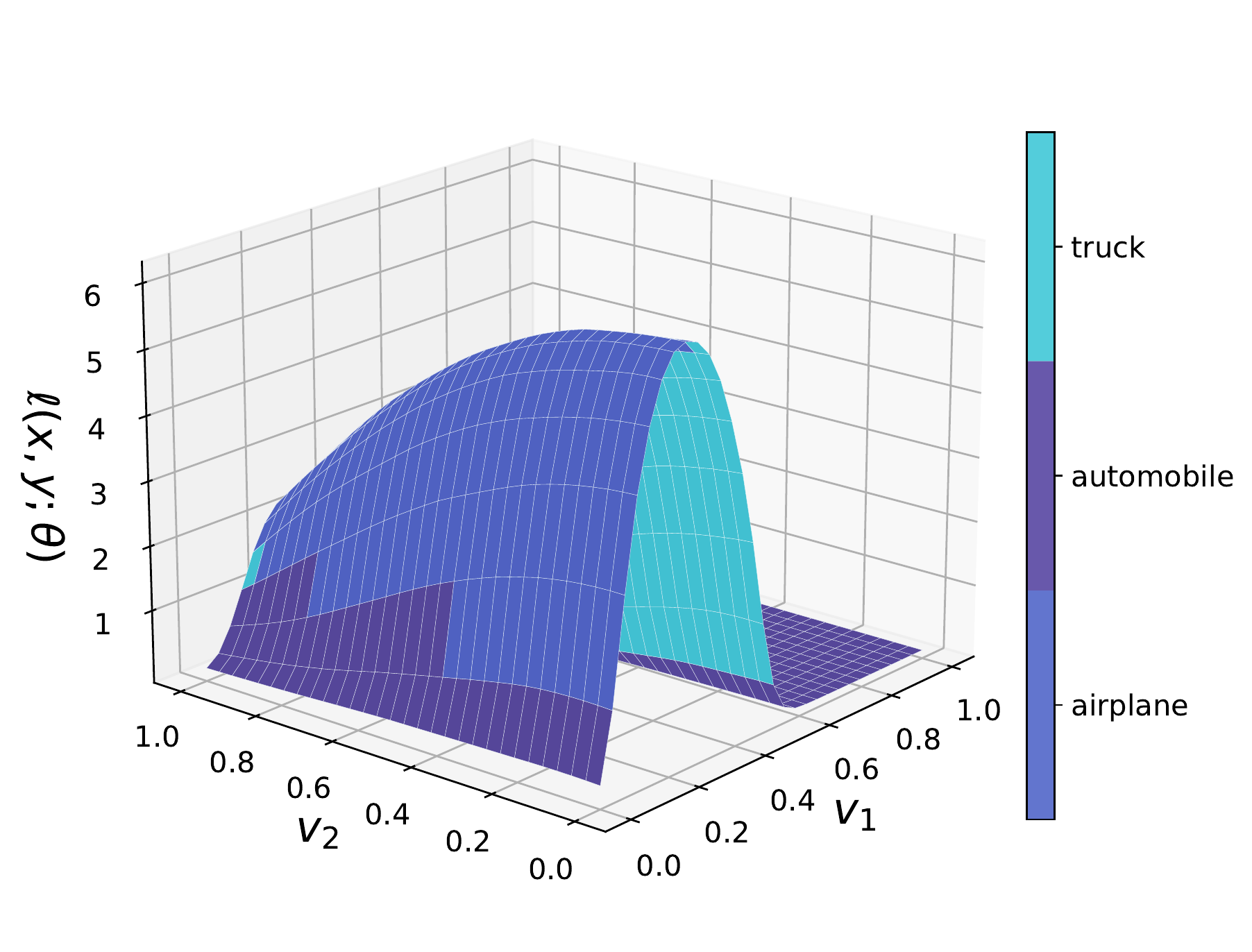} 
        \caption{Loss surface}
        \end{subfigure}
    \end{minipage}
\caption{(CIFAR10) Decision boundary distortion in FGSM adversarial training. Plot (a), the top plot shows robust accuracy of fast adversarial training against FGSM (red) and PGD7 (blue). In the bottom plot, distortion (green) is measured on the whole training images. Plot (b) shows loss surface with distorted decision boundary. Here, $v_1$ is a direction of the FGSM adversarial perturbation and $v_2$ is a random direction.}
\label{fig:fgsm}
\end{figure}

Although we mainly focus on fast adversarial training, we also found similar results on FGSM adversarial training. The left plot (a) of Figure \ref{fig:fgsm} shows distortion and robustness against FGSM and PGD7 on the training set. As shown in Figure \ref{fig:fgsm}, catastrophic overfitting occurs more frequently on FGSM adversarial training than fast adversarial training. Considering FGSM outputs the same adversarial images for the same inputs, the loss function is easier to be distorted because it is more difficult to train images within distorted intervals. The right plot (b) of Figure \ref{fig:fgsm} illustrates the existence of three different labels in the direction of adversarial perturbation.

\subsection{Varying the number of checkpoints $c$ and the maximum perturbation $\epsilon$}
\begin{figure}[t]
\centering
    \begin{minipage}{0.45\linewidth}
        \begin{subfigure}{\textwidth}
        \includegraphics[width=\linewidth]{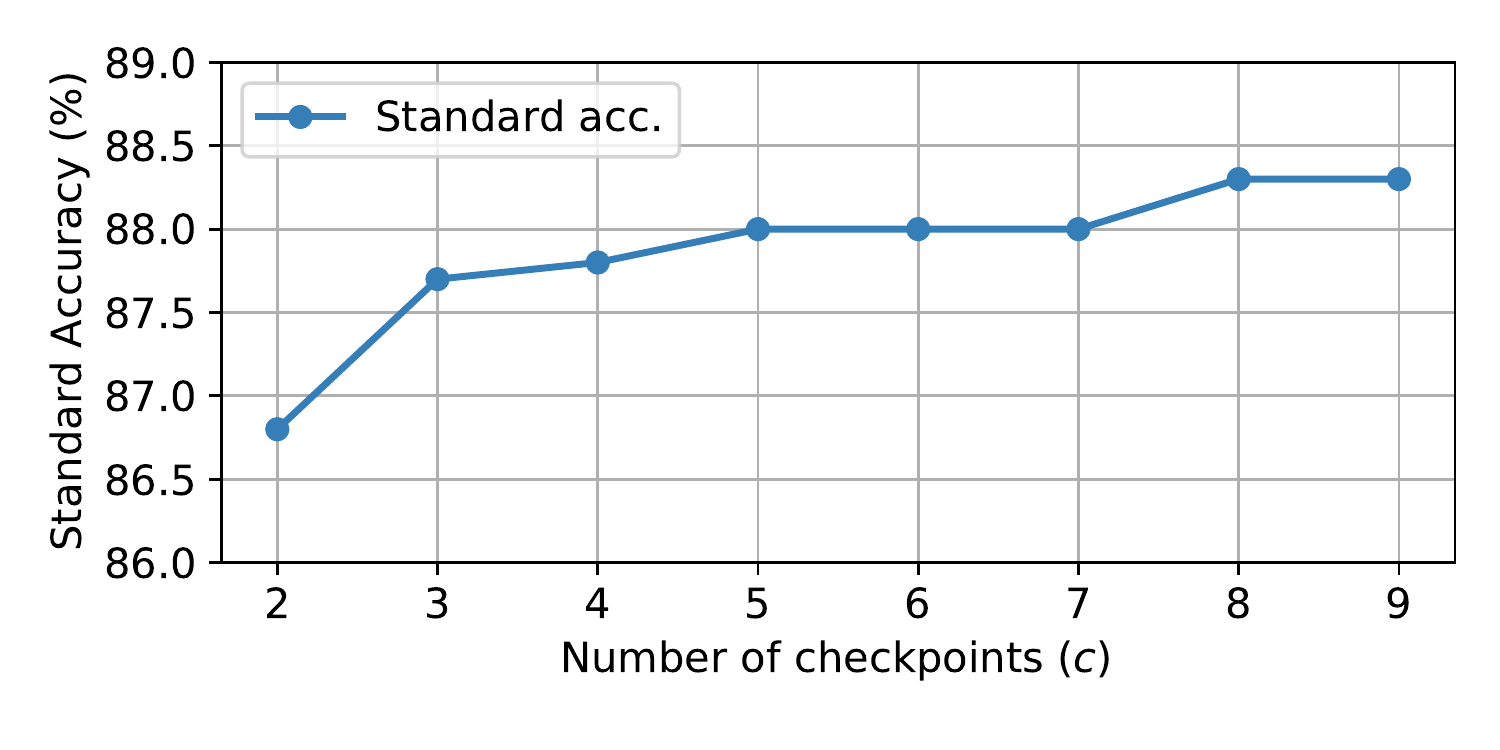} 
        \caption{Standard accuracy}
        \end{subfigure}
    \end{minipage}
    \begin{minipage}{0.45\linewidth}
            \begin{subfigure}{\textwidth}
            \includegraphics[width=\linewidth]{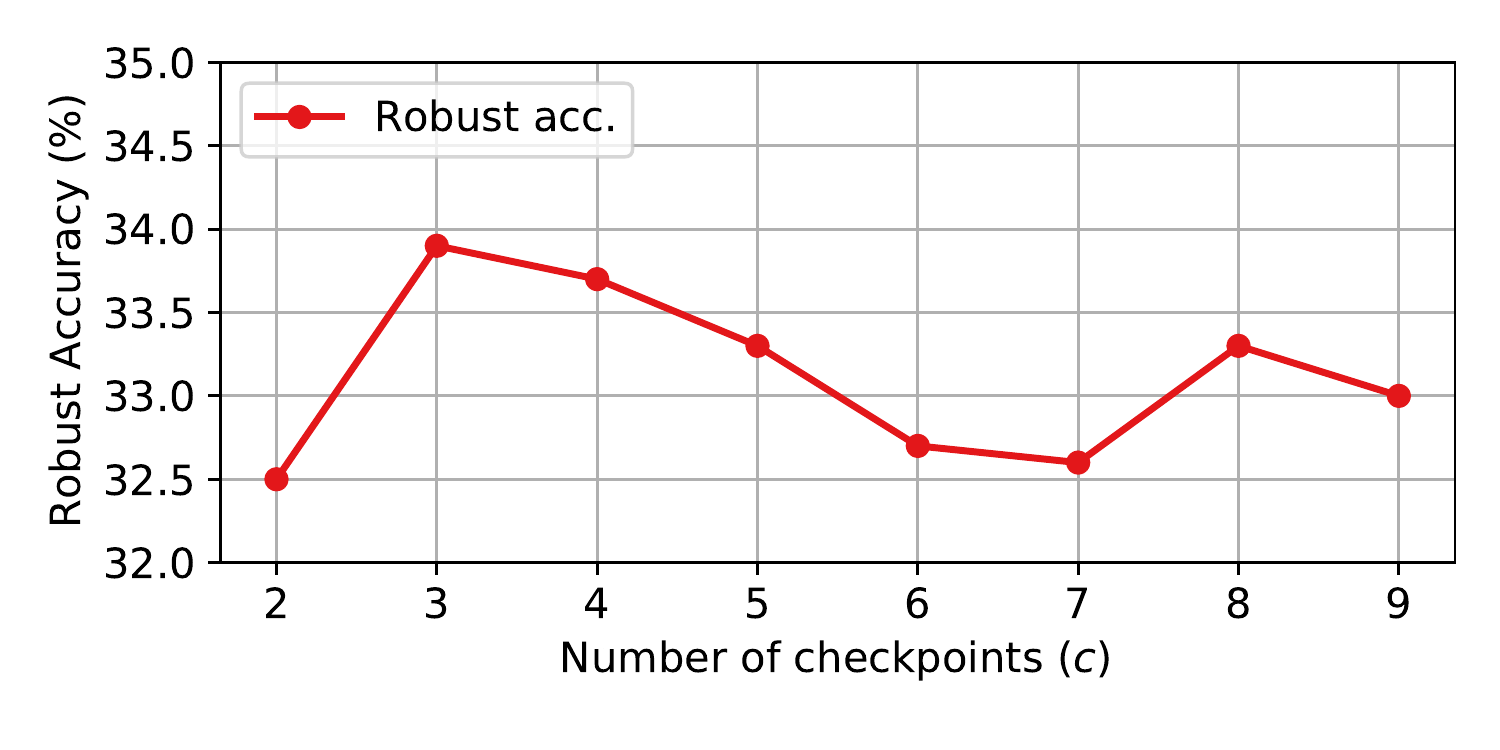} 
        \caption{Robust accuracy}
        \end{subfigure}
    \end{minipage}
\caption{(CIFAR10) Standard accuracy and robust accuracy against PGD50 with 10 random restarts under different number of checkpoints $c$.}
\label{fig:checkpoint}
\end{figure}

Here, we add the results of varying the number of checkpoints $c$. Figure \ref{fig:checkpoint} demonstrates standard accuracy and robust accuracy against PGD50 with 10 random restarts. First, the proposed method successfully avoids catastrophic overfitting compared to fast adversarial training. Again, we note that fast adversarial training shows 0\% accuracy against PGD. A higher $c$ shows a larger standard accuracy. However, we were unable to find a link between $c$ and robustness. Table \ref{table:checkpoints} shows the total training time for each number of checkpoints $c$.

\begin{table}[t]
\centering
\caption{Training time for each number of checkpoints $c$. The percentage in parentheses indicates the increased training time divided by the one of fast adversarial training.}
\begin{tabular}{c|c|c|c|c|c|c|c|c}
\hline
$\mathbf{c}$ & \textbf{2}   & \textbf{3}   & \textbf{4}   & \textbf{5}   & \textbf{6}   & \textbf{7}   & \textbf{8}   & \textbf{9} \\ \hline
\textbf{Time}                     & 3.5 & 3.8 & 4.4 & 4.6 & 4.8 & 5.1 & 5.2 & 5.5
\\
\textbf{(hour)}                     & (+9.4\%) & (+21.9\%) & (+37.5\%) & (+43.8\%) & (+50.0\%) & (+59.4\%) & (+62.5\%) & (+71.9\%) 
\\ \hline
\end{tabular}%
\label{table:checkpoints}
\end{table}

\begin{figure}[t]
\centering
    \begin{minipage}{0.23\linewidth}
        \begin{subfigure}{\textwidth}
        \includegraphics[width=\linewidth]{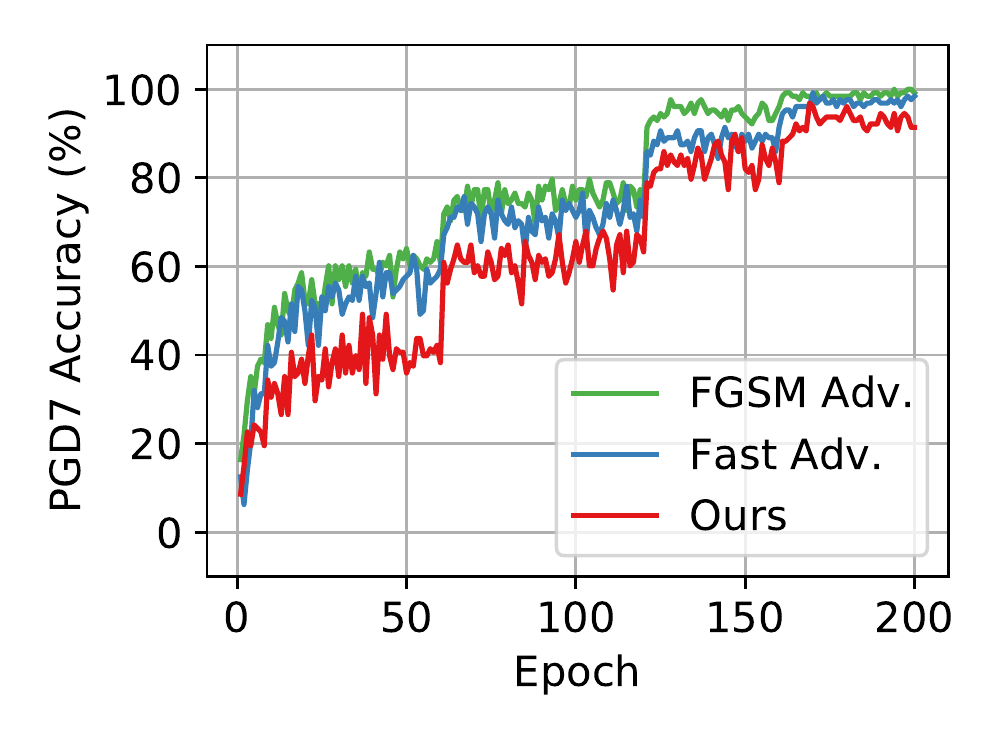} 
        \caption{$\epsilon=5/255$}
        \end{subfigure}
    \end{minipage}
    \begin{minipage}{0.23\linewidth}
            \begin{subfigure}{\textwidth}
            \includegraphics[width=\linewidth]{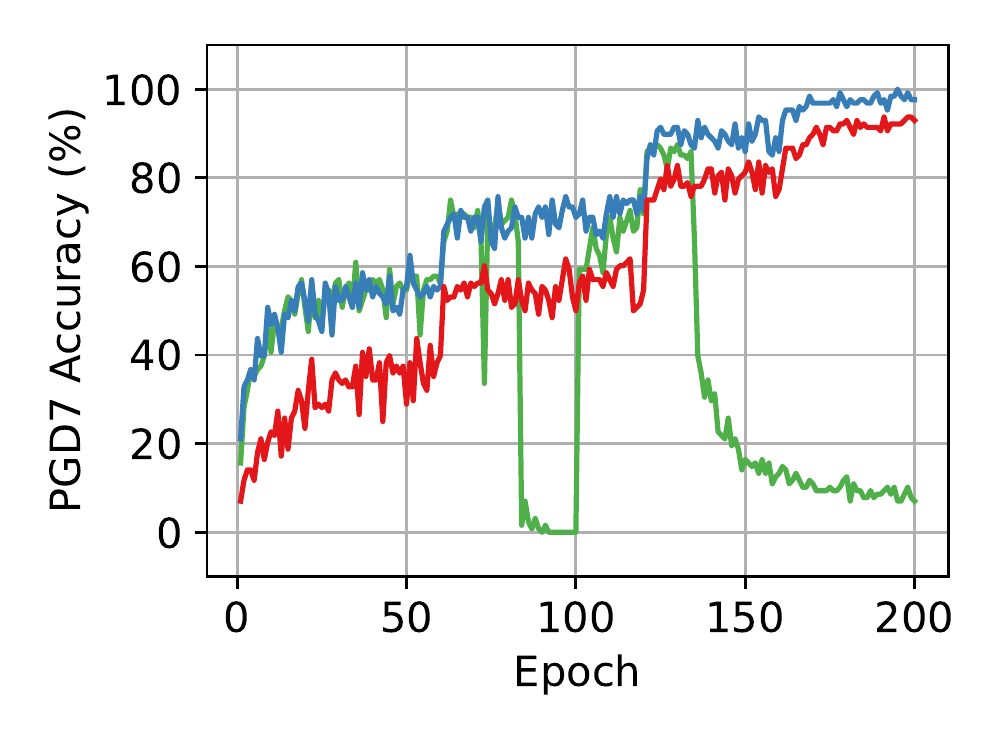} 
        \caption{$\epsilon=6/255$}
        \end{subfigure}
    \end{minipage}
    \begin{minipage}{0.23\linewidth}
            \begin{subfigure}{\textwidth}
            \includegraphics[width=\linewidth]{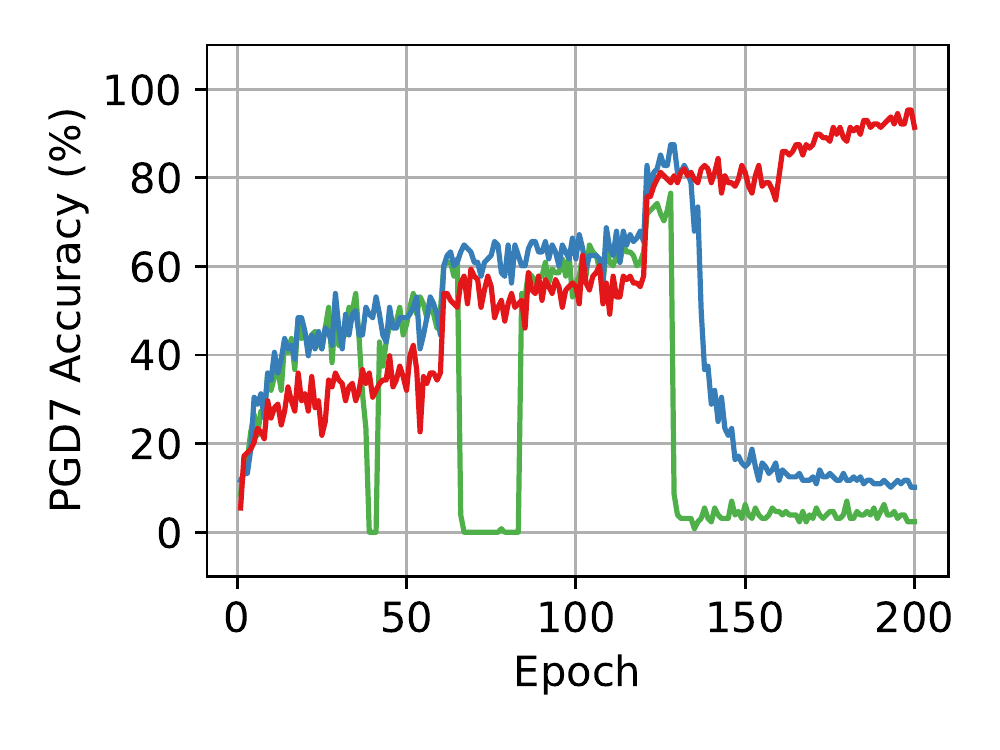} 
        \caption{$\epsilon=7/255$}
        \end{subfigure}
    \end{minipage}
    \begin{minipage}{0.23\linewidth}
            \begin{subfigure}{\textwidth}
            \includegraphics[width=\linewidth]{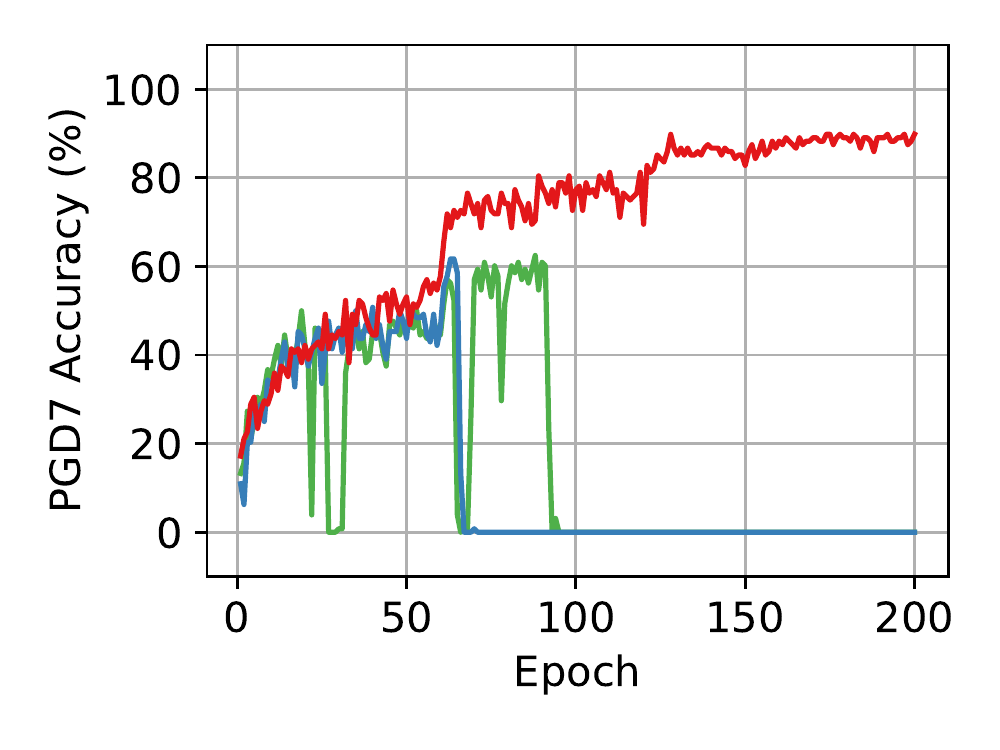} 
        \caption{$\epsilon=8/255$}
        \end{subfigure}
    \end{minipage}
\caption{(CIFAR10) Robust accuracy against PGD7 on the training batch under different maximum perturbation $\epsilon$.}
\label{fig:eps}
\end{figure}

We also tested various maximum perturbations $\epsilon$, as shown in Figure \ref{fig:eps}. All methods, including FGSM adversarial training, achieve nearly 100\% accuracy against PGD7 under $\epsilon = 5/255$. However, as $\epsilon$ increases, FGSM and fast adversarial training could not avoid catastrophic overfitting one after another. The proposed method shows stable training robust accuracy for all values of $\epsilon$. Here, we note that the proposed method also shows stable adversarial training for $\epsilon = 10/255$ with $m=5$.

\subsection{Cyclic learning rate schedules}

\begin{table*}[ht]
\centering
\caption{Standard and robust accuracy (\%) and training time (min) on CIFAR10.}
\resizebox{0.80\textwidth}{!}{%
\begin{tabular}{lcccccrr}
\hhline{========}
&\textbf{Method} & \textbf{Standard}     & \textbf{FGSM}         & \textbf{PGD50}        & \textbf{Black-box}   & \textbf{Time (min)} \\ \hhline{========}
\textbf{Multi-step} 
& PGD2 Adv.           & 85.5$\pm$0.1&54.6$\pm$0.1&45.0$\pm$0.0&84.4$\pm$0.1&41.0     \\
&PGD4 Adv.            & 85.3$\pm$0.1&54.9$\pm$0.3&45.3$\pm$0.1&84.4$\pm$0.1&62.0      \\
&PGD7 Adv.            & 81.5$\pm$0.0&\textbf{55.7$\pm$0.2}&\textbf{49.2$\pm$0.1}&80.8$\pm$0.1&92.0     \\
&TRADES          & \textbf{86.6$\pm$0.5}&52.2$\pm$0.1&42.6$\pm$0.6&\textbf{85.3$\pm$0.6}&125.0      \\ \hline
\textbf{Single-step}
&Fast Adv.           &84.3$\pm$0.2&\textbf{55.6$\pm$0.1}&45.5$\pm$0.1&83.3$\pm$0.4&28.0      \\
&GradAlign       &82.1$\pm$1.6&54.2$\pm$0.0&\textbf{46.9$\pm$0.9}&81.1$\pm$1.4&122.5      \\
&Ours ($c=2$)       &88.8$\pm$0.4&52.2$\pm$0.3&39.1$\pm$0.1&87.3$\pm$0.1&31.0       \\
&Ours ($c=3$)      & \textbf{89.1$\pm$0.0}&51.4$\pm$0.3&37.8$\pm$0.5&\textbf{87.6$\pm$0.1}&33.0
     \\ \hhline{========}
\end{tabular}%
}
\label{table:cyclic}
\end{table*}

Previously, step-wise learning rate schedule has been used at a large proportion; however, \citet{smith2019super} have discovered that cyclic learning rate schedule helps the model converge with small epochs. Because of its advantages, many researchers have adopted cyclic learning rate schedule. Thus, we conducted additional experiments using cyclic learning rate schedule. We use a maximum learning rate of $0.3$, and linearly continues to increase until 15 out of 30 epochs.

First, note that only a few epochs are normally used to train models using cyclic learning rate schedule and thus there is a small chance of catastrophic overfitting as \cite{wong2020fast} discovered. This indicates that the proposed method cannot show the strength of preventing catastrophic overfitting. As shown in Table \ref{table:cyclic}, among the single-step adversarial training methods, GradAlign shows the best performance at 46.9\% against PGD50. However, it is 3-times slower than PGD2 adversarial training. The proposed method shows lower performance than PGD2 and GradAlign. It is important to ensure that the average perturbation $\mathbb{E}[\vert\vert\delta\vert\vert_{\infty}]$ of the proposed method does not converge to the maximum perturbation $\epsilon$ unlike when using stepwise learning rate scheduling. Indeed, the proposed method achieved  performances of 40.2\% (+1.1\%p) and 39.5\% (+1.7\%p) for 60 training epochs. Considering that the recent methods to prevent catastrophic overfitting have shortcomings in time or have limitations in performance compared to multi-step adversarial training, better schemes for stable and robust single-step adversarial training is left as our future work.



\end{document}